\documentclass[final]{anthology-ch}         % for the submission

\usepackage{booktabs}
\usepackage{graphicx}
\usepackage{subcaption}
\usepackage[capitalise]{cleveref}
\usepackage{hyperref}
\hypersetup{
  colorlinks=true,
  linkcolor=blue,   % internal refs: \ref, \nameref, \autoref
  citecolor=blue,   % \cite
  urlcolor=blue     % \url, \href
}
\usepackage{longtable}
\usepackage[table]{xcolor}
\usepackage{tabularx}
\usepackage{ragged2e}
\usepackage{caption}
\usepackage{wrapfig}
\usepackage{ltablex}
\keepXColumns
\usepackage{fancyvrb}
\usepackage{fvextra}                                                                       \fvset{breaklines,breakanywhere,breaksymbolleft={},breakindent=2em}   \usepackage{csquotes}
\usepackage{xltabular}

\definecolor{skipred}{RGB}{232, 48, 70}

\title{Plot Twisting: Linearity and Uniformity\\in Literary Summaries}

\author[1]{Rebecca M.~M.~Hicke}[
  orcid=0009-0006-2074-8376
]

\author[2]{Sil Hamilton}[
  orcid=0000-0002-6579-4628
]

\author[2]{David Mimno}[
  orcid=0000-0001-7510-9404
]

\author[3, 4]{Ross Deans Kristensen-McLachlan}[
  orcid=0000-0001-8714-1911
]

\affiliation{1}{Department of Computer Science, Cornell University}
\affiliation{2}{Department of Information Science, Cornell University}
\affiliation{3}{Department of Linguistics, Cognitive Science, and Semiotics, Aarhus University}
\affiliation{4}{Center for Contemporary Cultures of Text, Aarhus University}

\keywords[eng]{summarization, narrative understanding, plot extraction, cultural analytics, language models}

\pubyear{2025}
\pubvolume{1}
\pagestart{1}
\pageend{1}
\paperorder{1}
\conferencename{Proceedings of Conference XXX}
\conferenceeditors{Editor1 Editor2}
\doi{00000/00000}  
\customhead{}  

\begin{document}
% \begin{wordcountsections}   
\maketitle

\begin{abstract}
Summaries of literary works such as novels are often used as stand-ins for the works themselves. Such substitutions are based on the assumption that summaries act as largely consistent distillations of those source works, but little existing empirical research has tested this assumption. We attempt to do so using a combination of manual and LLM-based annotation, from which we construct a dataset mapping sentences from 150 Wikipedia summaries to their respective source chapters. From these mappings, we measure summary \textit{linearity}, the degree to which it maintains the source's order of events; and \textit{uniformity}, the degree to which a summary spreads attention equally across its source material. We find that few summaries satisfy these constraints fully and that there exists a wide range of variation in both measures across our corpus. Examining when and how summaries break with linearity and uniformity, we identify differences in how novels and their summaries express plot, particularly in terms of how clearly and how prominently narrative details are presented.
\end{abstract}

\section{Introduction}
Researchers regularly use novel summaries as short proxies for literary texts in domains as varied as education \cite{dunbar2025writing, fu2026college}, computational literary studies \cite{bamman2013learning, madaan2017analyzing,madaan2018judging, aly2019hierarchical,chaturvedi2017unsupervised,kar2020multi,hatzel2024tell}, and content recommendation systems \cite{mooney2000content, wang2015collaborative, adhikari2025leveraging}. These studies often treat summaries as distillations of the underlying causal structure, or plot, of novels, an approach that has been supported by research from cognitive science \cite{LEHNERT1981293,TRABASSO1985612} and literary studies \cite{smith1980narrative,iser2014interaction}. In this work, we evaluate this view of summaries by examining the structural consistency of a set of summaries produced under a common style guide. Specifically, we study two properties of Wikipedia novel summaries: \textit{linearity}, or whether the summary refers to events in the order they occur in the original work; and \textit{uniformity}, or whether the summary allocates equal space to describing each part of the original work. 

There are many reasons why a summary might deviate from linearity and uniformity.
Indeed, the Wikipedia guidelines for plot summaries explicitly state that ``a plot summary is not a recap'' and breaks from both principles are permitted and sometimes encouraged. The guidelines stress that ``a plot summary... should not cover every scene or every moment of the story,'' and that many plots ``[include] a lengthy middle section... on their way to the climatic encounter'' full of events which ``often clutter a plot summary with excessive and repetitive detail.'' On the subject of linearity, the guidelines state that ``events can be reordered'' if it ``makes the plot easier to explain,'' implicitly acknowledging that narratives are not always optimized for coherence alone.\footnote{Quotes pulled from \url{https://en.wikipedia.org/wiki/Wikipedia:How_to_write_a_plot_summary}, current as of November 17, 2026.} The decisions summary authors make in interpreting these guidelines can cause variance in summary linearity and uniformity, meaning that summaries may not actually consistently distill plots.

\begin{figure}[t]
  \centering
  \includegraphics[width=0.6\linewidth]{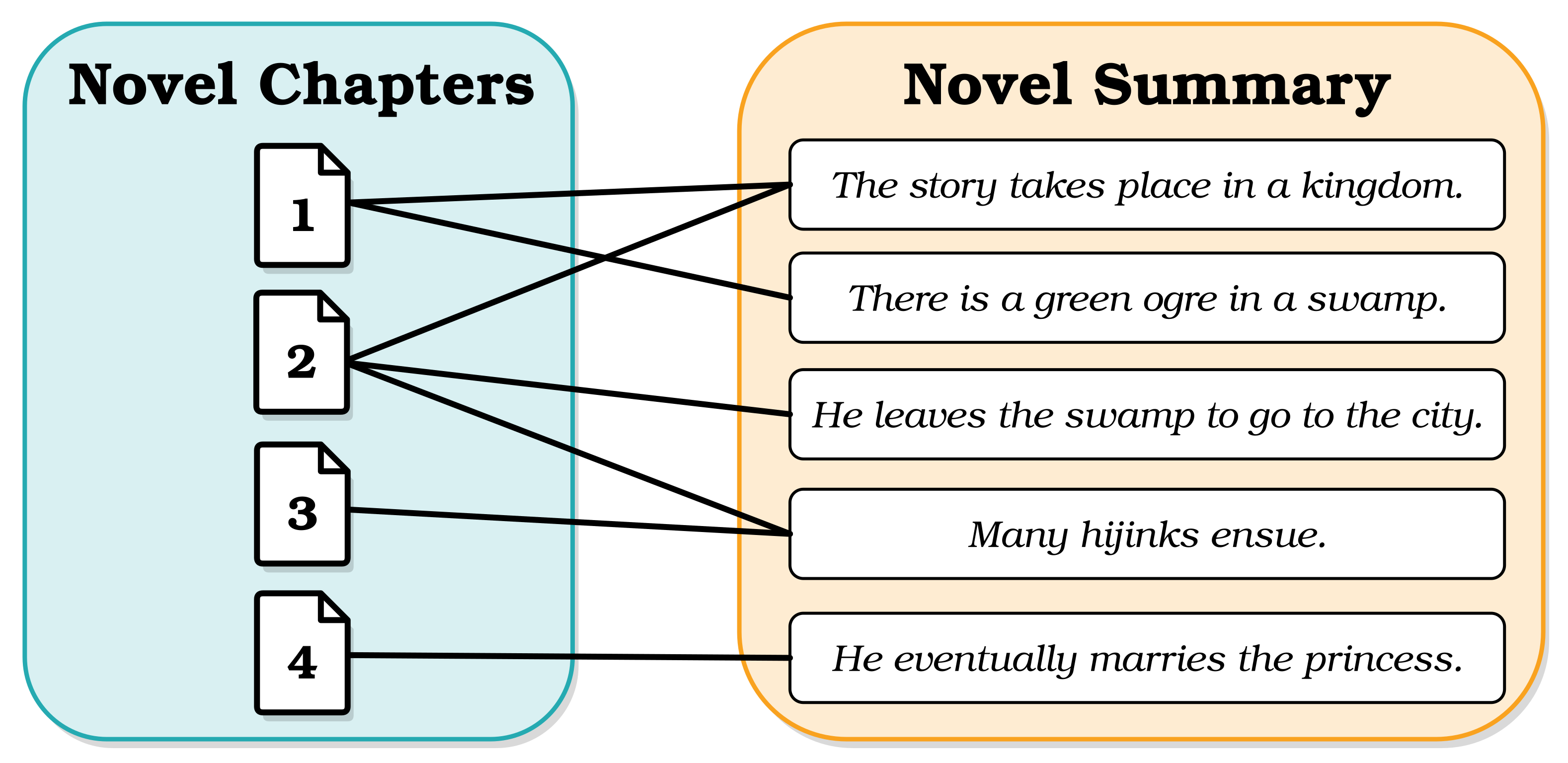}
  \caption{Our summary sentence to chapter alignment task. Novels are treated twice: they are first split into chapters, and then summarized. Summaries and chapters are then aligned.}
  \label{fig:fig1}
\end{figure}

This work presents a corpus of 150 human-written summaries from Wikipedia alongside the works they describe, as well as an automated method for aligning summary sentences with the chapters (or other delineated narrative segments) in the original work (cf. \autoref{fig:fig1}).\footnote{Our data is available at \url{https://github.com/c2-lab/plot-twisting}.}
This method is developed using model-in-the-loop prompt development and achieves $\sim$84\% F1 on a dataset of four human-aligned summaries and texts. We use a combination of quantitative metrics and close readings to analyze the aligned summaries and characterize patterns in summary writing.

Examining the aligned data, we find that summaries frequently violate linearity and uniformity to some degree, but differ in why and how these violations occur. For example, some narrative elements or information may be explicitly described in summaries but subtly conveyed in the original works; some events are more consequential and interesting and thus emphasized in summaries; and authors of literary works may choose to reveal information in ways that heighten emotion or drama where summarizers prioritize comprehension. Additionally, we note that, although summarization has long served as a task in NLP research, our findings suggest that contemporary frontier LLMs struggle to \textit{reverse} summarization, i.e. grounding summary sentences in their source texts. Overall, we provide evidence that literary summaries, even when written under a unified style guide, do not necessarily distill plots in similar ways. This suggests users of literary summaries should be cognizant of how they may vary when using them as proxies for literary works.

\section{Related Work}
\paragraph{Stories, discourse, and plot.}
Literary texts are not required to relate story events in a clear and unambiguous manner \cite{gius2022towards}.
In fact, they may intentionally impede narrative comprehension for artistic purposes by employing stylistic devices such as retelling events non-chronologically or unreliable narrators.
Implicit in this is the idea that the text and the story it describes may not always cleanly align with one another.
Russian formalists in the mid-20th century (and the narratologists who later built on their work) drew a distinction between the text's \textit{story}, or what happens, and its \textit{discourse}, or how what happens is relayed to the reader \cite{tomashevsky1965russian,genette1983narrative,prince_narratology_2012}.
% Those interested in narrative form may focus on the discourse, while those interested in narrative content can be understood to be interested in the story.
One important story element is the \textit{plot}, which is the causal ordering of character actions over chronological time \cite{Kukkonen2014Plot}.
The typical literary summary can be understood to be specifically describing these \textit{plot lines}, and so can reveal what readers believe to be the causal structure underlying a narrative \cite{LEHNERT1981293,TRABASSO1985612}.
Summaries have been used for a variety of research applications, including studying the differences between child and adult reading comprehension \cite{johnson2014you} and serving as transient structures for extracting plot-related entities like characters \cite{srinivasan2022character} among other uses \cite{piper2025towards,hatzel2024story,hayton2020narrative}. Previous research in the cognitive sciences has demonstrated the important role of narrative structure and coherence on human memory, particularly in the recall and summarization of information \cite{CohnSheehy2020NarrativesBT, Georgiou2023LargescaleSO}. However, little prior work has examined how literary summaries typically operationalize plot and why some story content is privileged over the rest \cite{mani2025plot}. We address this point throughout this paper.

\paragraph{Summaries and computational narratology.}
Early computational work engaging with summaries largely sought to create systems capable of automatically producing journalistic headlines and scientific abstracts \cite{mani1999advances,zhang2025systematic}.
Researchers hoped that such systems would eventually prove capable of parsing the semantic content of natural language more generally.
Early summarization datasets were therefore usually drawn from news sources like CNN and NYT \cite{sandhaus2008new,hermann2015teaching,grusky2018newsroom}.
Until the 2020s, other common target domains included medical records \cite{baumel2016topic}, legislative documents \cite{kornilova2019billsum}, and short stories \cite{kazantseva2010summarizing}.
These texts shared characteristics that made them convenient for early statistical and neural approaches to natural language processing such as short mean document lengths and copious availability.
With the introduction of the transformer architecture and the subsequent increase in tractable context lengths, machine learning researchers began using the summarization of long narrative texts as a test of text comprehension \cite{zhao2023narrasumlargescaledatasetabstractive,kim2024fables,wu2021recursively}.
However, this research largely treats summaries as transient artifacts produced in pursuit of a different task.
Works that do consider summary quality largely focus on logical consistency and clarity \cite{goyal2022snac,chang2024booookscore}.
Existing research on aligning summaries with literary texts either predominantly works on a sub-novel level due to technical limitations \cite{bamman2013new,chaudhury2019shmoop} or focuses on \textit{generating} summaries \cite{ladhak2020exploring}.
Other prior work has explored \textit{how much} of a book summaries cover, but reliance on n-gram overlap and BERT-based semantic similarity to align summaries with source texts limits utility \cite{kryscinski2021booksum}.
In contrast, we develop a method of aligning summaries to texts based on content, which permits more nuanced comparisons.

\section{Data}
% begin{figure}[t]
%   \centering
%   \includegraphics[width=1\linewidth]{images/process_chart.png}
%   \caption{A diagram of how summaries are aligned with the chapters of each novel on a many-to-many basis.}
%   \label{fig:dataChart}
% \end{figure}

% \subsection{Human Summaries}

Our dataset contains 150 novels in the public domain under US law and their Wikipedia plot summaries. We select Wikipedia summaries specifically because our interest is in understanding how summaries represent and reorganize narratives, not in assessing how any individual reader summarizes a story. Because Wikipedia summaries are typically co-written by multiple authors, they will not necessarily be biased towards the writing style of an individual. We select works from the public domain to allow the release of our data for reproducibility and future research.

We gathered full novel texts from Project Gutenberg.\footnote{\url{https://www.gutenberg.org}} To identify English-language novels available on Project Gutenberg with English-language summaries on Wikipedia, we first collected a candidate list of Gutenberg works in English. We then removed common single-word titles (which can overlap with generic Wikipedia pages) and searched for Wikipedia pages matching our candidate list containing subheadings like ``Plot'' and ``Summary.'' This process yielded 80 candidate texts. We then sourced an additional 70 novels and their corresponding Wikipedia summaries from the BookSum dataset \cite{kryscinski2021booksum}, selecting novels recently released into the public domain. The dataset includes several translated novels; when possible, we select standard English-language translations which should correspond to the English-language Wikipedia summaries.

We automatically split novels into chapters using the \texttt{chapterize} package and manually validate the resulting segmentation. We then standardized punctuation and removed illustration markers to ensure footnotes are placed in the correct chapter file. We extracted Wikipedia summaries manually,\footnote{All summaries were gathered between 02/18/2026 and 02/19/2026.} excluding paragraphs solely containing publication information and other similar meta-commentary. We finally split summaries into sentences using the stock \texttt{spaCy} sentence tokenizer. We provide all novels and their corresponding Project Gutenberg IDs in Appendix \ref{app:novels}.

The novels contain between 3 to 86 chapters (with an average of 32 per book) and are between 9,780 to 488,101 tokens long, with an average of 147,173 tokens and a mean of 4,519 tokens per chapter. The summaries are 992 tokens on average, ranging from 243 to 3,534 tokens long.\footnote{We find the total length of each novel and summary in tokens using the \texttt{tiktoken} package with the GPT-5 vocabulary.}
% These values are used to calculate the log-ratio of the summary length to book length. 

\section{Generating Sentence-to-Chapter Mappings}
\label{sec:alignments}

Having segmented books into chapters and summaries into sentences, our goal was next to create bipartite graphs where each edge connects one chapter to one sentence.
A summary sentence is mapped to a novel chapter if it describes some event that is dramatized in that chapter.
An individual sentence can map to any number of chapters, including zero, and vice versa.

We find these mappings are themselves a challenge to create.
We began with an evaluation mapping dataset of four novels randomly sampled from the corpus --- \textit{Rookwood} by William Harrison Ainsworth, \textit{Kazan} by James Oliver Curwood, \textit{Kim} by Rudyard Kipling, and \textit{Carmilla} by Joseph Sheridan Le Fanu --- and their Wikipedia summaries.
With two expert annotators, we found a high inter-annotator agreement (IAA) with a mean Cohen's $\kappa$ of 0.74.
Disagreements between annotators were resolved via discussion.
This resulted in a final evaluation dataset containing 172 sentence-to-chapter mappings across all four novels.
Despite high IAA, the annotators reported that the mapping task was challenging due to the length of both the summaries and the novel texts; each novel took approximately 1--4 hours per annotator to complete.
While two (\emph{Carmilla} and \emph{Kazan}) were relatively straightforward, \emph{Rookwood}'s complex plot and \emph{Kim}'s complex language made their annotations more challenging.

This dataset was then used to design and evaluate prompts for mapping summary sentences to novel chapters.
An initial prompt was drafted by one of the human annotators and then iteratively refined using Claude Opus 4.8 \cite{anthropic2026opus48}.
Each iteration of the mapping prompt or prompts was evaluated on the four-book annotated dataset.
The model was instructed to group and categorize the errors from each run, then propose prompt edits to fix selected error types.
The human annotator directed the model in choosing and designing prompt edits throughout this process.

\begin{table}
  \small
  \centering
  \caption{Performance of Qwen 3.6 27B on the sentence-to-chapter mapping task. Note that scores vary considerably between novels with the F1 floating between $\sim$70 and $\sim$92.}
  \label{tab:alignResults}
  \begin{tabular}{rccc}
    \textbf{Novel} & \textbf{Precision} & \textbf{Recall} & \textbf{F1} \\
    \midrule
    \emph{Kazan} & 92.3 & 92.3 & 92.3 \\
    \emph{Carmilla} & 93.2 & 87.3 & 90.2 \\
    \emph{Kim} & 78.9 & 90.9 & 84.5 \\
    \emph{Rookwood} & 70.3 & 70.3 & 70.3 \\
    \midrule
    \textbf{Overall} & 83.7 [$\pm$ 9.6] & 85.2 [$\pm$ 8.8] & 84.3  [$\pm$ 8.6] \\
    \bottomrule
  \end{tabular}
\end{table}

Iterative development yielded a mapping pipeline consisting of three prompts.
\nameref{Prompt1} performs the preliminary summary sentence to chapter mapping and is run once for every novel chapter.
It accepts the entire novel summary with sentences indexed from 1--$n$, one novel chapter, and the set of summary sentence indices matched to previous chapters.
We chose to pass the model a single chapter instead of the entire novel text because performance degraded with novel length.
This prompt instructs the model to match a sentence to a chapter if an event described in it occurs in the provided chapter \textit{or} if it describes a state, relationship, or characterization first introduced or directly dramatized in the chapter.
The prompt also advises the model to avoid sentence-to-chapter matches under some conditions (e.g.~an event is alluded to but not shown, a recurring character or setting merely appears, etc) and describes when a sentence may be matched to multiple chapters.

The next two prompts in the pipeline are run on the output of the first.
\nameref{Prompt2} corrects the off-by-one errors common in the first prompt's output and is only run on summary sentences already matched to one or two adjacent novel chapters.
It accepts a single summary sentence and its index, the chapters directly before and after the already matched chapter(s) as context, and the candidate chapter(s).
Using the provided context, it instructs the model to determine whether one, both, or none of the candidate chapter(s) are true matches to the provided summary sentence.
It returns a list of the chapter number(s) it determines are true matches which are used to update the existing sentence-to-chapter mapping.
Finally, \nameref{Prompt3} is run only on sentence-to-chapter matches confirmed by Prompt 2.
This prompt accepts a single summary sentence and its index, the nearby summary sentences as context, a single chapter the sentence is matched to, and the index of the chapter the summary sentence would be matched to if the summary were perfectly linear.
Given this information, the prompt double-checks whether this summary sentence matches the provided chapter.
This prompt returns a binary judgment on whether the sentence and chapter represent a true match.
Only matches confirmed by this prompt are included in the final summary sentence to chapter matches.
%The full text of all three prompts is provided in \cref{sec:alignPrompts}.

All prompts were run on Qwen 3.6 27B \cite{qwen3.6-27b} quantized to 6 bits and loaded on a DGX Spark. Temperature was set to 0 and nucleus sampling (\texttt{top\_p}) was set to 1.0. The complete mapping pipeline achieved an F1 of 84.3\% over all four novels in the evaluation dataset. However, performance varied dramatically by book (\cref{tab:alignResults}), suggesting that differences in summary or novel complexity may impact mapping accuracy. Using this pipeline, we mapped all 150 Wikipedia summaries to their respective source texts (see bipartite graphs of all mappings in \cref{sec:bipartiteGraphs}). We consider the performance strong enough to proceed with analysis, but caution that individual novels' alignments may be prone to error.

\begin{figure}[t]
  \centering
  \begin{subfigure}[t]{0.49\textwidth}
    \includegraphics[width=\linewidth]{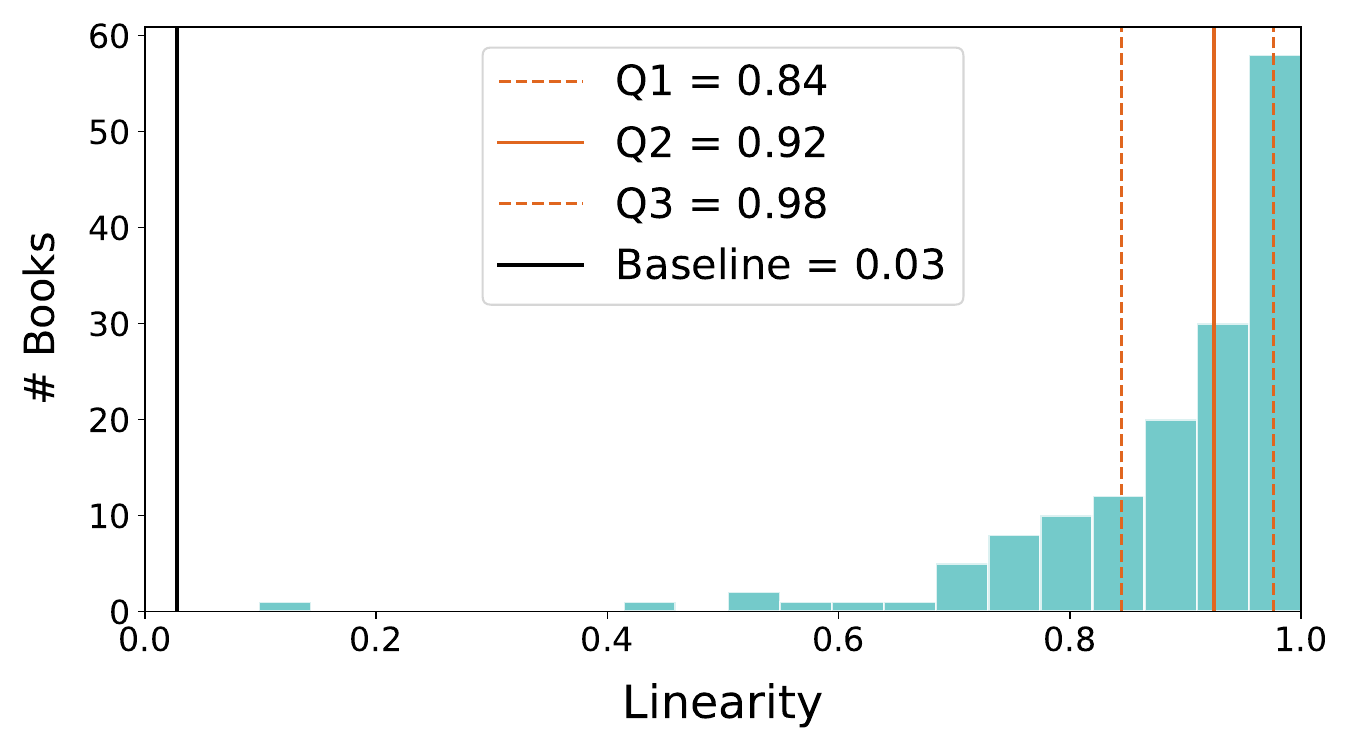}
    \phantomcaption
    \label{fig:hist-linearity}
  \end{subfigure}
  \hfill
  \begin{subfigure}[t]{0.49\textwidth}
    \includegraphics[width=\linewidth]{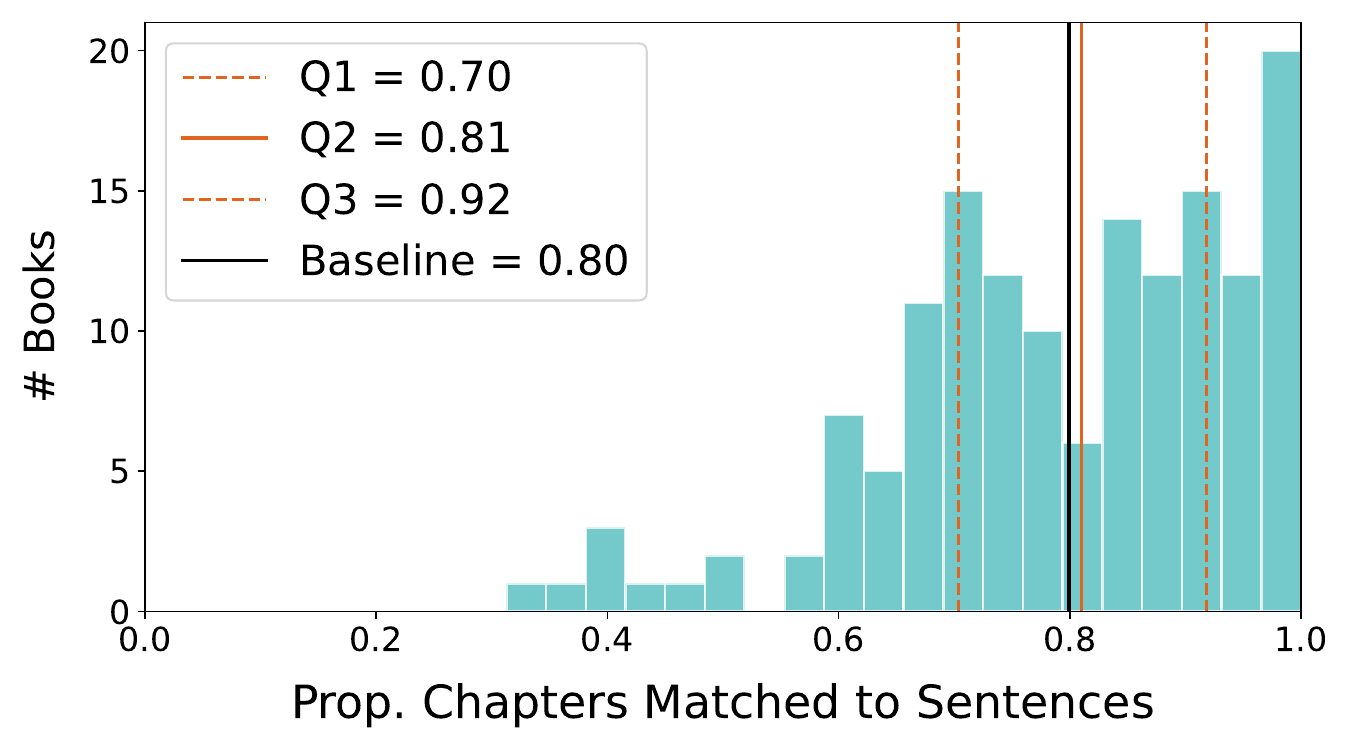}
    \phantomcaption
    \label{fig:hist-prop-chaps}
  \end{subfigure}
  \caption{Linearity (a) and the proportion of chapters matched to $\geq1$ summary sentence (b) across all 150 books. Solid orange lines mark the medians, dashed lines the quartiles. Solid black lines mark a random baseline, calculated from 200 random shuffles of the sentence and chapter matches respectively.}
  \label{fig:metric-histograms-1}
\end{figure}

\section{Summary Linearity}
We first evaluate adherence to the linearity principle.
Specifically, we order the sentence-to-chapter matches by summary sentence index, sorting the chapter indices matched to each sentence.
Then, we define linearity ($l$) as the Kendall's Tau correlation coefficient between the list of chapter indices ordered by the summary sentence they map to and the sorted version of this list.
Intuitively, this captures the extent to which the summary differs from its equivalent in perfect narrative order. 

Most summaries in the dataset are predominantly, but not perfectly, linear (\cref{fig:hist-linearity});
only 6.7\% (10 / 150) have $l=1.0$, but 62.7\% (94 / 150) have $l\geq0.9$.
Summaries with linearity values near the median (0.92) describe events in almost the same order as the original narrative with a small number of changes.
Further inspection reveals several reasons for these ``out-of-order'' sentence-to-chapter matches:
summaries may begin by framing the narrative using information only revealed later in the book (e.g.~Sentence 1 of \cref{sec:82-summary}; \cref{sec:bipartiteGraphs}, \cref{fig:pg82-bipartite});
single sentences may describe recurring events or a set of related events, which take place throughout the novel (e.g.~Sentence 8 of \cref{sec:70379-summary}; \cref{sec:bipartiteGraphs}, \cref{fig:pg70379-bipartite});
summaries may separate plot lines, even if they occur concurrently in the narrative (e.g.~Sentences 12 + 13 of \cref{sec:10007-summary}; \cref{sec:bipartiteGraphs}, \cref{fig:pg10007-bipartite});
or summaries may adjust the order in which information is revealed to increase clarity (e.g.~Sentence 50 of \cref{sec:10007-summary}; \cref{sec:bipartiteGraphs}, \cref{fig:pg10007-bipartite}).
Some out-of-order matches may also be artifacts of the mapping pipeline (e.g.~Sentence 14 of \cref{sec:10007-summary}; \cref{sec:bipartiteGraphs}, \cref{fig:pg10007-bipartite}).

\begin{figure}[t]
  \centering
  \begin{subfigure}[t]{0.49\textwidth}
    \includegraphics[width=\linewidth]{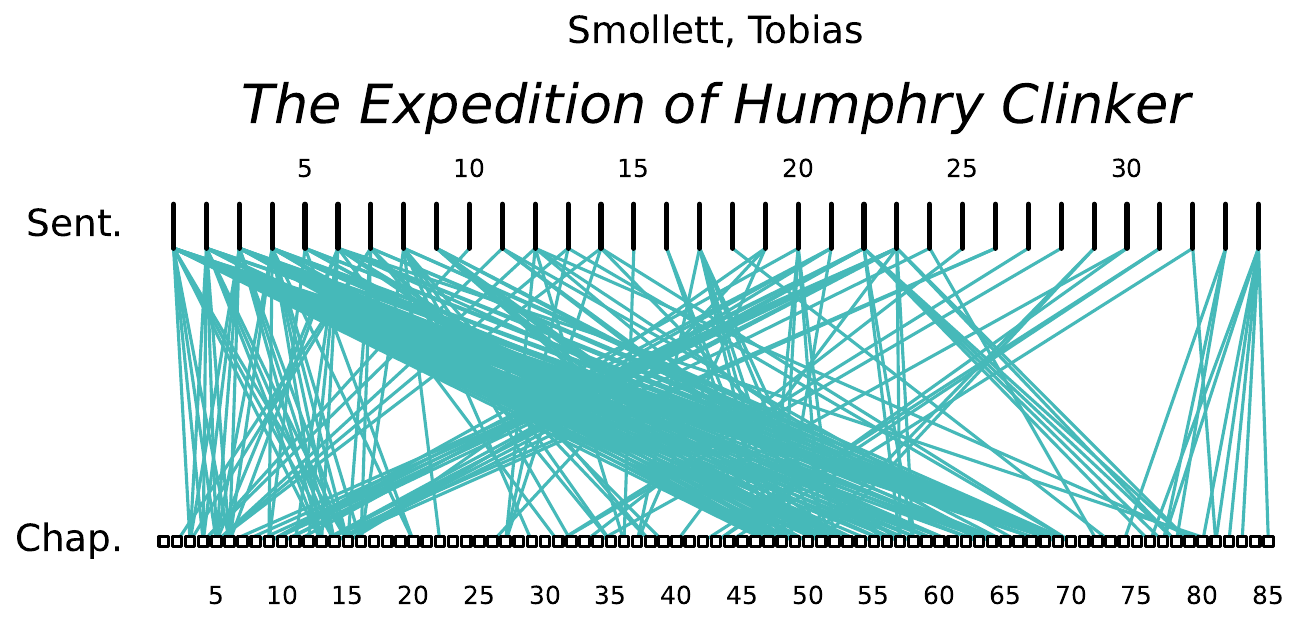}
    \phantomcaption
    \label{fig:2160-bipartite-main}
  \end{subfigure}
  \hfill
  \begin{subfigure}[t]{0.49\textwidth}
    \includegraphics[width=\linewidth]{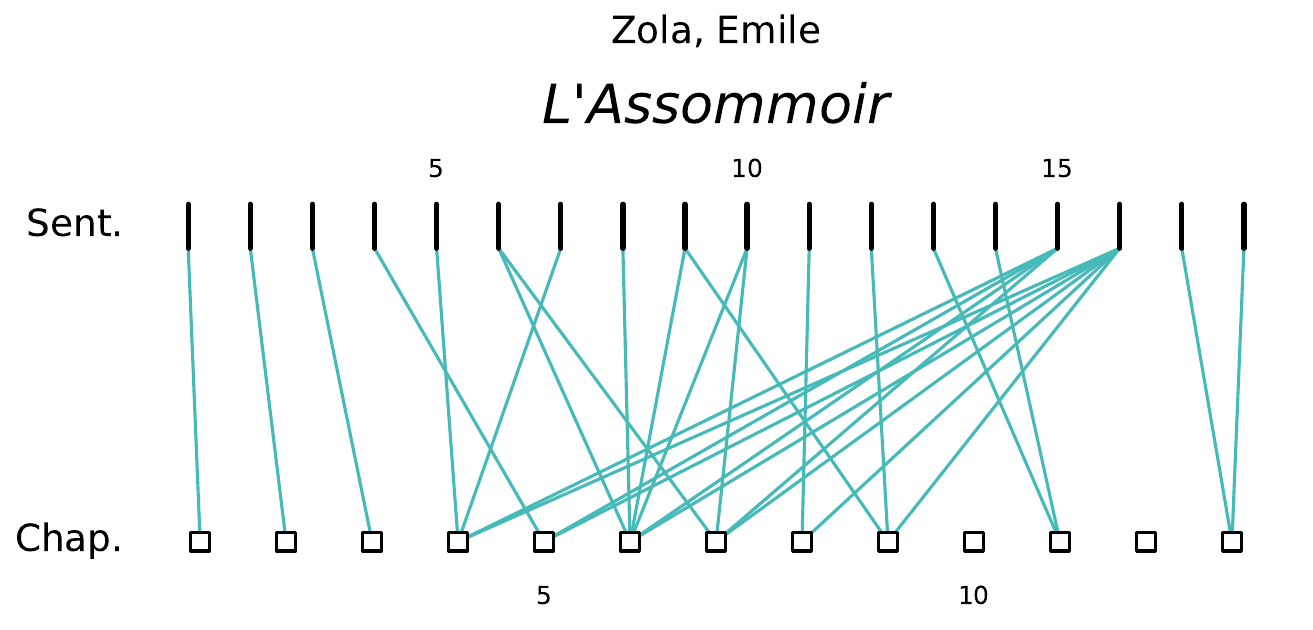}
    \phantomcaption
    \label{fig:8600-bipartite-main}
  \end{subfigure}
  \caption{Bipartite graphs representing the sentence-to-chapter mappings of two summaries with unusually low linearities. Note the highly intersecting lines, indicating nonlinearity.}
  \label{fig:bipartite-graphs-1}
\end{figure}

Although most summaries are primarily linear, the distribution is left-tailed (\cref{fig:hist-linearity}), with clear outliers. By far the least linear summary is of \emph{The Expedition of Humphry Clinker} by Tobias Smollett, with $l\approx0.10$ (\cref{fig:2160-bipartite-main}). Observation suggests two reasons for this summary's low linearity. First, it begins with several sentences that describe overarching plot points which cover the majority of the novel, and are therefore matched to a large number of chapters throughout the narrative. Second, the summary is, unusually, broken up by character, meaning that sequential sentences in the summary often reference events that take place in very different parts of the novel.

In contrast, the other outlier summaries often display more extreme versions of the same summary-writing choices identified in the more linear summaries.
For example, the summary of \emph{L'Assommoir} by \'{E}mile Zola ($l\approx0.46$, \cref{fig:8600-bipartite-main}) separates a secondary plotline from the main narrative (\cref{sec:8600-summary}, Sentences 15 and 16), causing a large number of out-of-order alignments. 
Similarly, the summary of \emph{The Marrow of Tradition} by Charles W.~Chesnutt ($l\approx0.51$; \cref{sec:bipartiteGraphs}, \cref{fig:pg11228-bipartite}) both describes overarching plot elements in single summary sentences and separates interwoven plotlines. 
The summary of \emph{Anne of Green Gables} by L.~M.~Montgomery ($l\approx0.58$; \cref{sec:bipartiteGraphs}, \cref{fig:pg45-bipartite}) contains many individual sentences that refer to recurring events or character traits,
and the summary of \emph{The Black Moth} by Georgette Heyer ($l\approx0.62$; \cref{sec:bipartiteGraphs}, \cref{fig:pg38703-bipartite}) begins with framing information only revealed much later in the narrative, separates out subplots, \emph{and} describes overarching plot elements in individual summary sentences.
The drop in linearity of another notable outlier, the summary of \emph{The Sound and the Fury} by William Faulkner ($l\approx0.52$; \cref{sec:bipartiteGraphs}, \cref{fig:pg75170-bipartite}), is likely due both to its famously complex narrative and to errors by the mapping pipeline, which struggled with Faulkner's experimental style.

Across the dataset, proportionally longer summaries tend to be more linear ---
linearity is moderately positively correlated with the log-ratio of summary length to book length ($r\approx0.32$, $p<10^{-3}$).\footnote{All reported correlations are Pearson's correlation coefficients with post-correction for multiple tests using Benjamini-Hochberg false discovery control.}
For example, the summary of \emph{Ulysses} by James Joyce is both unusually long and unusually linear ($l\approx0.93$; \cref{sec:bipartiteGraphs}, \cref{fig:pg4300-bipartite}). We note that books that receive considerable attention from Wikipedia editors may have longer summaries, which may contribute to this result.

Overall, we find that summaries usually follow the same ordering of events as the narratives they seek to describe. However, deviations from this ordering appear to be caused by a number of summarization techniques intended to make the narrative easier to parse for readers. These may therefore represent choices in summary writing.

\section{Summary Uniformity}

We next study whether the summaries adhere to the uniformity principle.
Since each summary sentence theoretically describes an important aspect of the novel's narrative, we assume that the proportion of summary sentences matched to an individual chapter reflects how much important or relevant information that chapter contains.
We proceed with this analysis using three metrics.

\begin{figure}[t]
  \centering
  \begin{subfigure}[t]{0.49\textwidth}
    \includegraphics[width=\linewidth]{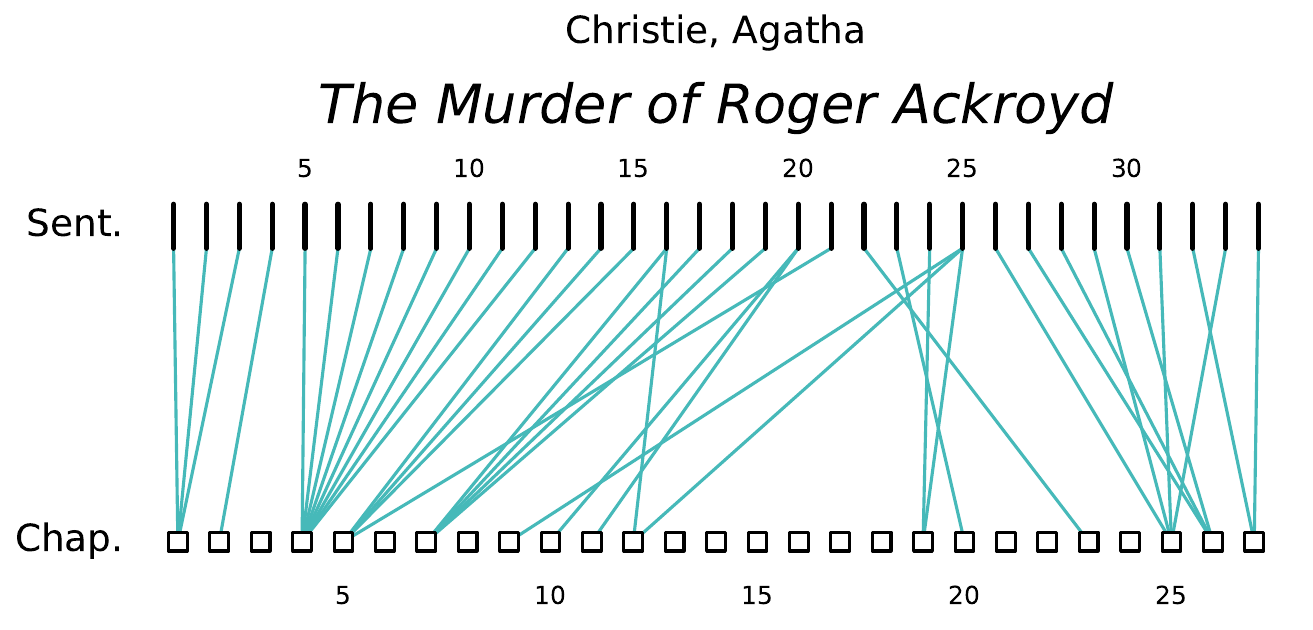}
    \phantomcaption
    \label{fig:69087-bipartite-main}
  \end{subfigure}
  \hfill
  \begin{subfigure}[t]{0.49\textwidth}
    \includegraphics[width=\linewidth]{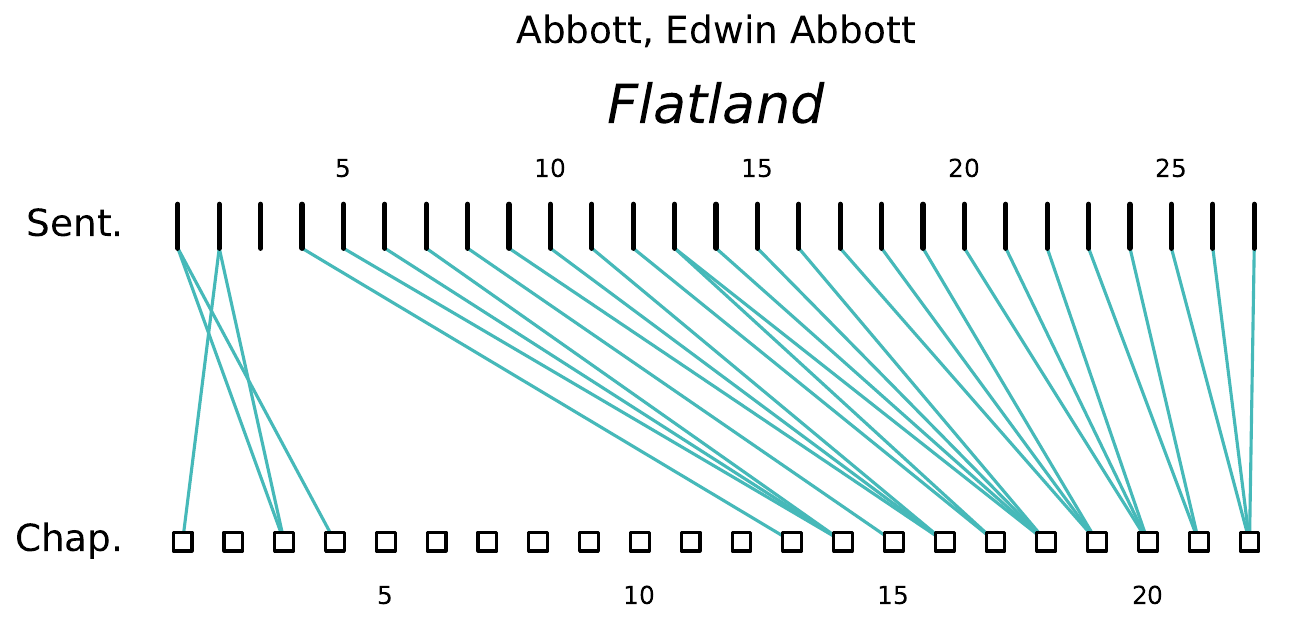}
    \phantomcaption
    \label{fig:45506-bipartite-main}
  \end{subfigure}
  \caption{Bipartite graphs representing the sentence-to-chapter mappings of two summaries with large log-ratios of summary-to-book length and a small proportion of matched chapters.}
  \label{fig:bipartite-graphs-2}
\end{figure}

\subsection{Proportion of Matched Chapters}

We first calculate the proportion of chapters in each novel matched to at least one summary sentence. We find that, on average, only 79\% of chapters are matched to $\geq$1 summary sentence and only 12\% (18 / 150) of summaries cover all novel chapters. The distribution of the proportion of chapters matched ($p_c$) is broad (\cref{fig:hist-prop-chaps}), ranging from 0.31 to 1.0.

Proportionally longer summaries are more likely to match a larger proportion of chapters;
there is a strong positive correlation between $p_c$ and the log-ratio of the summary length to book length ($r\approx0.55$, $p<10^{-11}$). It is intuitive that summaries with more ``space'' to discuss a book's contents are likely to cover a greater proportion of novel chapters. Summary length may also reflect the information density of the novel itself --- a summary may need to be comparatively longer to accurately describe a denser narrative. Both explanations likely contribute to this relationship. 

However, summary-to-book density does not entirely explain variation in $p_c$; linear regression shows that the log-ratio of summary-to-book length explains only $R^2\approx0.31$ of the variation in the proportion of novel chapters matched ($p<10^{-12}$). Some summaries have comparatively high summary-to-book length ratios but low $p_c$ values. One example is the summary of \textit{The Murder of Roger Ackroyd} by Agatha Christie, which maps sentences to only 56\% of the novel's chapters (\cref{fig:69087-bipartite-main}).
This work is, famously, characterized by a revelation at the end that calls into question the reliability of the narrator. 
Indeed, many of the un-matched chapters depict the detective revealing clues to the narrator but withholding their meaning; the summary either ignores smaller clues or presents them with their true meaning, revealed later in the novel.
Similarly, the summary of \emph{Flatland: A Romance of Many Dimensions} by Edwin Abbott Abbott has a high summary-to-book length log-ratio but only maps to 59\% of the novel's chapters (\cref{fig:45506-bipartite-main}). However, many of these skipped chapters involve the narrator describing aspects of life in his world to the readers. Although the summary says that the narrator ``guides the readers through some of the implications of life in two dimensions,'' (\cref{sec:45506-summary}) the alignment algorithm does not successfully match this sentence to the relevant chapters, perhaps because the summary description is so broad.
These examples suggest that some chapters are skipped not because they are irrelevant, but because the summary introduces information differently, either by streamlining (as in \emph{The Murder of Roger Ackroyd}) or by abstracting to the intention of the chapters (as for \emph{Flatland}).

\subsection{Uniformity of Sentence-to-Chapter Match Distribution}

In addition to skipping chapters, another violation of uniformity is to dwell on particular chapters.
To quantify this, we calculate the Gini coefficient \cite{gini1912variabilita} of the proportion of summary sentences matched to each chapter ($G_c$).
Values closer to 0.0 indicate greater uniformity, and values closer to 1.0 indicate unequal focus.
For this metric, we only consider chapters linked to at least one sentence; otherwise the results are mostly indistinguishable from the proportion of matched chapters.

\begin{figure}[t]
  \centering
  \begin{subfigure}[t]{0.49\textwidth}
    \includegraphics[width=\linewidth]{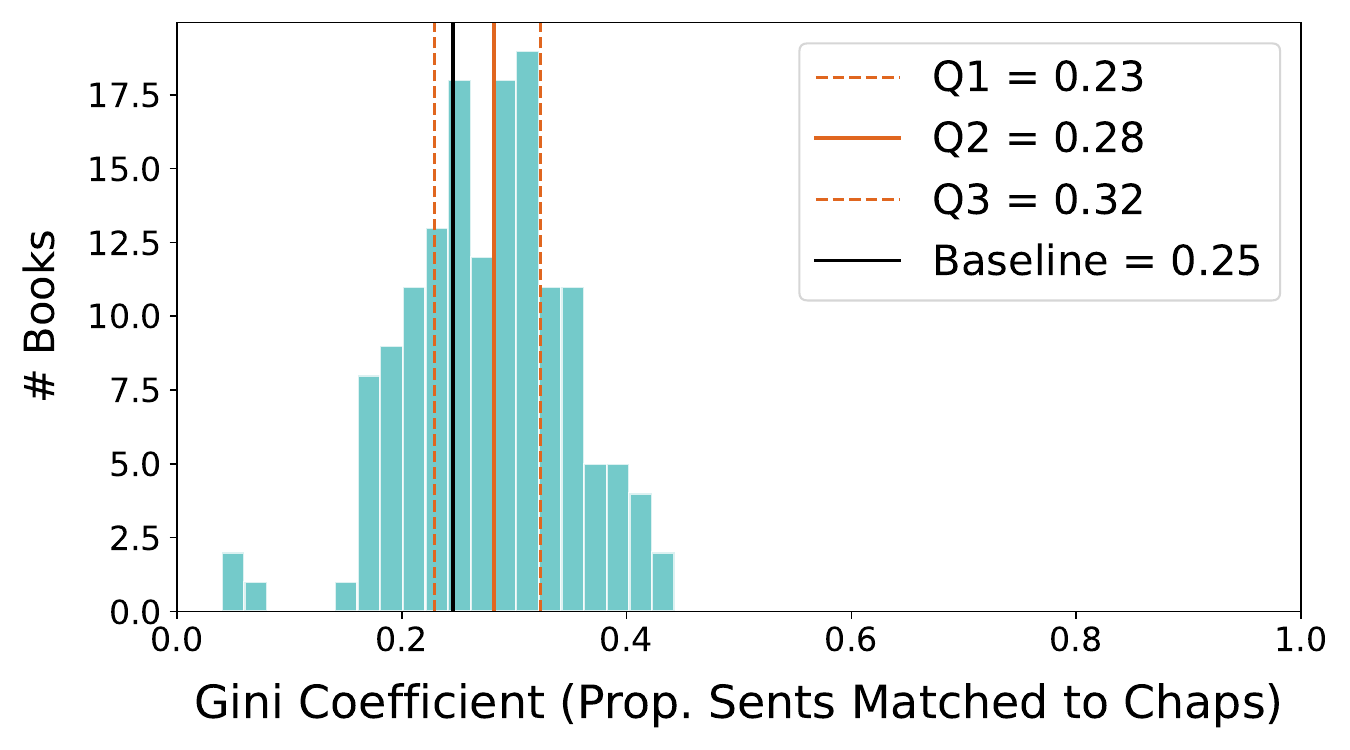}
    \phantomcaption
    \label{fig:hist-gini-chaps}
  \end{subfigure}
  \hfill
  \begin{subfigure}[t]{0.49\textwidth}
    \includegraphics[width=\linewidth]{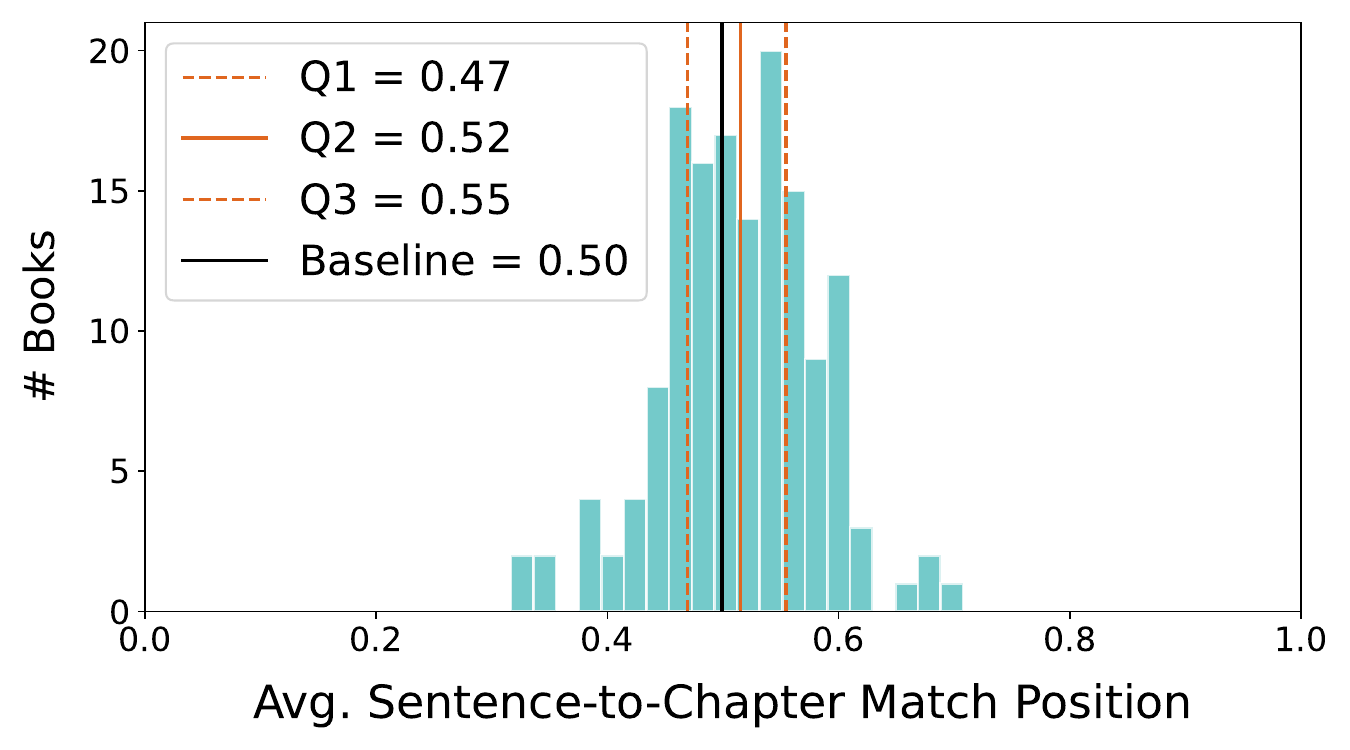}
    \phantomcaption
    \label{fig:hist-avg-dist-center}
  \end{subfigure}
\caption{Gini coefficient for sentence-to-chapter matches (a) and the average position of sentence-to-chapter matches (b) across all 150 novels. Baselines calculated from 200 random shuffles of the chapter matches.}
  \label{fig:metric-histograms-2}
\end{figure}

\begin{figure}[b]
  \centering
    \includegraphics[width=0.6\textwidth]{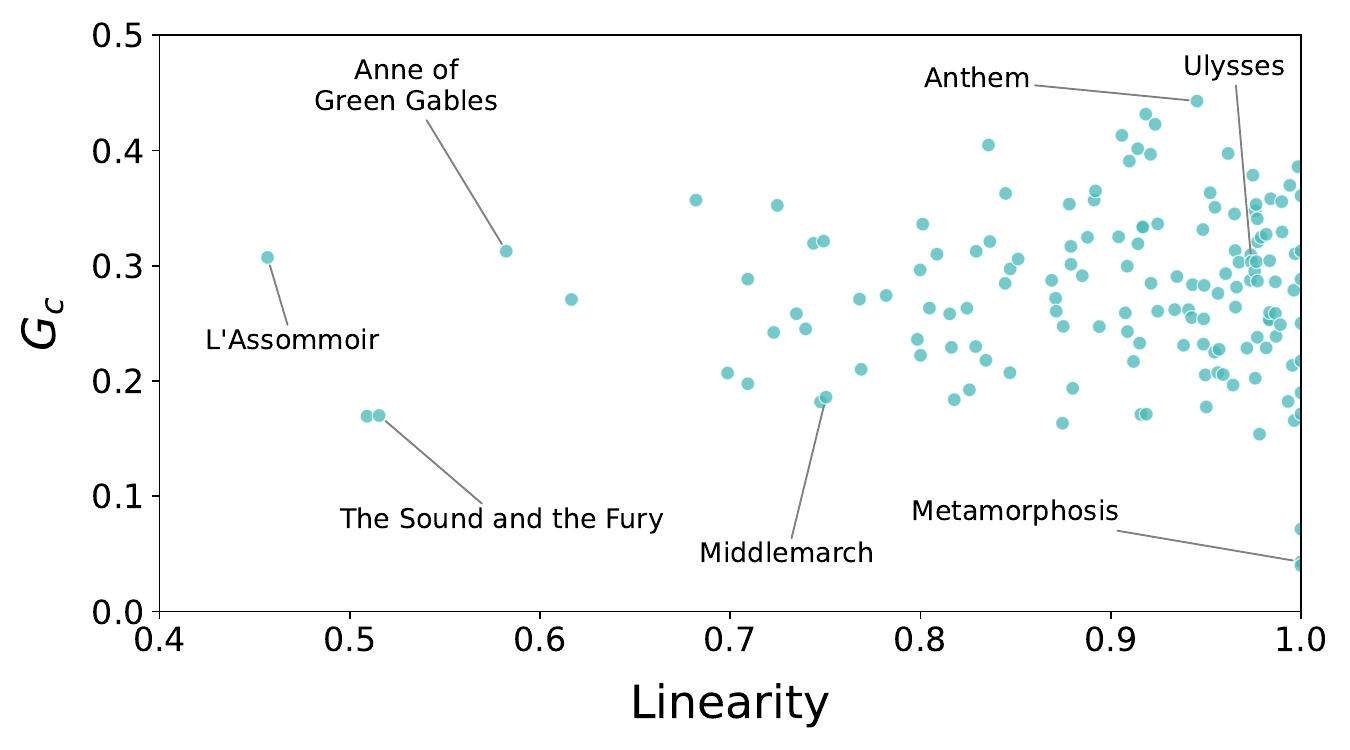}
  \caption{The relationship between linearity ($l$) and the Gini coefficient of sentence-to-chapter matches ($G_c$). One summary with $l<0.5$ is cut from the figure. Note the lack of correlation.}
  \label{fig:linearity-vs-gini-chaps}
\end{figure}

\begin{figure}[t]
  \centering
  \begin{subfigure}[t]{0.49\textwidth}
    \includegraphics[width=\linewidth]{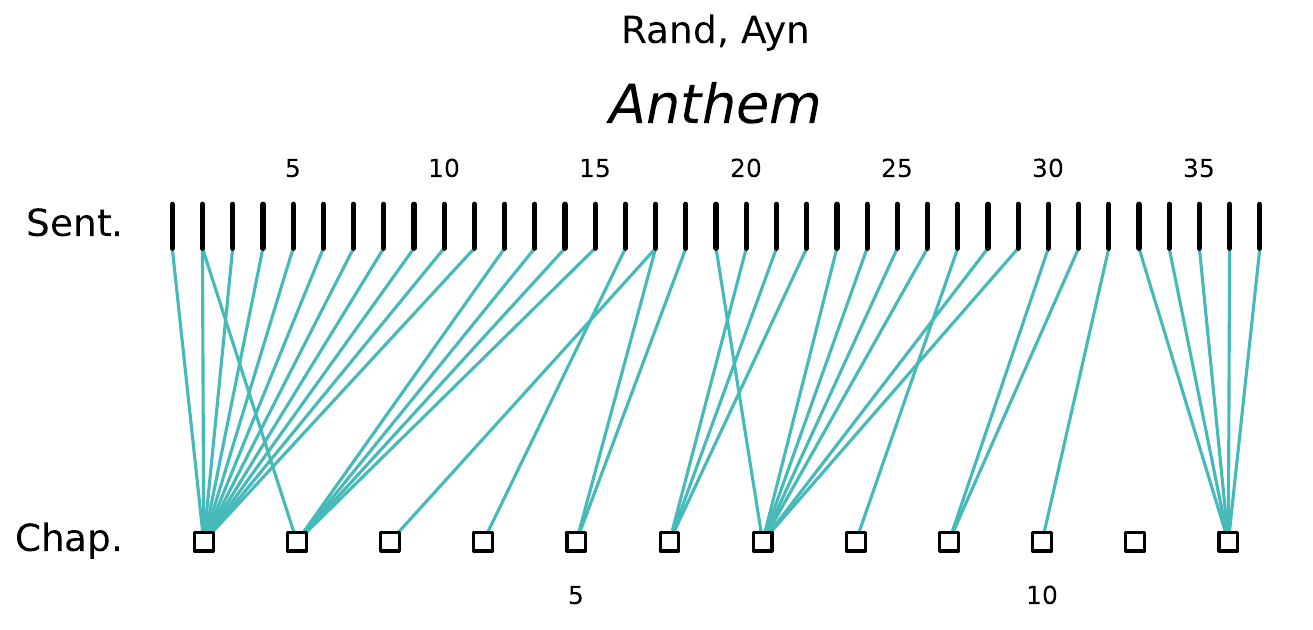}
    \phantomcaption
    \label{fig:1250-bipartite-main}
  \end{subfigure}
  \hfill
  \begin{subfigure}[t]{0.49\textwidth}
    \includegraphics[width=\linewidth]{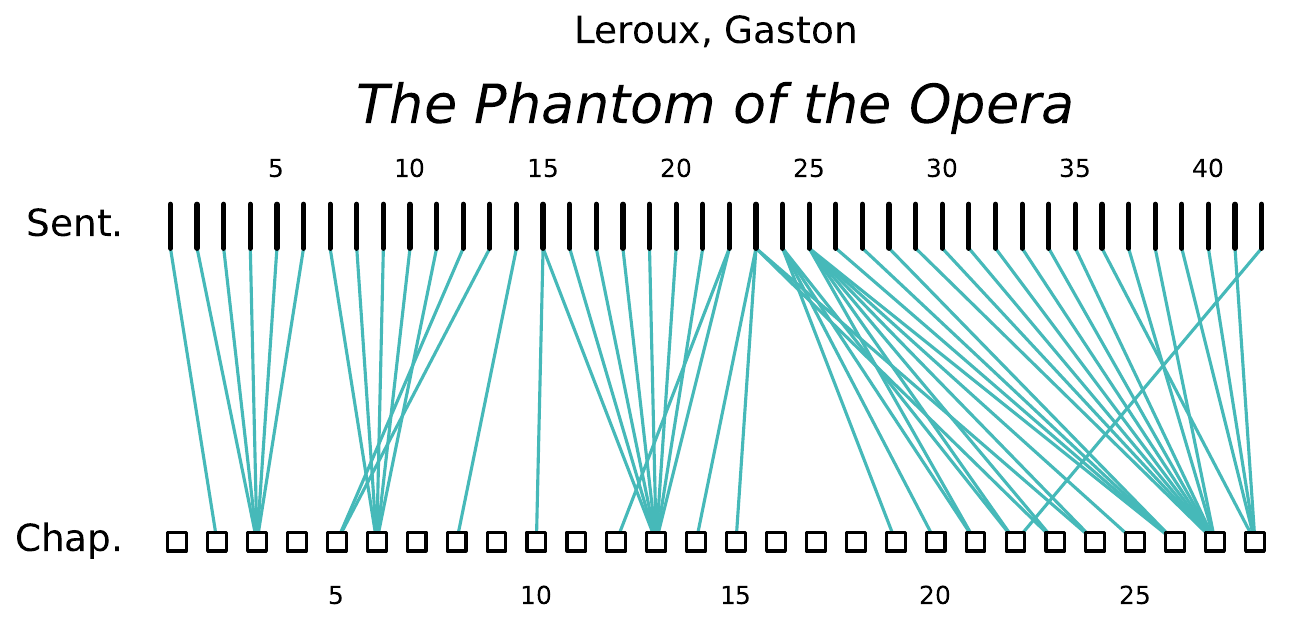}
    \phantomcaption
    \label{fig:175-bipartite-main}
  \end{subfigure}
  \caption{Bipartite graphs representing the sentence-to-chapter mappings of two summaries with large $G_c$ values. Note the relative lack of intersecting lines, indicating linearity and proportionality.}
  \label{fig:bipartite-graphs-3}
\end{figure}

The median $G_c$ is 0.28 with a relatively normal distribution (\cref{fig:hist-gini-chaps}); all novels have $G_c<0.5$. We find that summaries with $G_c$ values near the median have an uneven but largely continuous distribution of sentence-to-chapter matches, with only small differences between the most and least matched chapters (e.g.~\Cref{sec:bipartiteGraphs}, \cref{fig:pg25581-bipartite,fig:pg559-bipartite,fig:pg77600-bipartite}).

\begin{figure}[b]
  \centering
  \begin{subfigure}[t]{0.49\textwidth}
    \includegraphics[width=\linewidth]{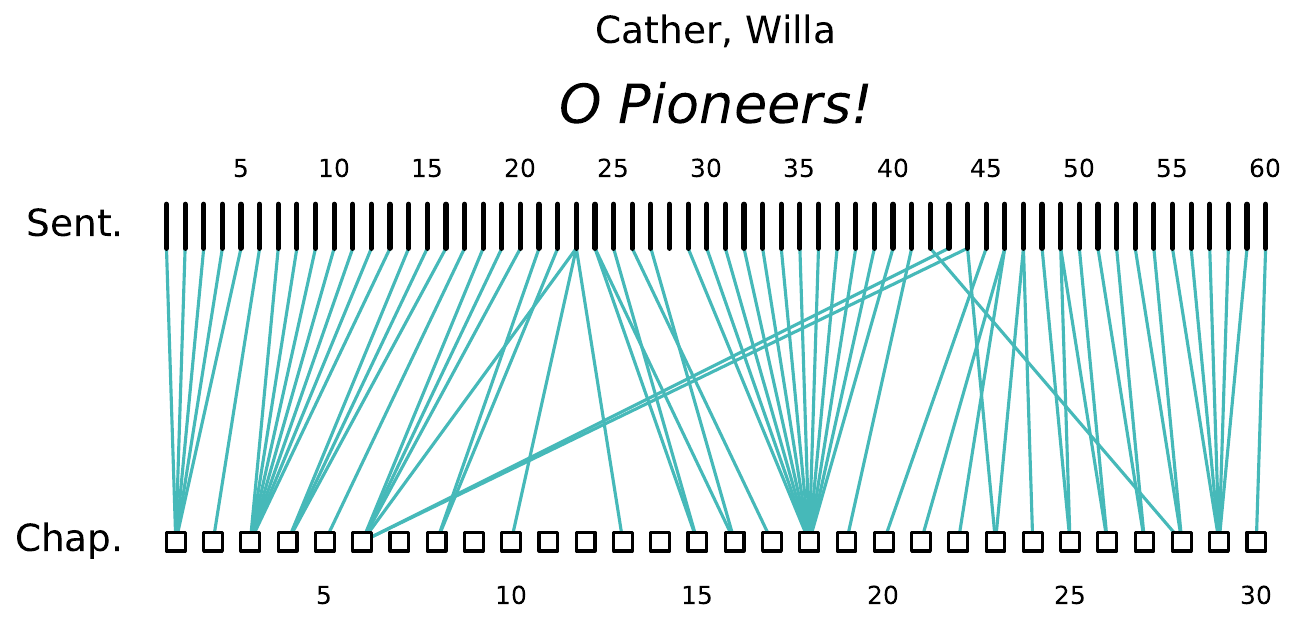}
    \phantomcaption
    \label{fig:24-bipartite-main}
  \end{subfigure}
  \caption{Bipartite graphs representing the sentence-to-chapter mappings of \textit{O Pioneers!} which features larger $G_c$ values. Note a few chapters receive a disproportionate number of sentences.}
  \label{fig:bipartite-graphs-4}
\end{figure}

On the most uniform end of the distribution, three summaries have $G_c < 0.1$.
These are unusual for different reasons.
The summary of \emph{Adam Bede} by George Eliot ($G_c\approx0.04$) only refers to 41\% of the novel's 56 chapters, but all but one of those chapters matches to exactly one of the 15 summary sentences (\cref{sec:507-summary}).
These sentences are in strictly linear order ($l=1.0$).
\emph{Metamorphosis} by Franz Kafka ($G_c\approx0.04$) and \emph{A Christmas Carol} by Charles Dickens ($G_c\approx0.07$) are both short texts (split into three and five sections respectively) each of which is paired with a similar number of summary sentences (\cref{sec:bipartiteGraphs}, \cref{fig:pg5200-bipartite} + \cref{fig:pg19337-bipartite}). Again, both of these summaries are completely linear ($l=1.0$), although linearity and $G_c$ are not significantly correlated more broadly (\cref{fig:linearity-vs-gini-chaps}).

In contrast, the three summaries with the largest $G_c$ values --- \emph{Anthem} by Ayn Rand ($G_c\approx0.44$), \emph{O Pioneers!} by Willa Cather ($G_c\approx0.43$), and \emph{The Phantom of the Opera} by Gaston Leroux ($G_c\approx0.42$) --- all map disproportionately many sentences to a subset of chapters. Some of these chapters contain important information or significant parts of the narrative. For example, the first 11 sentences of \emph{Anthem}'s 37-sentence summary are narrative framing, all of which is introduced in the novella's first chapter (\cref{fig:1250-bipartite-main}; \cref{sec:1250-summary}). 20 of the 42 summary sentences for \emph{The Phantom of the Opera} describe Chapter 13, Christine's first abduction by the Phantom, and roughly a further 20\% describe the second-to-last chapter (27), which contains much of the climax (\cref{fig:175-bipartite-main}; \cref{sec:175-summary}). However, disproportionate focus on a single chapter is often simply a quirk. In the summary of \textit{O Pioneers!}, 12 of 60 sentences are matched only with Chapter 18; however, 11 of these sentences are an unmarked direct quote of the chapter's beginning (\cref{fig:24-bipartite-main}; \cref{sec:24-summary}).

Some evidence suggests summaries refer to events that are uniformly distributed over \textit{tokens} rather than chapters: longer chapters are somewhat more likely to have summary sentences matched to them.
In 127 books there is a slight positive Pearson's R correlation between chapter length and the proportion of summary sentences matched to that chapter, with an average of $r\approx0.24$. After Benjamini-Hochberg correction \cite{benjamini1995controlling}, 14\% (21 / 150) of the correlations are significant (at $p<0.01$). For example, the first chapter of \emph{Anthem} has three times more tokens than the novel's mean (6,357 vs.~2,004). Similarly, Chapter 13 of \emph{The Phantom of the Opera} is over double the novel average (9,644 vs.~4,092). 
However, Chapter 27 of the same book is only slightly longer than average (4,228 tokens) yet is matched to more sentences, indicating many noteworthy events.

In contrast to linearity, we find $G_c$ is moderately positively correlated with summary length ($r\approx0.42$, $p<10^{-6}$) --- that is to say, longer summaries are somewhat more likely to dwell on certain chapters. Interestingly, $G_c$ is not significantly correlated with the log-ratio of summary-to-book length or novel length, only summary length.

\subsection{Position of Sentence-to-Chapter Matches Across Novels}

\begin{figure}[t]
  \centering
  \begin{subfigure}[t]{0.49\textwidth}
    \includegraphics[width=\linewidth]{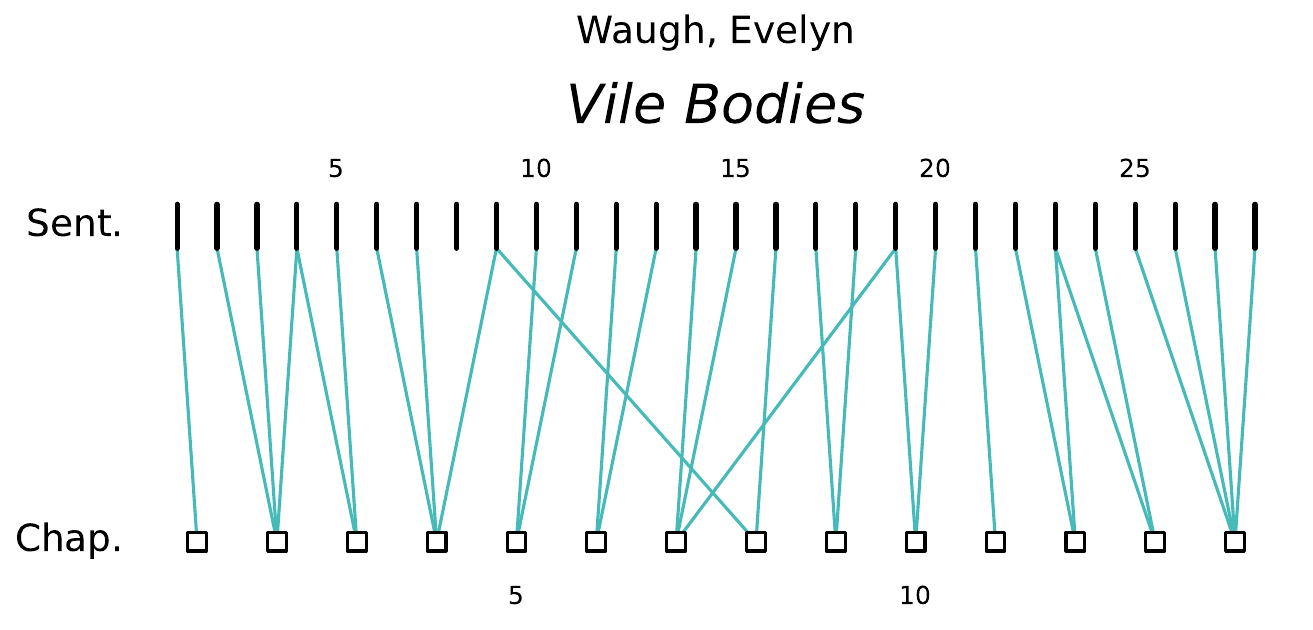}
    \phantomcaption
    \label{fig:77900-bipartite-main}
  \end{subfigure}
  \begin{subfigure}[t]{0.49\textwidth}
    \includegraphics[width=\linewidth]{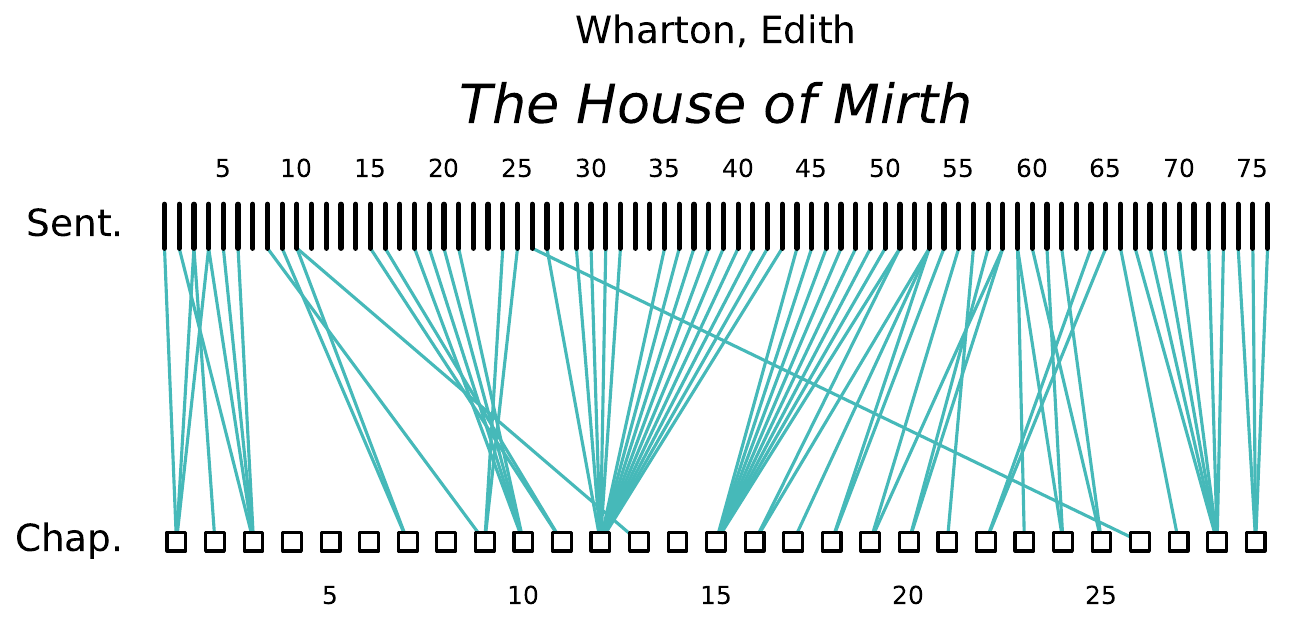}
    \phantomcaption
    \label{fig:284-bipartite-main}
  \end{subfigure}
  \caption{Bipartite graphs representing the sentence-to-chapter mappings of two summaries with near median $\mu_m$ values. Note the Gini coefficient appears to be correlated with summary length.}
  \label{fig:bipartite-graphs-5}
\end{figure}

Finally, having confirmed that summary sentence-to-chapter matches are not equally distributed throughout novels, we look at where in the narrative the matches are concentrated. For each summary, we calculate the position in the novel of the average sentence-to-chapter match. Specifically, if there are $n_c$ chapters and the set $M$ of sentence-to-chapter matches, where each match $m$ consists of the sentence index ($s_m$) and the chapter index ($c_m$), we calculate:
\begin{equation*}
    \mu_m = \frac{\Sigma_{m\in M}\frac{c_m-1}{n_c-1}}{|M|}.
\end{equation*}
A summary that only references the first chapter would have a value close to 0.0, while a summary that only references the final chapter would be close to 1.0.

\begin{figure}[b]
  \centering
  \begin{subfigure}[t]{0.49\textwidth}
    \includegraphics[width=\linewidth]{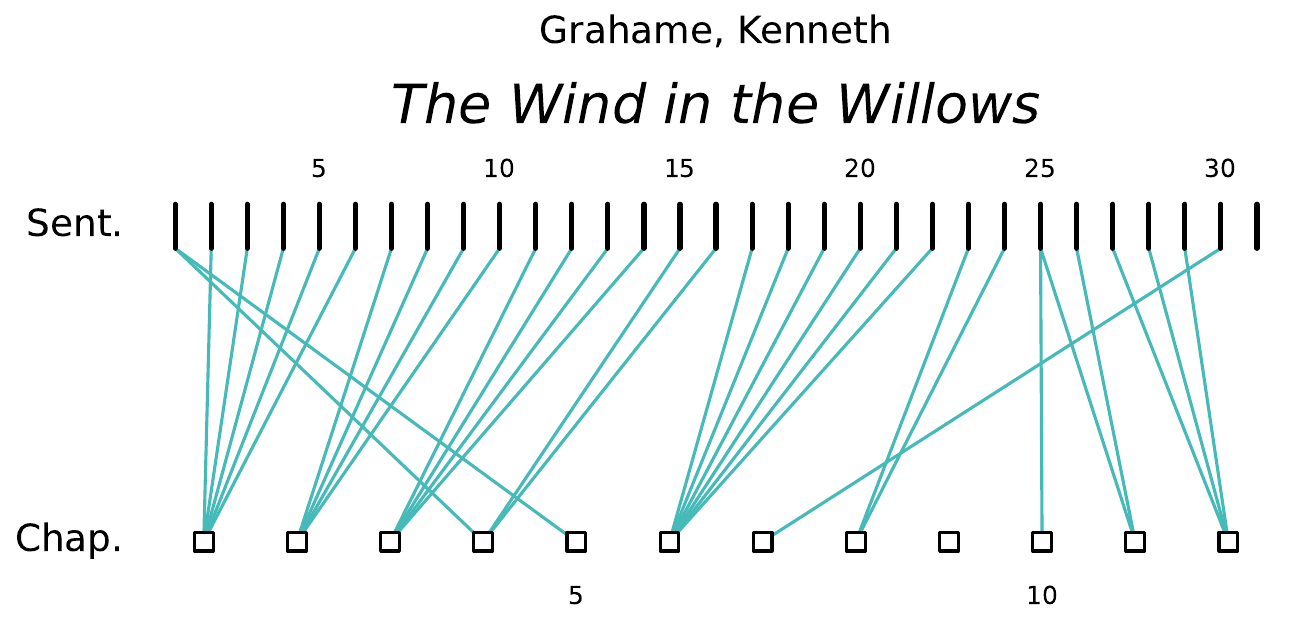}
    \phantomcaption
    \label{fig:289-bipartite-main}
  \end{subfigure}
  \caption{Bipartite graph representing the sentence-to-chapter mapping of one summary with low $\mu_m$ value. Note the summary appears to place more focus on the first half of the novel.}
  \label{fig:bipartite-graphs-6}
\end{figure}

In this dataset, $\mu_m$ ranges from 0.32 to 0.71 with a median of 0.52, suggesting that on average sentence-to-chapter matches are evenly spread between novel halves, with slightly more focus placed on novels' second halves (\cref{fig:hist-avg-dist-center}). Novels with $\mu_m$ values near the median still differ in their distribution of summary attention. For example, the summary of \emph{Vile Bodies} by Evelyn Waugh ($\mu_m\approx0.52$) distributes summary attention mostly evenly across each chapter ($G_c\approx0.18$; \cref{fig:77900-bipartite-main}). In contrast, the summary of \emph{The House of Mirth} by Edith Wharton ($\mu_m\approx0.51$) distributes attention very unevenly per-chapter ($G_c\approx0.41$; \cref{fig:284-bipartite-main}), but evenly across the novel halves. 

\begin{figure}[t]
  \centering
  \begin{subfigure}[t]{0.49\textwidth}
    \includegraphics[width=\linewidth]{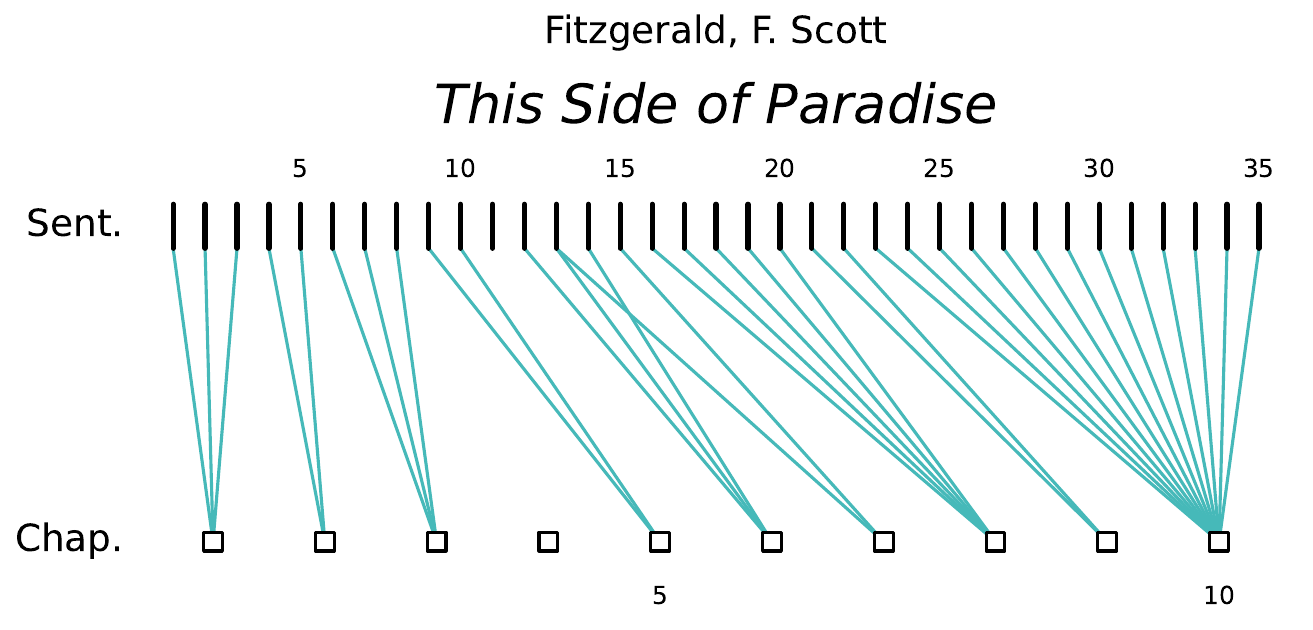}
    \phantomcaption
    \label{fig:805-bipartite-main}
  \end{subfigure}
  \begin{subfigure}[t]{0.49\textwidth}
    \includegraphics[width=\linewidth]{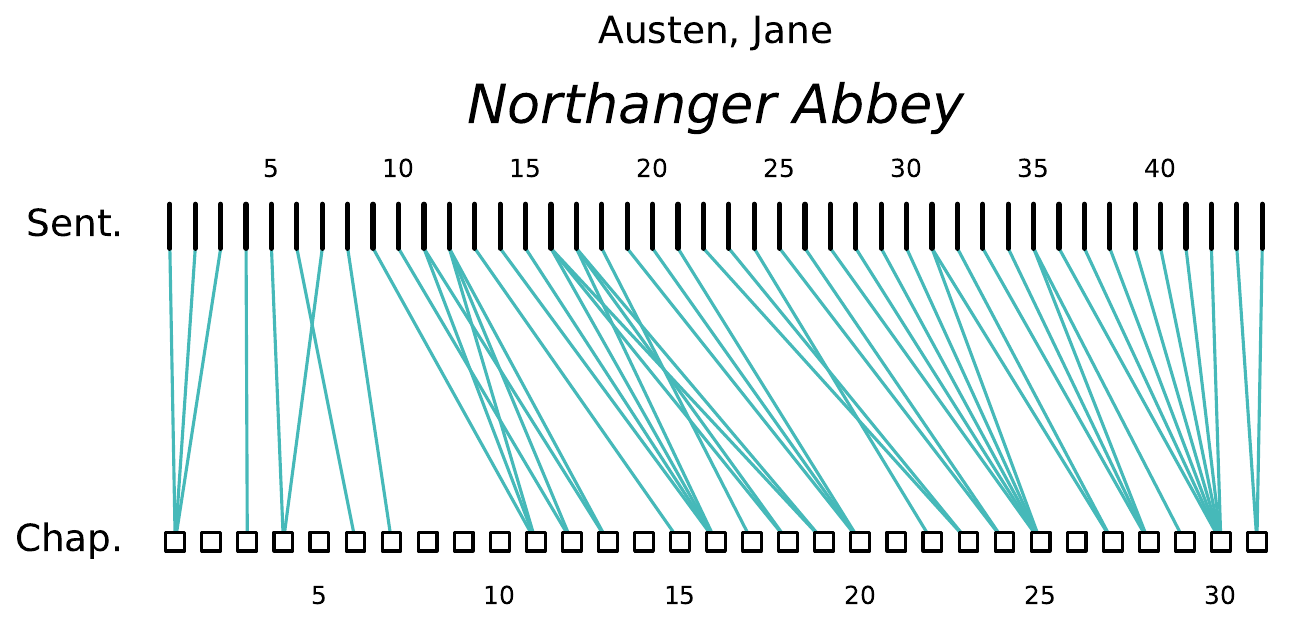}
    \phantomcaption
    \label{fig:121-bipartite-main}
  \end{subfigure}
  \caption{Bipartite graphs representing the sentence-to-chapter mappings of two summaries with high $\mu_m$ values. Note the relative uniform assortment of sentence to chapter pairs.}
  \label{fig:bipartite-graphs-7}
\end{figure}

Summaries with low $\mu_m$ values place more focus on the first halves of novels. We find that some of these summaries skip the novels' endings. For example, the summaries of \emph{The Triumph of the Scarlet Pimpernel} by Baroness Orczy ($\mu_m\approx0.32$), \emph{Beauvallet} by Georgette Heyer ($\mu_m\approx0.35$), and \emph{The Golovlyov Family} by Mikhail Saltykov-Shchedrin ($\mu_m\approx0.35$) do not cover the last 12, 3, and 6 chapters of the novels respectively (\cref{sec:bipartiteGraphs}, \cref{fig:pg65695-bipartite,fig:pg75547-bipartite,fig:pg44237-bipartite}). All of these summaries end with open-ended sentences (\cref{sec:65695-summary,sec:75547-summary,sec:44237-summary}) --- the summary of \emph{The Triumph of the Scarlet Pimpernel} ends with ``Lady Blakeney is kidnapped yet again and taken to France and imprisoned as bait for Sir Percy.'' --- highlighting an unexpected summarization trend. Another group of summaries with low $\mu_m$ values include descriptions of novels' endings, but skip some other series of chapters in the novels' second halves. These include the summaries of \emph{Anne of Green Gables} by L.~M.~Montgomery ($\mu_m\approx0.38$), \emph{David Copperfield} by Charles Dickens ($\mu_m\approx0.38$), \emph{Main Street} by Sinclair Lewis ($\mu_m\approx0.40$), and \emph{Captain Blood} by Rafael Sabatini ($\mu_m\approx0.33$) (\cref{sec:bipartiteGraphs}, \cref{fig:pg45-bipartite,fig:pg766-bipartite,fig:pg543-bipartite,fig:pg1965-bipartite}). In these summaries, there is often a summary sentence matched to a chapter(s) preceding the unmatched stretch that describes a series of events or a period of time (\cref{sec:45-summary,sec:766-summary,sec:543-summary,sec:1965-summary}); for example, the ninth sentence of the summary of \emph{Main  Street}, which is matched to chapters 8--10, states that ``despite her efforts, these ventures are ineffective and she is constantly derided by the leading cliques.'' After this, six novel chapters go unmatched. These unmatched novel sections may represent a stretch of narrative that subtly repeats events described in these broad summary sentences or may simply capture less ``eventful'' or ``interesting'' swathes of the novels. A final category of front-weighted summaries include extensive description of the novels' first chapters either because they include detailed novel framing (e.g.~\emph{Anthem} by Ayn Rand --- \cref{fig:1250-bipartite-main}, \cref{sec:1250-summary}) or are particularly eventful (e.g.~\emph{The Wind in the Willows} by Kenneth  Grahame --- \cref{fig:289-bipartite-main}, \cref{sec:289-summary}). Low $\mu_m$ values may also be impacted by the existence of other eventful chapters in the novels' first halves (again, e.g.~\emph{The Wind in the Willows}) and the novel properties described here may coexist.

Summaries often have high $\mu_m$ values for the inverse of the reasons described above. Some summaries ignore chapters at or near novels' beginnings. For example, in \emph{Flatland} by Edwin Abbott Abbott ($\mu_m\approx0.71$), chapters 5--12 are not matched with any part of the summary (\cref{fig:45506-bipartite-main}). The second summary sentence (\cref{sec:45506-summary}) --- ``The narrator is a square, a member of the caste of gentlemen and professionals, who guides the readers through some of the implications of life in two dimensions.'' --- broadly describes these chapters' content, but in too general terms for the sentence mapping pipeline to identify. Similarly, the first four chapters of \emph{Nostromo: A Tale of the Seaboard} by Joseph Conrad ($\mu_m\approx0.67$) are not matched with any summary sentence due to mistakes by the mapping pipeline (\cref{sec:bipartiteGraphs}, \cref{fig:pg2021-bipartite}); although several summary sentences  describe events from these chapters (\cref{sec:2021-summary}) they are presented with different context and in a different order than in the novel, likely causing the errors. Another group of summaries, including those of \emph{This Side of Paradise} by F.~Scott Fitzgerald ($\mu_m\approx0.67$), \emph{Heart of Darkness} by Joseph Conrad ($\mu_m\approx0.62$), and \emph{Fanshawe} by Nathaniel Hawthorne ($\mu_m\approx0.62$), pays considerable attention to novels' last chapters (\cref{fig:805-bipartite-main}; \cref{sec:bipartiteGraphs}, \cref{fig:pg219-bipartite,fig:pg7085-bipartite}), which appear to contain a large portion of the summarized events (\cref{sec:805-summary,sec:219-summary,sec:7085-summary}). Finally, some summaries appear to broadly describe the novels' second halves in more detail. These include the summaries of \emph{A Tale of Two Cities} by Charles Dickens ($\mu_m\approx0.61$), \emph{Northanger Abbey} by Jane Austen ($\mu_m\approx0.61$), \emph{The Good Soldier} by Ford Madox Ford ($\mu_m\approx0.61$), and \emph{A High Wind in Jamaica} by Richard Hughes ($\mu_m\approx0.62$) (\cref{fig:121-bipartite-main}; \cref{sec:bipartiteGraphs}, \cref{fig:pg98-bipartite,fig:pg2775-bipartite,fig:pg75530-bipartite}). Again, many summaries share multiple of the described characteristics, increasing their $\mu_m$ value.

\begin{figure}[t]
  \centering
  \begin{subfigure}[t]{0.49\textwidth}
    \includegraphics[width=\linewidth]{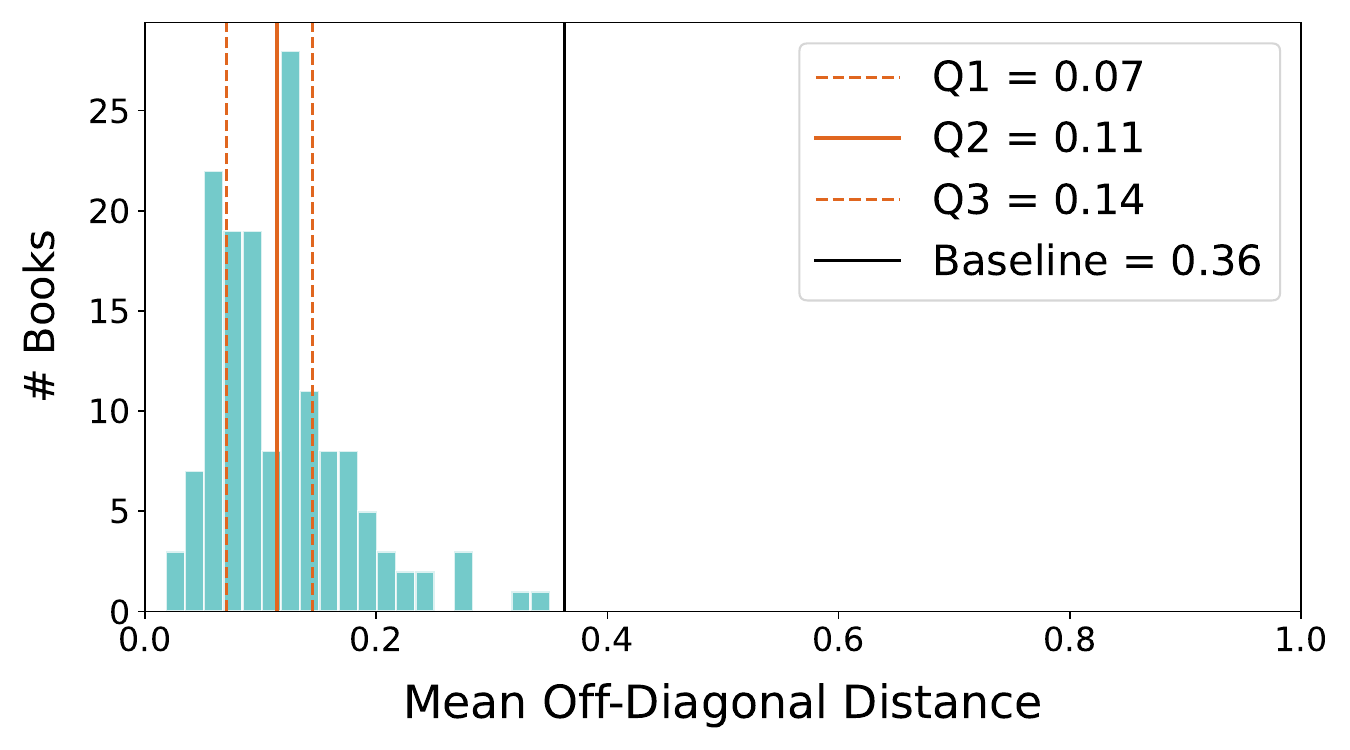}
  \end{subfigure}
    \caption{Mean off-diagonal distance for all 150 novels. Note measured off-diagonal distances nearly all fall below the random baseline of 0.36, indicating plot editing. Baseline calculated from 200 random shuffles of the sentence matches.}
  \label{fig:hist-off-diagonal}
\end{figure}

These three metrics --- the proportion of matched chapters, the Gini coefficient of sentence-to-chapter matches, and the average sentence-to-chapter match position --- reveal that attention is not distributed equally in most summaries. Some variance in these metrics can be attributed to errors in the sentence-to-chapter mappings, as the specific examples cited above demonstrate, but many reveal true summarization choices. Some novel chapters are clearly more ``eventful'' according to the summaries, even when they are not longer texts. Often, the distribution of summary attention seems to reflect choices made by the summarizers or aspects of novel construction. Overall, we find that summaries do make choices about what text represents ``interesting'' narrative information, as the Wikipedia instructions encourage; novel text is not of equal importance.

\section{Combining Linearity and Uniformity}

We next describe each summary's distance from our naive uniform and linear summary in a single metric. While this flattens some of the nuance the previous metrics allowed, it facilitates a more direct comparison between summaries. Specifically, we look at the mean off-diagonal distance ($\mu_{ODD}$) of each summary by averaging the distance of each sentence-to-chapter match from the perfectly linear, even matching of summary sentences to novel chapters --- represented as the diagonal of the sentence-to-chapter matrix. Mean off-diagonal distance is negatively correlated with linearity ($r\approx-0.52$, $p<10^{-9}$) and the proportion of chapters matched ($r\approx-0.43$, $p<10^{-6}$), confirming that summaries with higher $\mu_{ODD}$ are generally both less linear and uniform.

The median off-diagonal distance for all human-written summaries is relatively low ($\mu_{ODD}\approx0.11$), but ranges from 0.02 to 0.35 (\cref{fig:hist-off-diagonal}). Summaries with low $\mu_{ODD}$ values resemble the naive baseline, and are mostly linear with relatively even distribution of summary attention across the novel. Examples of this type of summary include those of \emph{Mike} by P.~G.~Wodehouse ($\mu_{ODD}\approx0.02$), \emph{The Cave Girl} by Edgar Rice Burroughs ($\mu_{ODD}\approx0.03$), \emph{Sister Carrie: A Novel} by Theodore Dreiser ($\mu_{ODD}\approx0.03$), \emph{Around the World in Eighty Days} by Jules Verne ($\mu_{ODD}\approx0.04$), and \emph{Greenmantle} by John Buchan ($\mu_{ODD}\approx0.04$) (\cref{sec:bipartiteGraphs}, \cref{fig:pg7423-bipartite,fig:pg69191-bipartite,fig:pg233-bipartite,fig:pg103-bipartite,fig:pg559-bipartite}). The summaries with comparatively large $\mu_{ODD}$ values, in contrast, are either non-linear, distribute attention unevenly across the novel they summarize, or combine both. Many of these summaries were highlighted as outliers or extremes in the previously discussed metrics: the summary of \emph{The Expedition of Humphry Clinker} ($\mu_{ODD}\approx0.35$) is extremely non-linear (\cref{fig:2160-bipartite-main}), whereas the summaries of \emph{Beauvallet} ($\mu_{ODD}\approx0.32$), \emph{The Triumph of the Scarlet Pimpernel} ($\mu_{ODD}\approx0.28$), \emph{Captain Blood} ($\mu_{ODD}\approx0.27$), and \emph{Main Street} ($\mu_{ODD}\approx0.27$) are all predominantly linear but heavily non-uniform (\cref{sec:bipartiteGraphs}, \cref{fig:pg75547-bipartite,fig:pg65695-bipartite,fig:pg1965-bipartite,fig:pg543-bipartite}).

We find that $\mu_{ODD}$ is also moderately negatively correlated with the log-ratio of summary-to-book length ($r\approx-0.36$, $p<10^{-4}$), suggesting that the less ``space'' the summary has to describe the book, the less linear and even the summaries will be. It seems that the more a summary condenses the novel, the more changes they make to how the narrative is delivered.

\section{Differences in Summary Sentences}

\begin{figure}[t]
  \centering
  \begin{subfigure}[t]{0.49\textwidth}
    \includegraphics[width=\linewidth]{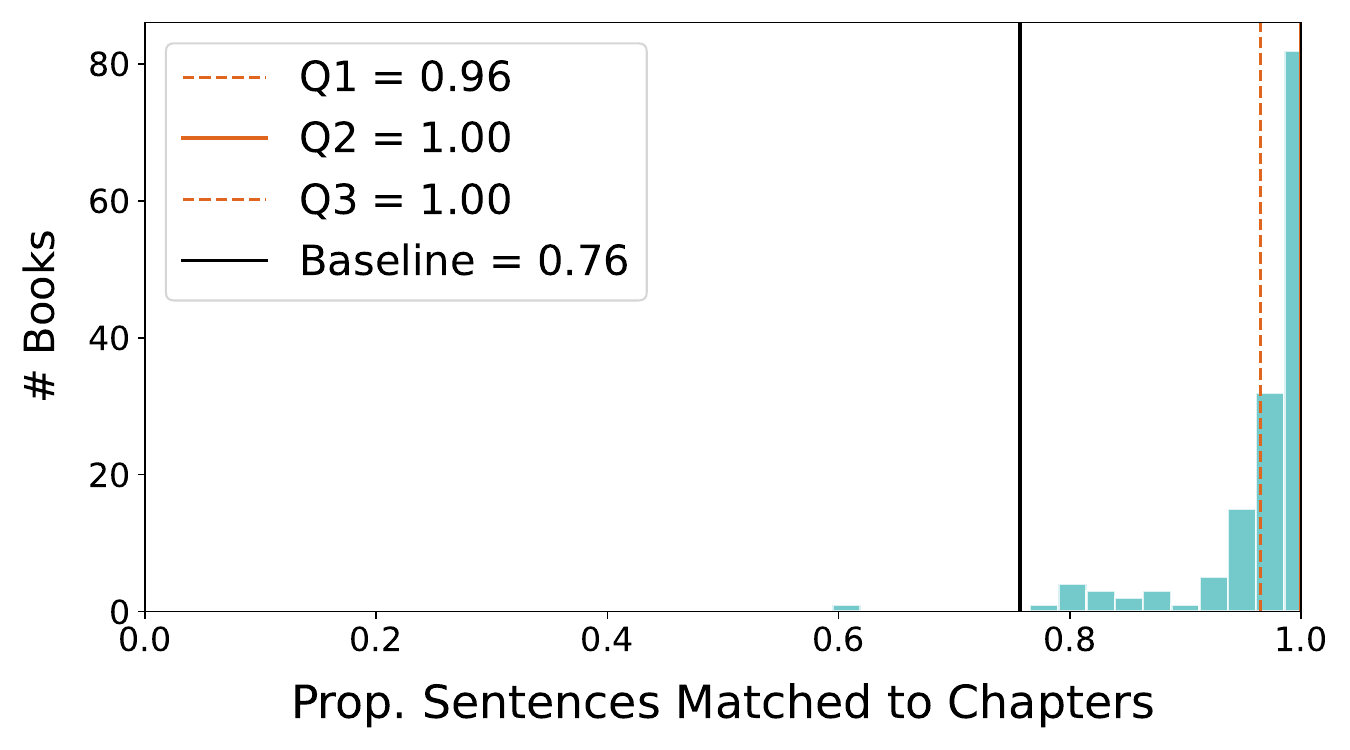}
    \phantomcaption
    \label{fig:hist-prop-sents}
  \end{subfigure}
  \hfill
  \begin{subfigure}[t]{0.49\textwidth}
    \includegraphics[width=\linewidth]{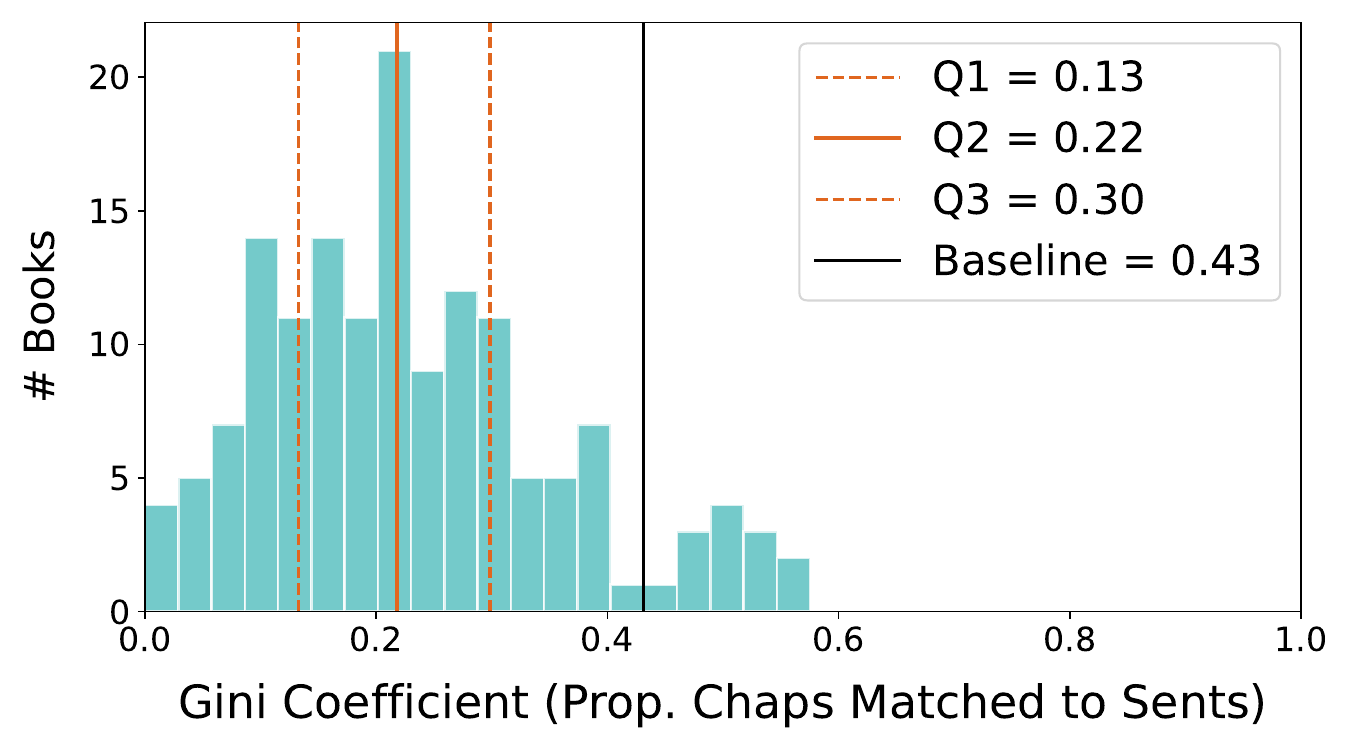}
    \phantomcaption
    \label{fig:hist-gini-sents}
  \end{subfigure}
\caption{Proportions of matched summary sentences (a) and the Gini coefficient of chapter-to-sentence matches (b) for all 150 human-authored summaries. Note nearly all sentences were matched with chapters, indicating the pipeline found success in alignment. Baselines calculated from 200 random shuffles of the chapter and sentence matches respectively.}
  \label{fig:metric-histograms-3}
\end{figure}

Having examined how summaries project onto novels, we now study the reverse. Here, we are interested in whether summary sentences are comparable units of information. We look at two metrics: the proportion of summary sentences matched to at least one novel chapter ($p_s$) and the Gini coefficient of the proportion of novel chapters matched to each summary sentence ($G_s$).

\begin{figure}[b]
  \centering
  \begin{subfigure}[t]{0.49\textwidth}
    \includegraphics[width=\linewidth]{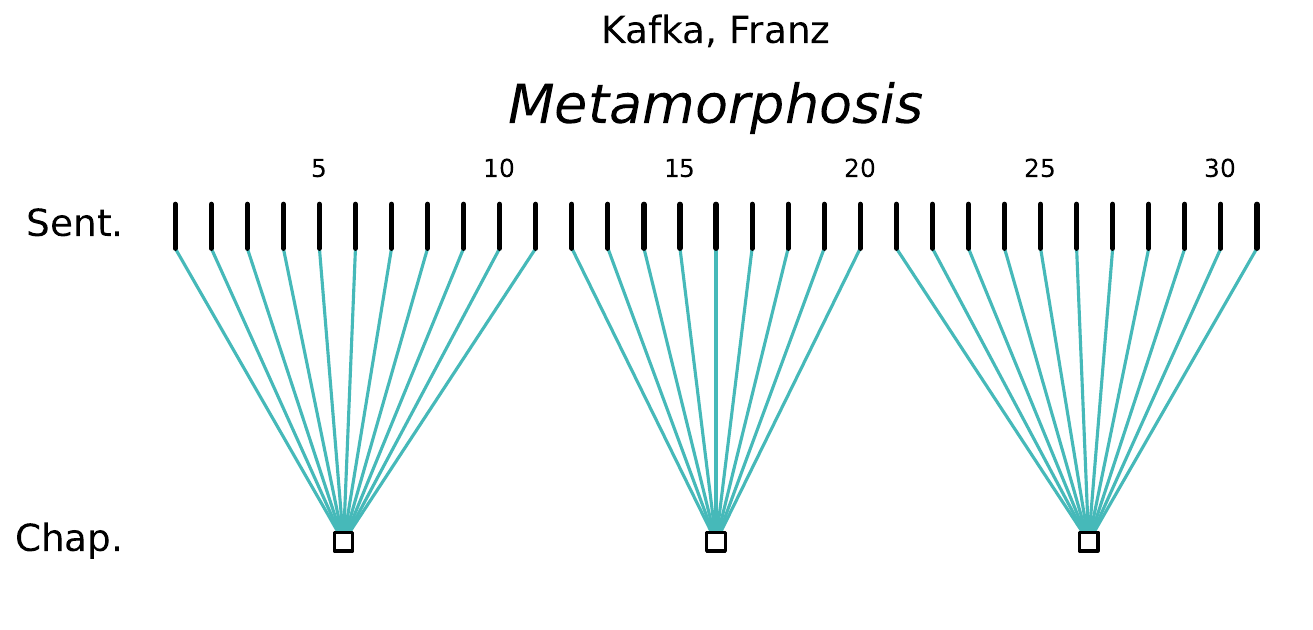}
    \phantomcaption
    \label{fig:5200-bipartite-main}
  \end{subfigure}
  \begin{subfigure}[t]{0.49\textwidth}
    \includegraphics[width=\linewidth]{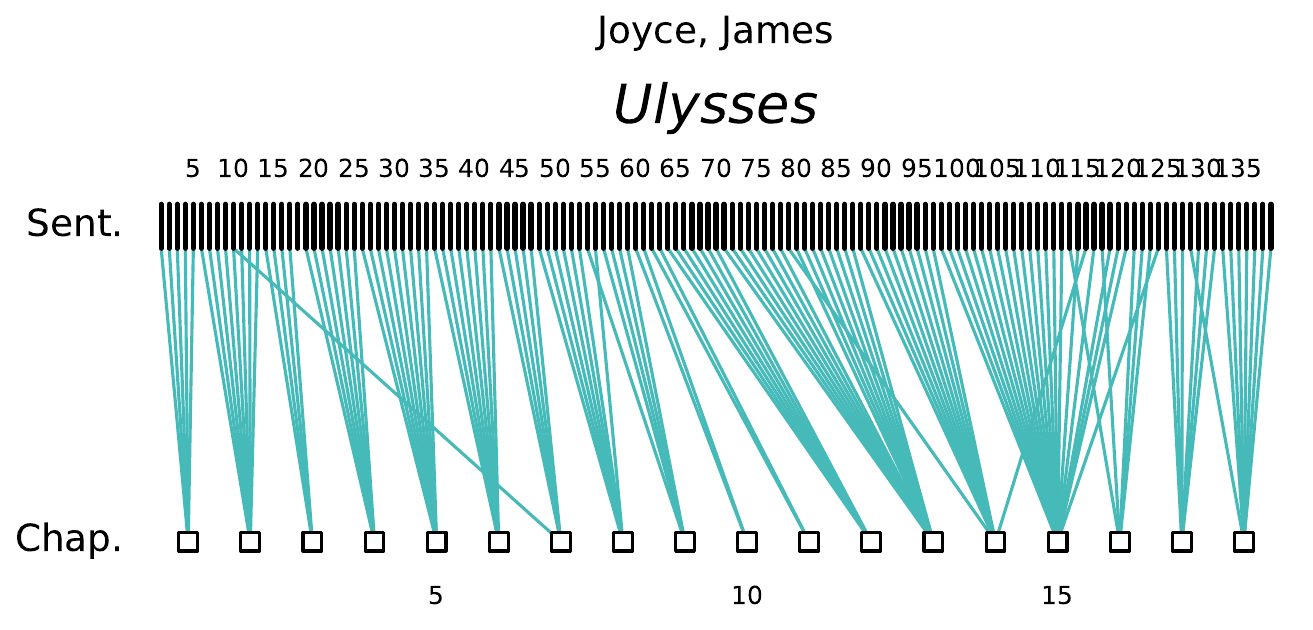}
    \phantomcaption
    \label{fig:4300-bipartite-main}
  \end{subfigure}
  \caption{Bipartite graphs representing the sentence-to-chapter mappings of two summaries with low $G_s$ values.}
  \label{fig:bipartite-graphs-8}
\end{figure}

We find all sentences are matched to at least one novel chapter ($p_s=1.0$) in 52\% (78 / 150) of summaries and at least 90\% of sentences are matched in 91\% (136 / 150) of summaries. However, the distribution of $p_s$ is again left-tailed (\cref{fig:hist-prop-sents}). Unmatched summary sentences are typically mapping errors, e.g. most of the unmatched sentences in the summary \emph{The Sound and the Fury} by William Faulkner ($p_s\approx0.60$; \cref{sec:bipartiteGraphs}, \cref{fig:pg75170-bipartite}) describe matchable events (\cref{sec:75170-summary}) but the stylistic complexity of the novel appears to confound the mapping pipeline. Similarly, the unmatched sentences in the summaries of \emph{Pollyanna} by Eleanor H.~Porter ($p_s\approx0.79$), \emph{The Seven Dials Mystery} by Agatha Christie ($p_s\approx0.80$), \emph{Adam Bede} by George Eliot ($p_s\approx0.80$), and Wolfbane by Frederik Pohl and C.~M.~Kornbluth ($p_s\approx0.81$) all appear to be alignment errors (\cref{sec:1450-summary,sec:75288-summary,sec:507-summary,sec:51845-summary}). However, a smaller subset of the unmatched sentences deliver meta-commentary about the novel and its history or framing information never directly stated in the novel. For example, the three unmatched sentences in the summary of \emph{Germinal} by \'{E}mile Zola (\cref{sec:56528-summary}) all fall into this category, as do the single unmatched sentences in the summaries of \emph{Ulysses} by James Joyce (\cref{sec:4300-summary}) and \emph{Micromegas} by Voltaire (\cref{sec:30123-summary}). We see that summaries may contain information more than a mere synthesis of novel events despite the Wikipedia plot summary guideline suggesting summaries ``avoid commentary.''

$G_s$ values are distributed more widely than the $G_c$ values (\cref{fig:hist-gini-sents}, \cref{fig:hist-gini-chaps}), ranging from 0.00 to 0.58, although the median $G_s$ value is less than the median $G_c$ (0.22 vs.~0.28).  
Summaries with very low $G_s$ values, like those of \emph{Metamorphosis} by Franz Kafka ($G_s\approx0.00$), \emph{Heart of Darkness} by Joseph Conrad ($G_s\approx0.00$), and \emph{Ulysses} by James Joyce ($G_s\approx0.01$) (\cref{fig:5200-bipartite-main,fig:4300-bipartite-main}; \cref{sec:bipartiteGraphs}, \cref{fig:pg219-bipartite}), usually match each sentence with 1--2 chapters and are often strongly linear. Indeed, we find there is negative correlation between linearity and $G_s$ ($r\approx-0.50$, $p<10^{-8}$), suggesting summaries with even distributions of novel chapters across sentences are usually more linear. Summaries with low $G_s$ values also tend to be more uniform; $G_s$ is negatively correlated with the proportion of novel chapters matched ($r\approx-0.40$, $p<10^{-5}$). It is also positively correlated with mean off-diagonal distance ($r\approx0.43$, $p<10^{-6}$) (the combination of linearity and uniformity) suggesting a uniform distribution of novel chapters across summary sentences reflects summary's adherence to a naive summary structure. Finally, $G_s$ is strongly negatively correlated with the log-ratio of summary-to-book length ($r\approx-0.60$, $p<10^{-13}$), again suggesting the more summary `space' there is per book text, the more standard the summaries will be.

\section{Conclusion}

In this work, we examine whether summaries serve as consistent distillations of plot.
We find that summaries vary in form and structure, seemingly impacted by both summarization choices driven by stylistic desire and the internal structure of the works they summarize.
Summaries often break from linearity (the expectation that summaries describe events in the same order as source works) and uniformity (the expectation that summaries attend equally to all of the source), but that how and why they do so varies. 
For example, while many summaries we examined adhere strictly to the order of events described in their original texts, we find summarizers occasionally reorder events to foreground narrative framing or separate concurrent plot lines.
Therefore, the order of events as given in a summary can falsely represent the source text's chronology.
Broadly, our results suggest that researchers making use of summaries as proxies for their source texts should consider what properties of literary texts need to be preserved and the likelihood that summaries they have access to \textit{actually} contain or signal these narrative qualities.

\paragraph{Next steps.}
Our work hints at a number of productive research threads to follow on. In constructing our dataset, we found that mapping summary sentences to their origins in the source texts was surprisingly challenging for both human annotators and frontier LLMs. While our work contributes a small evaluation dataset and examples of automated mappings, we have identified opportunities to expand both the dataset and mapping pipeline. Moreover, further formalizations of summary-to-source mapping may allow for more fine-grained analysis of this topic.

Additionally, while we restrict our attention to Wikipedia summaries, expanding to solo human-authored or AI-authored summaries would be a promising way to characterize the distribution of summarization choices. Moreover, further research should consider more carefully the collaborative production of these summaries and the effect this has on their structure, both in the context of Human-Human and Human-AI co-creation.
Finally, while our close-reading analysis of the alignment results offers a reason why summary writers violate linearity and uniformity, we believe there is an opportunity for further work studying the choices summarizers make in condensing texts.

\section*{Acknowledgments}
We would like to thank Axel Bax, Federica Bologna, Kiara Liu, Sanghoon Oh, Andrea Wang, Matthew Wilkens, and Shengqi Zhu for their thoughtful feedback.
This work was supported in part by the Cornell Foundational AI PhD Fellowship, the Schmidt Sciences Humanities and AI Virtual Institute, and the Danish National Research Foundation via TEXT: Center for Contemporary Cultures of Text (grant number DNRF193).
We additionally acknowledge the support of the Natural Sciences and Engineering Research Council of Canada (NSERC); nous remercions le Conseil de recherches en sciences naturelles et en génie du Canada (CRSNG) de son soutien.

% Print the bibliography at the end. Keep this line after the main
% text of your paper, and before an appendix. 
\printbibliography
\newpage

% You can include an appendix using the following command
\appendix

\section{Novel Dataset}\label{app:novels}

\begin{center}
\newcolumntype{R}{>{\RaggedRight\arraybackslash}X}
\renewcommand{\arraystretch}{1.3}
\rowcolors{2}{gray!15}{white}
\begin{tabularx}{\linewidth}{RRrr}
    \toprule
    \textbf{Author} & \textbf{Title} & \textbf{Publication Date} & \textbf{ID} \\
    \midrule
    \endhead
    \bottomrule
    \endfoot
    Abbott, Edwin Abbott & Flatland: A Romance of Many Dimensions & 1884 & 45506 \\
    Ainsworth, William Harrison & Rookwood & 1834 & 23564 \\
    Alcott, Louisa May & Little Women & 1868--9 & 514  \\
    Alcott, Louisa May & Moods & 1864 & 28203 \\
    Alcott, Louisa May & Work: A Story of Experience & 1873 & 4770 \\
    Austen, Jane & Emma & 1816 & 158 \\
    Austen, Jane & Mansfield Park & 1814 & 141 \\
    Austen, Jane & Northanger Abbey & 1818 & 121 \\
    Austen, Jane & Persuasion & 1818 & 105 \\
    Austen, Jane & Pride and Prejudice & 1813 & 1342 \\
    Baum, L.\ Frank & Rinkitink in Oz & 1916 & 25581 \\
    Baum, L.\ Frank & The Wonderful Wizard of Oz & 1900 & 55 \\
    Bellamy, Edward & Equality & 1897 & 7303 \\
    Braddon, M.\ E.\ & Lady Audley's Secret & 1862 & 8954 \\
    Bront\"e, Charlotte & Jane Eyre & 1847 & 1260 \\
    Bront\"e, Charlotte & Shirley & 1849 & 30486 \\
    Bront\"e, Charlotte & Villette & 1853 & 9182 \\
    Bront\"e, Emily & Wuthering Heights & 1847 & 768 \\
    Buchan, John & Greenmantle & 1916 & 559 \\
    Buchan, John & Huntingtower & 1922 & 3782 \\
    Burroughs, Edgar Rice & The Cave Girl & 1925 & 69191 \\
    Burnett, Frances Hodgson & A Little Princess & 1905 & 146 \\
    Burney, Fanny & Evelina & 1778 & 6053 \\
    Butler, Samuel & The Way of All Flesh & 1903 & 2084 \\
    Carroll, Lewis & Alice's Adventures in Wonderland & 1865 & 11 \\
    Cather, Willa & My \'{A}ntonia & 1918 & 242\\
    Cather, Willa & O Pioneers! & 1913 & 24 \\
    Chesnutt, Charles W.\ & The Marrow of Tradition & 1901 & 11228 \\
    Christie, Agatha & The Murder of Roger Ackroyd & 1926 & 69087 \\
    Christie, Agatha & The Seven Dials Mystery & 1929 & 75288 \\
    Collins, Wilkie & Armadale & 1866 & 1895 \\
    Cooper, James Fenimore & The Deerslayer & 1841 & 3285 \\
    Cooper, James Fenimore & The Last of the Mohicans & 1826 & 940 \\
    Conrad, Joseph & Heart of Darkness & 1902 & 219 \\
    Conrad, Joseph & Nostromo: A Tale of the Seaboard & 1904 & 2021 \\
    Conrad, Joseph & The Secret Agent: A Simple Tale & 1907 & 974 \\
    Conrad, Joseph & Victory: An Island Tale & 1915 & 6378 \\
    Crane, Stephen & Maggie: A Girl of the Streets & 1893 & 447 \\
    Crane, Stephen & The Red Badge of Courage & 1895 & 73 \\
    Curwood, James Oliver  & Kazan & 1914 & 10084 \\
    Defoe, Daniel & The Further Adventures of Robinson Crusoe & 1719 & 561 \\
    Dickens, Charles & A Christmas Carol & 1843 & 19337 \\
    Dickens, Charles & David Copperfield & 1850 & 766 \\
    Dickens, Charles & Little Dorrit & 1857 & 963 \\
    Dickens, Charles & Oliver Twist & 1838 & 730 \\
    Dickens, Charles & A Tale of Two Cities & 1859 & 98 \\
    Doyle, Arthur Conan & The Hound of the Baskervilles & 1902 & 2852 \\
    Doyle, Arthur Conan & A Study in Scarlet & 1888 & 244 \\
    Doyle, Arthur Conan & The Valley of Fear & 1914 & 3289 \\
    Dreiser, Theodore & Sister Carrie: A Novel & 1900 & 233 \\
    Du Maurier, George & Trilby & 1894 & 39858 \\
    Dumas, Alexandre & The Three Musketeers & 1844 & 1257 \\
    Eliot, George & Adam Bede & 1859 & 507 \\
    Eliot, George & Middlemarch & 1871--2 & 145 \\
    Eliot, George & The Mill on the Floss & 1860 & 6688 \\
    Eliot, George & Romola & 1862--3 & 24020 \\
    Falkner, John Meade & Moonfleet & 1898 & 10743 \\
    Faulkner, William & The Sound and the Fury & 1929 & 75170 \\
    Fitzgerald, F.\ Scott & This Side of Paradise & 1920 & 805 \\
    Flaubert, Gustave & Madame Bovary & 1857 & 2413 \\
    Ford, Ford Madox & The Good Soldier & 1915 & 2775 \\
    Forster, E.\ M.\ & Howards End & 1910 & 2891 \\
    Forster, E.\ M.\ & A Room with a View & 1908 & 2641 \\
    Forster, E.\ M.\ & Where Angels Fear to Tread & 1905 & 2948 \\
    Galdos, Benita Perez & Marianela & 1878 & 48818 \\
    Gaskell, Elizabeth & Mary Barton & 1848 & 2153 \\
    Gilman, Charlotte Perkins & Herland & 1979 & 32 \\
    Gissing, George & Demos & 1886 & 4309 \\
    Glasgow, Ellen & Virginia & 1913 & 26316 \\
    Goldsmith, Oliver & The Vicar of Wakefield & 1766 & 2667 \\
    Goncharov, Ivan & Oblomov & 1859 & 54700 \\
    Grahame, Kenneth & The Wind in the Willows & 1908 & 289 \\
    Haggard, H.\ Rider & She: A History of Adventure & 1887 & 3155 \\
    Hammett, Dashiell & The Maltese Falcon & 1930 & 77600 \\
    Hardy, Thomas & Jude the Obscure & 1895 & 153 \\
    Hardy, Thomas & The Return of the Native & 1878 & 122 \\
    Harrison, Harry & Deathworld & 1960 & 28346 \\
    Hawthorne, Nathaniel & Fanshawe & 1828 & 7085 \\
    Hawthorne, Nathaniel & The House of the Seven Gables & 1851 & 77 \\
    Hawthorne, Nathaniel & The Scarlet Letter & 1850 & 25344 \\
    Hemingway, Ernest & A Farewell to Arms & 1929 & 75201 \\
    Hesse, Hermann & Siddhartha & 1922 & 2500 \\
    Heyer, Georgette & Beauvallet & 1929 & 75547 \\
    Heyer, Georgette & The Black Moth: A Romance of the XVIIIth Century & 1921 & 38703 \\
    Howells, William Dean & The Rise of Silas Lapham & 1885 & 154 \\
    Hudson, W.\ H.\ & Green Mansions: A Romance of the Tropical Forest & 1904 & 942 \\
    Hughes, Richard & A High Wind in Jamaica & 1929 & 75530 \\
    Hugo, Victor & Ninety-Three & 1874 & 49372 \\
    Jackson, Helen Hunt & Ramona & 1884 & 2802 \\
    James, Henry & The Turn of the Screw & 1898 & 209 \\
    James, Henry & What Maisie Knew & 1897 & 7118 \\
    Joyce, James & Ulysses & 1922 & 4300 \\
    Kafka, Franz & Metamorphosis & 1915 & 5200 \\
    Keene, Carolyn & The Hidden Staircase & 1930 & 77602 \\
    Kipling, Rudyard & Kim & 1901 & 2226 \\
    Le Fanu, Joseph Sheridan & Carmilla & 1872 & 10007 \\
    Lawrence, D.\ H.\ & Sons and Lovers & 1913 & 217 \\
    Lindsay, David & A Voyage to Arcturus & 1920 & 1329 \\
    Leroux, Gaston & The Phantom of the Opera & 1910 & 175 \\
    Lewis, Sinclair & Arrowsmith & 1925 & 70875 \\
    Lewis, Sinclair & Babbitt & 1922 & 1156 \\
    Lewis, Sinclair & Main Street & 1920 & 543 \\
    Lofting, Hugh & The Voyages of Doctor Dolittle & 1922 & 1154 \\
    London, Jack & The Valley of the Moon & 1913 & 1449 \\
    London, Jack & White Fang & 1906 & 910 \\
    Mason, A.\ E.\ W.\ & Clementina & 1901 & 13567 \\
    Montgomery, L.\ M.\ & Anne of Green Gables & 1908 & 45 \\
    Orczy, Baroness & The Triumph of the Scarlet Pimpernel & 1922 & 65695 \\
    Peacock, Thomas Love & Gryll Grange & 1861 & 21514 \\
    Poe, Edgar Allan & The Narrative of Arthur Gordon Pym of Nantucket & 1838 & 51060 \\
    Pohl, Frederik \& C.\ M.\ Kornbluth & Wolfbane & 1959 & 51845 \\
    Porter, Eleanor H.\ & Pollyanna & 1913 & 1450 \\
    Radcliffe, Ann & The Mysteries of Udolpho & 1794 & 3268 \\
    Rand, Ayn & Anthem & 1938 & 1250 \\
    Rice, Alice Hegan & Sandy & 1905 & 14079 \\
    Sabatini, Rafael & Captain Blood & 1922 & 1965 \\
    Salten, Felix & Bambi, a Life in the Woods & 1923 & 63849 \\
    Saltykov-Shchedrin, Mikhail & The Golovlyov Family & 1880 & 44237 \\
    Sand, George & Indiana & 1832 & 63445 \\
    Sayers, Dorothy L.\ \& Robert Eustace & The Documents in the Case & 1930 & 77601 \\
    Scott, Walter & Ivanhoe: A Romance & 1819 & 82 \\
    Scott, Walter & Kenilworth & 1821 & 1606 \\
    Shelley, Mary Wollstonecraft & Frankenstein & 1818 & 84 \\
    Sinclair, Upton & Oil! & 1926--7 & 70379 \\
    Smollett, Tobias & The Expedition of Humphry Clinker & 1771 & 2160 \\
    Spyri, Johanna & Heidi & 1880--1 & 1448 \\
    Stevenson, Robert Louis & Catriona & 1893 & 589 \\
    Stevenson, Robert Louis & Kidnapped & 1886 & 421 \\
    Stevenson, Robert Louis & Treasure Island & 1883 & 120 \\
    Stoker, Bram & Dracula & 1897 & 45839 \\
    Stratton-Porter, Gene & Freckles & 1904 & 111 \\
    Thackeray, William Makepeace & Vanity Fair & 1848 & 599 \\
    Verne, Jules & Around the World in Eighty Days & 1873 & 103 \\
    Verne, Jules & From the Earth to the Moon & 1865 & 83 \\
    Verne, Jules & A Journey to the Centre of the Earth & 1864 & 18857 \\
    Verne, Jules & Twenty Thousand Leagues Under the Sea & 1870 & 164 \\
    Voltaire & Candide & 1759 & 19942 \\
    Voltaire & Microm\'{e}gas & 1752 & 30123 \\
    Warner, Gertrude Chandler & The Box-Car Children & 1924 & 42796 \\
    Waugh, Evelyn & Vile Bodies & 1930 & 77900 \\
    Webb, Frank J.\ & The Garies and Their Friends & 1857 & 11214 \\
    Wells, H.\ G.\ & Bealby: A Holiday & 1915 & 59769 \\
    Wells, H.\ G.\ & The Invisible Man & 1897 & 5230 \\
    Wells, H.\ G.\ & The Time Machine & 1895 & 35 \\
    Wharton, Edith & The House of Mirth & 1905 & 284 \\
    Wilkins, Mary E.\ & The Jamesons & 1899 & 17792 \\
    Wodehouse, P.\ G.\ & Mike & 1909 & 7423 \\
    Zamiatin, Evgenii Ivanonich & We & 1924 & 61963 \\
    Zola, \'{E}mile & L'Assommoir & 1877 & 8600 \\
    Zola, \'{E}mile & Germinal & 1885 & 56528 \\
\end{tabularx}
\end{center}

\section{Alignment Prompts}
\label{sec:alignPrompts}

\subsection{Prompt 1} \label{Prompt1}

\begin{Verbatim}[frame=single,fontsize=\small]
You are a literary alignment assistant. You will be given a set of numbered 
summary sentences and the full text of ONE chapter from a novel. For each 
summary sentence, decide whether it should be matched to this chapter.

Judge only from the text provided. Do not rely on any prior knowledge of this 
novel, its plot, or its characters; base every decision solely on the chapter 
text and summary below.

––––––––––––––––––––––––––––––––––––––––
INPUTS

SUMMARY SENTENCES (one per line, each prefixed with its integer id, e.g. 
"1. ..."):
[summary]

CHAPTER TEXT:
[text]

ALREADY-MATCHED IDS (a JSON array of sentence ids that were matched to EARLIER 
chapters of this same novel; may be empty):
[old_ids]

––––––––––––––––––––––––––––––––––––––––
WHAT COUNTS AS A MATCH

A sentence matches this chapter if at least one thing it describes is realized
here:
(a) An event it describes actually occurs in this chapter — it is dramatized in 
    the chapter's narrative (including inside a flashback or scene that is 
    shown, not merely referred to); OR
(b) For sentences that describe a state, relationship, or characterization 
    rather than an event (e.g. "the village is poor and isolated," "Mara 
    distrusts her brother"), this chapter is where that state is first 
    established or directly shown.

Match on identity of the event or state, NOT on wording. The summary will 
describe things more abstractly or in different words than the chapter. You are 
matching the same underlying event/state, not identical phrasing. Do not 
require a literal or word-for-word match.

WHAT DOES NOT COUNT

Do NOT match a sentence to this chapter if the relevant event is only:
  - recalled, summarized, or discussed by characters after the fact,
  - foreshadowed or anticipated, or
  - alluded to without being shown.
Also do NOT match on presence or framing alone:
  - A recurring character or place merely APPEARING here is NOT a match. Only 
    the specific event or state the sentence describes counts — not the 
    character's presence. A character introduced earlier who simply turns up 
    again in a separate, later scene is NOT a re-match unless the sentence's 
    own event actually happens here.
  - A sentence that merely asserts setting, time/date, or who narrates (e.g. 
    "the story is set in England in 1737," "Laura narrates her childhood") and 
    dramatizes no event here matches NO chapter on setting alone.
The test is whether the event or state the sentence describes is actually 
happening or shown in this chapter — not whether the chapter mentions it or 
merely features the same character.

MATCHING ACROSS MULTIPLE CHAPTERS

A summary sentence may match more than one chapter. Match this chapter if 
EITHER:
  - it bundles several distinct events that may occur in DIFFERENT, possibly 
    NON-ADJACENT chapters, and at least one of them actually occurs here. 
    Match EVERY chapter where any of its events occurs, even chapters far apart 
    — e.g. "X happens, and later Y happens" matches BOTH X's chapter and Y's 
    chapter; OR
  - it describes a single continuing event (a journey, a search, an illness, an 
    evolving relationship) that is ACTIVELY depicted — advancing, developing, 
    or being shown — here. A continuing event matches the FULL run of 
    consecutive chapters in which it is actively shown, not only the chapter 
    where it begins.

Match a chapter only where one of the sentence's OWN events is actually 
occurring or shown — NOT where the same character merely reappears in a
separate matter, and NOT where the event is only continuing in the background, 
recalled, or referred to.

For ids listed in the ALREADY-MATCHED IDS above: an event they describe
occurred in an earlier chapter. Match such an id again here if that SAME event 
is still actively unfolding (the next consecutive step of an ongoing event), 
or if a DIFFERENT event from the sentence occurs here. Answer NO if this 
chapter merely recalls, refers back to, foreshadows, or revisits the event in 
a separate later scene without it actively unfolding now.

BORDERLINE CASES

When a match is genuinely uncertain, decide by the test above: answer YES if 
the chapter shows the event or state itself, and NO only if the chapter merely 
refers to it or does not contain it. Do not default to NO when the event is 
plausibly shown here.

––––––––––––––––––––––––––––––––––––––––
EXAMPLES OF THE DISTINCTION

- Summary "Tom leaves home and travels to the city," in a chapter that shows 
  Tom packing and boarding the train → YES.
- The same sentence, in a later chapter where Tom only tells a friend how he 
  once left home → NO (recalled, not happening).
- Summary "Tom leaves home, and years later returns for his father's
  funeral." The departure occurred in an earlier chapter (its id is already 
  matched); this chapter dramatizes the funeral return → YES (a distinct 
  second event occurs here).

––––––––––––––––––––––––––––––––––––––––
OUTPUT

For EVERY summary sentence id, output "YES" or "NO". Return ONLY a single valid 
JSON object mapping every id (in ascending order) to "YES" or "NO" — no other 
text, commentary, or reasoning.

Example:
{ "1": "YES", "2": "NO", "3": "YES" }
\end{Verbatim}

\subsection{Prompt 2} \label{Prompt2}

\begin{Verbatim}[frame=single,fontsize=\small]
You are an intelligent literary assistant. A summary sentence has been 
tentatively matched to a chapter, but a neighbouring chapter can look similar 
(it may lead up to or follow the event). Decide which chapter(s) the event 
ACTUALLY takes place in, using **only** the chapter texts provided.

SUMMARY SENTENCE (id [sentence-id]):
[sentence]

NEARBY SUMMARY CONTEXT (for reference only; do NOT judge these):
[context-chapters]

CANDIDATE CHAPTERS (adjacent):
[chapters]

TASK:
Among chapters {choices}, return EVERY chapter in which the event the sentence 
describes ACTUALLY happens — i.e. is dramatized or actively advancing in the 
narrative. Often this is a single chapter; but when the event genuinely unfolds 
across two ADJACENT chapters, include BOTH — do NOT force a single choice when 
the event truly continues across the  chapter boundary. Exclude only a chapter 
that merely leads up to, recalls, foreshadows, or follows the event without it 
actually occurring there.

OUTPUT FORMAT:
Return ONLY {"chapters": [...]} — a JSON list of the chosen chapter number(s) 
from {choices}: one number, or two adjacent numbers if the event truly spans 
them.
\end{Verbatim}

\subsection{Prompt 3} \label{Prompt3}

\begin{Verbatim}[frame=single,fontsize=\small]
You are an intelligent literary assistant. Re-check ONE summary-sentence-to-
chapter match using **only** the chapter text provided.

SUMMARY SENTENCE (id [sentence-id]):
[sentence]

NEARBY SUMMARY CONTEXT (for reference only; do NOT judge these):
[context-chapters]

CHAPTER [chapter-index]:
[chapter-text]

NOTE: Summaries are usually written in chapter order, so by position this 
sentence would fall around chapter [expected-index]; this is chapter [chapter-
index], which is out of that order. That does NOT make it wrong — flashbacks, 
recurring events, and foreshadowing fulfilled later are genuine. Judge ONLY 
from the chapter text above, not from the order.

TASK:
Answer YES if the event the sentence describes is ACTUALLY happening —
dramatized or actively advancing — in THIS chapter. Answer NO if the chapter 
merely mentions, recalls, foreshadows, or leads up to it, or if the event does 
not appear at all.

The match must be the SPECIFIC occurrence the sentence describes — the same 
participants doing the same thing. Answer NO if this chapter shows only a 
similar or parallel event involving DIFFERENT characters, or shows the same 
characters merely present or continuing/pursuing an ongoing situation whose 
defining event (e.g. its initiation) happens elsewhere.

OUTPUT FORMAT:
Return ONLY {"match": "YES"} or {"match": "NO"}.
\end{Verbatim}

\section{Bipartite Summary Sentence to Chapter Graphs}
\label{sec:bipartiteGraphs}

\noindent
  \begin{minipage}[t]{0.48\linewidth}\centering
    \includegraphics[width=\linewidth]{images/pg45506_bipartite.pdf}
    {\captionsetup{hypcap=false, labelformat=simple, labelsep=none, textformat=empty}\captionof{figure}{\label{fig:pg45506-bipartite}}}
  \end{minipage}\hfill
  \begin{minipage}[t]{0.48\linewidth}\centering
    \includegraphics[width=\linewidth]{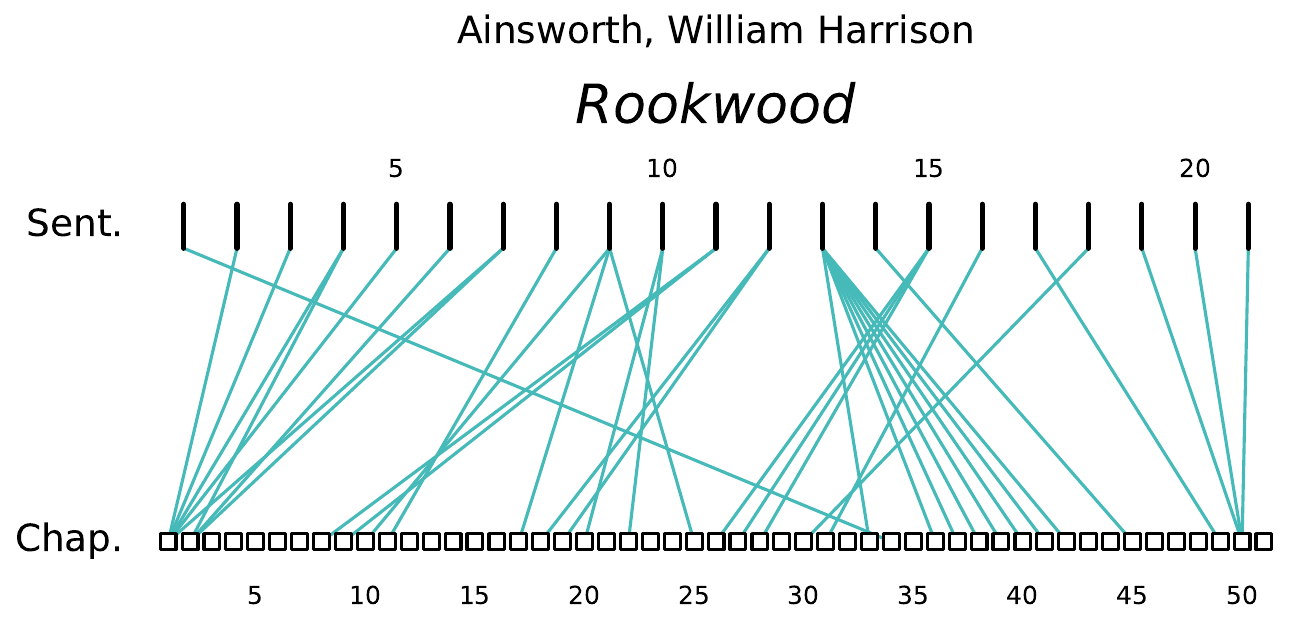}
    {\captionsetup{hypcap=false, labelformat=simple, labelsep=none, textformat=empty}\captionof{figure}{\label{fig:pg23564-bipartite}}}
  \end{minipage}\bigskip

\noindent
  \begin{minipage}[t]{0.48\linewidth}\centering
    \includegraphics[width=\linewidth]{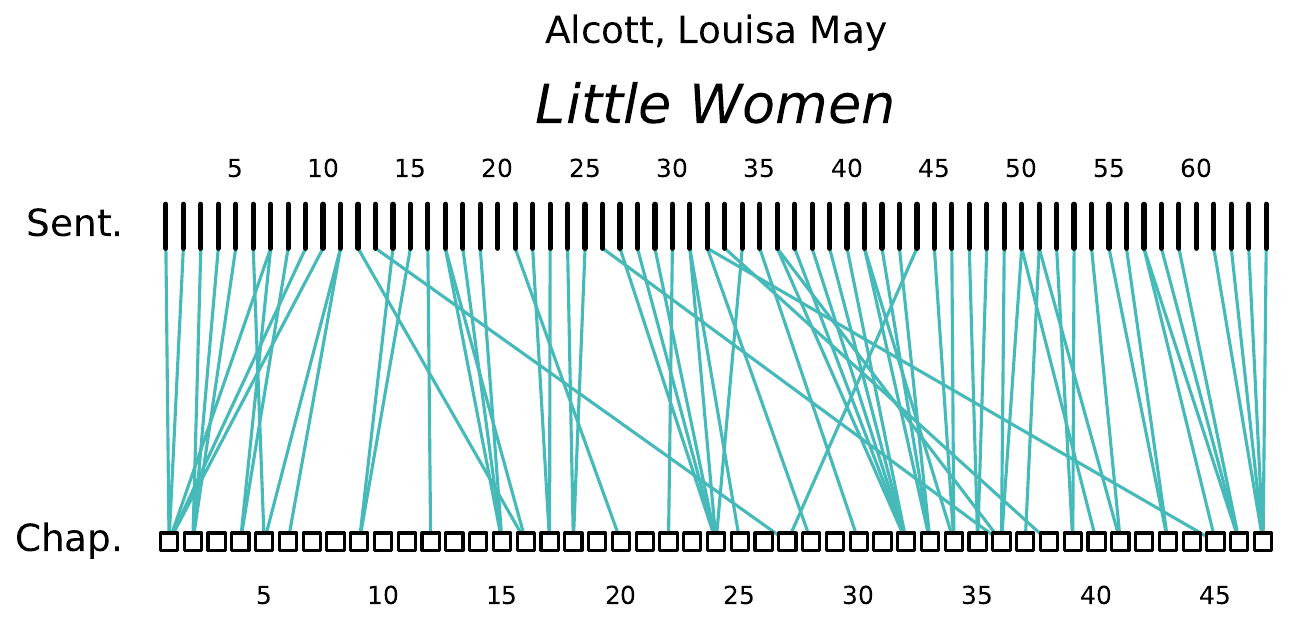}
    {\captionsetup{hypcap=false, labelformat=simple, labelsep=none, textformat=empty}\captionof{figure}{\label{fig:pg514-bipartite}}}
  \end{minipage}\hfill
  \begin{minipage}[t]{0.48\linewidth}\centering
    \includegraphics[width=\linewidth]{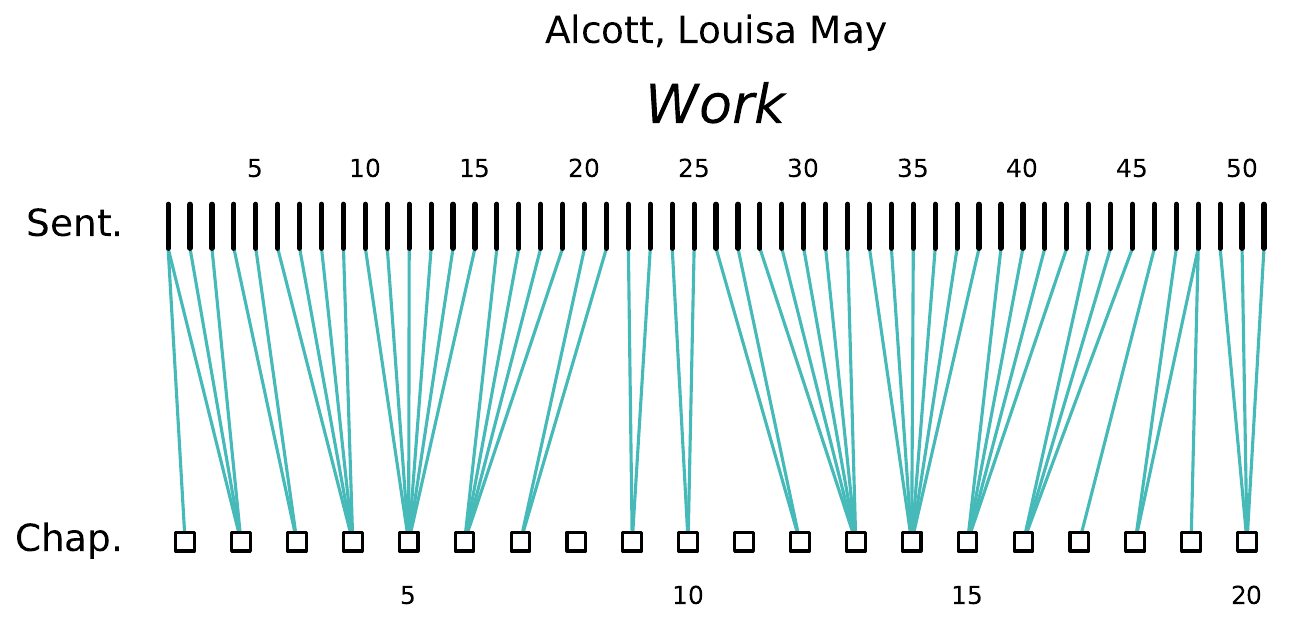}
    {\captionsetup{hypcap=false, labelformat=simple, labelsep=none, textformat=empty}\captionof{figure}{\label{fig:pg4770-bipartite}}}
  \end{minipage}\bigskip

\noindent
  \begin{minipage}[t]{0.48\linewidth}\centering
    \includegraphics[width=\linewidth]{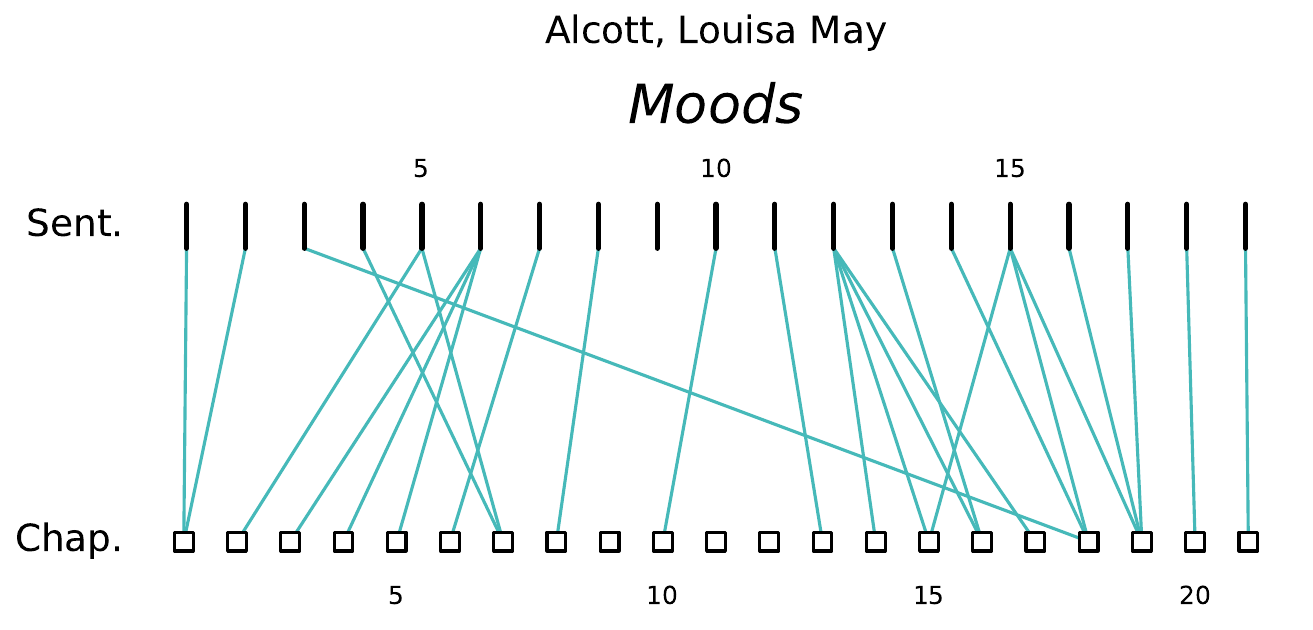}
    {\captionsetup{hypcap=false, labelformat=simple, labelsep=none, textformat=empty}\captionof{figure}{\label{fig:pg28203-bipartite}}}
  \end{minipage}\hfill
  \begin{minipage}[t]{0.48\linewidth}\centering
    \includegraphics[width=\linewidth]{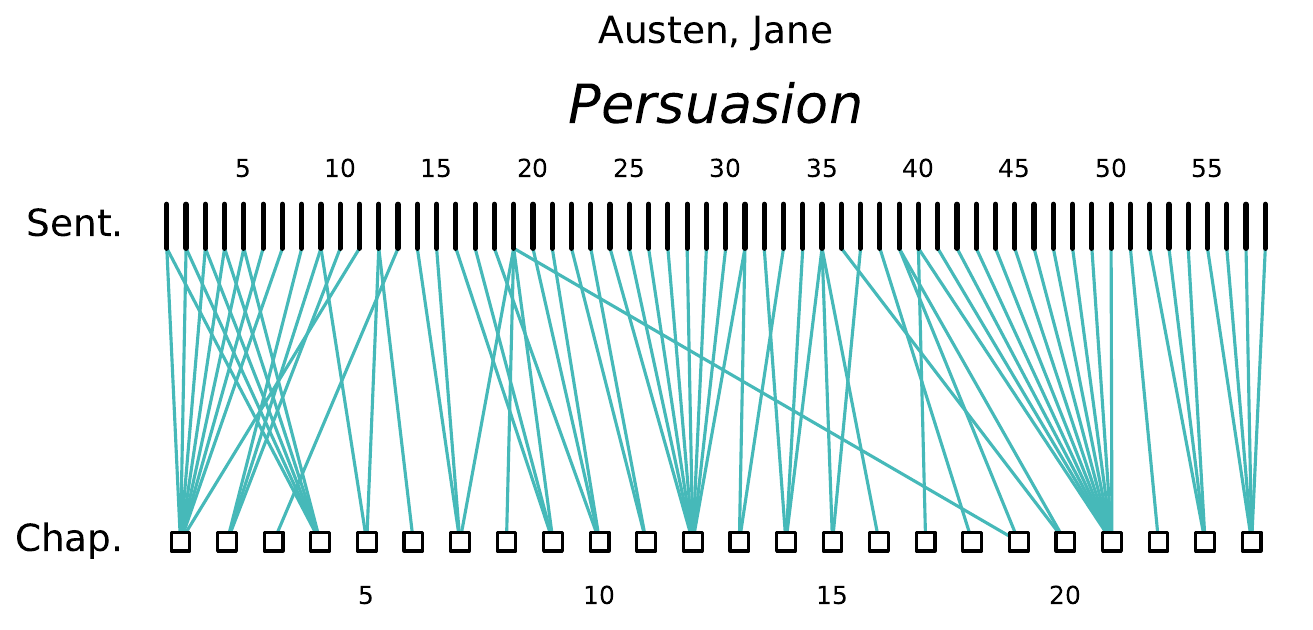}
    {\captionsetup{hypcap=false, labelformat=simple, labelsep=none, textformat=empty}\captionof{figure}{\label{fig:pg105-bipartite}}}
  \end{minipage}\bigskip

\noindent
  \begin{minipage}[t]{0.48\linewidth}\centering
    \includegraphics[width=\linewidth]{images/pg121_bipartite.pdf}
    {\captionsetup{hypcap=false, labelformat=simple, labelsep=none, textformat=empty}\captionof{figure}{\label{fig:pg121-bipartite}}}
  \end{minipage}\hfill
  \begin{minipage}[t]{0.48\linewidth}\centering
    \includegraphics[width=\linewidth]{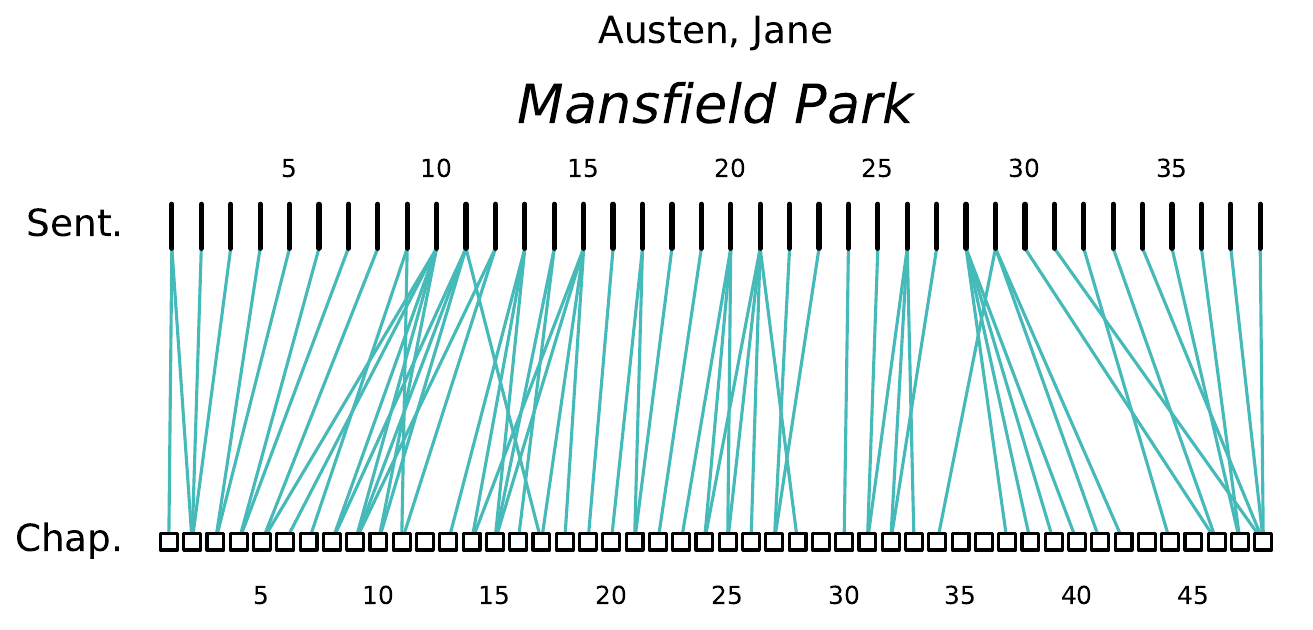}
    {\captionsetup{hypcap=false, labelformat=simple, labelsep=none, textformat=empty}\captionof{figure}{\label{fig:pg141-bipartite}}}
  \end{minipage}\bigskip

\noindent
  \begin{minipage}[t]{0.48\linewidth}\centering
    \includegraphics[width=\linewidth]{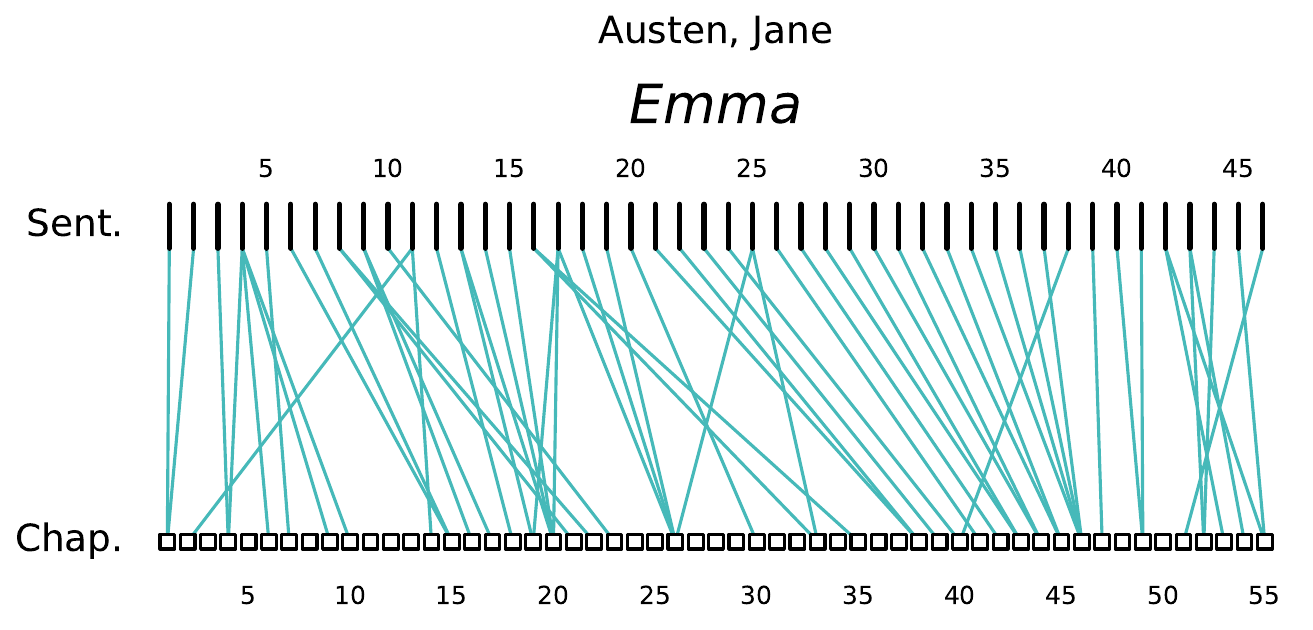}
    {\captionsetup{hypcap=false, labelformat=simple, labelsep=none, textformat=empty}\captionof{figure}{\label{fig:pg158-bipartite}}}
  \end{minipage}\hfill
  \begin{minipage}[t]{0.48\linewidth}\centering
    \includegraphics[width=\linewidth]{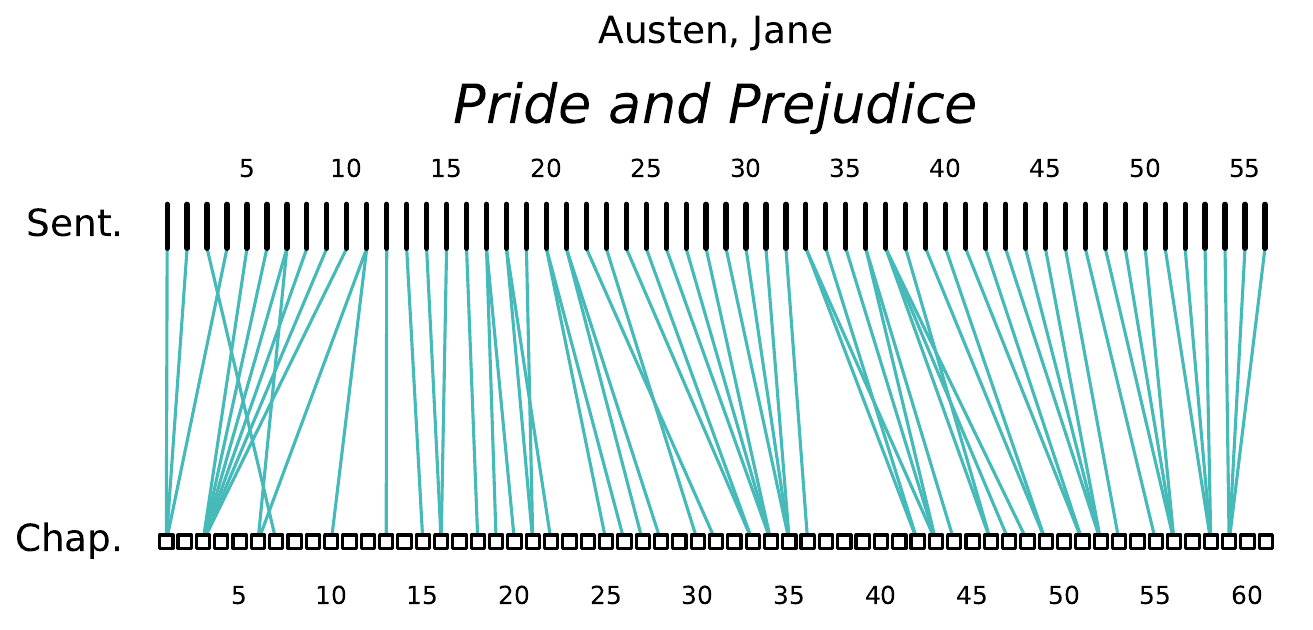}
    {\captionsetup{hypcap=false, labelformat=simple, labelsep=none, textformat=empty}\captionof{figure}{\label{fig:pg1342-bipartite}}}
  \end{minipage}\bigskip

\noindent
  \begin{minipage}[t]{0.48\linewidth}\centering
    \includegraphics[width=\linewidth]{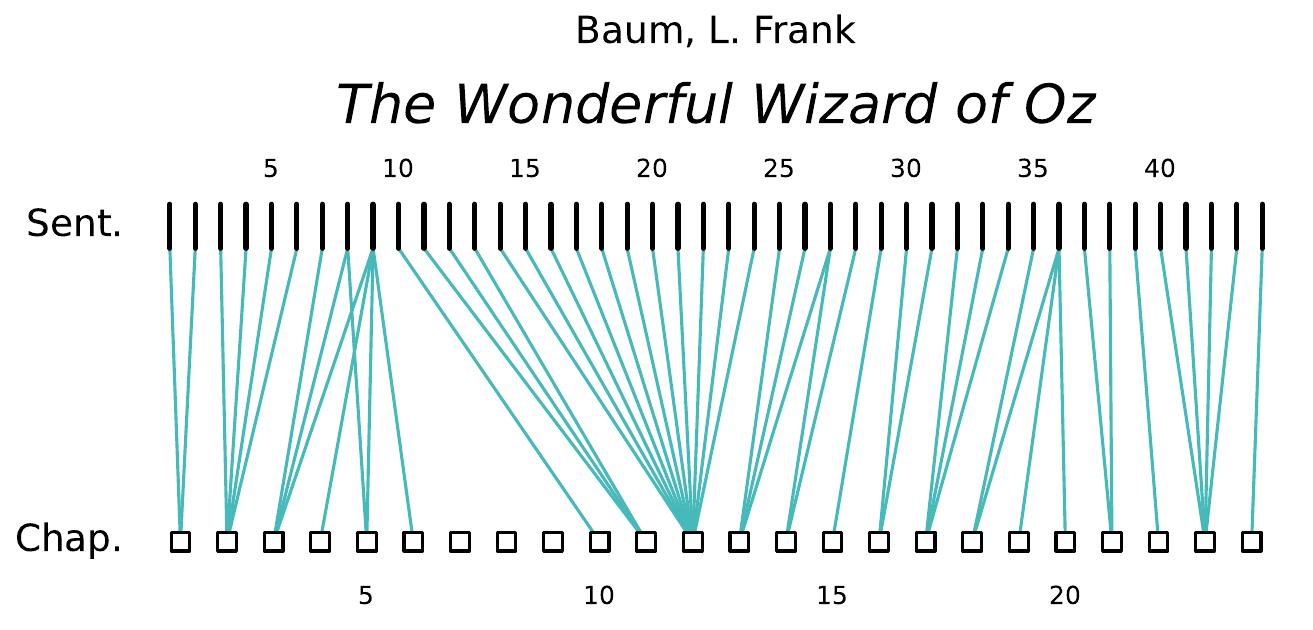}
    {\captionsetup{hypcap=false, labelformat=simple, labelsep=none, textformat=empty}\captionof{figure}{\label{fig:pg55-bipartite}}}
  \end{minipage}\hfill
  \begin{minipage}[t]{0.48\linewidth}\centering
    \includegraphics[width=\linewidth]{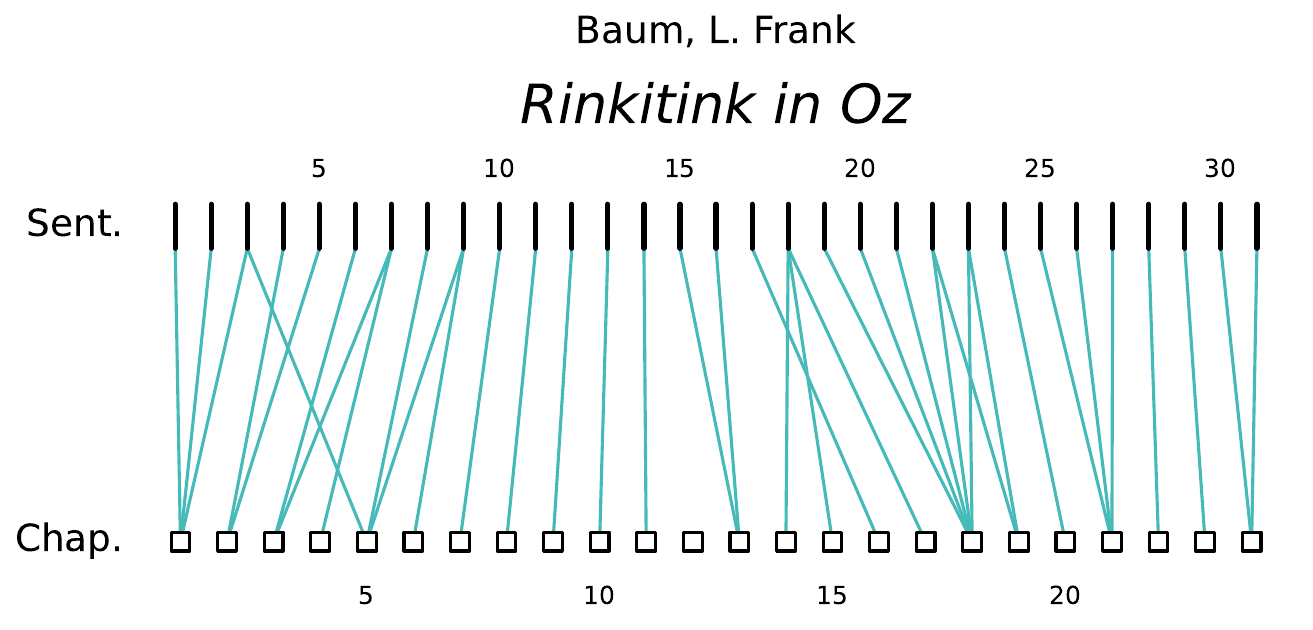}
    {\captionsetup{hypcap=false, labelformat=simple, labelsep=none, textformat=empty}\captionof{figure}{\label{fig:pg25581-bipartite}}}
  \end{minipage}\bigskip

\noindent
  \begin{minipage}[t]{0.48\linewidth}\centering
    \includegraphics[width=\linewidth]{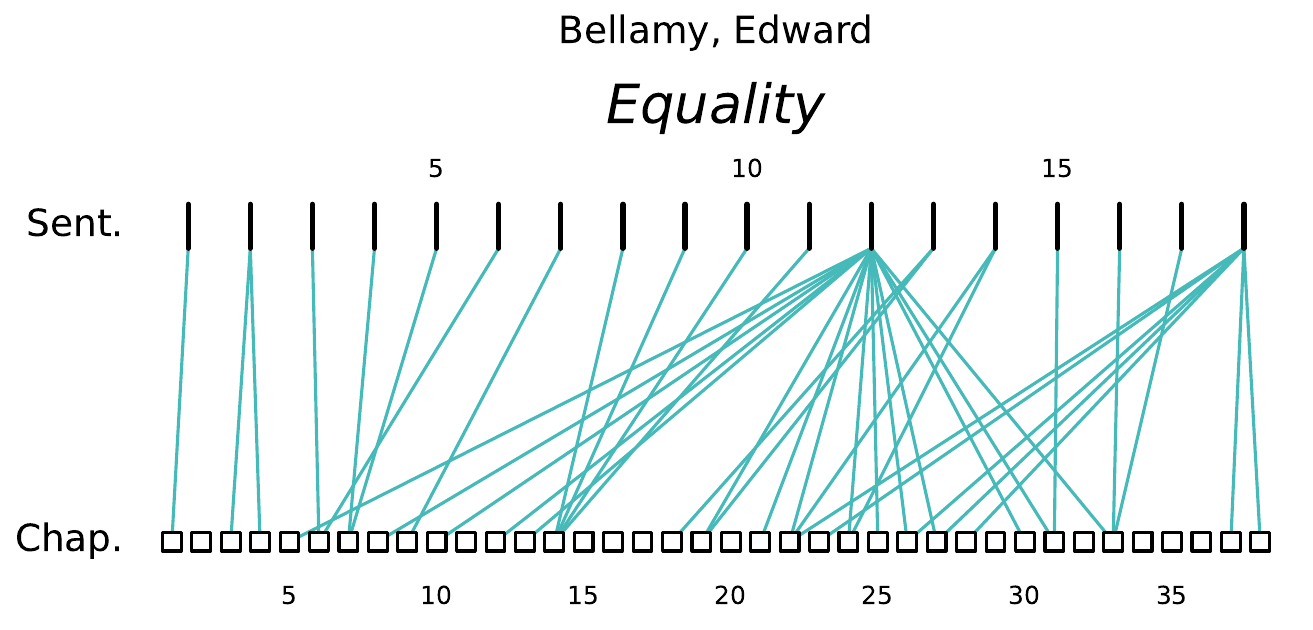}
    {\captionsetup{hypcap=false, labelformat=simple, labelsep=none, textformat=empty}\captionof{figure}{\label{fig:pg7303-bipartite}}}
  \end{minipage}\hfill
  \begin{minipage}[t]{0.48\linewidth}\centering
    \includegraphics[width=\linewidth]{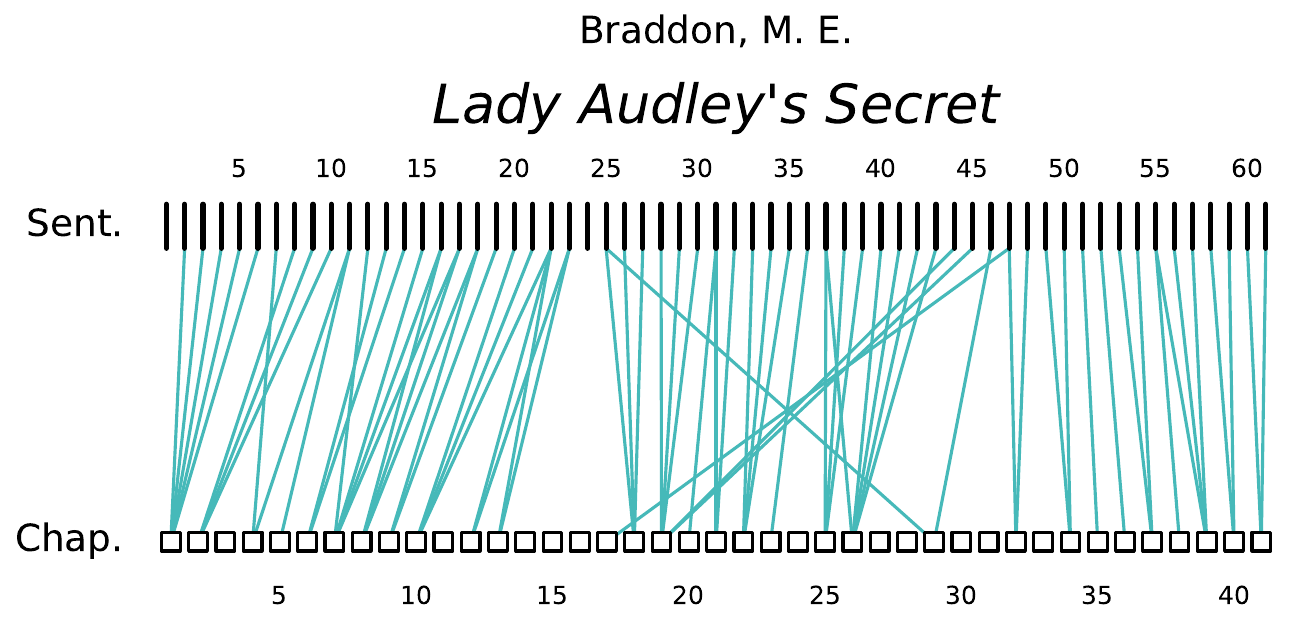}
    {\captionsetup{hypcap=false, labelformat=simple, labelsep=none, textformat=empty}\captionof{figure}{\label{fig:pg8954-bipartite}}}
  \end{minipage}\bigskip

\noindent
  \begin{minipage}[t]{0.48\linewidth}\centering
    \includegraphics[width=\linewidth]{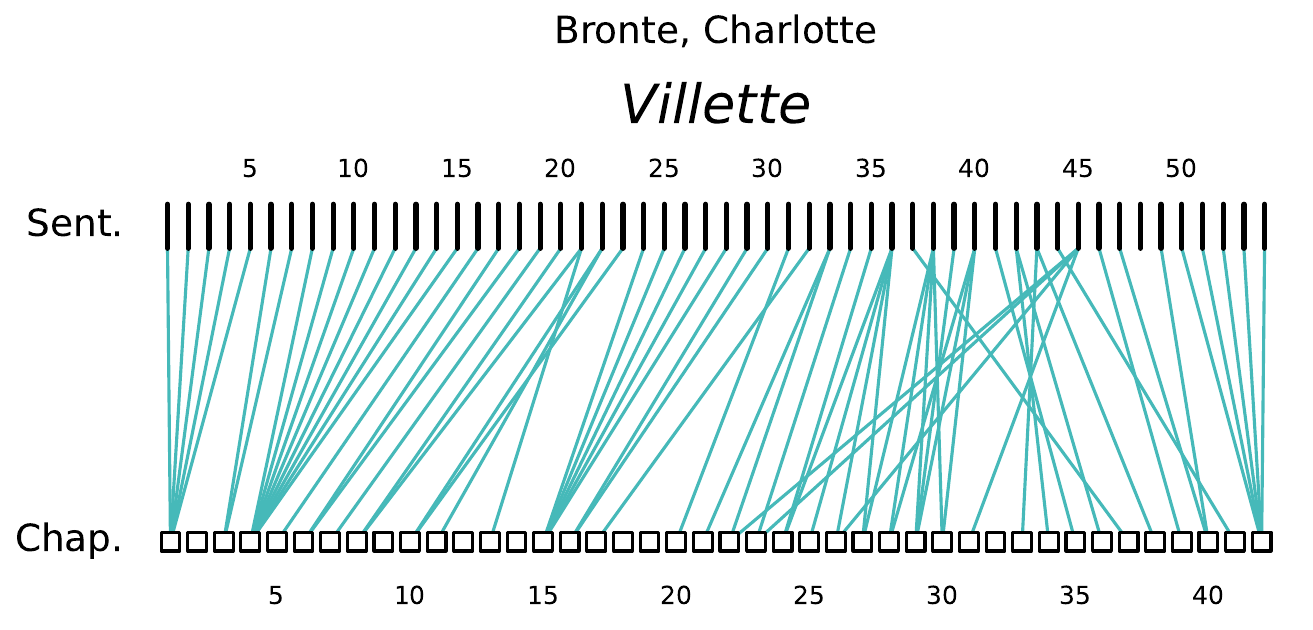}
    {\captionsetup{hypcap=false, labelformat=simple, labelsep=none, textformat=empty}\captionof{figure}{\label{fig:pg9182-bipartite}}}
  \end{minipage}\hfill
  \begin{minipage}[t]{0.48\linewidth}\centering
    \includegraphics[width=\linewidth]{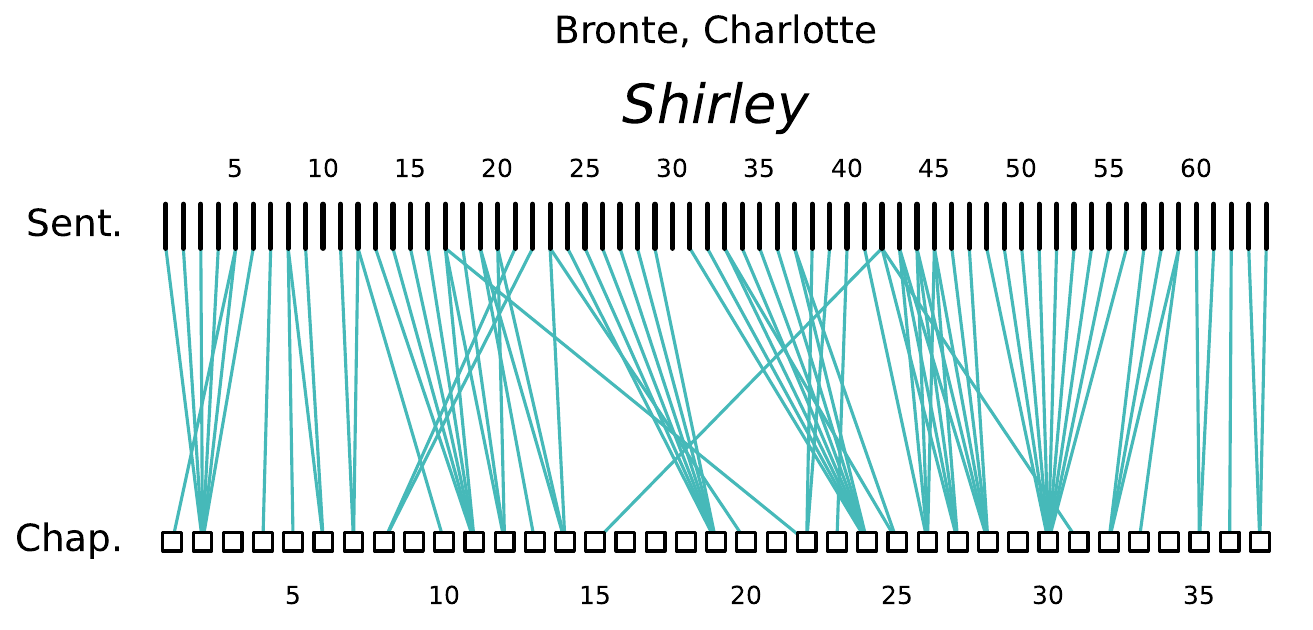}
    {\captionsetup{hypcap=false, labelformat=simple, labelsep=none, textformat=empty}\captionof{figure}{\label{fig:pg30486-bipartite}}}
  \end{minipage}\bigskip

\noindent
  \begin{minipage}[t]{0.48\linewidth}\centering
    \includegraphics[width=\linewidth]{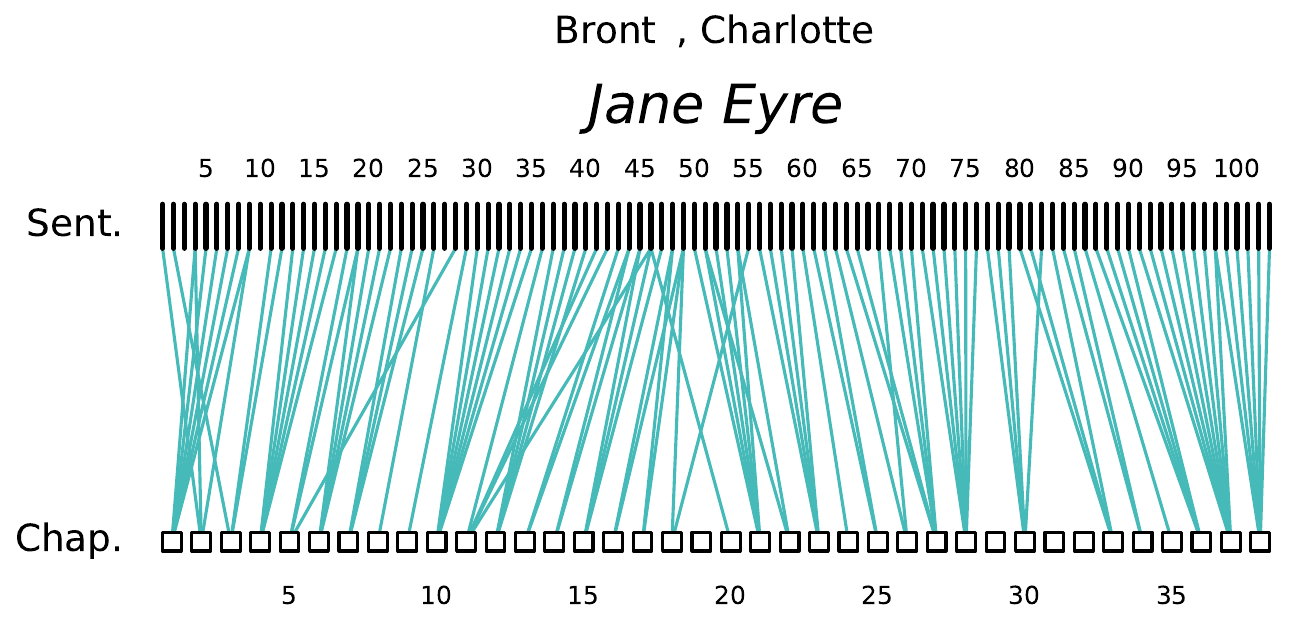}
    {\captionsetup{hypcap=false, labelformat=simple, labelsep=none, textformat=empty}\captionof{figure}{\label{fig:pg1260-bipartite}}}
  \end{minipage}\hfill
  \begin{minipage}[t]{0.48\linewidth}\centering
    \includegraphics[width=\linewidth]{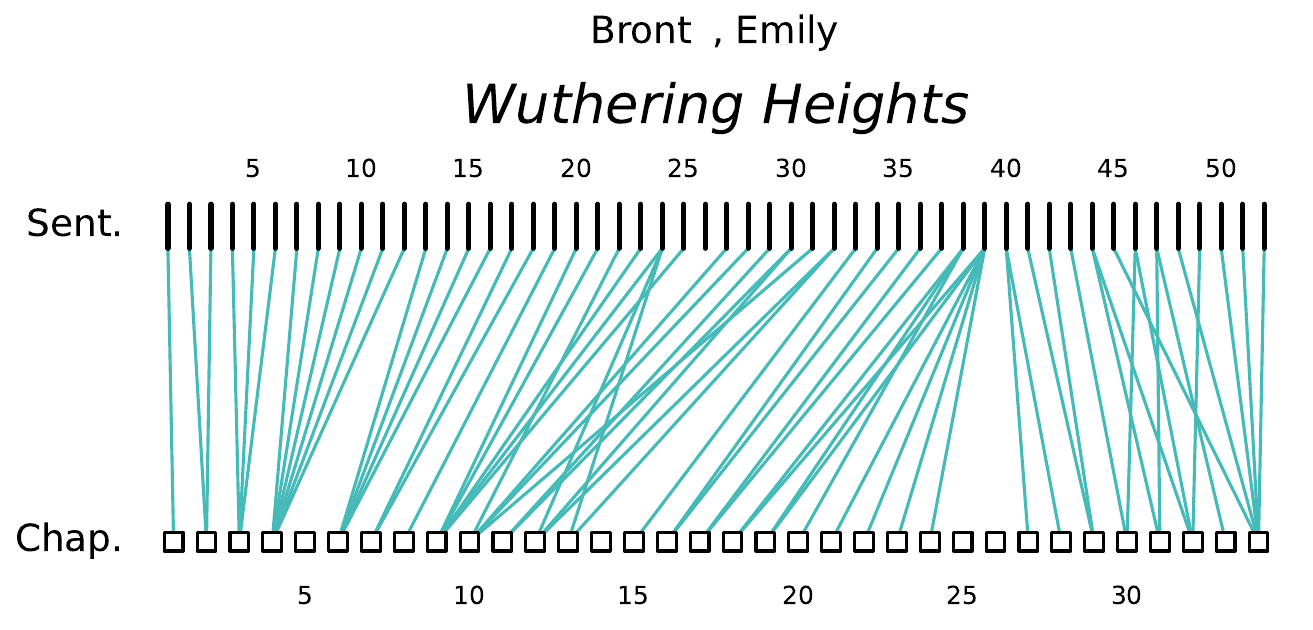}
    {\captionsetup{hypcap=false, labelformat=simple, labelsep=none, textformat=empty}\captionof{figure}{\label{fig:pg768-bipartite}}}
  \end{minipage}\bigskip

\noindent
  \begin{minipage}[t]{0.48\linewidth}\centering
    \includegraphics[width=\linewidth]{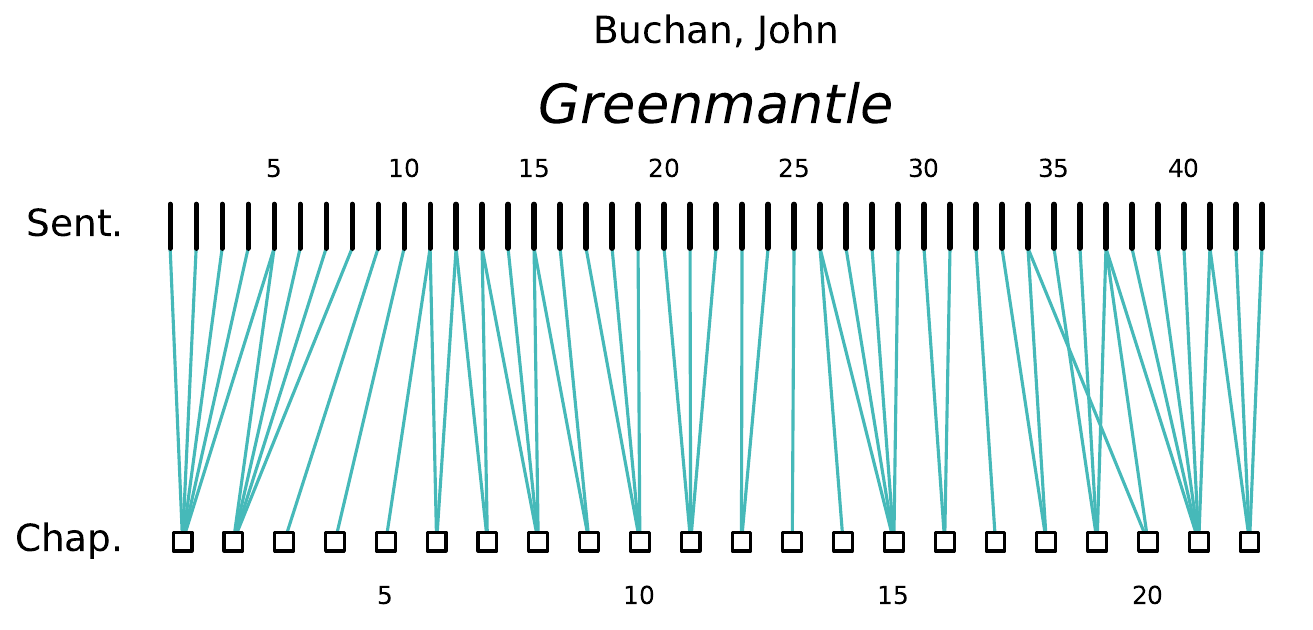}
    {\captionsetup{hypcap=false, labelformat=simple, labelsep=none, textformat=empty}\captionof{figure}{\label{fig:pg559-bipartite}}}
  \end{minipage}\hfill
  \begin{minipage}[t]{0.48\linewidth}\centering
    \includegraphics[width=\linewidth]{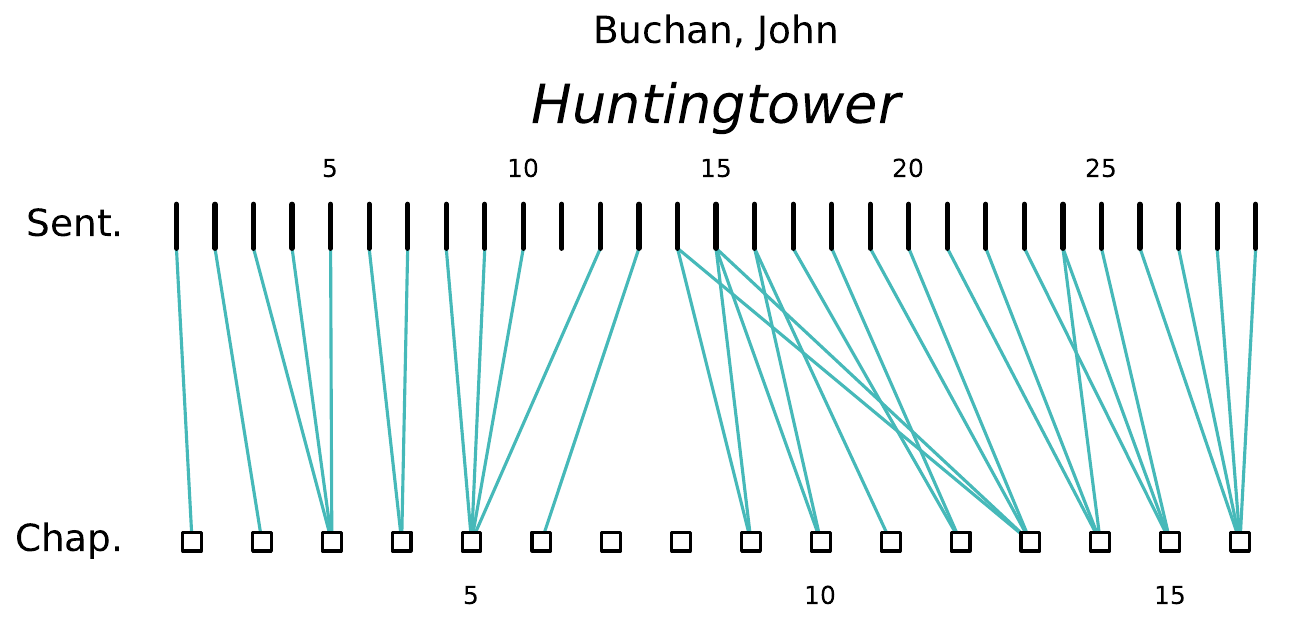}
    {\captionsetup{hypcap=false, labelformat=simple, labelsep=none, textformat=empty}\captionof{figure}{\label{fig:pg3782-bipartite}}}
  \end{minipage}\bigskip

\noindent
  \begin{minipage}[t]{0.48\linewidth}\centering
    \includegraphics[width=\linewidth]{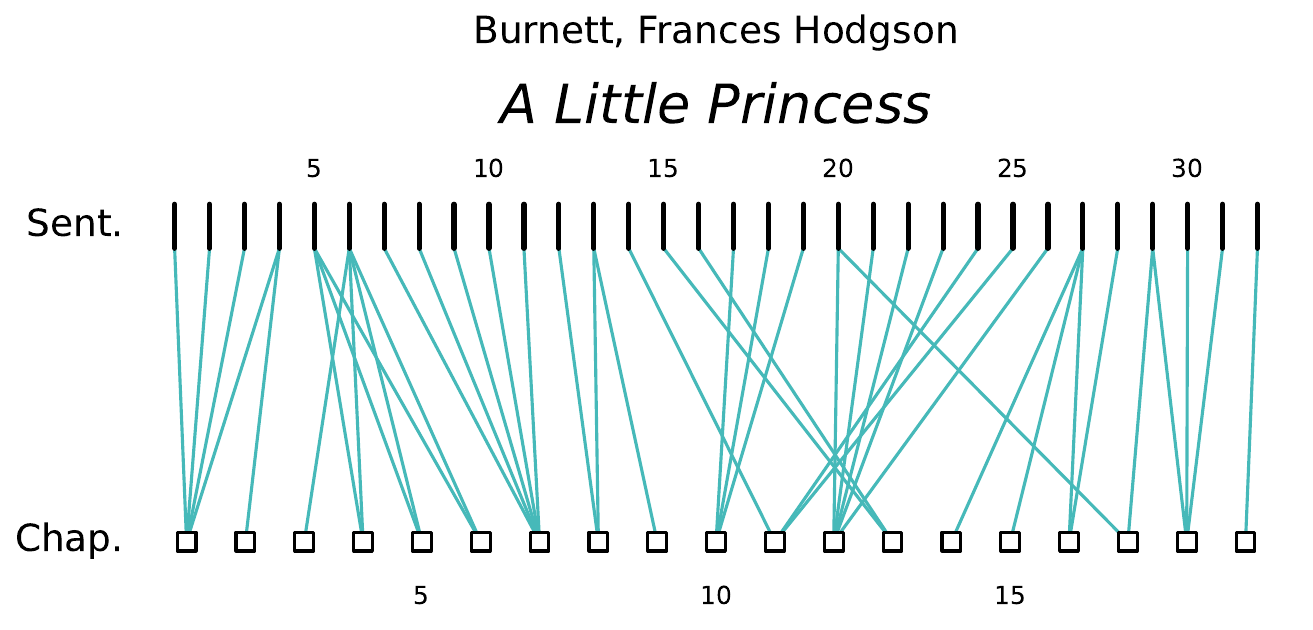}
    {\captionsetup{hypcap=false, labelformat=simple, labelsep=none, textformat=empty}\captionof{figure}{\label{fig:pg146-bipartite}}}
  \end{minipage}\hfill
  \begin{minipage}[t]{0.48\linewidth}\centering
    \includegraphics[width=\linewidth]{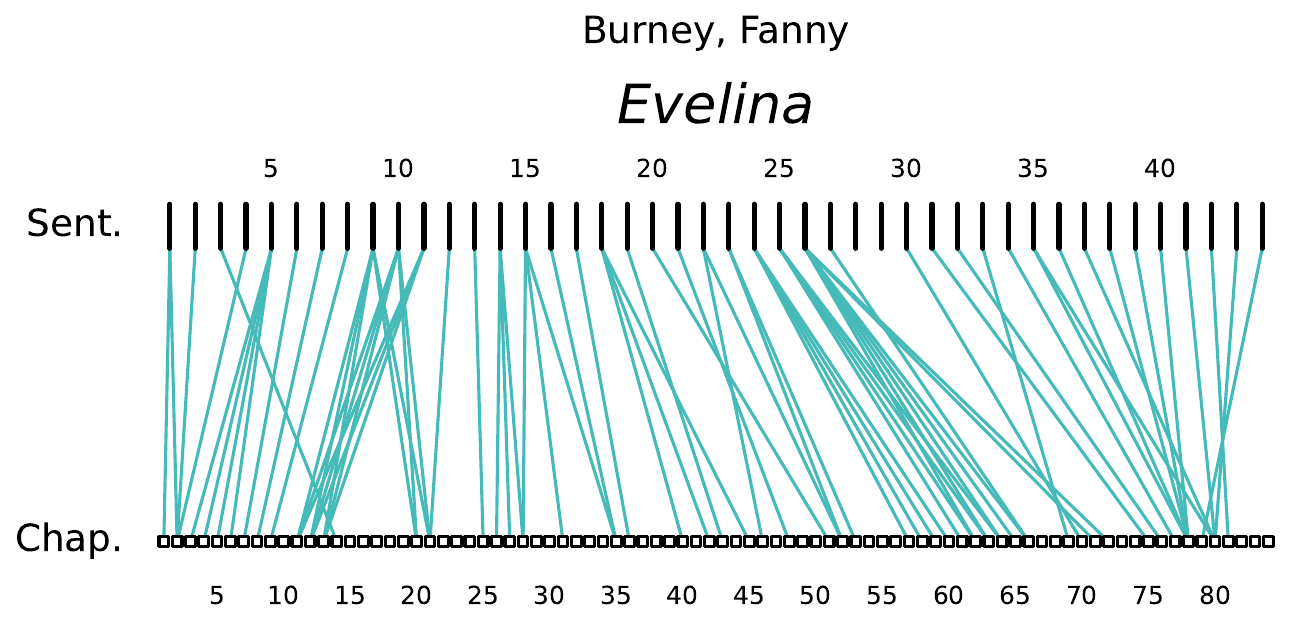}
    {\captionsetup{hypcap=false, labelformat=simple, labelsep=none, textformat=empty}\captionof{figure}{\label{fig:pg6053-bipartite}}}
  \end{minipage}\bigskip

\noindent
  \begin{minipage}[t]{0.48\linewidth}\centering
    \includegraphics[width=\linewidth]{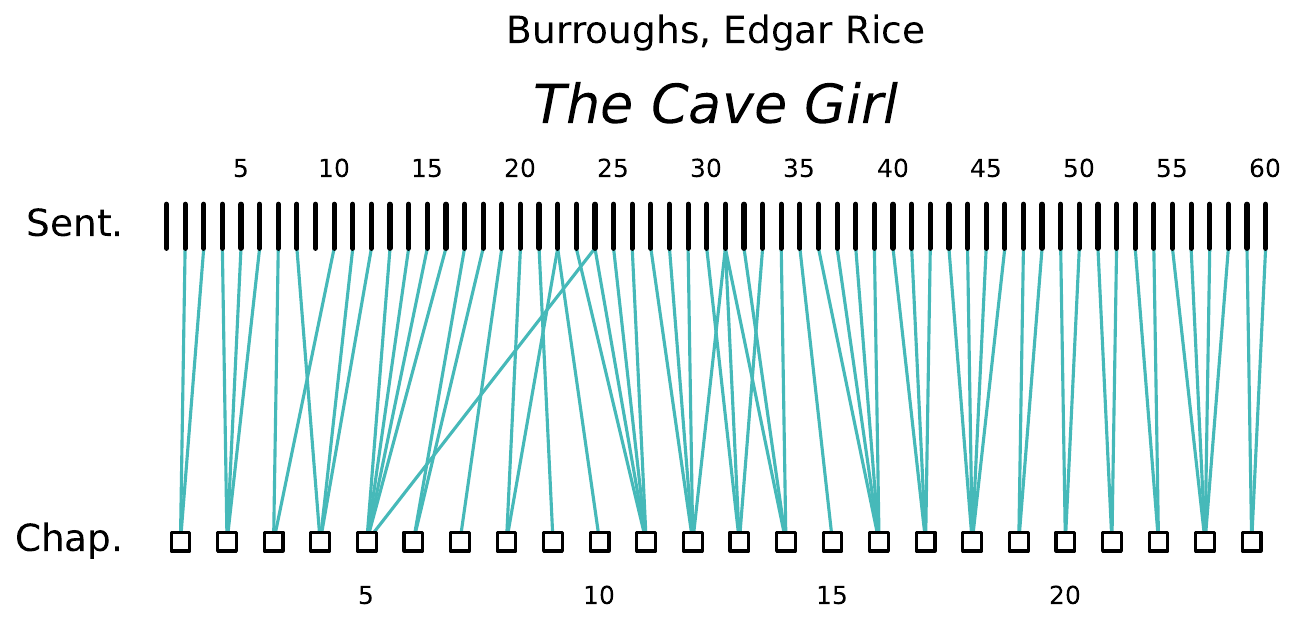}
    {\captionsetup{hypcap=false, labelformat=simple, labelsep=none, textformat=empty}\captionof{figure}{\label{fig:pg69191-bipartite}}}
  \end{minipage}\hfill
  \begin{minipage}[t]{0.48\linewidth}\centering
    \includegraphics[width=\linewidth]{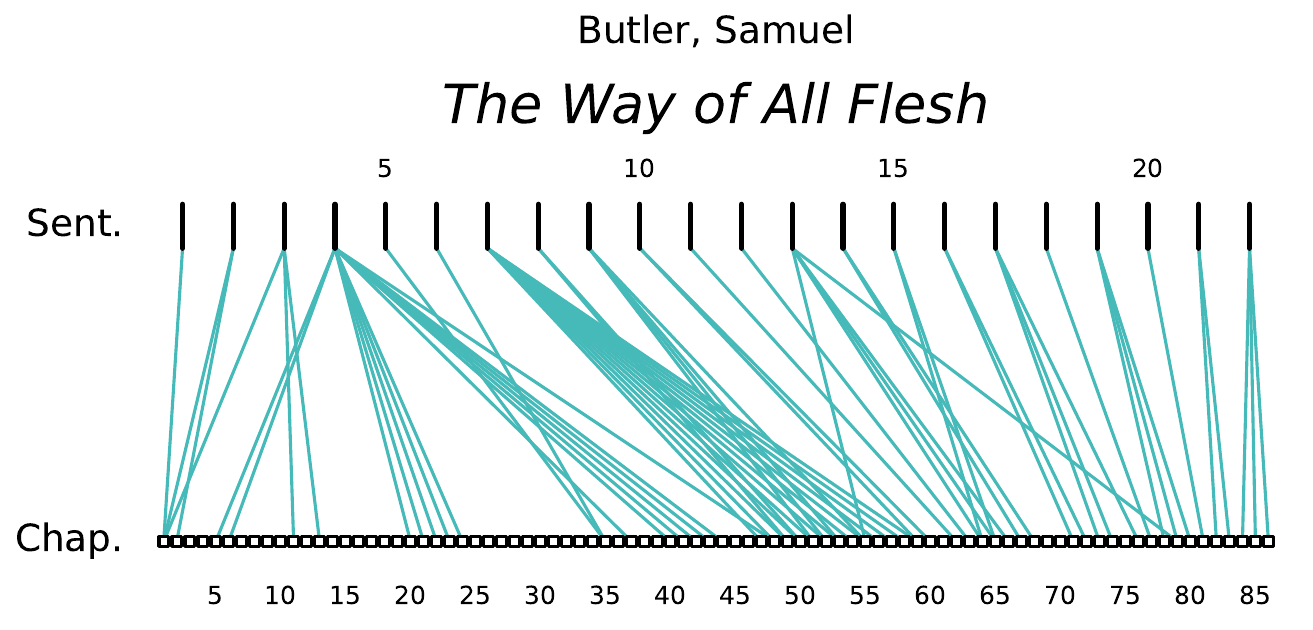}
    {\captionsetup{hypcap=false, labelformat=simple, labelsep=none, textformat=empty}\captionof{figure}{\label{fig:pg2084-bipartite}}}
  \end{minipage}\bigskip

\noindent
  \begin{minipage}[t]{0.48\linewidth}\centering
    \includegraphics[width=\linewidth]{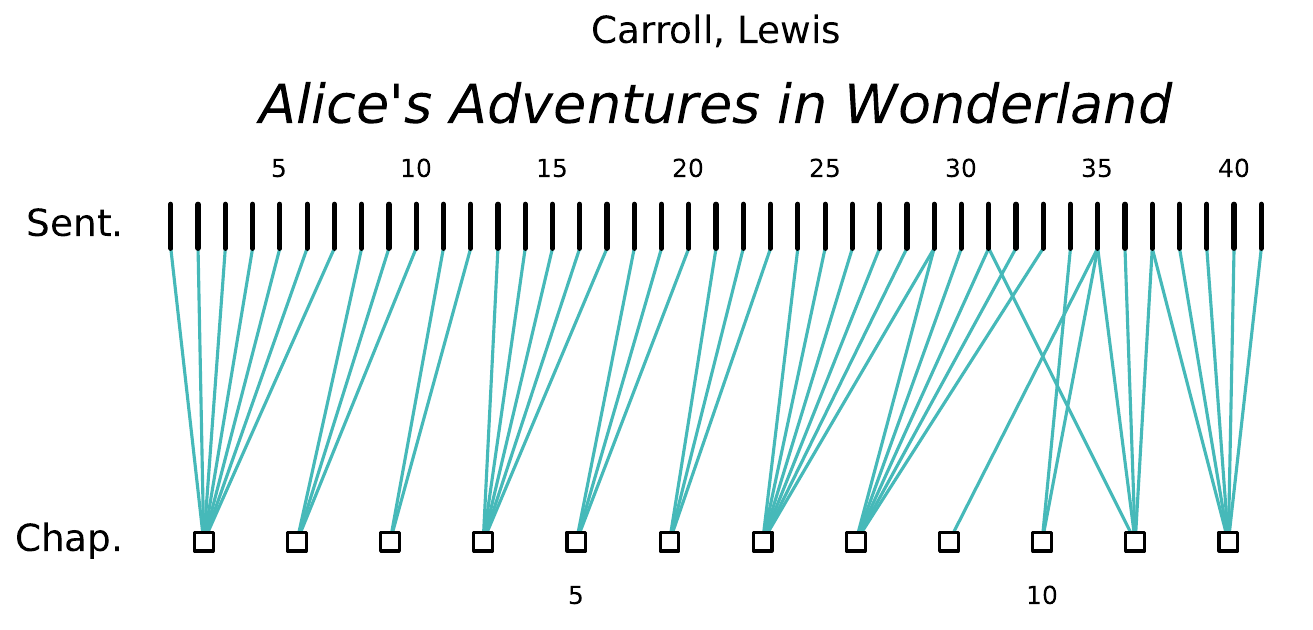}
    {\captionsetup{hypcap=false, labelformat=simple, labelsep=none, textformat=empty}\captionof{figure}{\label{fig:pg11-bipartite}}}
  \end{minipage}\hfill
  \begin{minipage}[t]{0.48\linewidth}\centering
    \includegraphics[width=\linewidth]{images/pg24_bipartite.pdf}
    {\captionsetup{hypcap=false, labelformat=simple, labelsep=none, textformat=empty}\captionof{figure}{\label{fig:pg24-bipartite}}}
  \end{minipage}\bigskip

\noindent
  \begin{minipage}[t]{0.48\linewidth}\centering
    \includegraphics[width=\linewidth]{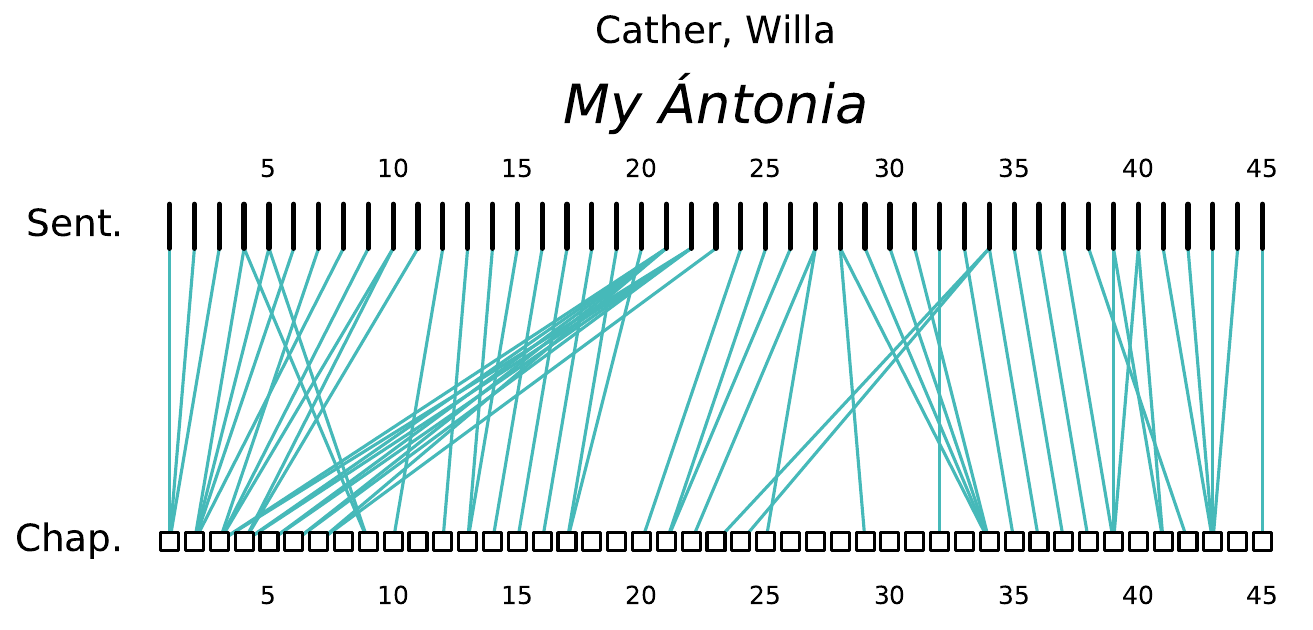}
    {\captionsetup{hypcap=false, labelformat=simple, labelsep=none, textformat=empty}\captionof{figure}{\label{fig:pg242-bipartite}}}
  \end{minipage}\hfill
  \begin{minipage}[t]{0.48\linewidth}\centering
    \includegraphics[width=\linewidth]{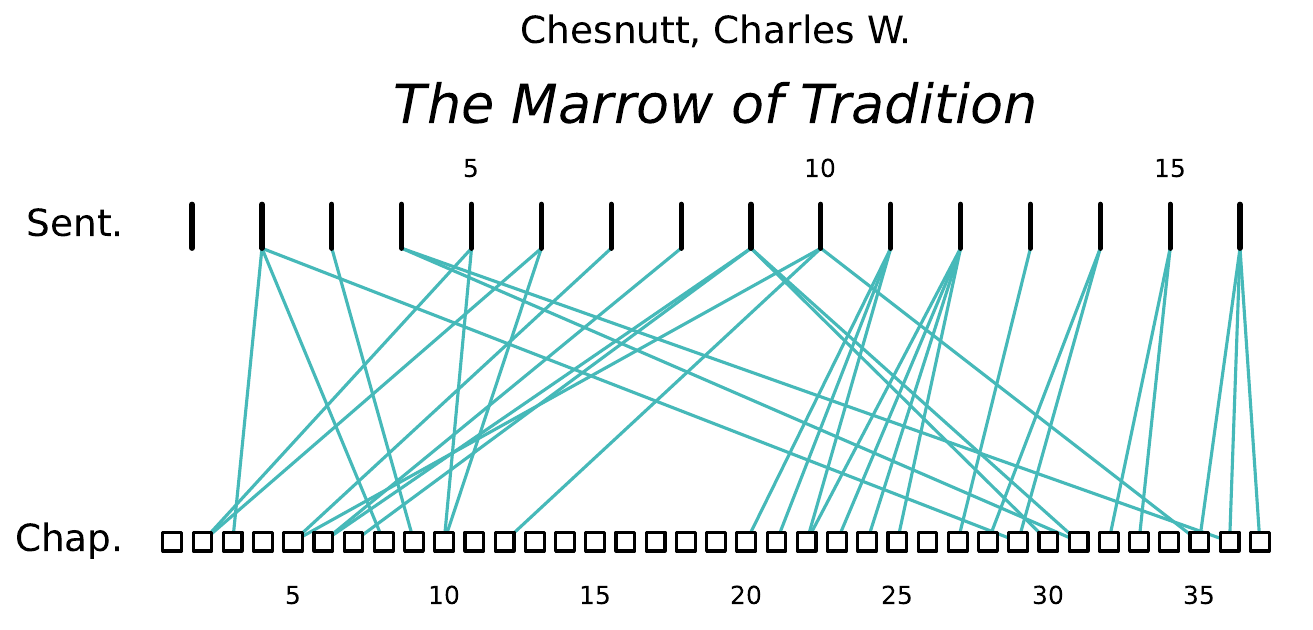}
    {\captionsetup{hypcap=false, labelformat=simple, labelsep=none, textformat=empty}\captionof{figure}{\label{fig:pg11228-bipartite}}}
  \end{minipage}\bigskip

\noindent
  \begin{minipage}[t]{0.48\linewidth}\centering
    \includegraphics[width=\linewidth]{images/pg69087_bipartite.pdf}
    {\captionsetup{hypcap=false, labelformat=simple, labelsep=none, textformat=empty}\captionof{figure}{\label{fig:pg69087-bipartite}}}
  \end{minipage}\hfill
  \begin{minipage}[t]{0.48\linewidth}\centering
    \includegraphics[width=\linewidth]{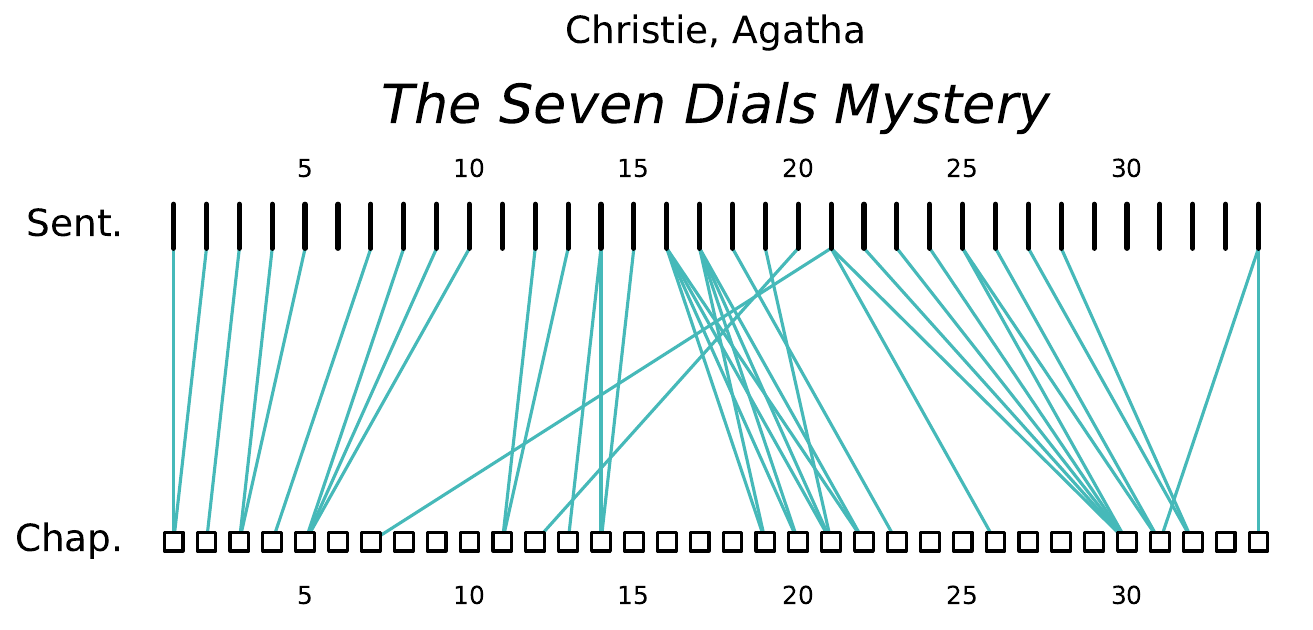}
    {\captionsetup{hypcap=false, labelformat=simple, labelsep=none, textformat=empty}\captionof{figure}{\label{fig:pg75288-bipartite}}}
  \end{minipage}\bigskip

\noindent
  \begin{minipage}[t]{0.48\linewidth}\centering
    \includegraphics[width=\linewidth]{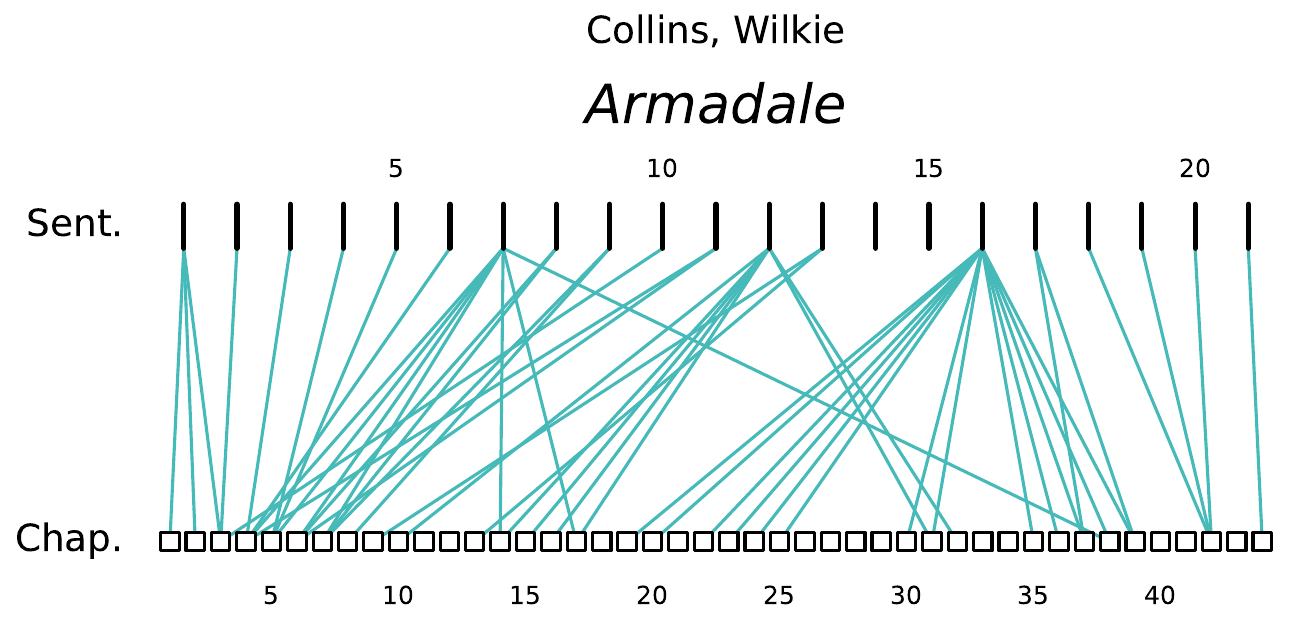}
    {\captionsetup{hypcap=false, labelformat=simple, labelsep=none, textformat=empty}\captionof{figure}{\label{fig:pg1895-bipartite}}}
  \end{minipage}\hfill
  \begin{minipage}[t]{0.48\linewidth}\centering
    \includegraphics[width=\linewidth]{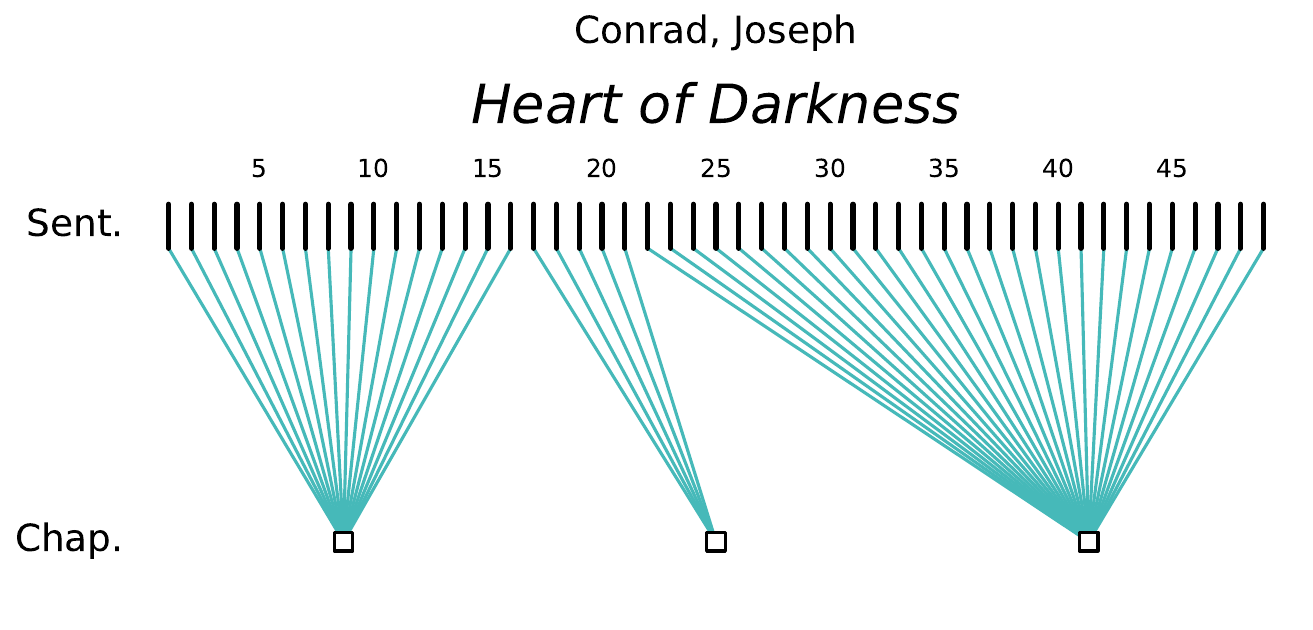}
    {\captionsetup{hypcap=false, labelformat=simple, labelsep=none, textformat=empty}\captionof{figure}{\label{fig:pg219-bipartite}}}
  \end{minipage}\bigskip

\noindent
  \begin{minipage}[t]{0.48\linewidth}\centering
    \includegraphics[width=\linewidth]{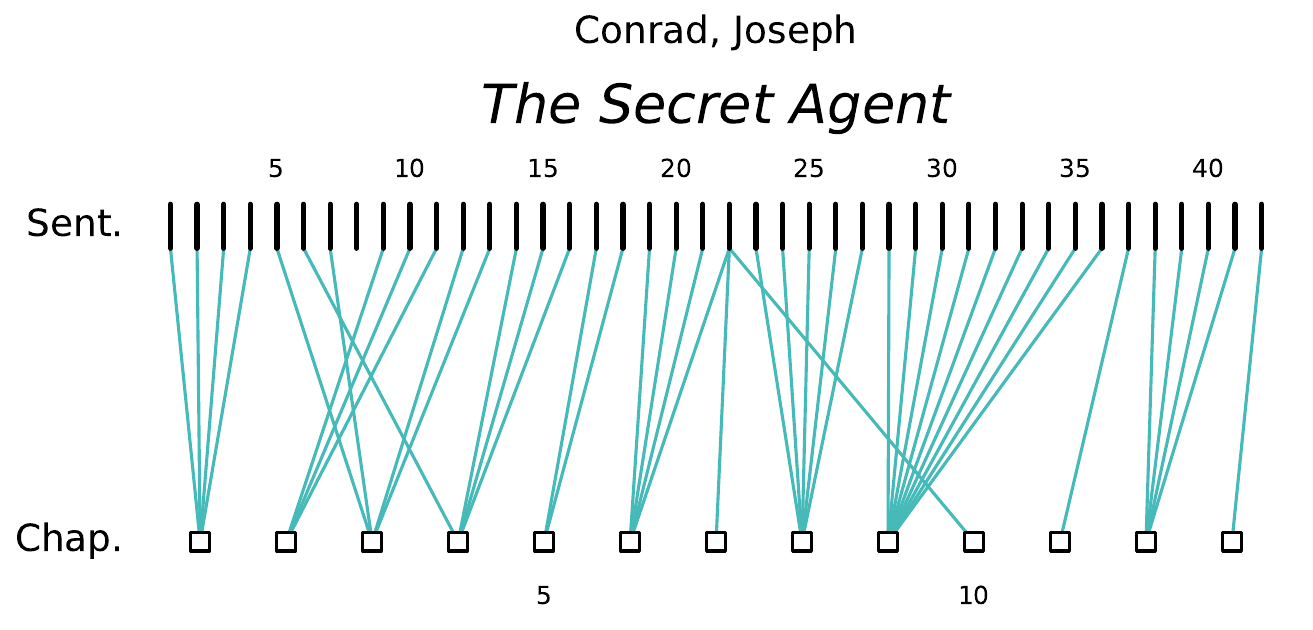}
    {\captionsetup{hypcap=false, labelformat=simple, labelsep=none, textformat=empty}\captionof{figure}{\label{fig:pg974-bipartite}}}
  \end{minipage}\hfill
  \begin{minipage}[t]{0.48\linewidth}\centering
    \includegraphics[width=\linewidth]{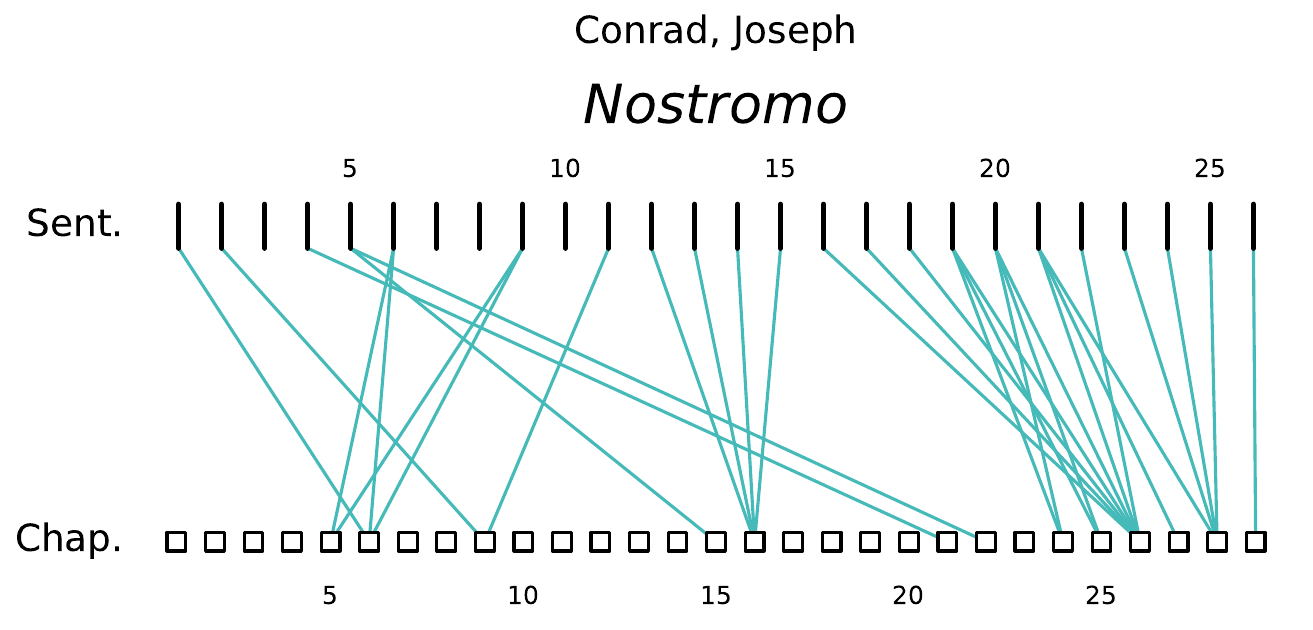}
    {\captionsetup{hypcap=false, labelformat=simple, labelsep=none, textformat=empty}\captionof{figure}{\label{fig:pg2021-bipartite}}}
  \end{minipage}\bigskip

\noindent
  \begin{minipage}[t]{0.48\linewidth}\centering
    \includegraphics[width=\linewidth]{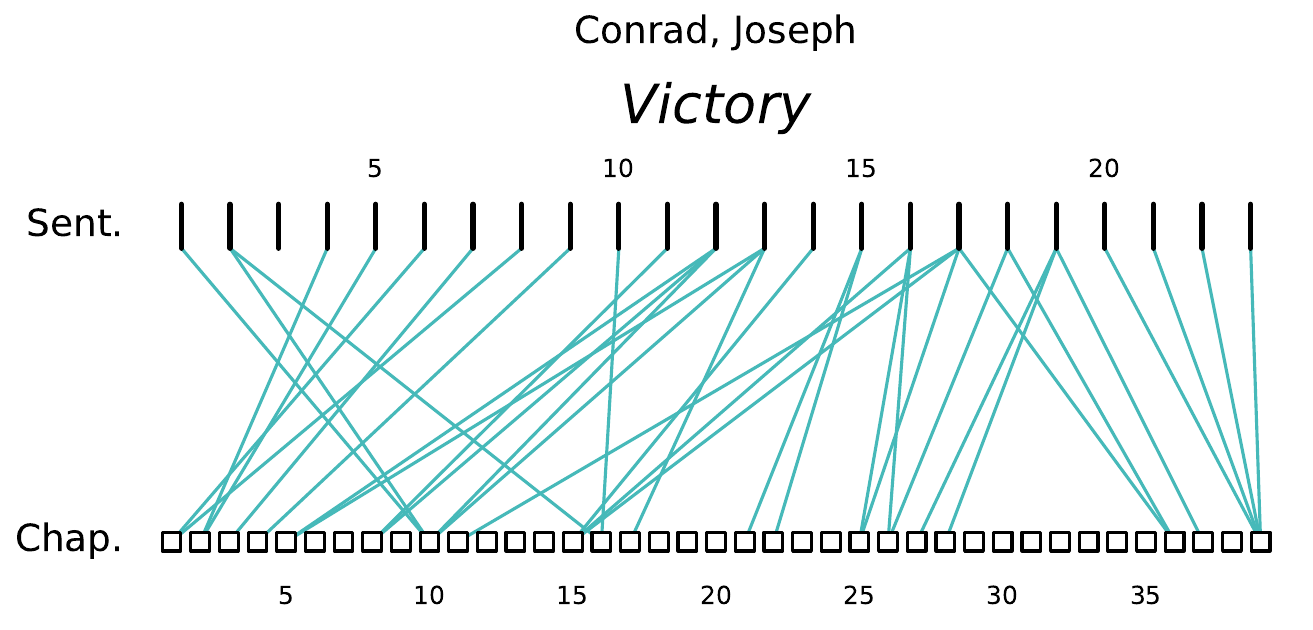}
    {\captionsetup{hypcap=false, labelformat=simple, labelsep=none, textformat=empty}\captionof{figure}{\label{fig:pg6378-bipartite}}}
  \end{minipage}\hfill
  \begin{minipage}[t]{0.48\linewidth}\centering
    \includegraphics[width=\linewidth]{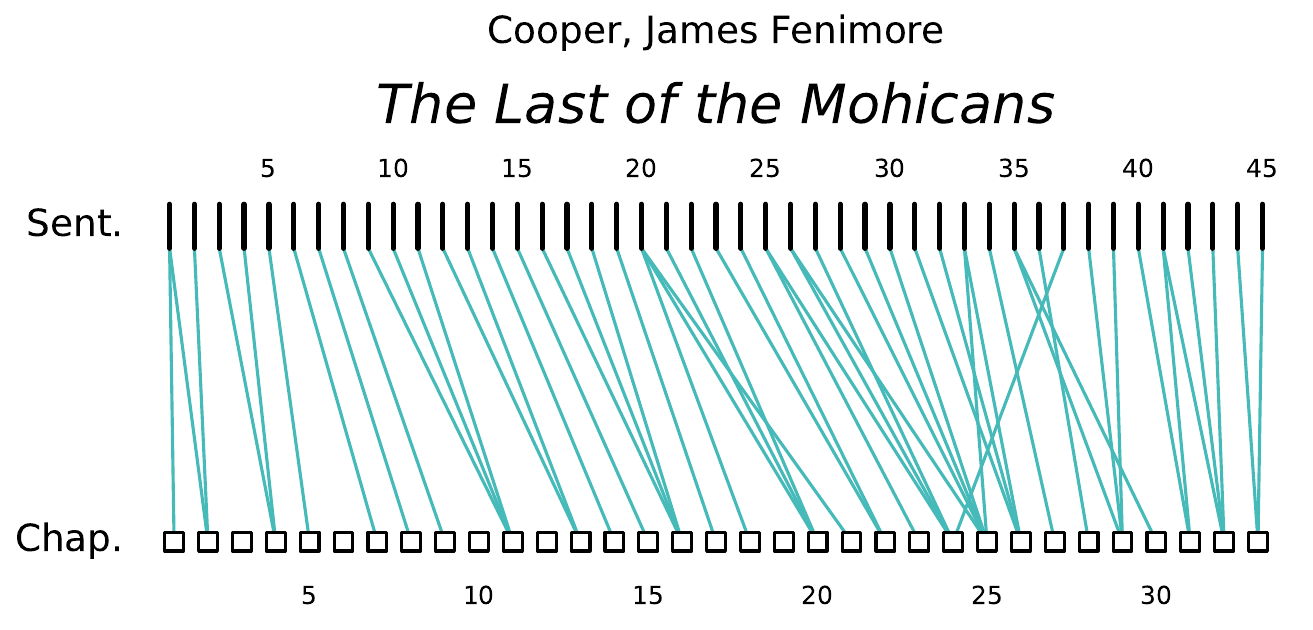}
    {\captionsetup{hypcap=false, labelformat=simple, labelsep=none, textformat=empty}\captionof{figure}{\label{fig:pg940-bipartite}}}
  \end{minipage}\bigskip

\noindent
  \begin{minipage}[t]{0.48\linewidth}\centering
    \includegraphics[width=\linewidth]{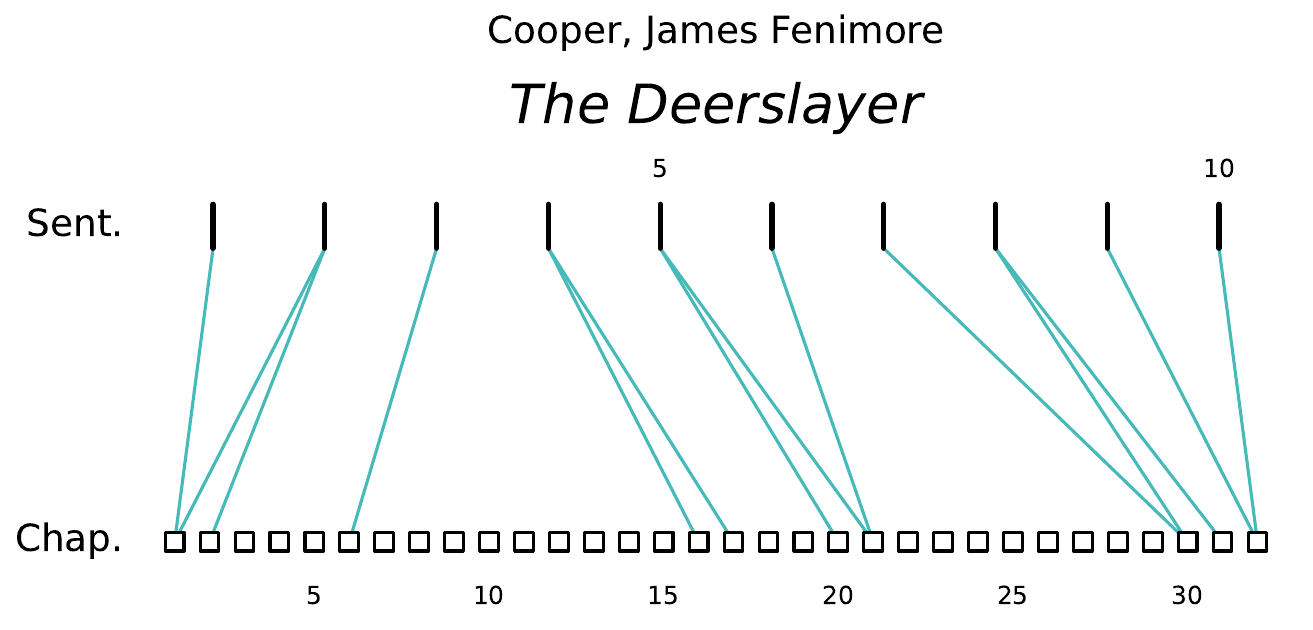}
    {\captionsetup{hypcap=false, labelformat=simple, labelsep=none, textformat=empty}\captionof{figure}{\label{fig:pg3285-bipartite}}}
  \end{minipage}\hfill
  \begin{minipage}[t]{0.48\linewidth}\centering
    \includegraphics[width=\linewidth]{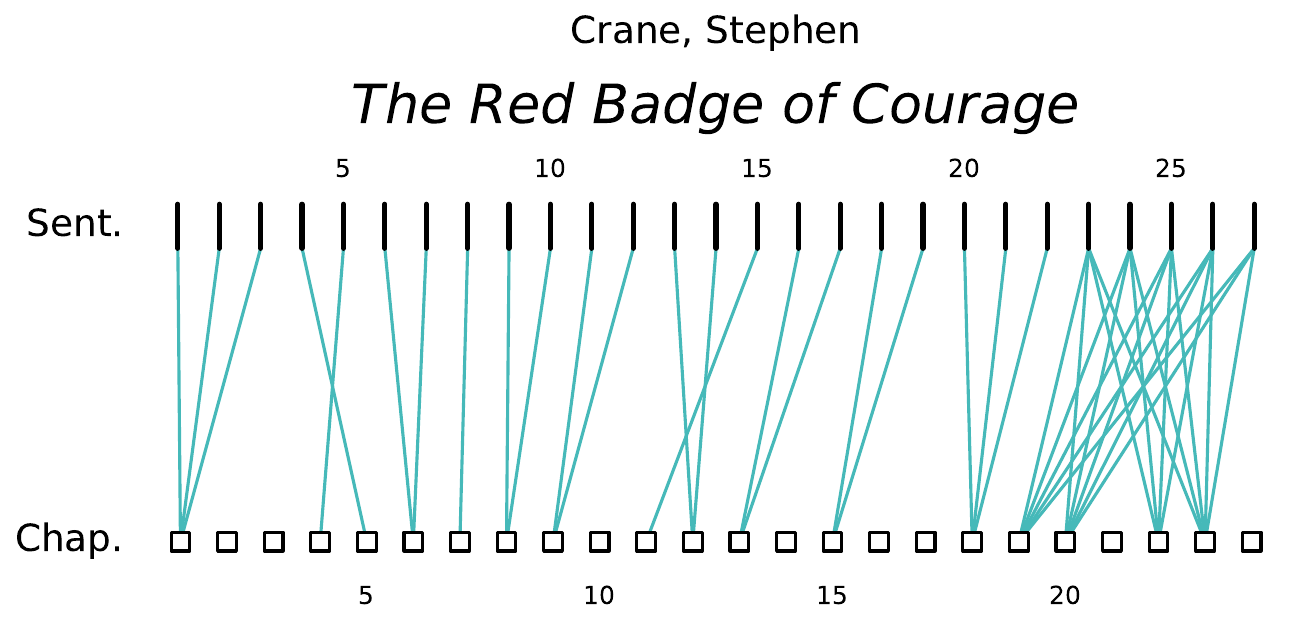}
    {\captionsetup{hypcap=false, labelformat=simple, labelsep=none, textformat=empty}\captionof{figure}{\label{fig:pg73-bipartite}}}
  \end{minipage}\bigskip

\noindent
  \begin{minipage}[t]{0.48\linewidth}\centering
    \includegraphics[width=\linewidth]{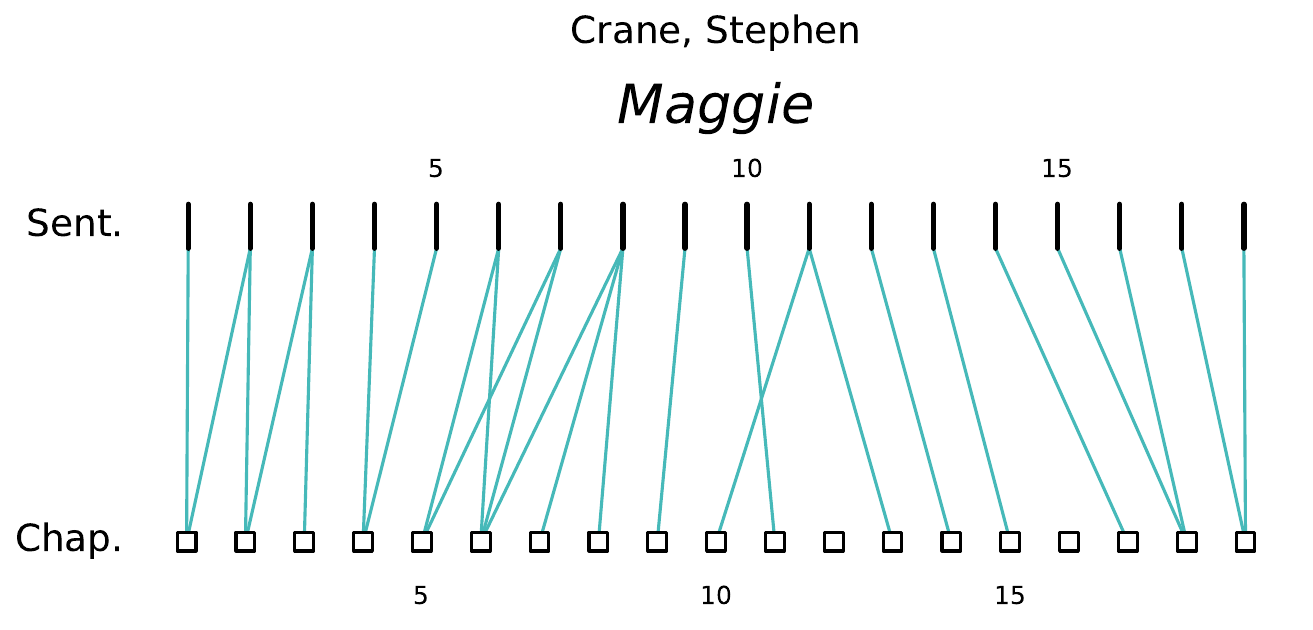}
    {\captionsetup{hypcap=false, labelformat=simple, labelsep=none, textformat=empty}\captionof{figure}{\label{fig:pg447-bipartite}}}
  \end{minipage}\hfill
  \begin{minipage}[t]{0.48\linewidth}\centering
    \includegraphics[width=\linewidth]{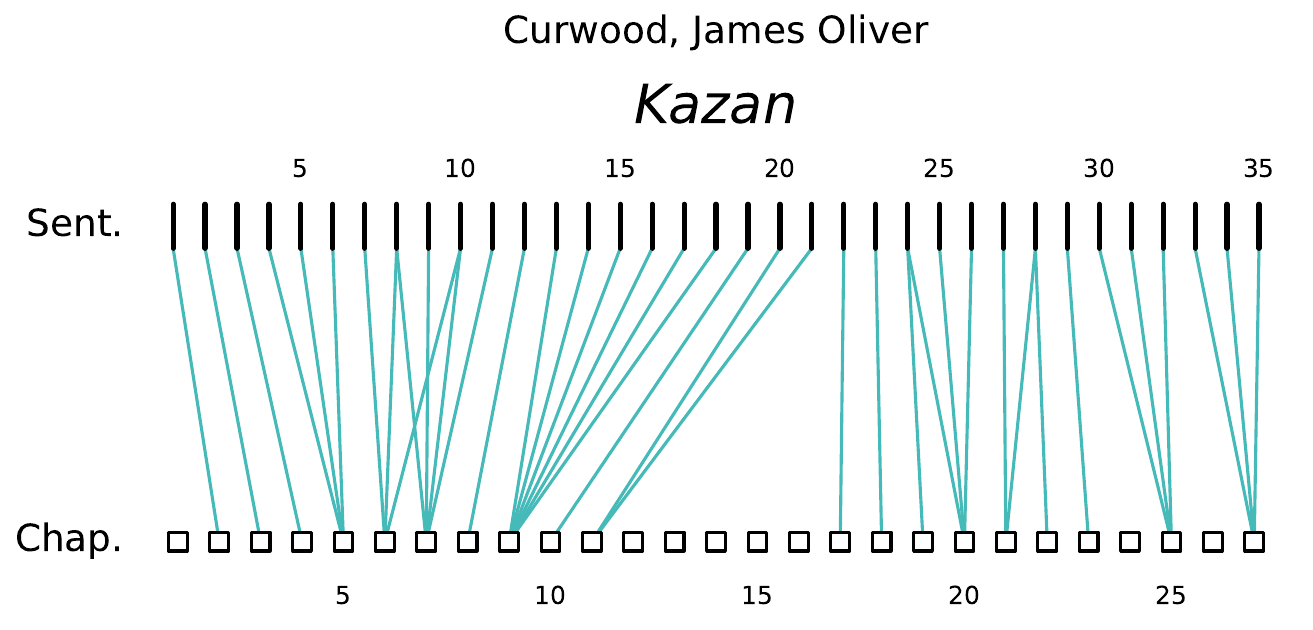}
    {\captionsetup{hypcap=false, labelformat=simple, labelsep=none, textformat=empty}\captionof{figure}{\label{fig:pg10084-bipartite}}}
  \end{minipage}\bigskip

\noindent
  \begin{minipage}[t]{0.48\linewidth}\centering
    \includegraphics[width=\linewidth]{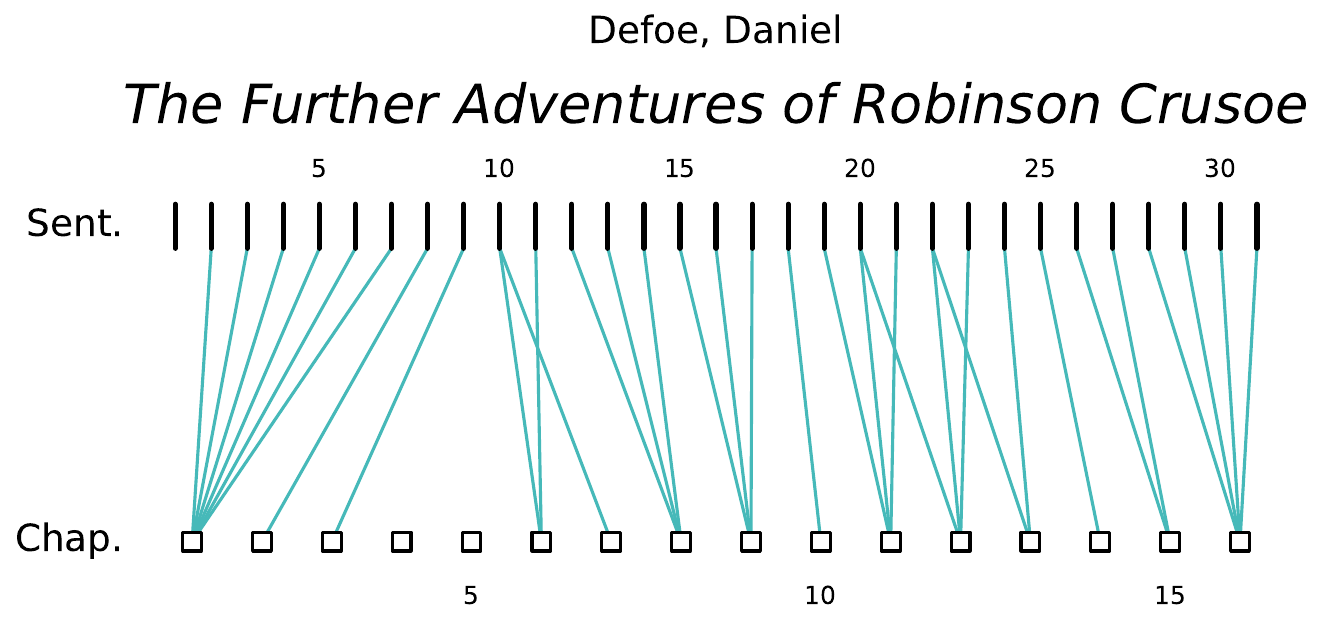}
    {\captionsetup{hypcap=false, labelformat=simple, labelsep=none, textformat=empty}\captionof{figure}{\label{fig:pg561-bipartite}}}
  \end{minipage}\hfill
  \begin{minipage}[t]{0.48\linewidth}\centering
    \includegraphics[width=\linewidth]{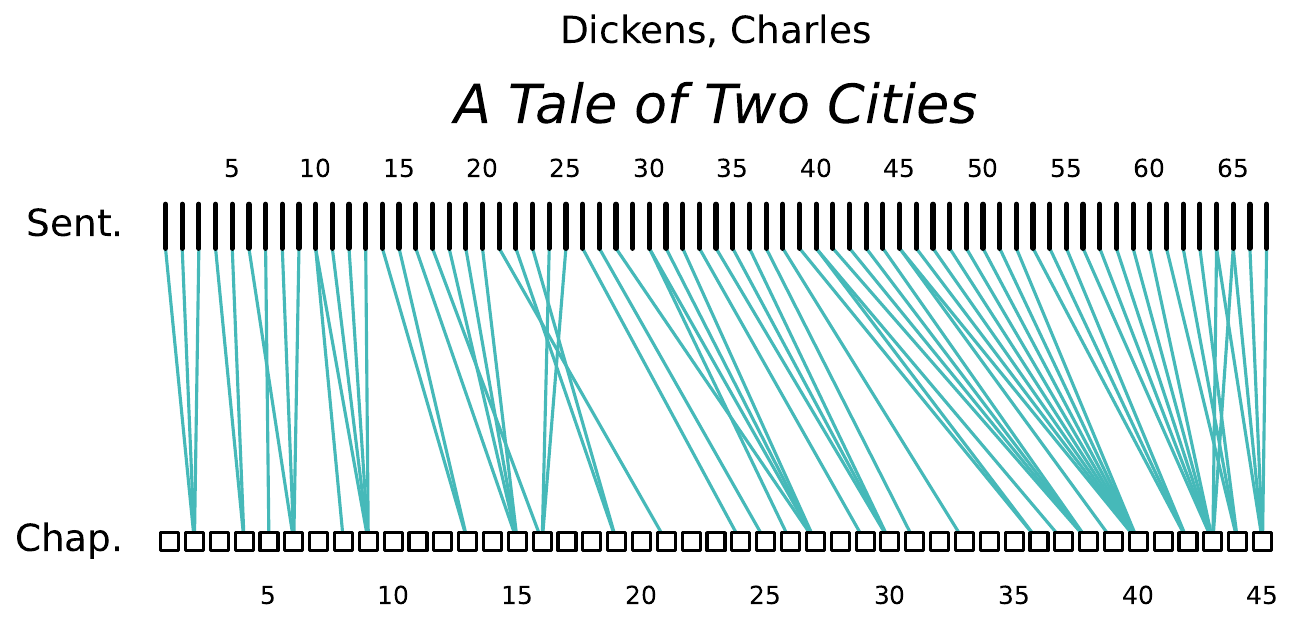}
    {\captionsetup{hypcap=false, labelformat=simple, labelsep=none, textformat=empty}\captionof{figure}{\label{fig:pg98-bipartite}}}
  \end{minipage}\bigskip

\noindent
  \begin{minipage}[t]{0.48\linewidth}\centering
    \includegraphics[width=\linewidth]{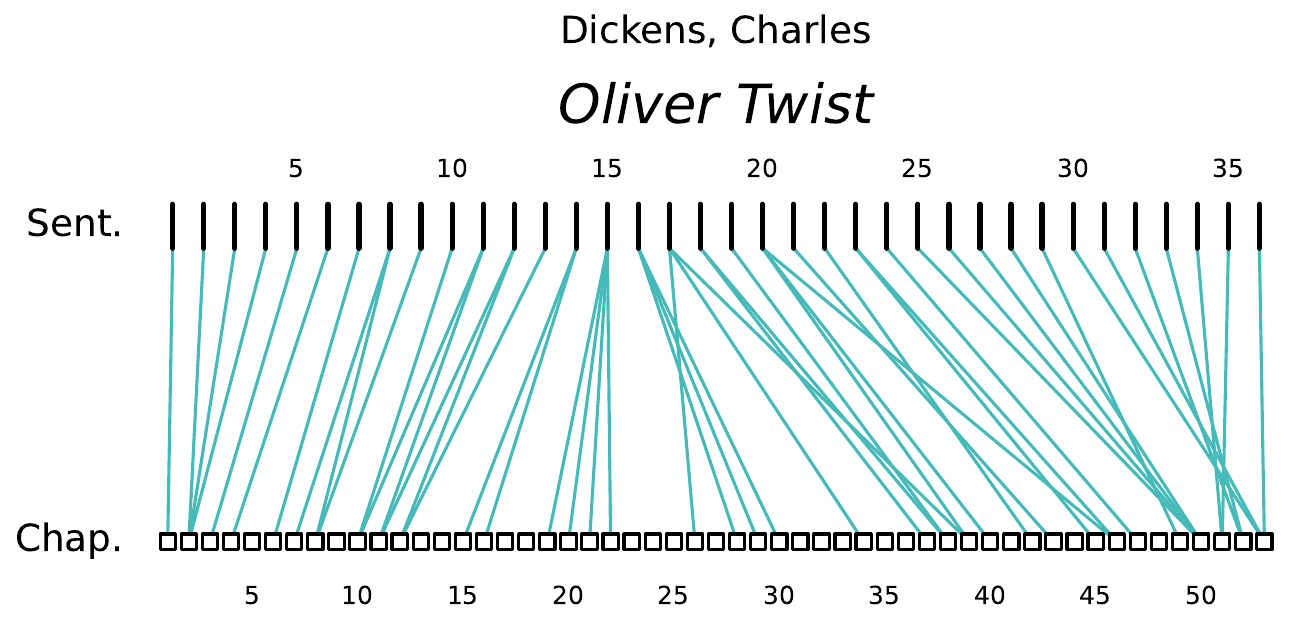}
    {\captionsetup{hypcap=false, labelformat=simple, labelsep=none, textformat=empty}\captionof{figure}{\label{fig:pg730-bipartite}}}
  \end{minipage}\hfill
  \begin{minipage}[t]{0.48\linewidth}\centering
    \includegraphics[width=\linewidth]{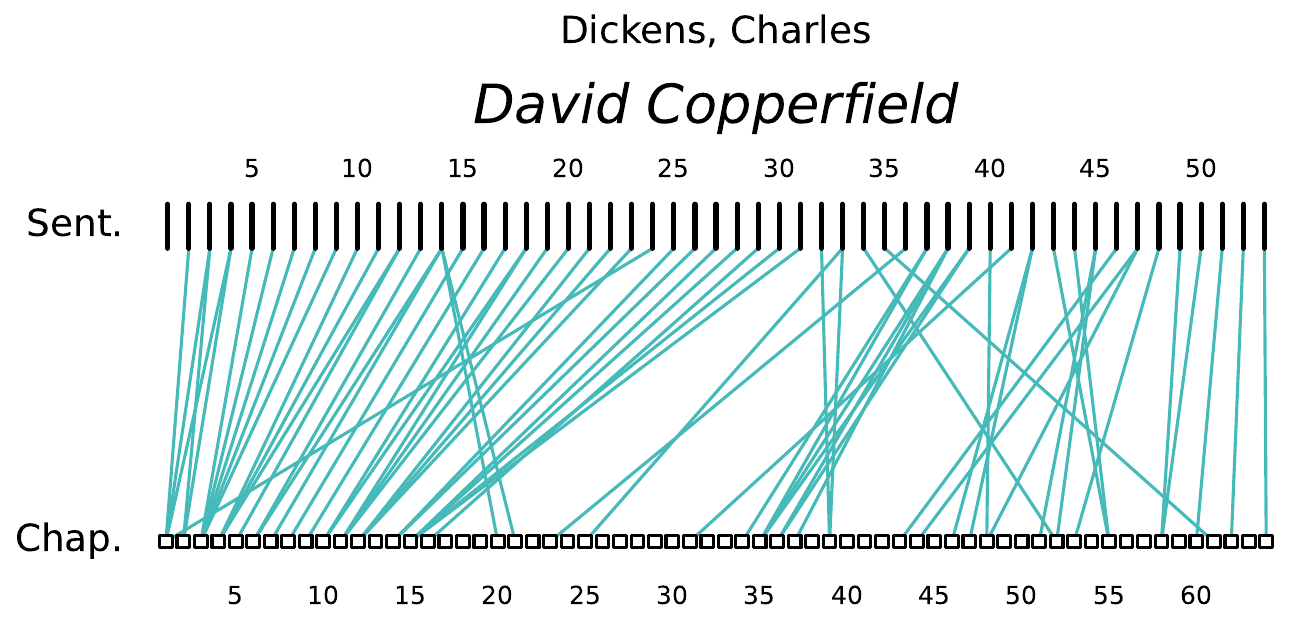}
    {\captionsetup{hypcap=false, labelformat=simple, labelsep=none, textformat=empty}\captionof{figure}{\label{fig:pg766-bipartite}}}
  \end{minipage}\bigskip

\noindent
  \begin{minipage}[t]{0.48\linewidth}\centering
    \includegraphics[width=\linewidth]{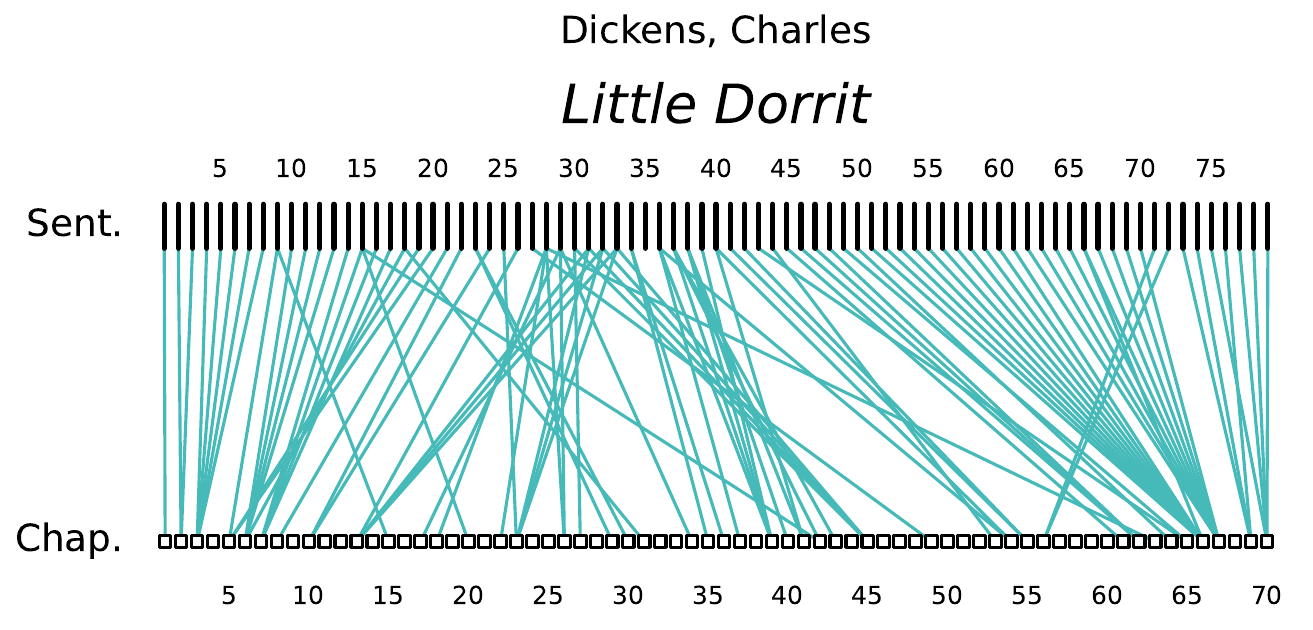}
    {\captionsetup{hypcap=false, labelformat=simple, labelsep=none, textformat=empty}\captionof{figure}{\label{fig:pg963-bipartite}}}
  \end{minipage}\hfill
  \begin{minipage}[t]{0.48\linewidth}\centering
    \includegraphics[width=\linewidth]{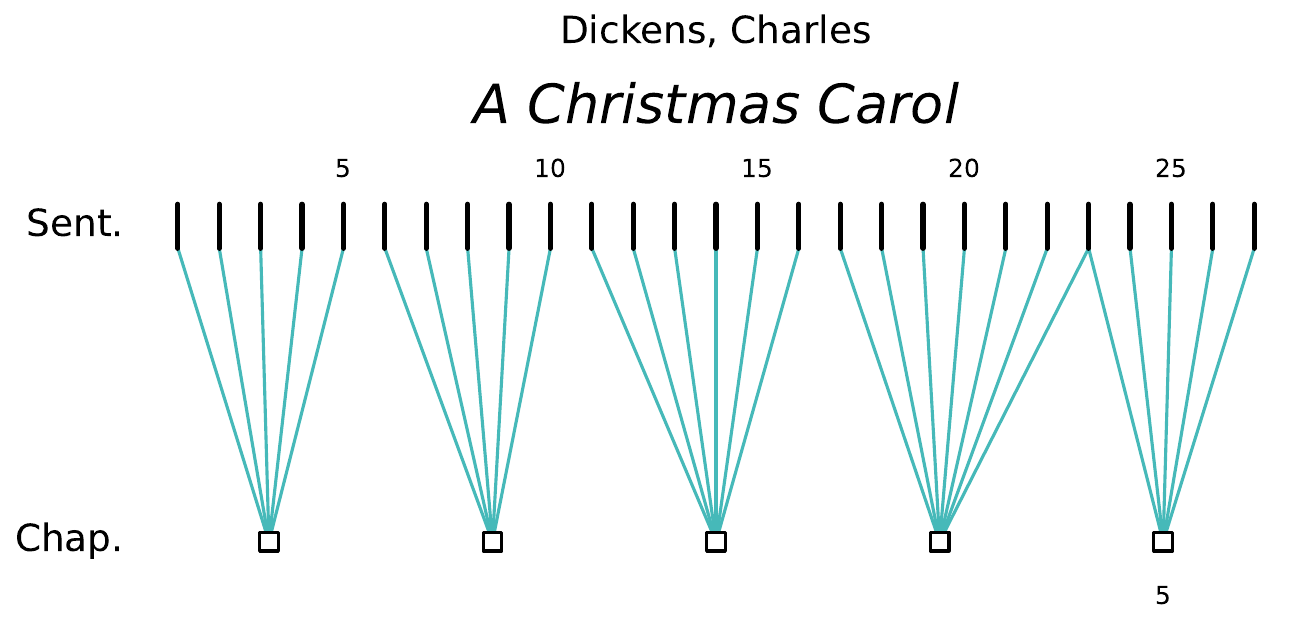}
    {\captionsetup{hypcap=false, labelformat=simple, labelsep=none, textformat=empty}\captionof{figure}{\label{fig:pg19337-bipartite}}}
  \end{minipage}\bigskip

\noindent
  \begin{minipage}[t]{0.48\linewidth}\centering
    \includegraphics[width=\linewidth]{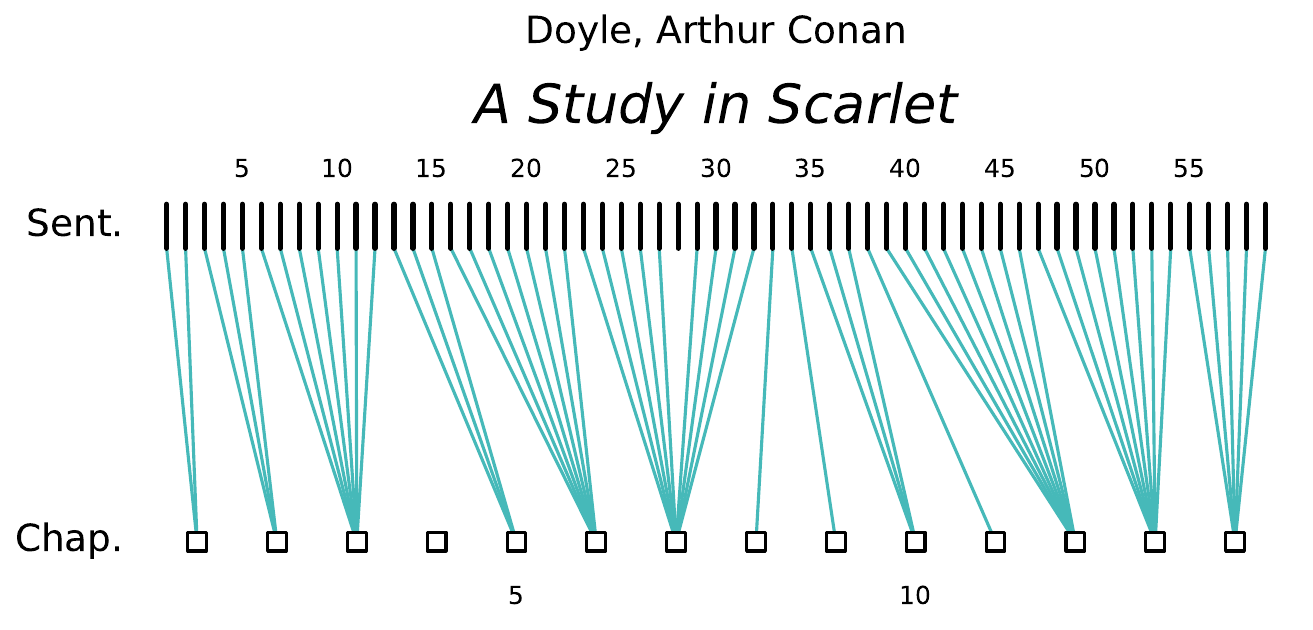}
    {\captionsetup{hypcap=false, labelformat=simple, labelsep=none, textformat=empty}\captionof{figure}{\label{fig:pg244-bipartite}}}
  \end{minipage}\hfill
  \begin{minipage}[t]{0.48\linewidth}\centering
    \includegraphics[width=\linewidth]{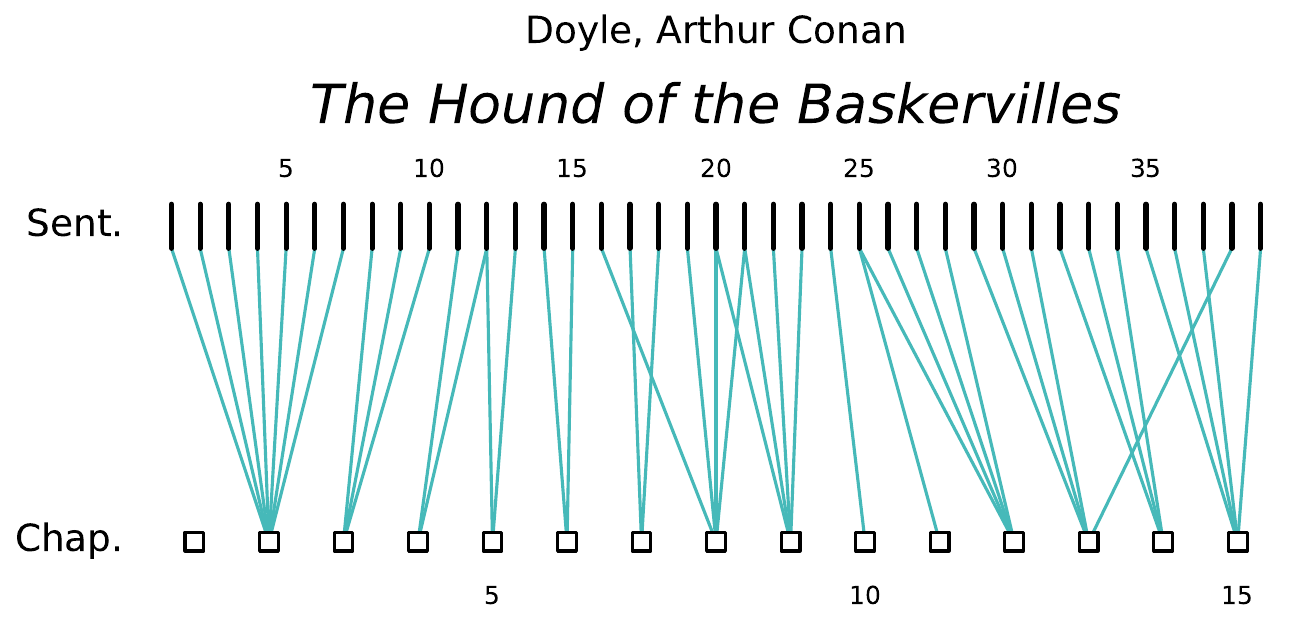}
    {\captionsetup{hypcap=false, labelformat=simple, labelsep=none, textformat=empty}\captionof{figure}{\label{fig:pg2852-bipartite}}}
  \end{minipage}\bigskip

\noindent
  \begin{minipage}[t]{0.48\linewidth}\centering
    \includegraphics[width=\linewidth]{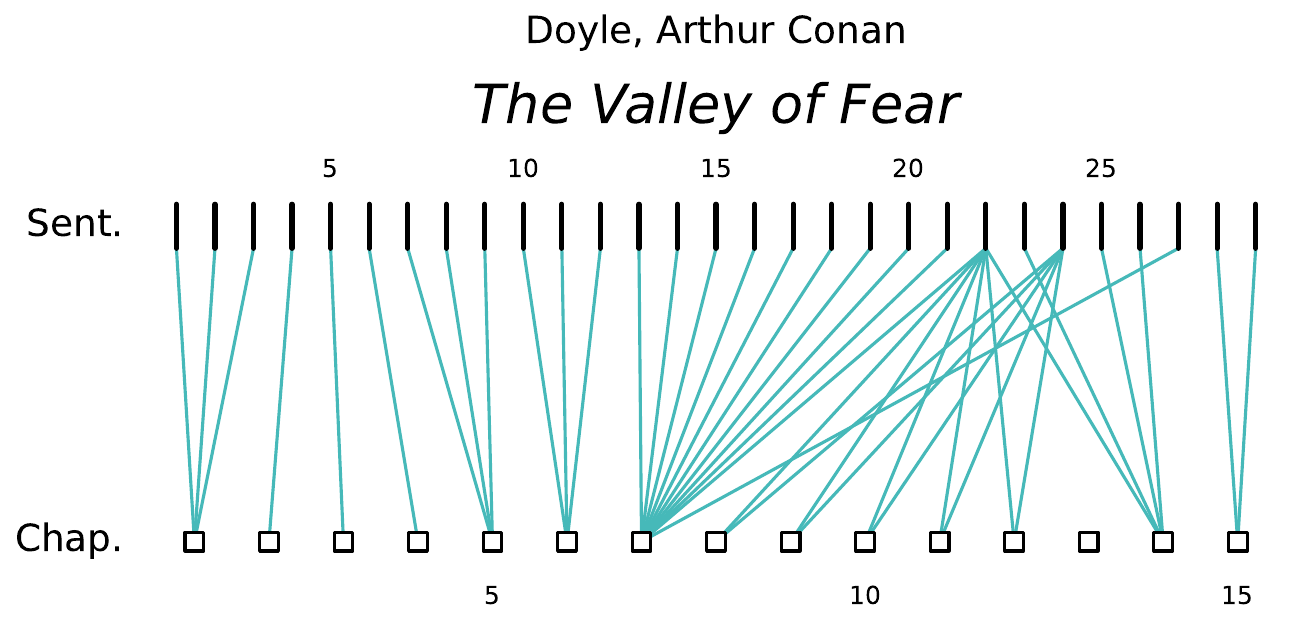}
    {\captionsetup{hypcap=false, labelformat=simple, labelsep=none, textformat=empty}\captionof{figure}{\label{fig:pg3289-bipartite}}}
  \end{minipage}\hfill
  \begin{minipage}[t]{0.48\linewidth}\centering
    \includegraphics[width=\linewidth]{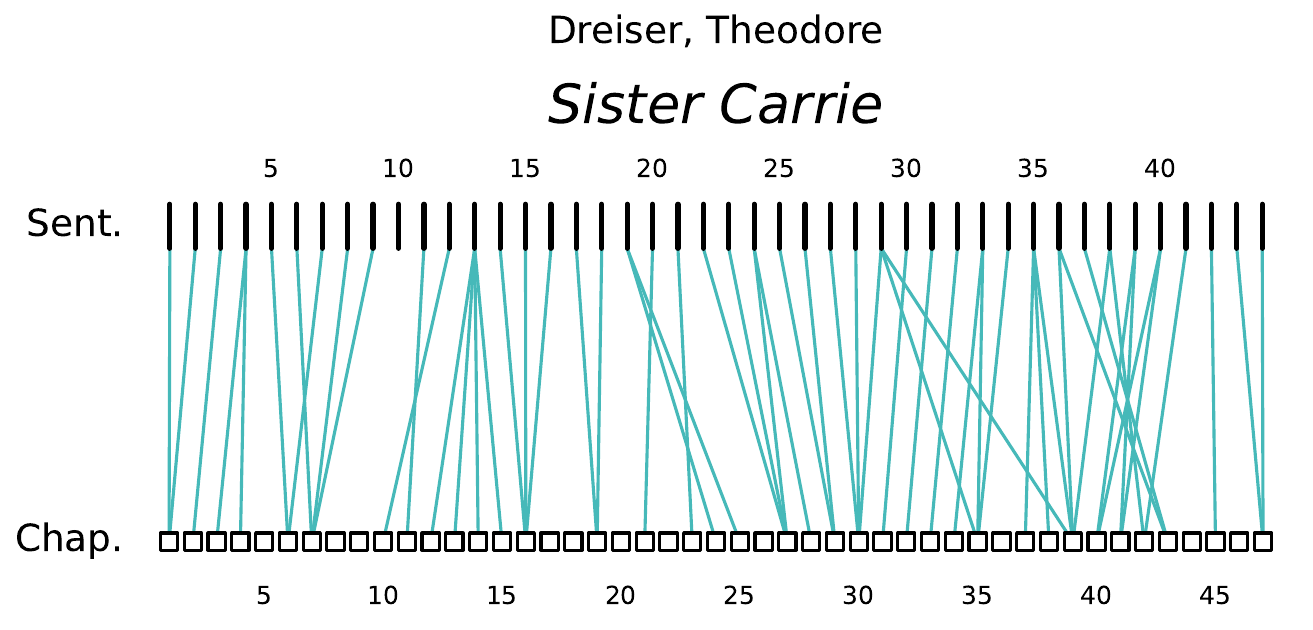}
    {\captionsetup{hypcap=false, labelformat=simple, labelsep=none, textformat=empty}\captionof{figure}{\label{fig:pg233-bipartite}}}
  \end{minipage}\bigskip

\noindent
  \begin{minipage}[t]{0.48\linewidth}\centering
    \includegraphics[width=\linewidth]{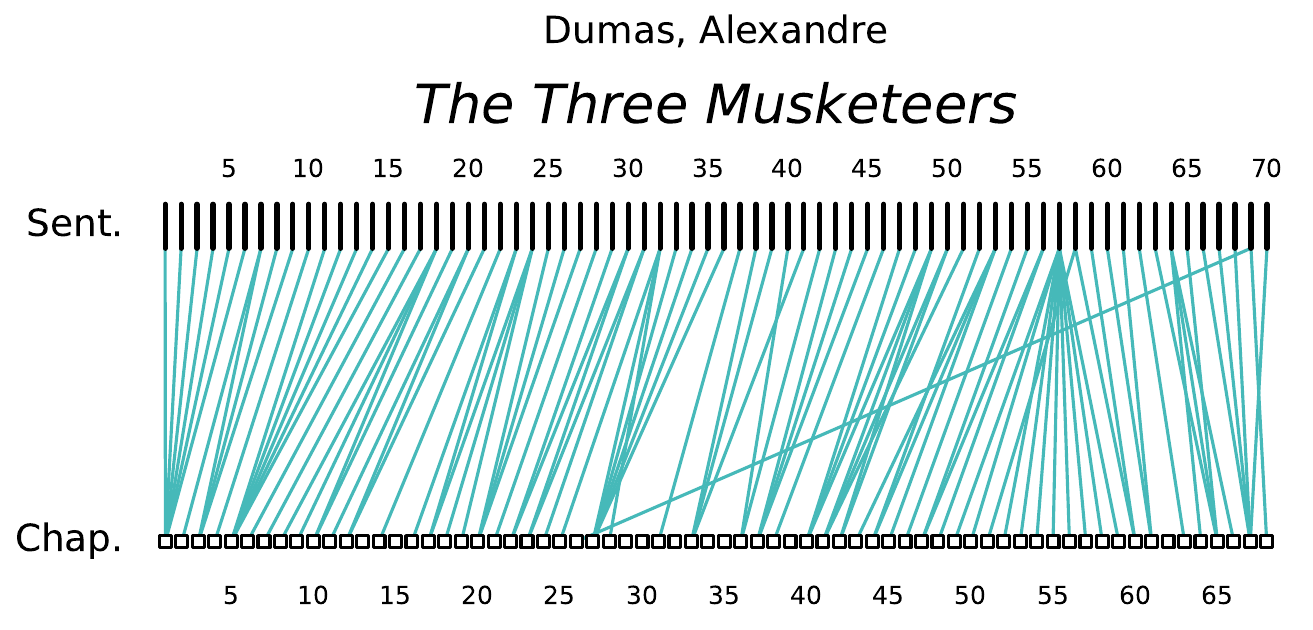}
    {\captionsetup{hypcap=false, labelformat=simple, labelsep=none, textformat=empty}\captionof{figure}{\label{fig:pg1257-bipartite}}}
  \end{minipage}\hfill
  \begin{minipage}[t]{0.48\linewidth}\centering
    \includegraphics[width=\linewidth]{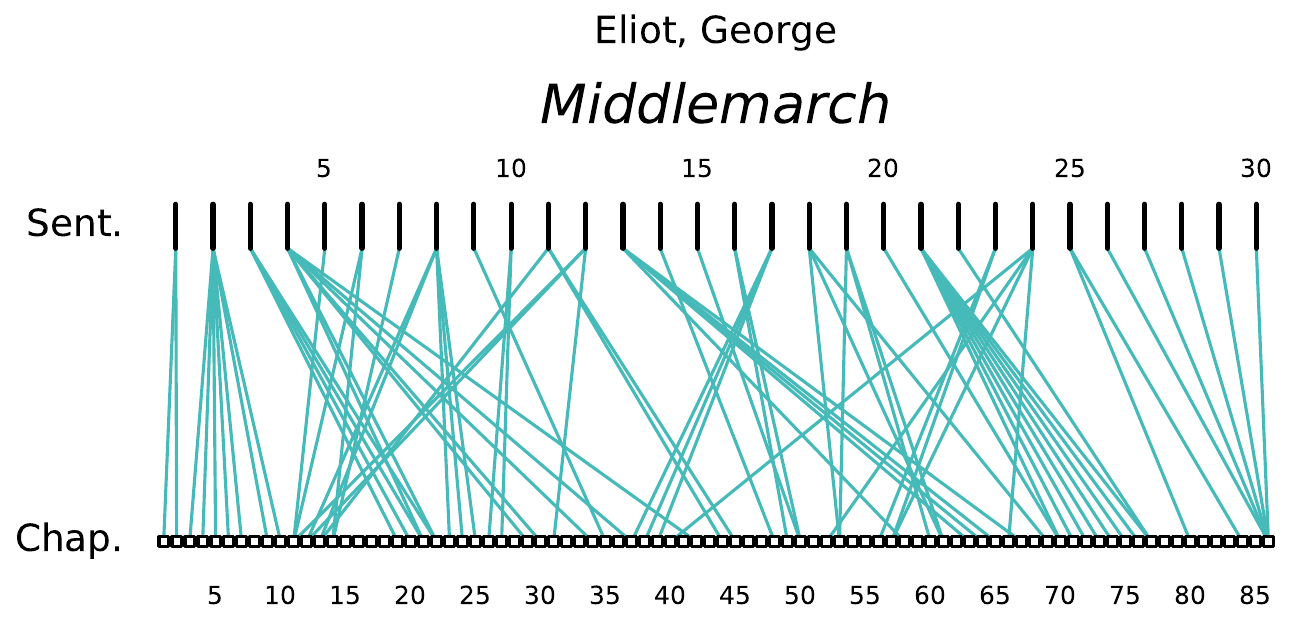}
    {\captionsetup{hypcap=false, labelformat=simple, labelsep=none, textformat=empty}\captionof{figure}{\label{fig:pg145-bipartite}}}
  \end{minipage}\bigskip

\noindent
  \begin{minipage}[t]{0.48\linewidth}\centering
    \includegraphics[width=\linewidth]{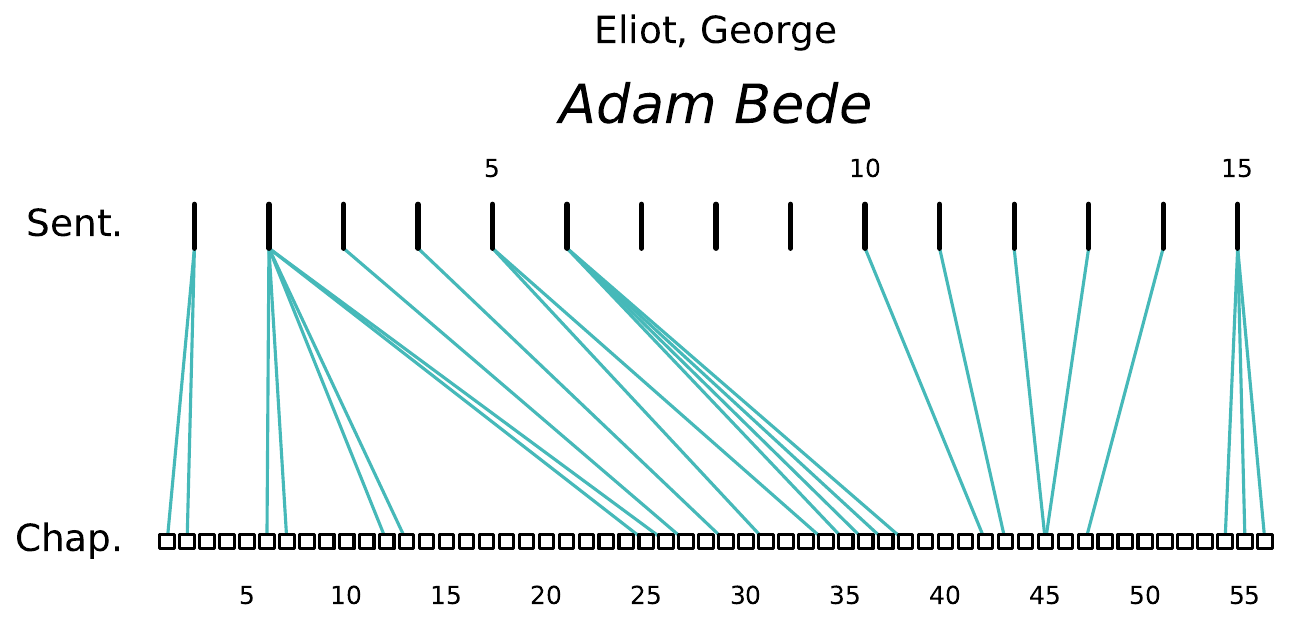}
    {\captionsetup{hypcap=false, labelformat=simple, labelsep=none, textformat=empty}\captionof{figure}{\label{fig:pg507-bipartite}}}
  \end{minipage}\hfill
  \begin{minipage}[t]{0.48\linewidth}\centering
    \includegraphics[width=\linewidth]{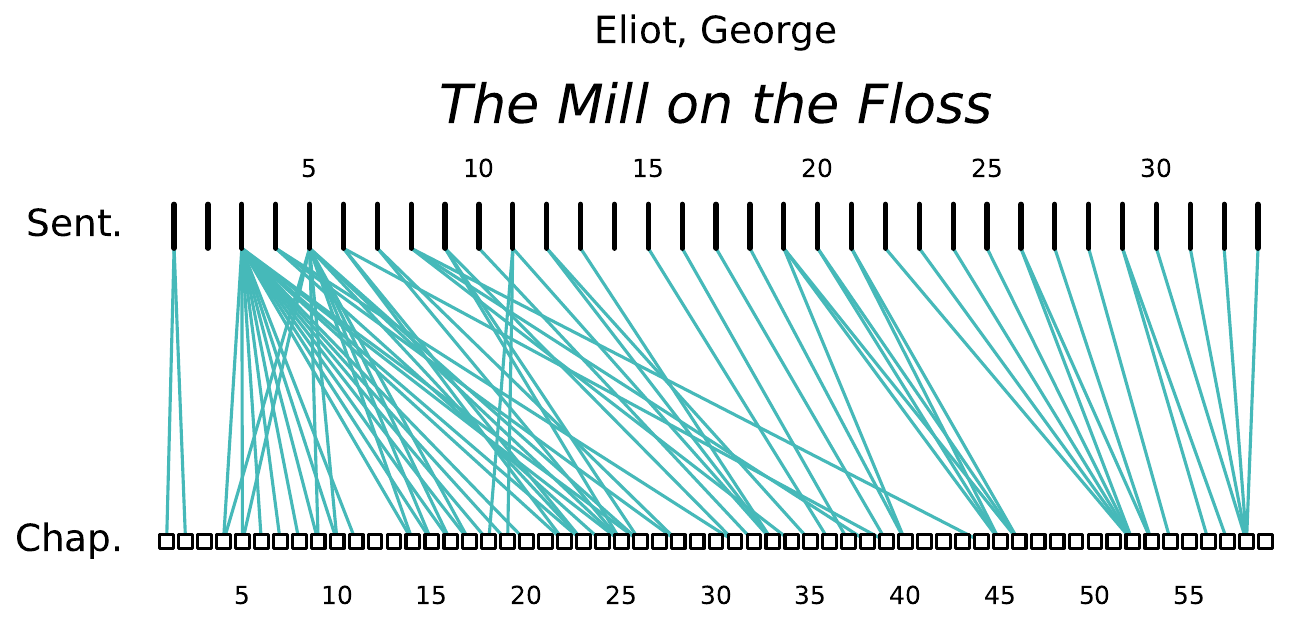}
    {\captionsetup{hypcap=false, labelformat=simple, labelsep=none, textformat=empty}\captionof{figure}{\label{fig:pg6688-bipartite}}}
  \end{minipage}\bigskip

\noindent
  \begin{minipage}[t]{0.48\linewidth}\centering
    \includegraphics[width=\linewidth]{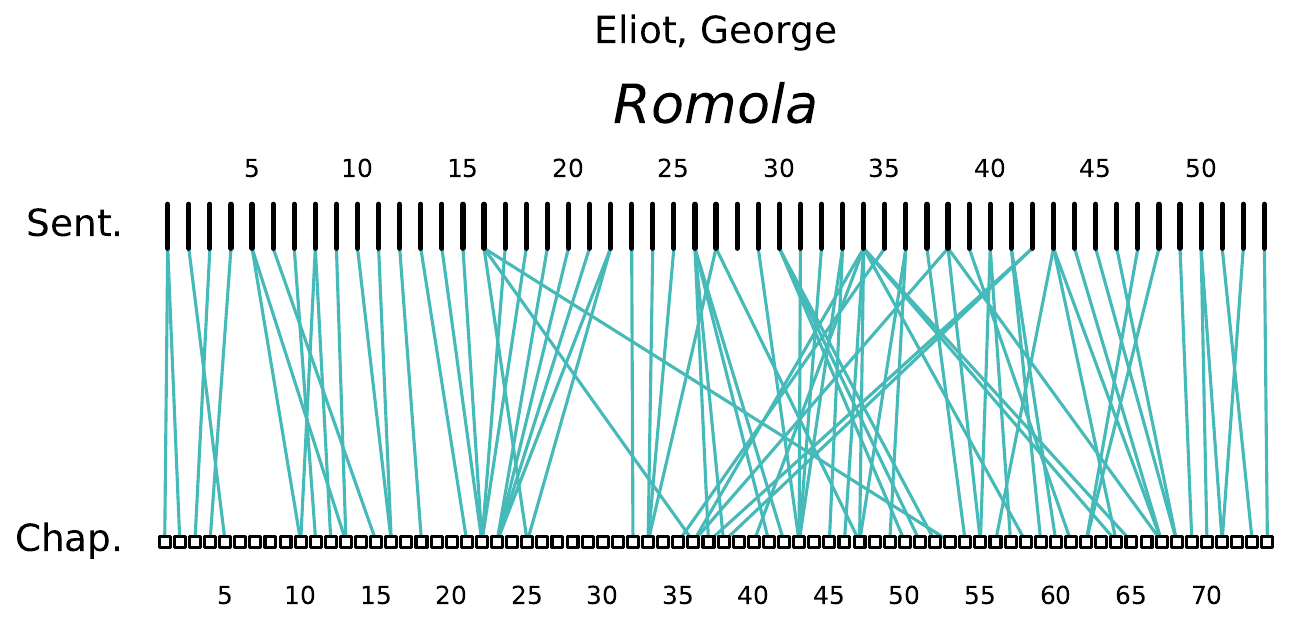}
    {\captionsetup{hypcap=false, labelformat=simple, labelsep=none, textformat=empty}\captionof{figure}{\label{fig:pg24020-bipartite}}}
  \end{minipage}\hfill
  \begin{minipage}[t]{0.48\linewidth}\centering
    \includegraphics[width=\linewidth]{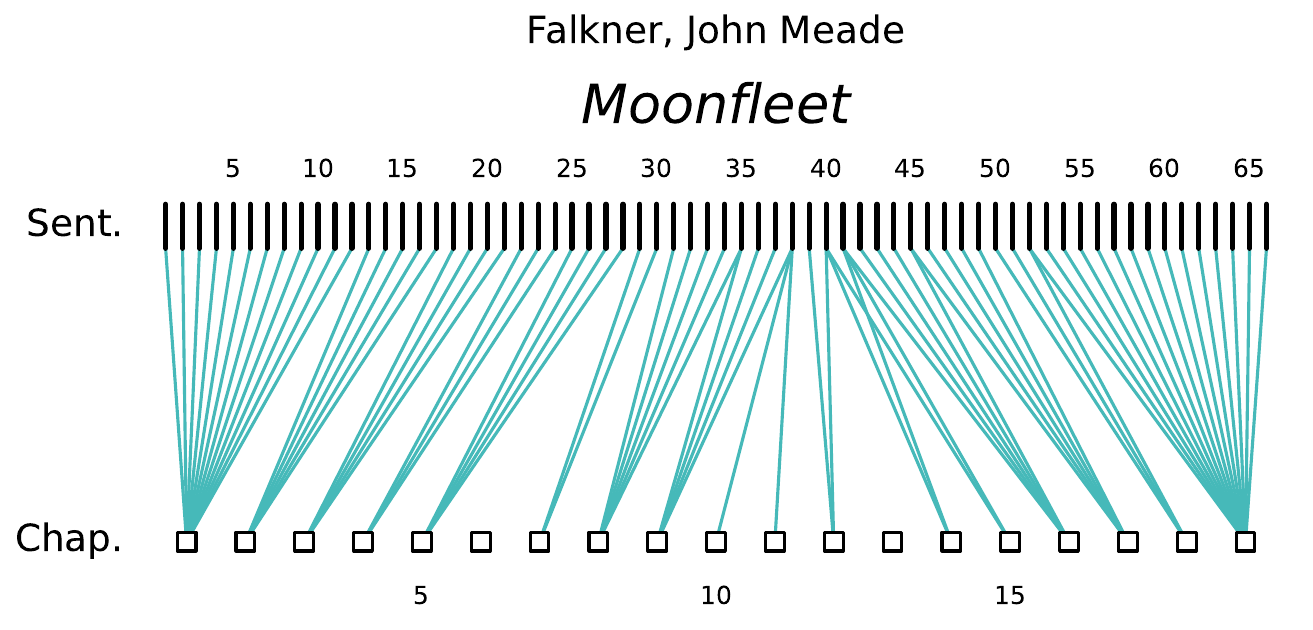}
    {\captionsetup{hypcap=false, labelformat=simple, labelsep=none, textformat=empty}\captionof{figure}{\label{fig:pg10743-bipartite}}}
  \end{minipage}\bigskip

\noindent
  \begin{minipage}[t]{0.48\linewidth}\centering
    \includegraphics[width=\linewidth]{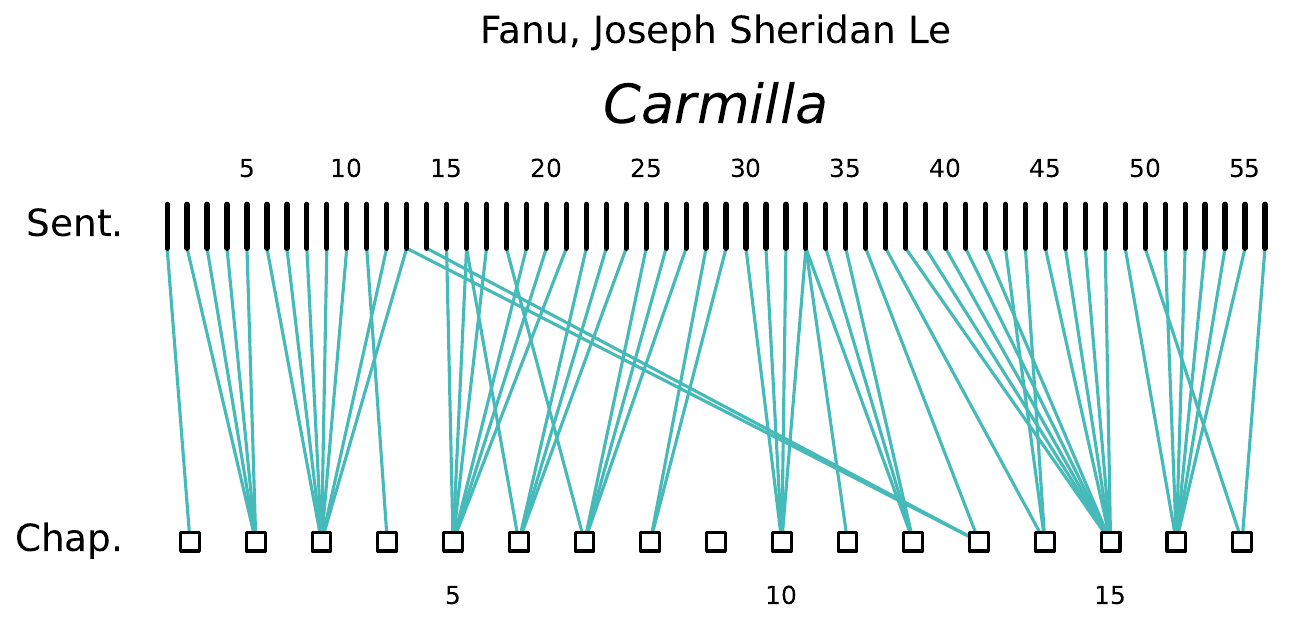}
    {\captionsetup{hypcap=false, labelformat=simple, labelsep=none, textformat=empty}\captionof{figure}{\label{fig:pg10007-bipartite}}}
  \end{minipage}\hfill
  \begin{minipage}[t]{0.48\linewidth}\centering
    \includegraphics[width=\linewidth]{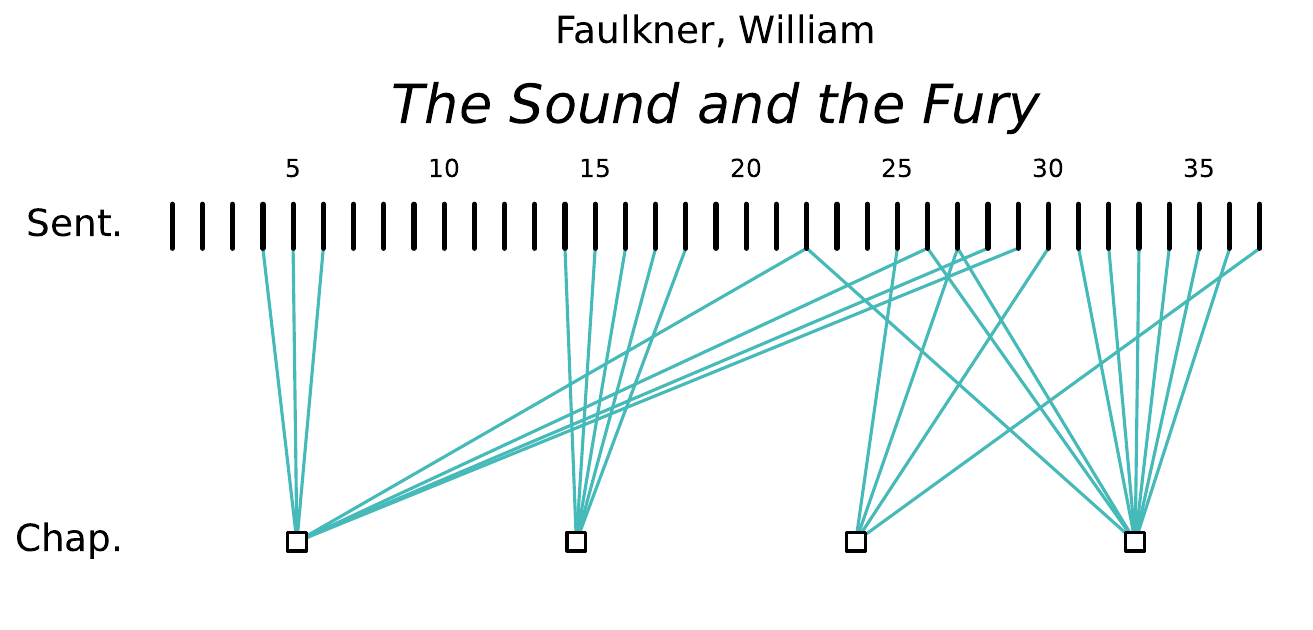}
    {\captionsetup{hypcap=false, labelformat=simple, labelsep=none, textformat=empty}\captionof{figure}{\label{fig:pg75170-bipartite}}}
  \end{minipage}\bigskip

\noindent
  \begin{minipage}[t]{0.48\linewidth}\centering
    \includegraphics[width=\linewidth]{images/pg805_bipartite.pdf}
    {\captionsetup{hypcap=false, labelformat=simple, labelsep=none, textformat=empty}\captionof{figure}{\label{fig:pg805-bipartite}}}
  \end{minipage}\hfill
  \begin{minipage}[t]{0.48\linewidth}\centering
    \includegraphics[width=\linewidth]{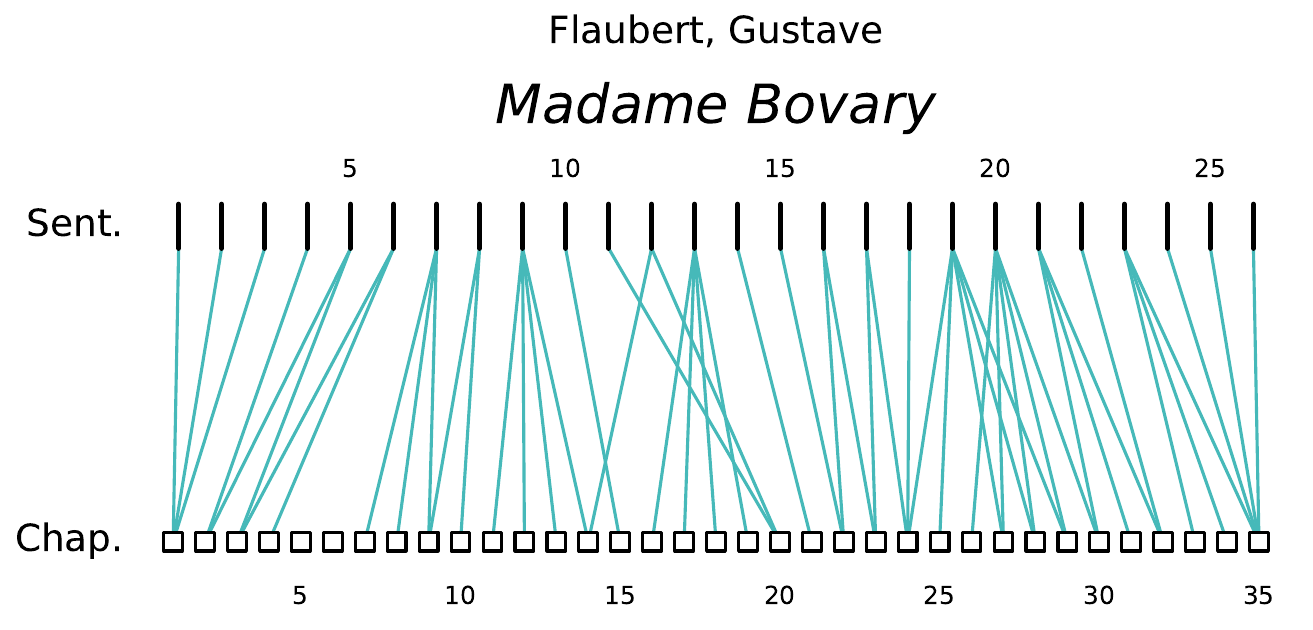}
    {\captionsetup{hypcap=false, labelformat=simple, labelsep=none, textformat=empty}\captionof{figure}{\label{fig:pg2413-bipartite}}}
  \end{minipage}\bigskip

\noindent
  \begin{minipage}[t]{0.48\linewidth}\centering
    \includegraphics[width=\linewidth]{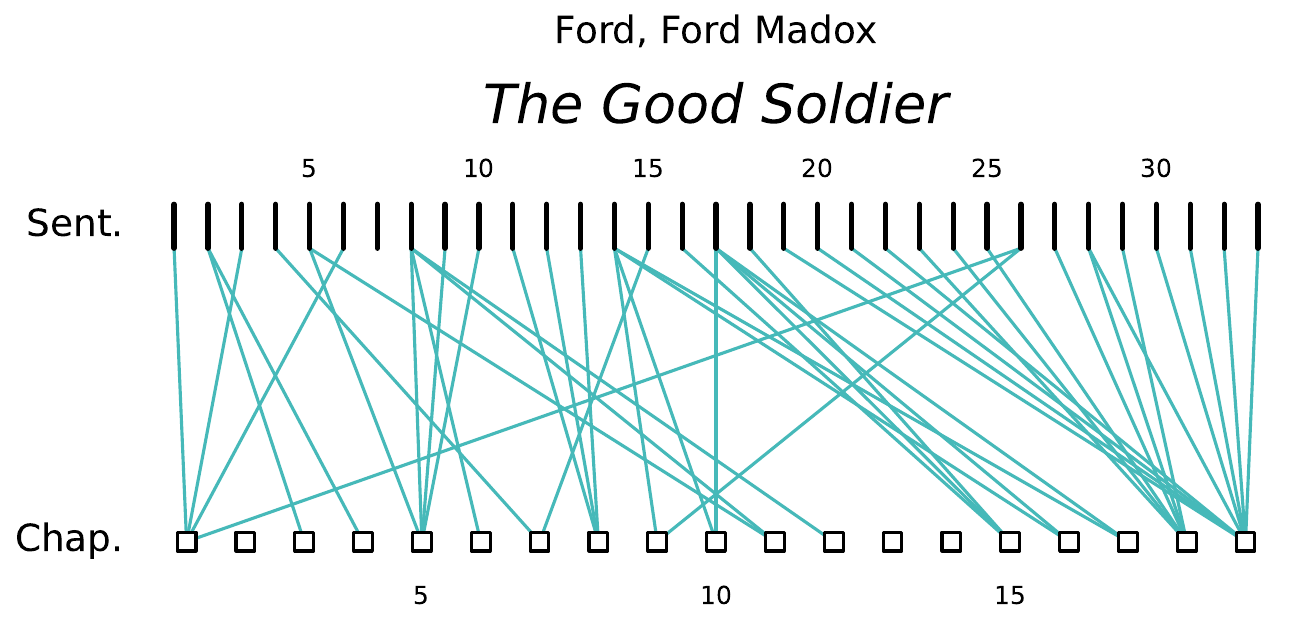}
    {\captionsetup{hypcap=false, labelformat=simple, labelsep=none, textformat=empty}\captionof{figure}{\label{fig:pg2775-bipartite}}}
  \end{minipage}\hfill
  \begin{minipage}[t]{0.48\linewidth}\centering
    \includegraphics[width=\linewidth]{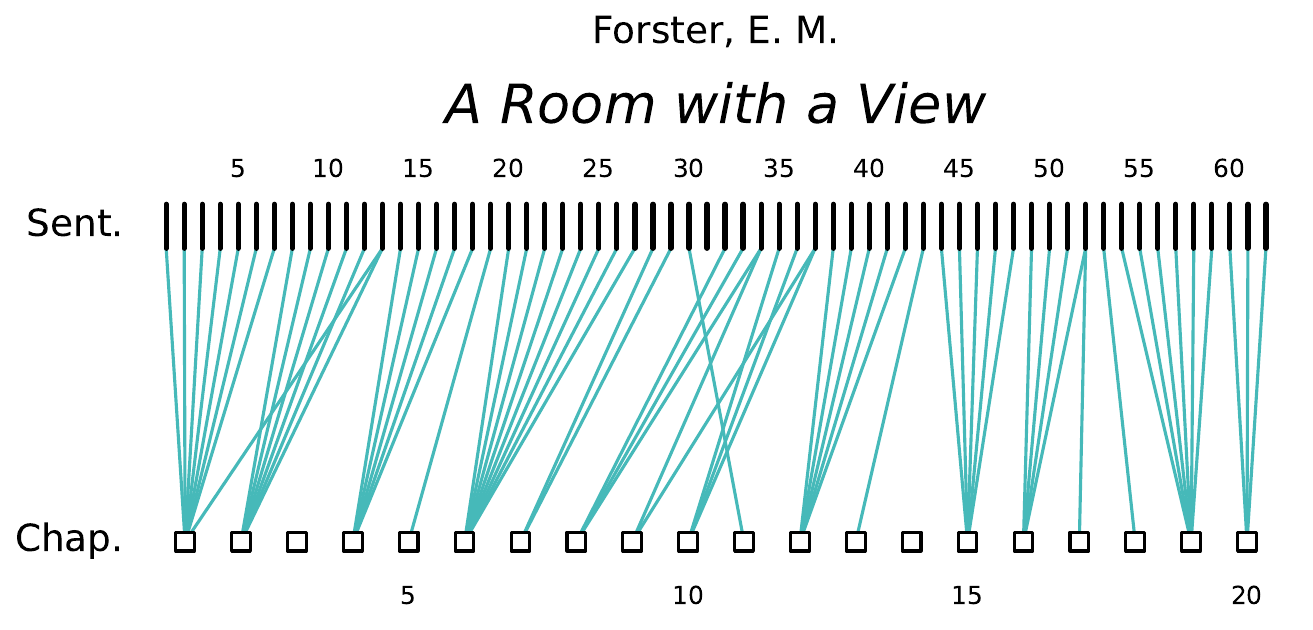}
    {\captionsetup{hypcap=false, labelformat=simple, labelsep=none, textformat=empty}\captionof{figure}{\label{fig:pg2641-bipartite}}}
  \end{minipage}\bigskip

\noindent
  \begin{minipage}[t]{0.48\linewidth}\centering
    \includegraphics[width=\linewidth]{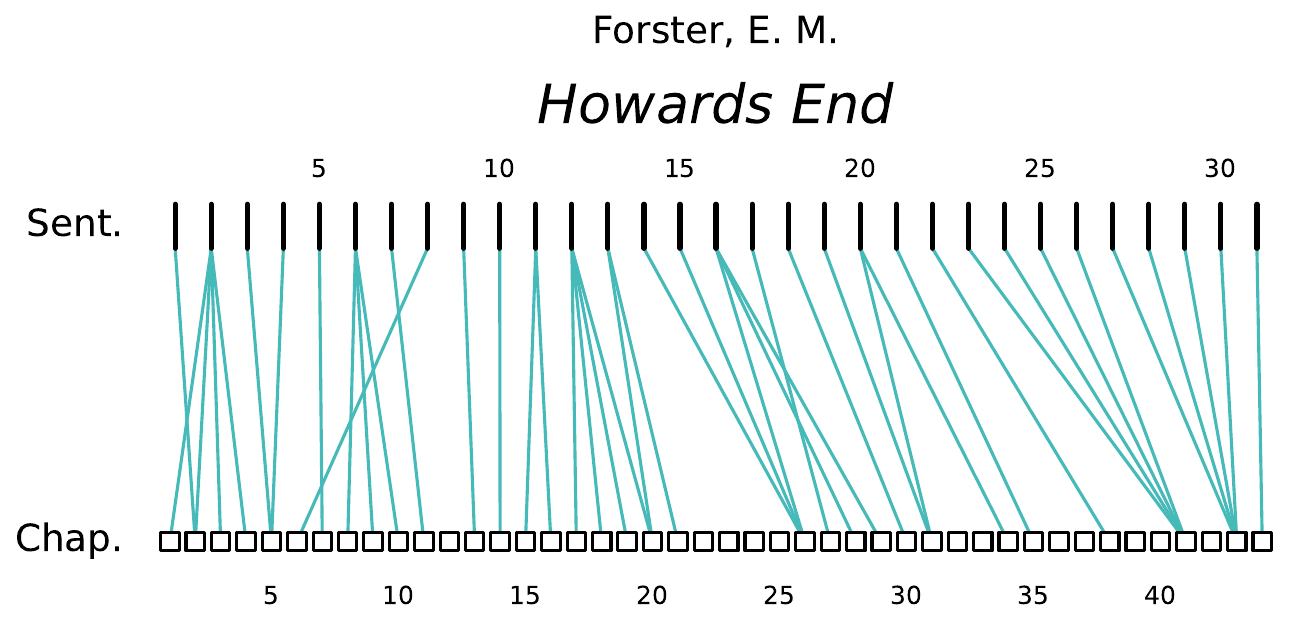}
    {\captionsetup{hypcap=false, labelformat=simple, labelsep=none, textformat=empty}\captionof{figure}{\label{fig:pg2891-bipartite}}}
  \end{minipage}\hfill
  \begin{minipage}[t]{0.48\linewidth}\centering
    \includegraphics[width=\linewidth]{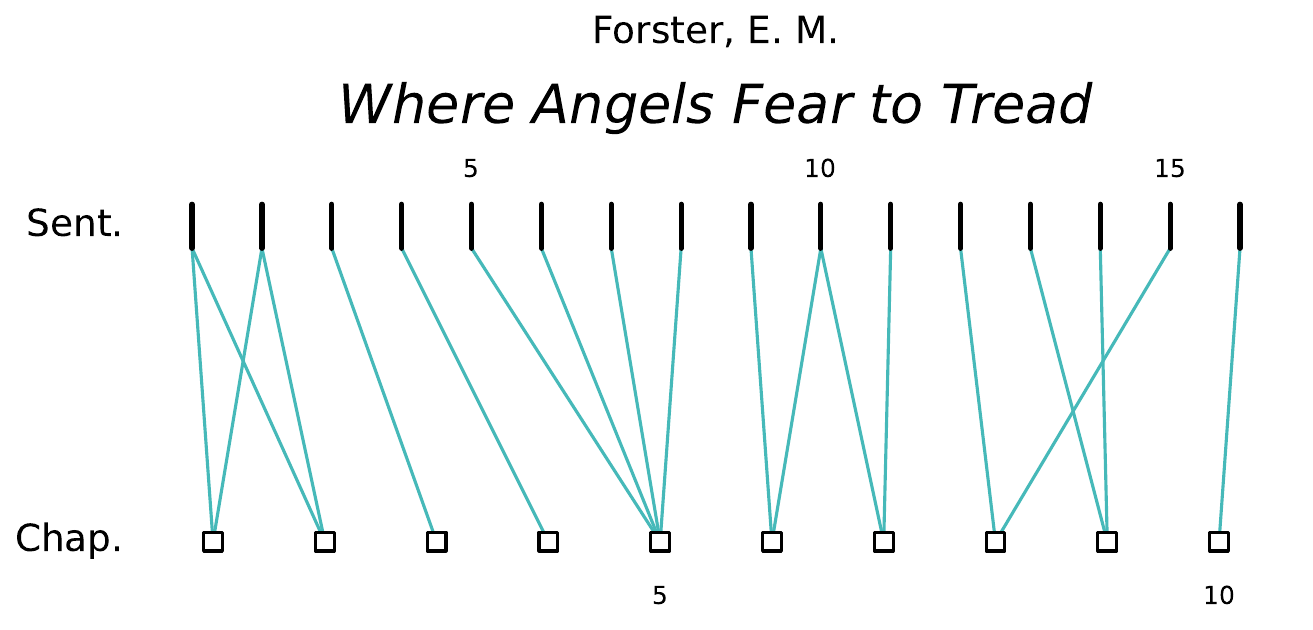}
    {\captionsetup{hypcap=false, labelformat=simple, labelsep=none, textformat=empty}\captionof{figure}{\label{fig:pg2948-bipartite}}}
  \end{minipage}\bigskip

\noindent
  \begin{minipage}[t]{0.48\linewidth}\centering
    \includegraphics[width=\linewidth]{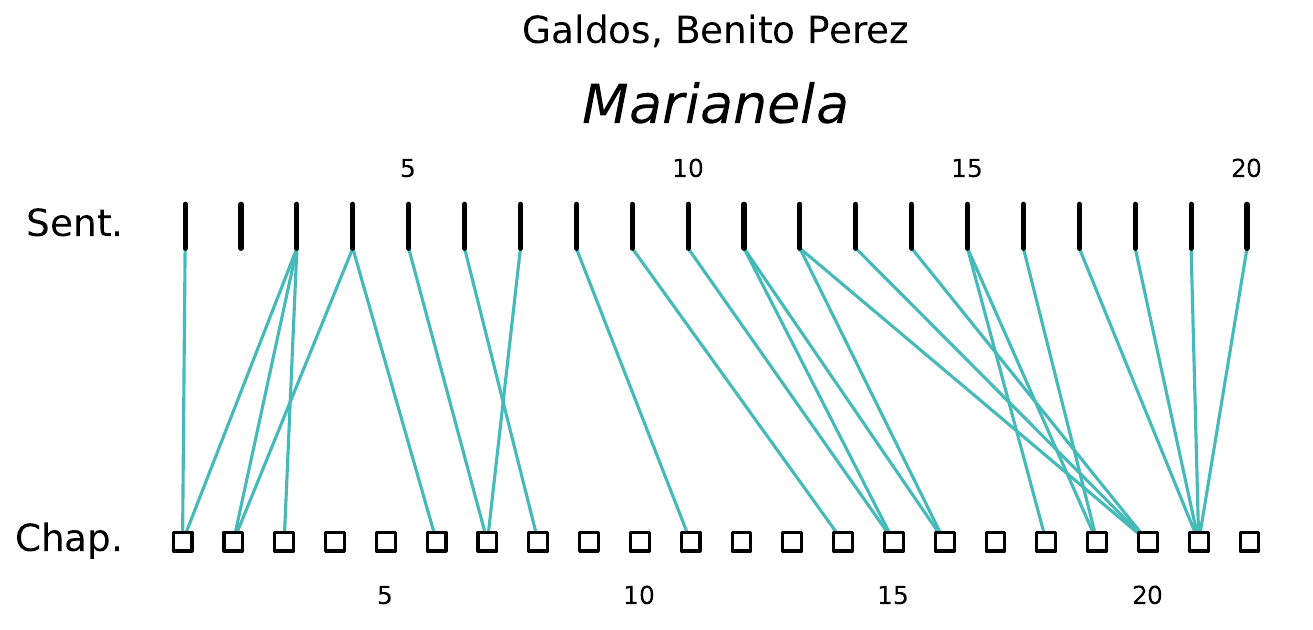}
    {\captionsetup{hypcap=false, labelformat=simple, labelsep=none, textformat=empty}\captionof{figure}{\label{fig:pg48818-bipartite}}}
  \end{minipage}\hfill
  \begin{minipage}[t]{0.48\linewidth}\centering
    \includegraphics[width=\linewidth]{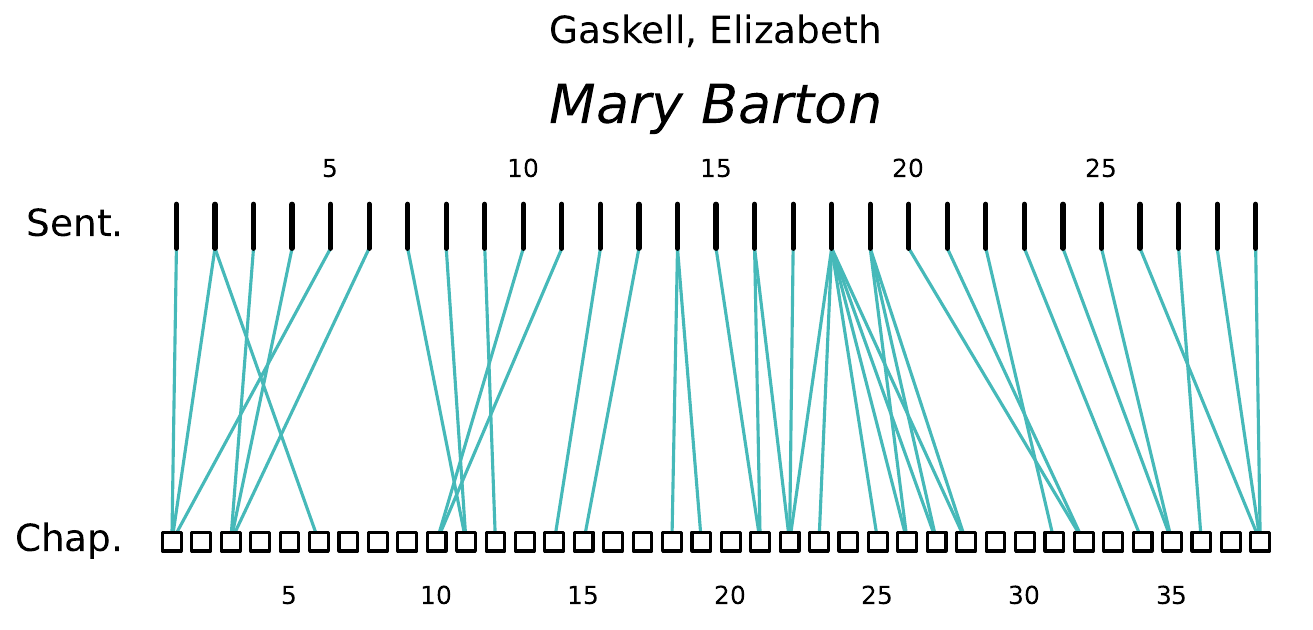}
    {\captionsetup{hypcap=false, labelformat=simple, labelsep=none, textformat=empty}\captionof{figure}{\label{fig:pg2153-bipartite}}}
  \end{minipage}\bigskip

\noindent
  \begin{minipage}[t]{0.48\linewidth}\centering
    \includegraphics[width=\linewidth]{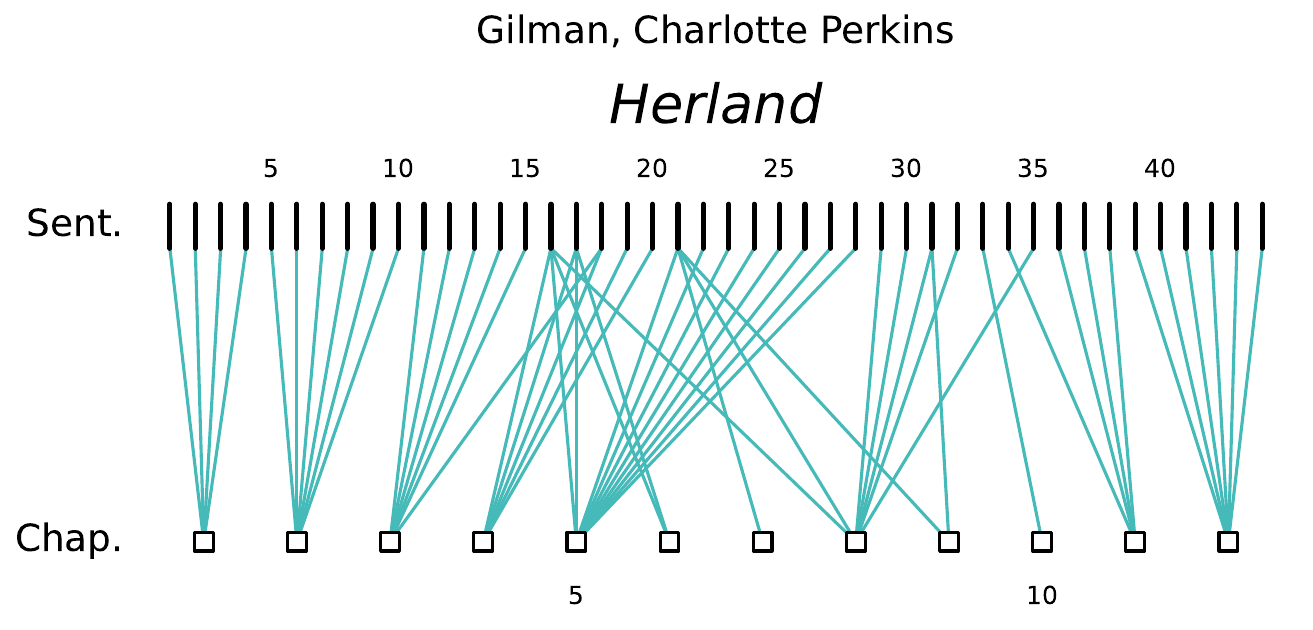}
    {\captionsetup{hypcap=false, labelformat=simple, labelsep=none, textformat=empty}\captionof{figure}{\label{fig:pg32-bipartite}}}
  \end{minipage}\hfill
  \begin{minipage}[t]{0.48\linewidth}\centering
    \includegraphics[width=\linewidth]{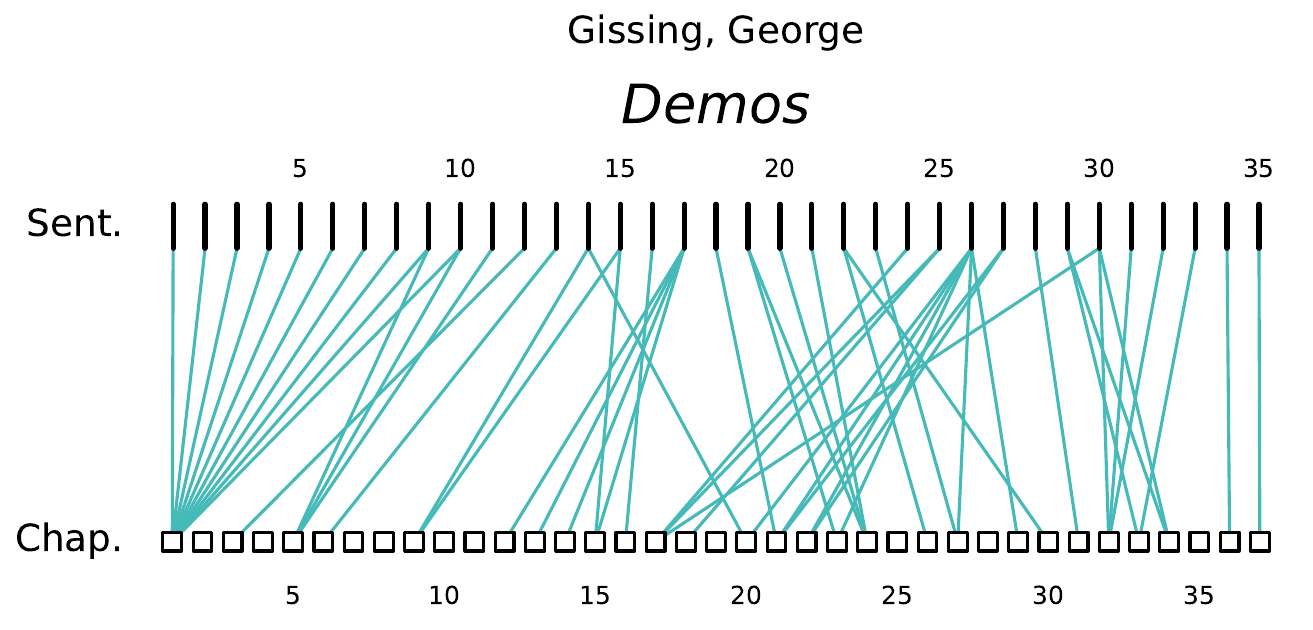}
    {\captionsetup{hypcap=false, labelformat=simple, labelsep=none, textformat=empty}\captionof{figure}{\label{fig:pg4309-bipartite}}}
  \end{minipage}\bigskip

\noindent
  \begin{minipage}[t]{0.48\linewidth}\centering
    \includegraphics[width=\linewidth]{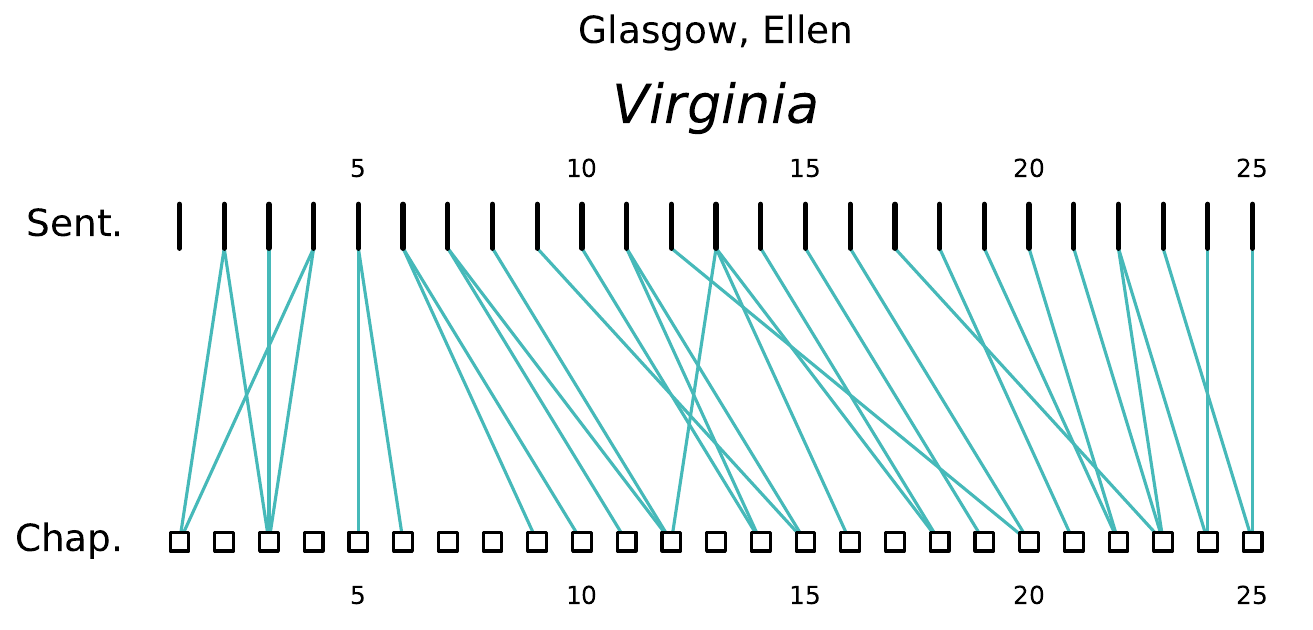}
    {\captionsetup{hypcap=false, labelformat=simple, labelsep=none, textformat=empty}\captionof{figure}{\label{fig:pg26316-bipartite}}}
  \end{minipage}\hfill
  \begin{minipage}[t]{0.48\linewidth}\centering
    \includegraphics[width=\linewidth]{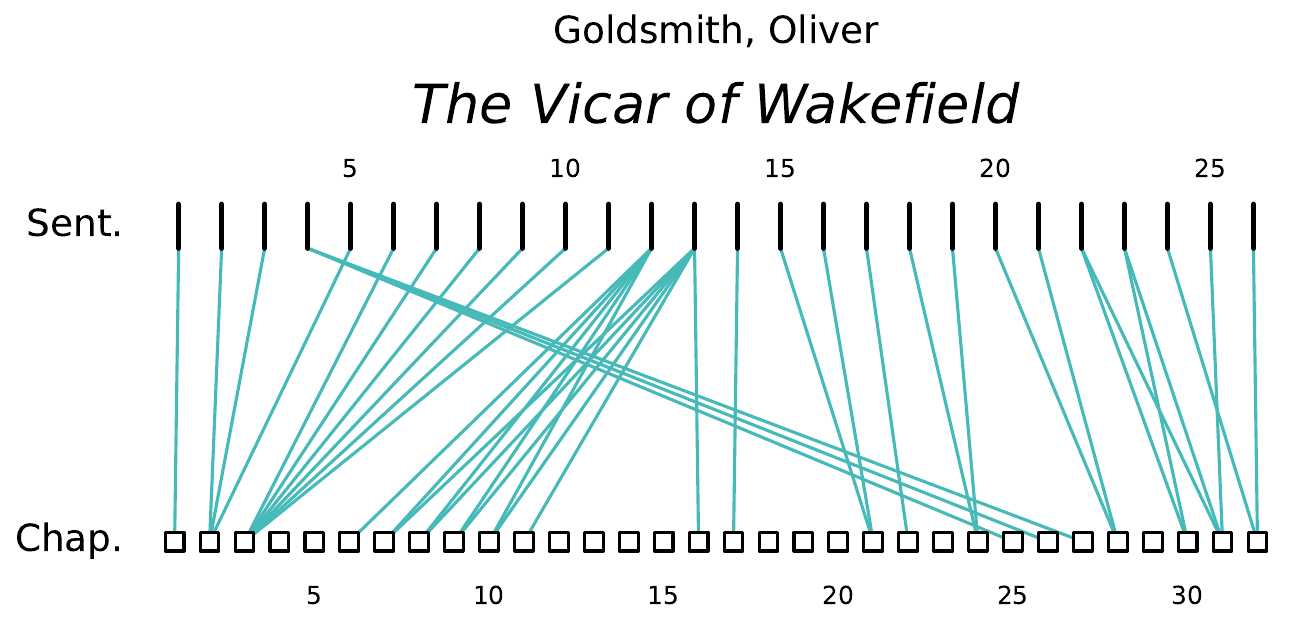}
    {\captionsetup{hypcap=false, labelformat=simple, labelsep=none, textformat=empty}\captionof{figure}{\label{fig:pg2667-bipartite}}}
  \end{minipage}\bigskip

\noindent
  \begin{minipage}[t]{0.48\linewidth}\centering
    \includegraphics[width=\linewidth]{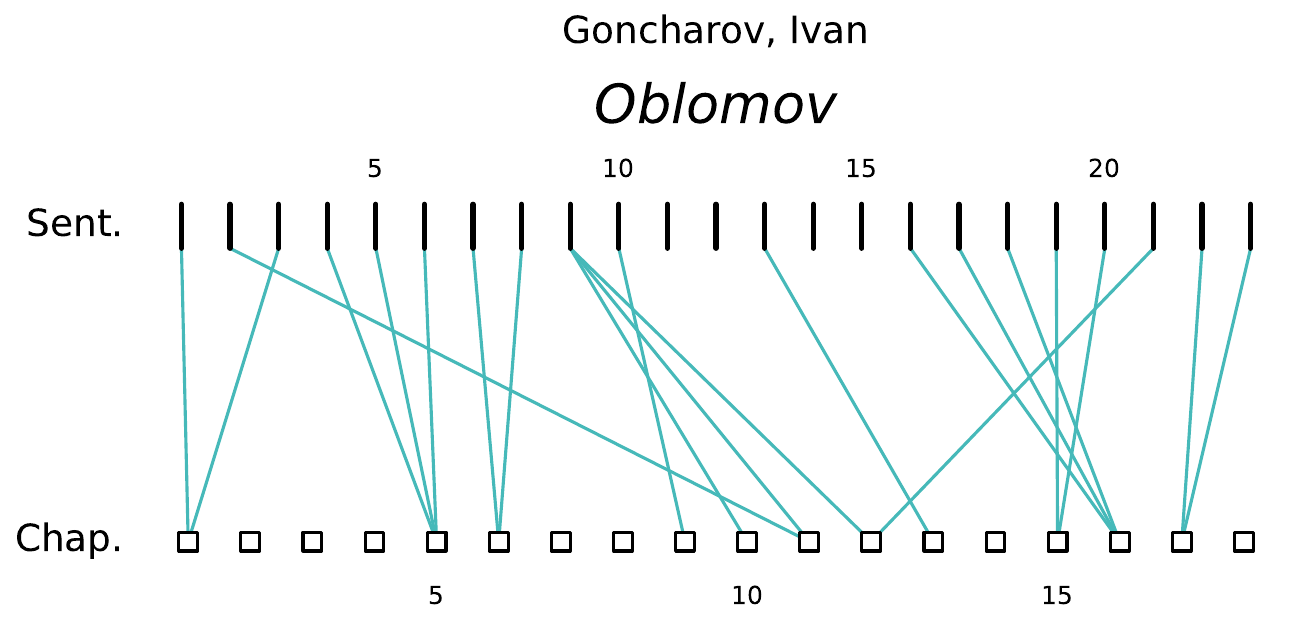}
    {\captionsetup{hypcap=false, labelformat=simple, labelsep=none, textformat=empty}\captionof{figure}{\label{fig:pg54700-bipartite}}}
  \end{minipage}\hfill
  \begin{minipage}[t]{0.48\linewidth}\centering
    \includegraphics[width=\linewidth]{images/pg289_bipartite.pdf}
    {\captionsetup{hypcap=false, labelformat=simple, labelsep=none, textformat=empty}\captionof{figure}{\label{fig:pg289-bipartite}}}
  \end{minipage}\bigskip

\noindent
  \begin{minipage}[t]{0.48\linewidth}\centering
    \includegraphics[width=\linewidth]{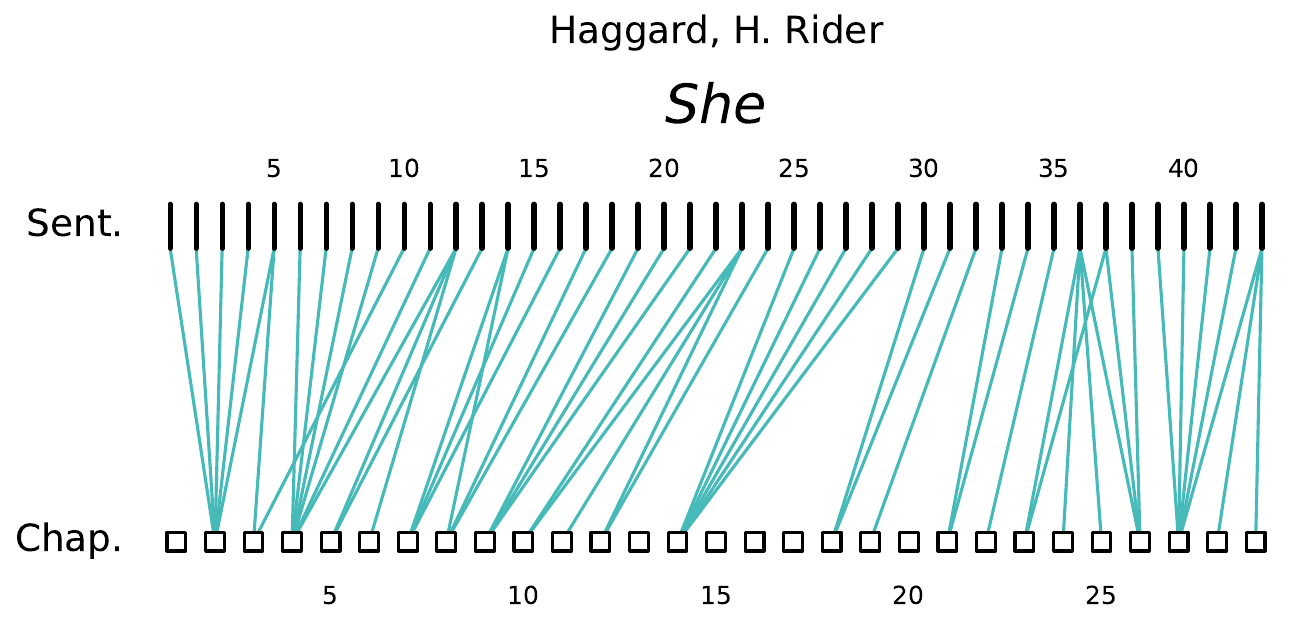}
    {\captionsetup{hypcap=false, labelformat=simple, labelsep=none, textformat=empty}\captionof{figure}{\label{fig:pg3155-bipartite}}}
  \end{minipage}\hfill
  \begin{minipage}[t]{0.48\linewidth}\centering
    \includegraphics[width=\linewidth]{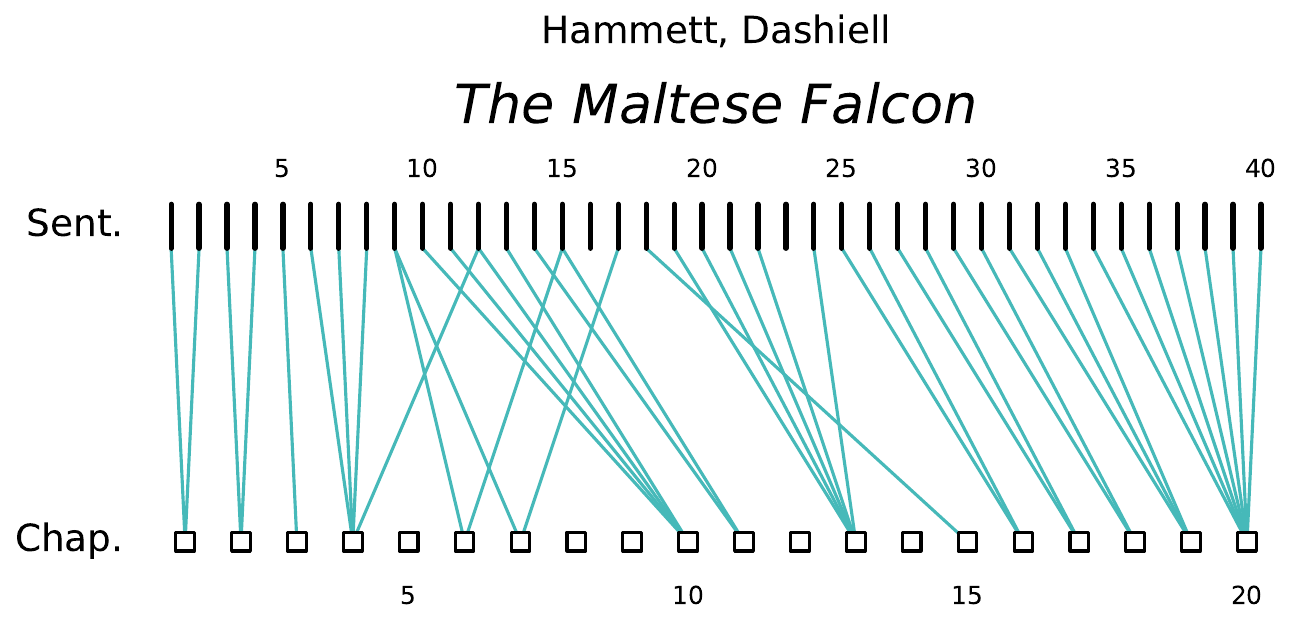}
    {\captionsetup{hypcap=false, labelformat=simple, labelsep=none, textformat=empty}\captionof{figure}{\label{fig:pg77600-bipartite}}}
  \end{minipage}\bigskip

\noindent
  \begin{minipage}[t]{0.48\linewidth}\centering
    \includegraphics[width=\linewidth]{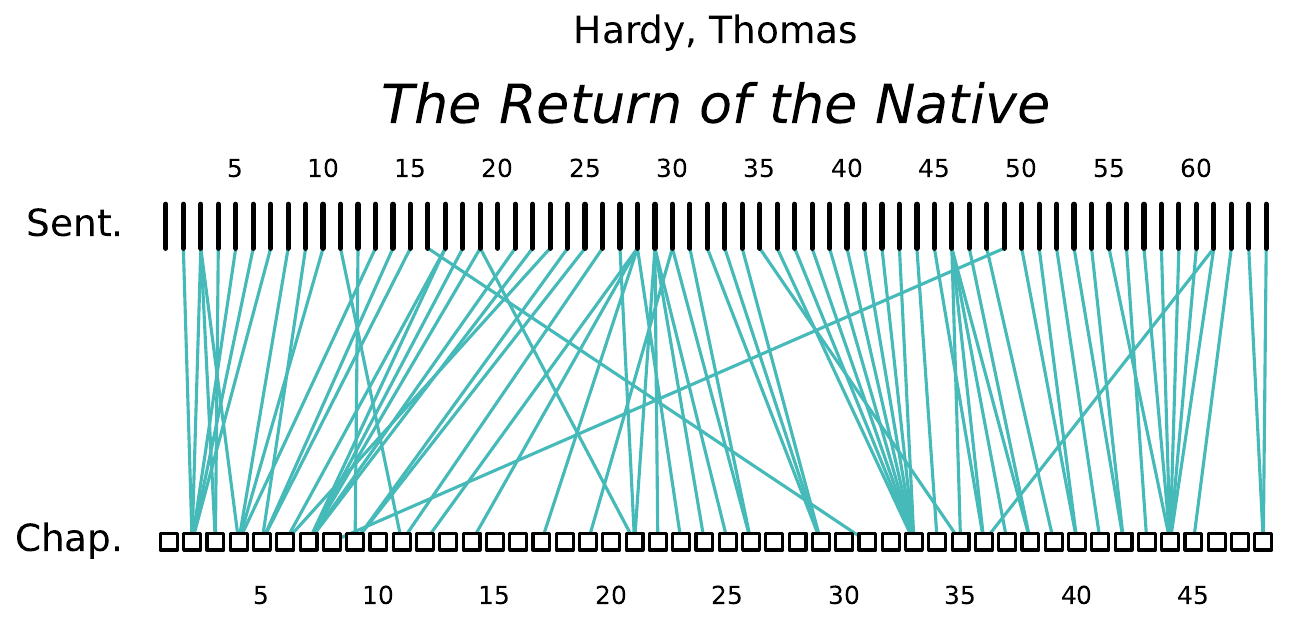}
    {\captionsetup{hypcap=false, labelformat=simple, labelsep=none, textformat=empty}\captionof{figure}{\label{fig:pg122-bipartite}}}
  \end{minipage}\hfill
  \begin{minipage}[t]{0.48\linewidth}\centering
    \includegraphics[width=\linewidth]{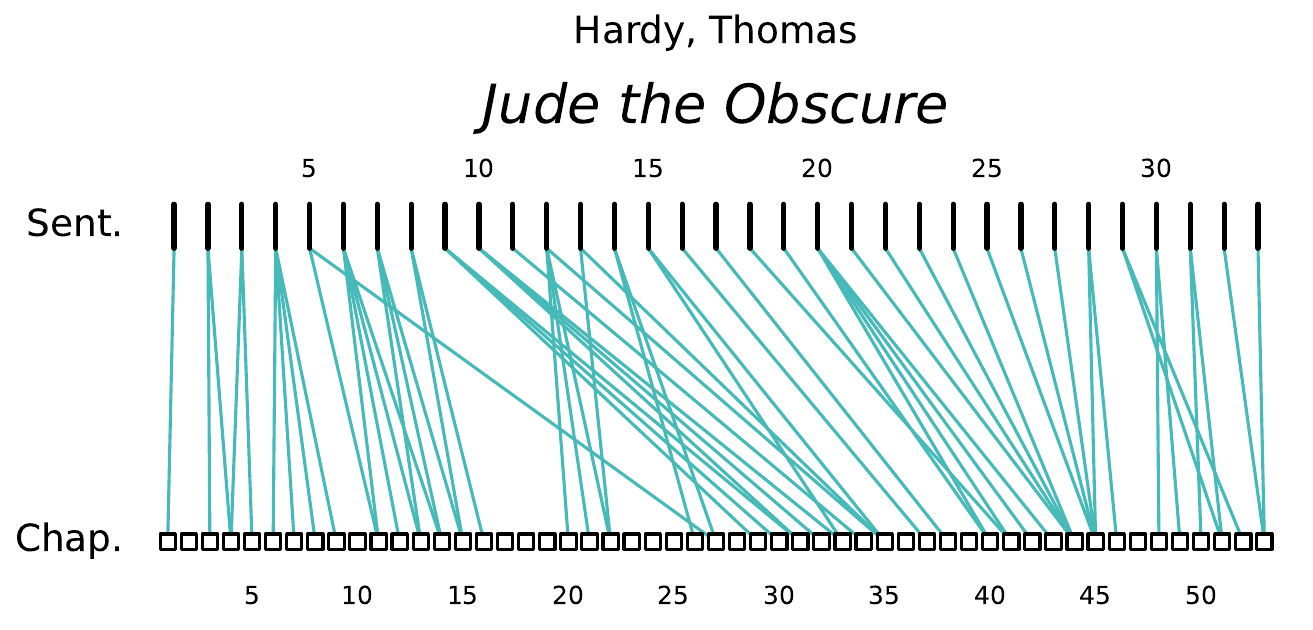}
    {\captionsetup{hypcap=false, labelformat=simple, labelsep=none, textformat=empty}\captionof{figure}{\label{fig:pg153-bipartite}}}
  \end{minipage}\bigskip

\noindent
  \begin{minipage}[t]{0.48\linewidth}\centering
    \includegraphics[width=\linewidth]{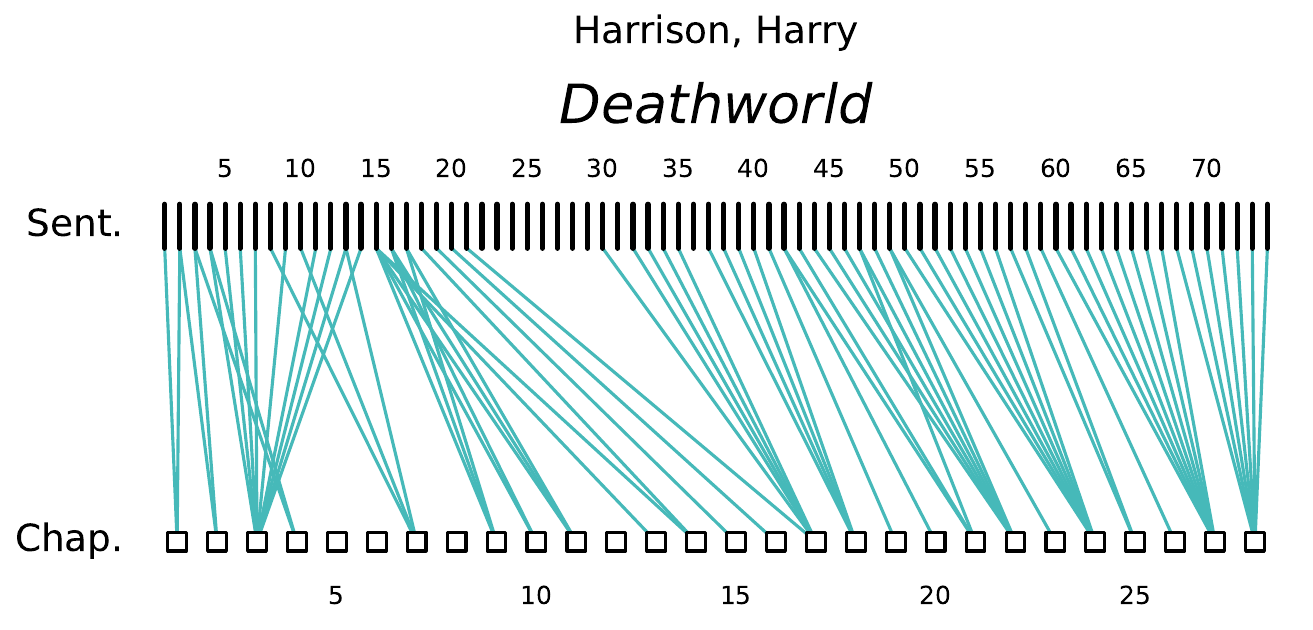}
    {\captionsetup{hypcap=false, labelformat=simple, labelsep=none, textformat=empty}\captionof{figure}{\label{fig:pg28346-bipartite}}}
  \end{minipage}\hfill
  \begin{minipage}[t]{0.48\linewidth}\centering
    \includegraphics[width=\linewidth]{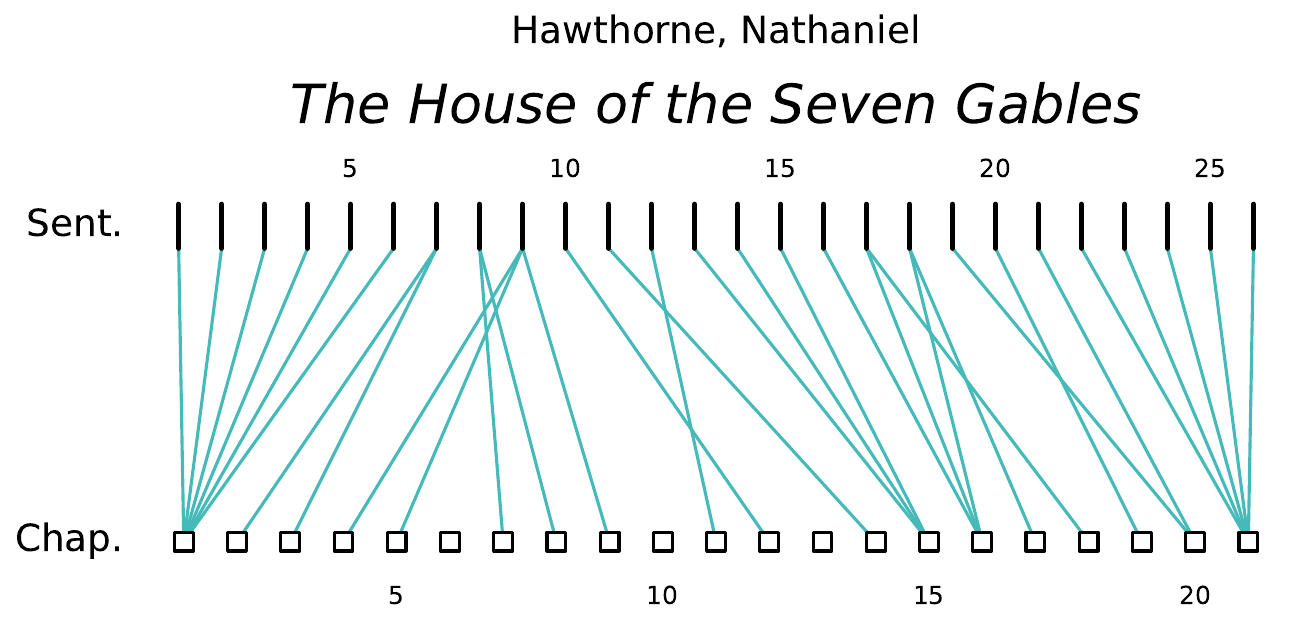}
    {\captionsetup{hypcap=false, labelformat=simple, labelsep=none, textformat=empty}\captionof{figure}{\label{fig:pg77-bipartite}}}
  \end{minipage}\bigskip

\noindent
  \begin{minipage}[t]{0.48\linewidth}\centering
    \includegraphics[width=\linewidth]{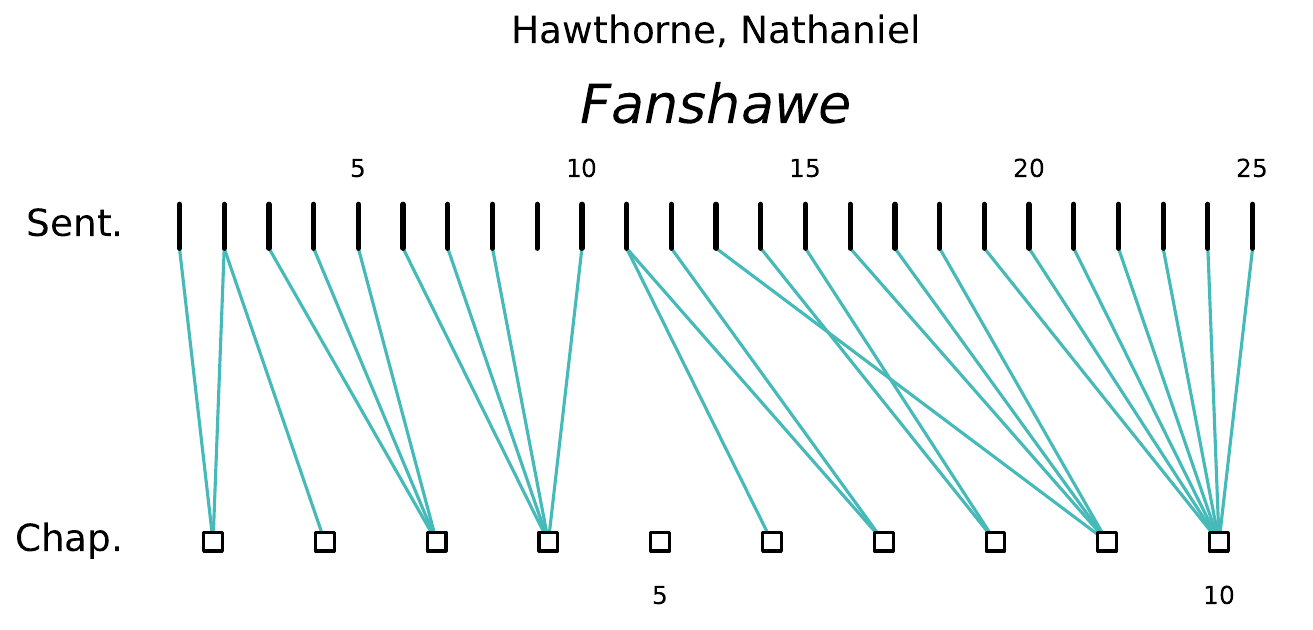}
    {\captionsetup{hypcap=false, labelformat=simple, labelsep=none, textformat=empty}\captionof{figure}{\label{fig:pg7085-bipartite}}}
  \end{minipage}\hfill
  \begin{minipage}[t]{0.48\linewidth}\centering
    \includegraphics[width=\linewidth]{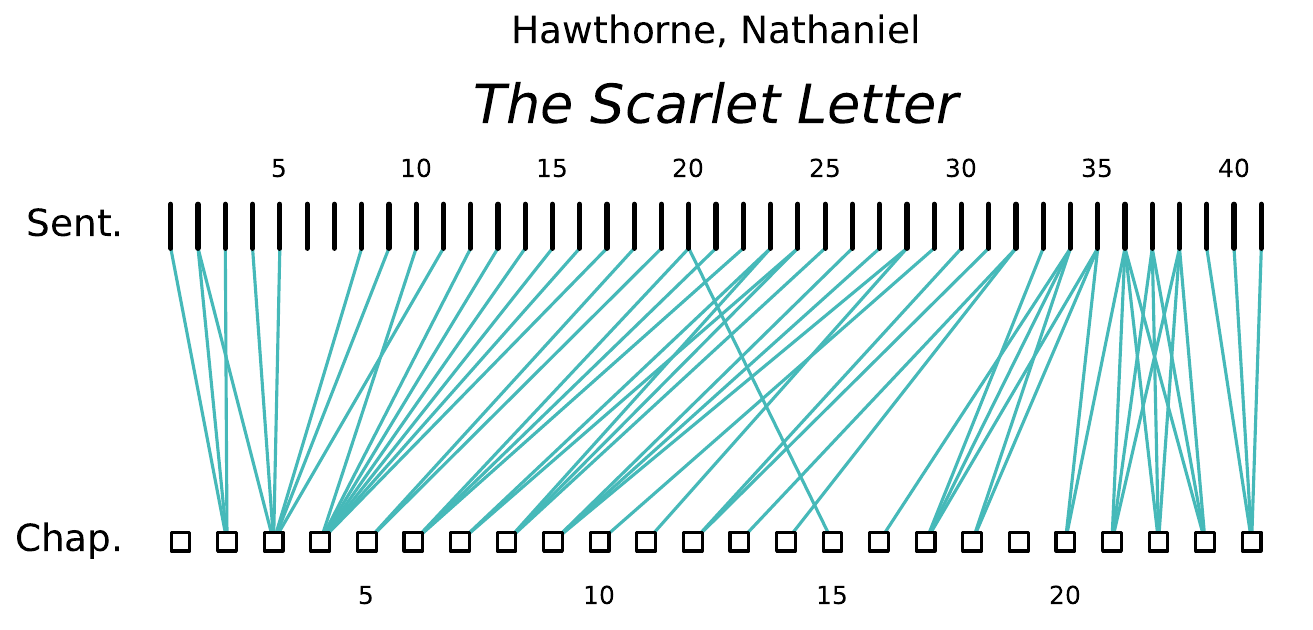}
    {\captionsetup{hypcap=false, labelformat=simple, labelsep=none, textformat=empty}\captionof{figure}{\label{fig:pg25344-bipartite}}}
  \end{minipage}\bigskip

\noindent
  \begin{minipage}[t]{0.48\linewidth}\centering
    \includegraphics[width=\linewidth]{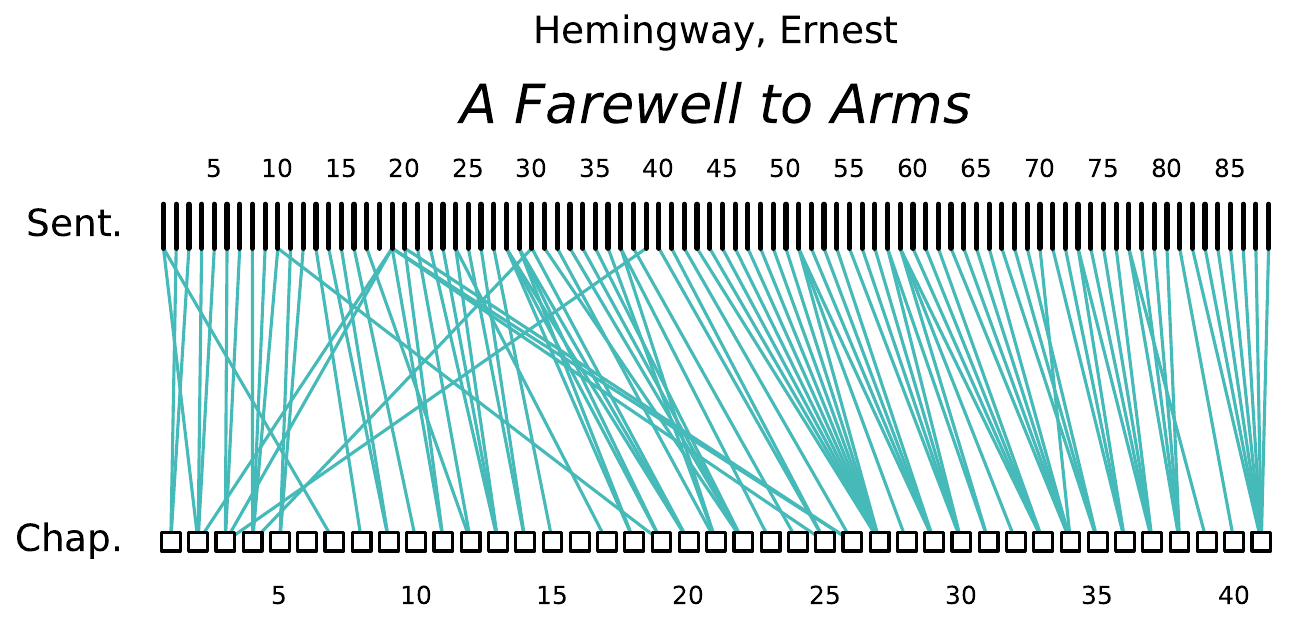}
    {\captionsetup{hypcap=false, labelformat=simple, labelsep=none, textformat=empty}\captionof{figure}{\label{fig:pg75201-bipartite}}}
  \end{minipage}\hfill
  \begin{minipage}[t]{0.48\linewidth}\centering
    \includegraphics[width=\linewidth]{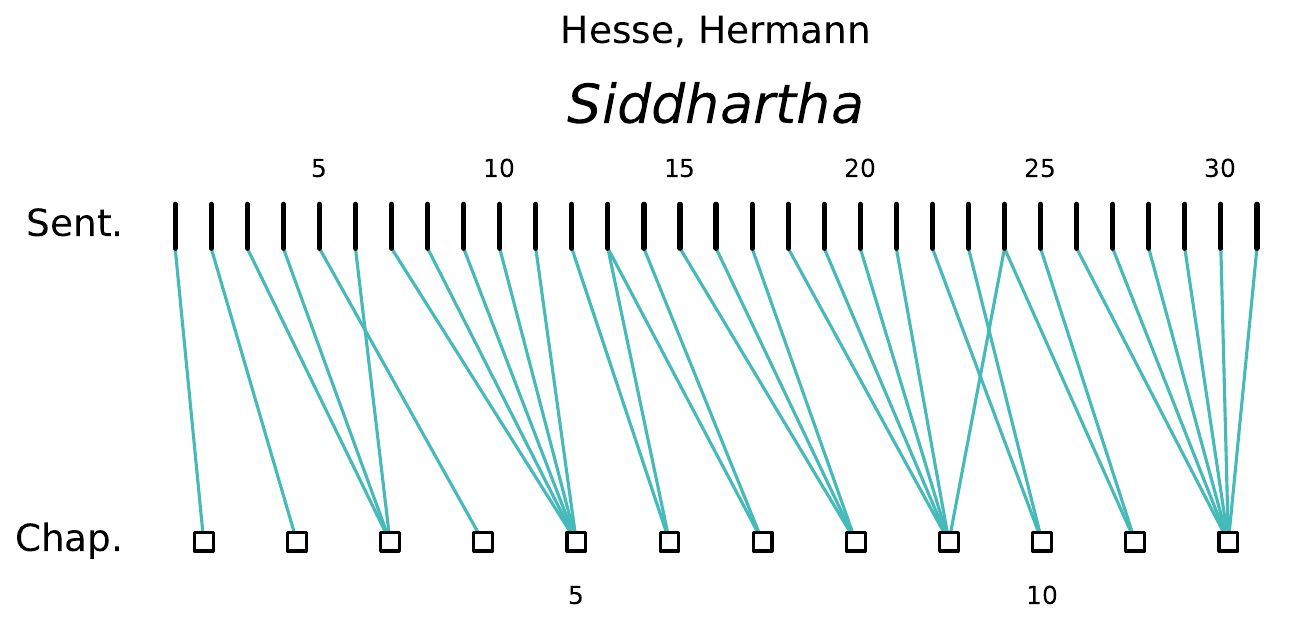}
    {\captionsetup{hypcap=false, labelformat=simple, labelsep=none, textformat=empty}\captionof{figure}{\label{fig:pg2500-bipartite}}}
  \end{minipage}\bigskip

\noindent
  \begin{minipage}[t]{0.48\linewidth}\centering
    \includegraphics[width=\linewidth]{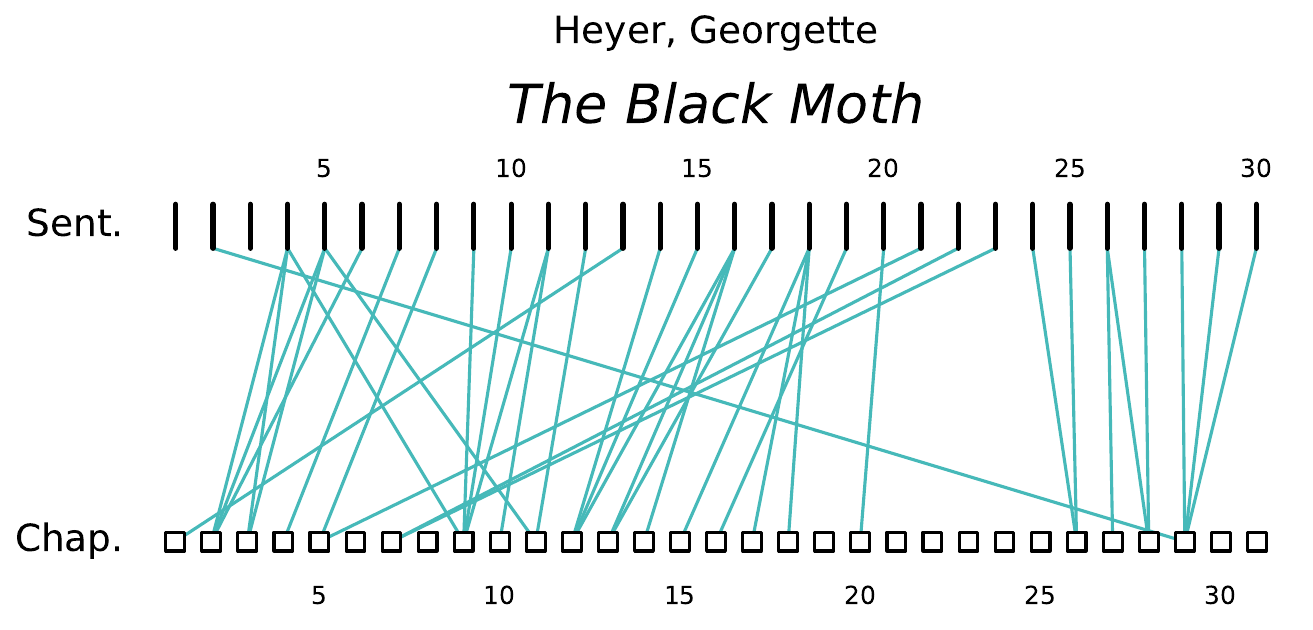}
    {\captionsetup{hypcap=false, labelformat=simple, labelsep=none, textformat=empty}\captionof{figure}{\label{fig:pg38703-bipartite}}}
  \end{minipage}\hfill
  \begin{minipage}[t]{0.48\linewidth}\centering
    \includegraphics[width=\linewidth]{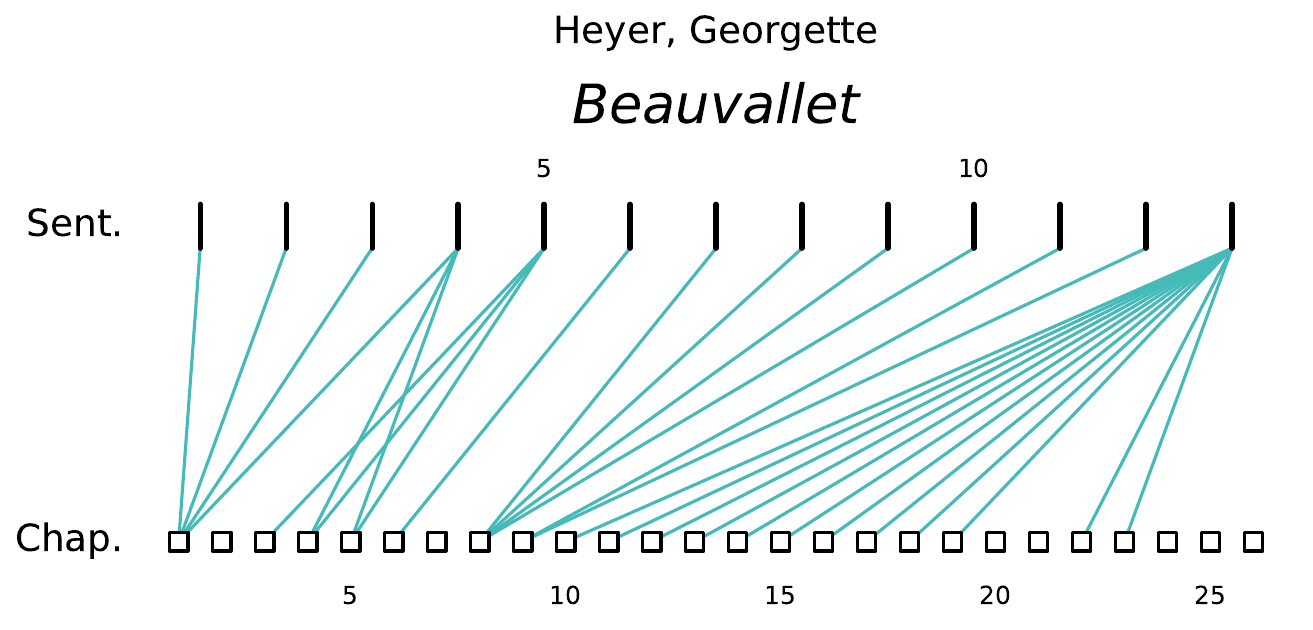}
    {\captionsetup{hypcap=false, labelformat=simple, labelsep=none, textformat=empty}\captionof{figure}{\label{fig:pg75547-bipartite}}}
  \end{minipage}\bigskip

\noindent
  \begin{minipage}[t]{0.48\linewidth}\centering
    \includegraphics[width=\linewidth]{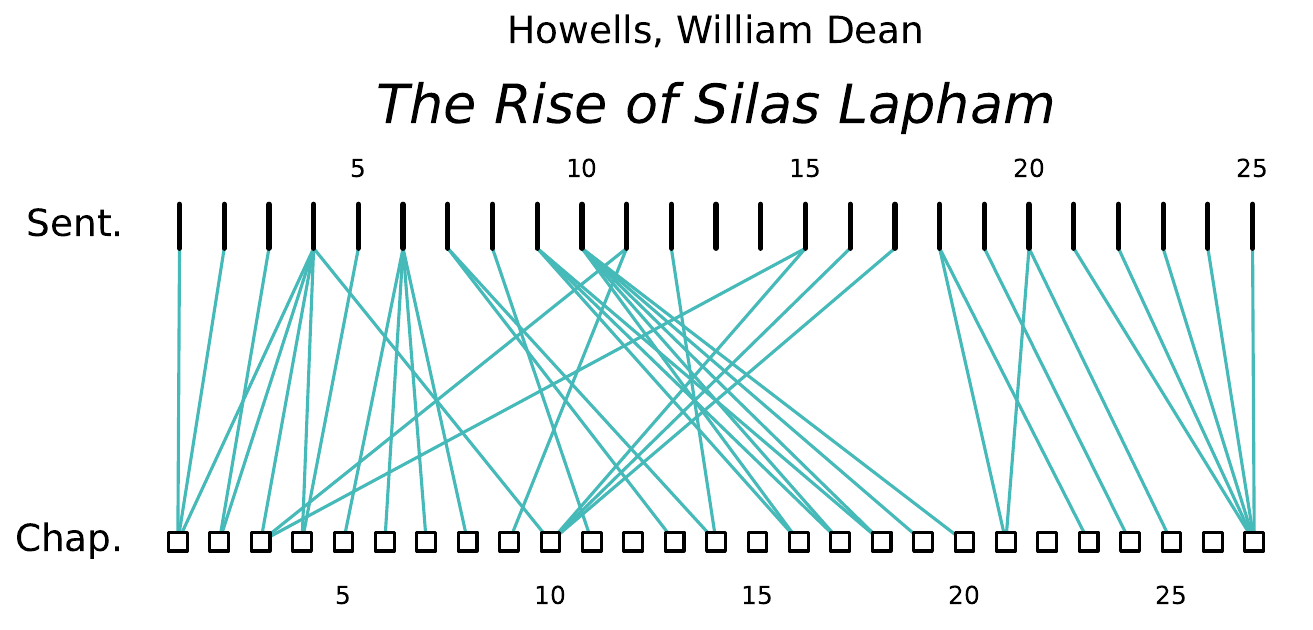}
    {\captionsetup{hypcap=false, labelformat=simple, labelsep=none, textformat=empty}\captionof{figure}{\label{fig:pg154-bipartite}}}
  \end{minipage}\hfill
  \begin{minipage}[t]{0.48\linewidth}\centering
    \includegraphics[width=\linewidth]{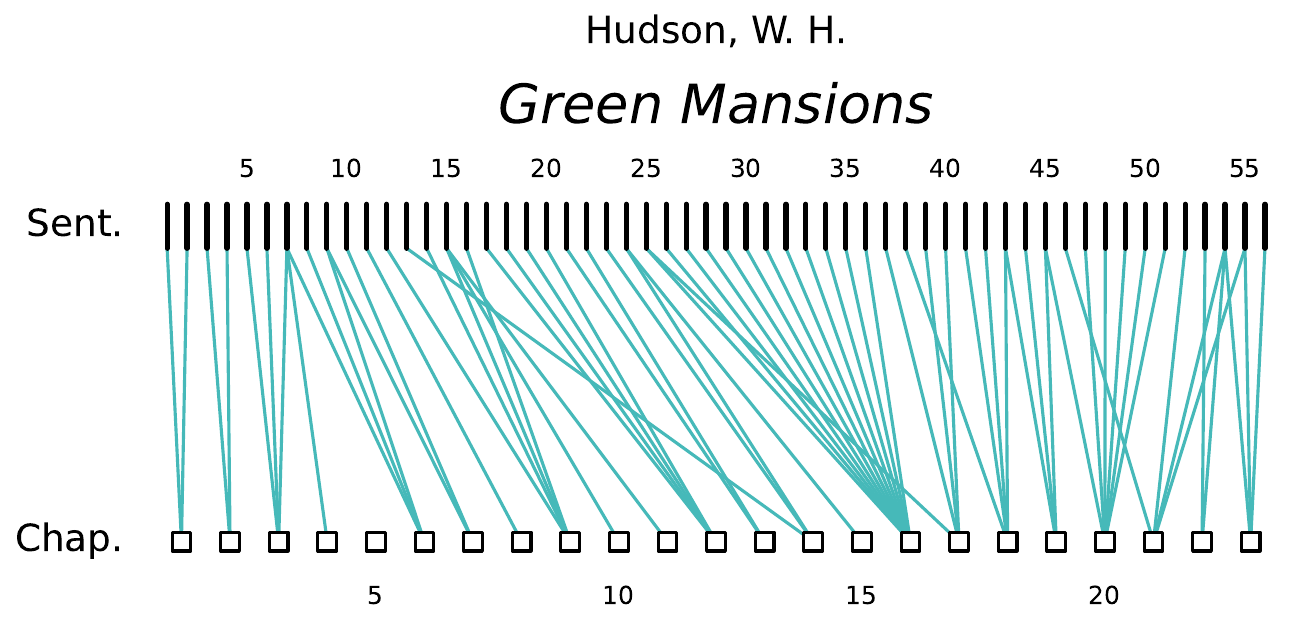}
    {\captionsetup{hypcap=false, labelformat=simple, labelsep=none, textformat=empty}\captionof{figure}{\label{fig:pg942-bipartite}}}
  \end{minipage}\bigskip

\noindent
  \begin{minipage}[t]{0.48\linewidth}\centering
    \includegraphics[width=\linewidth]{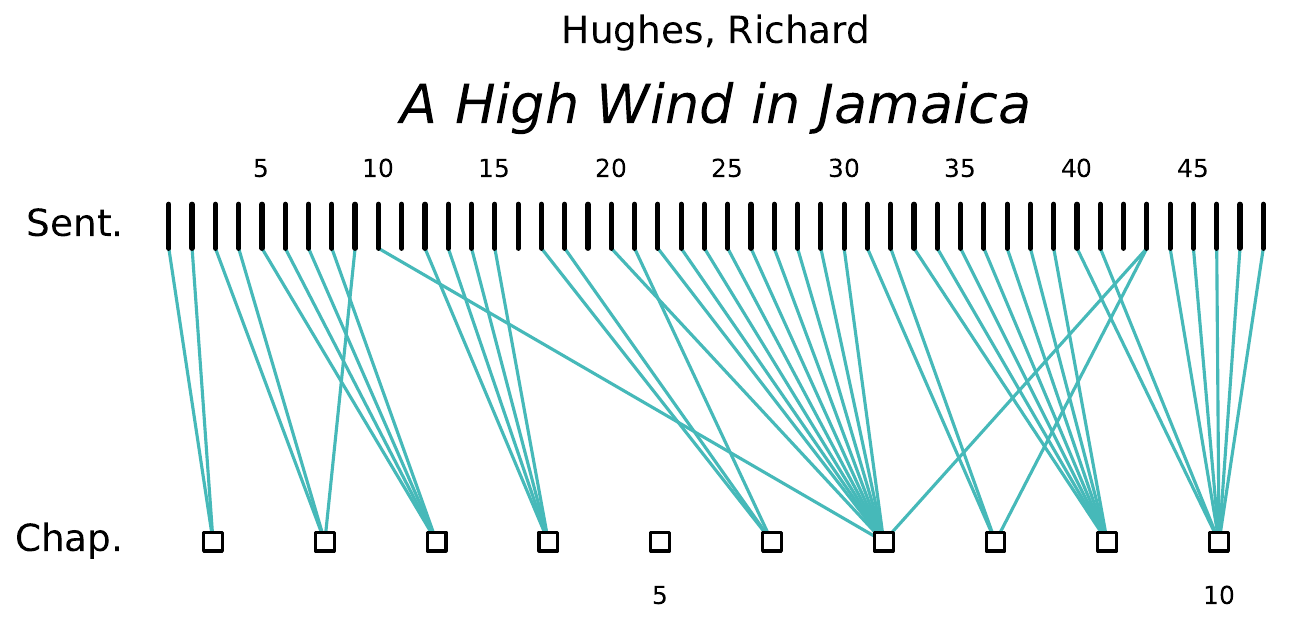}
    {\captionsetup{hypcap=false, labelformat=simple, labelsep=none, textformat=empty}\captionof{figure}{\label{fig:pg75530-bipartite}}}
  \end{minipage}\hfill
  \begin{minipage}[t]{0.48\linewidth}\centering
    \includegraphics[width=\linewidth]{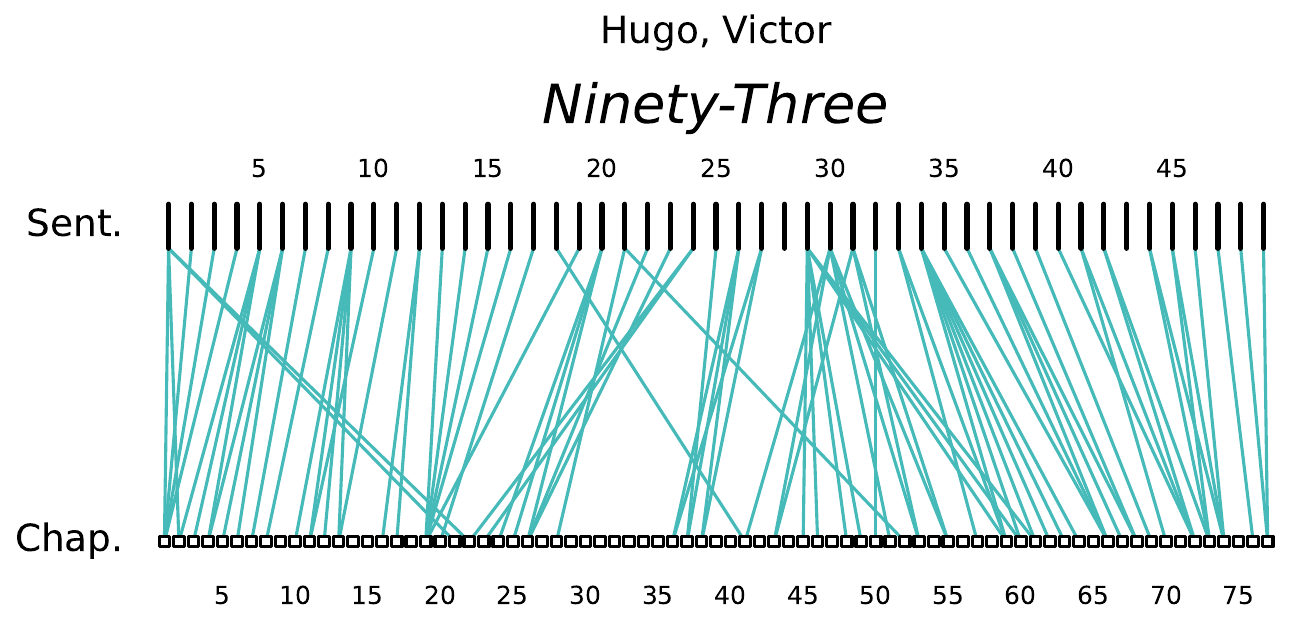}
    {\captionsetup{hypcap=false, labelformat=simple, labelsep=none, textformat=empty}\captionof{figure}{\label{fig:pg49372-bipartite}}}
  \end{minipage}\bigskip

\noindent
  \begin{minipage}[t]{0.48\linewidth}\centering
    \includegraphics[width=\linewidth]{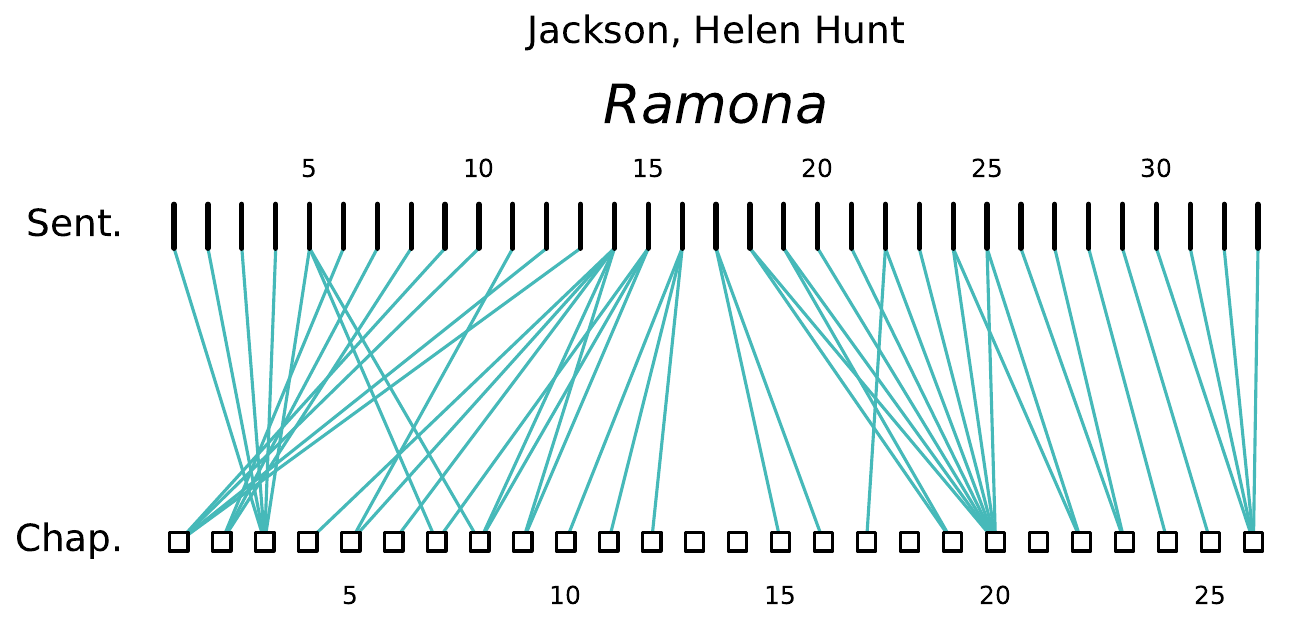}
    {\captionsetup{hypcap=false, labelformat=simple, labelsep=none, textformat=empty}\captionof{figure}{\label{fig:pg2802-bipartite}}}
  \end{minipage}\hfill
  \begin{minipage}[t]{0.48\linewidth}\centering
    \includegraphics[width=\linewidth]{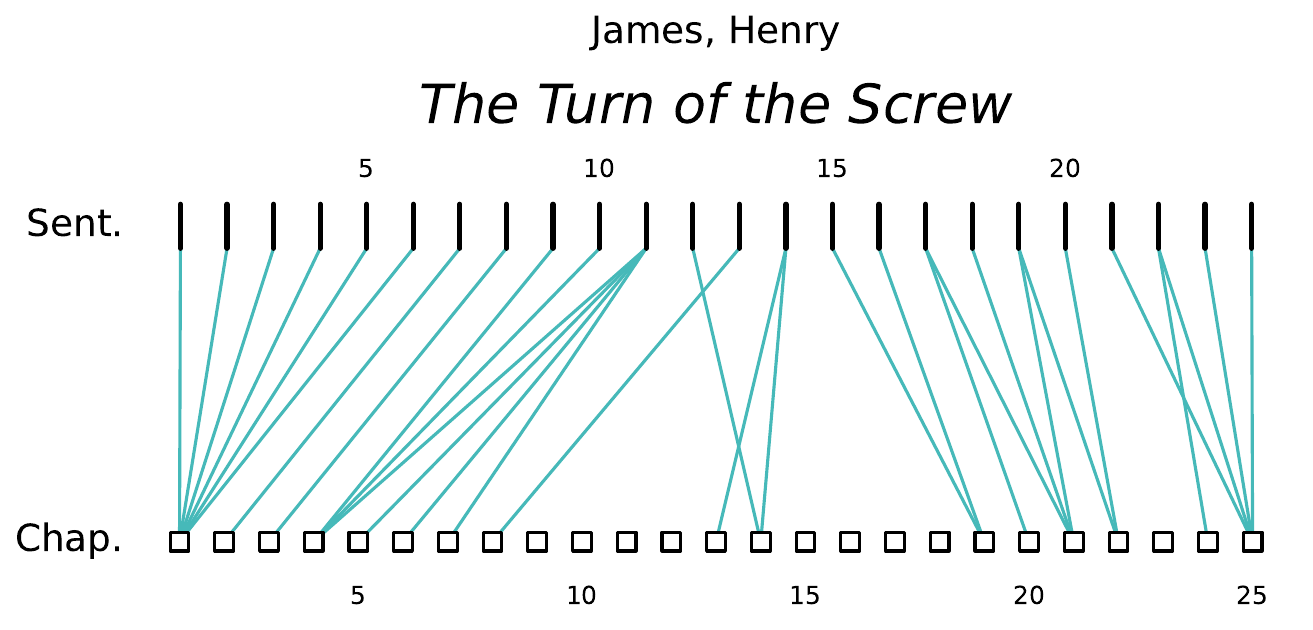}
    {\captionsetup{hypcap=false, labelformat=simple, labelsep=none, textformat=empty}\captionof{figure}{\label{fig:pg209-bipartite}}}
  \end{minipage}\bigskip

\noindent
  \begin{minipage}[t]{0.48\linewidth}\centering
    \includegraphics[width=\linewidth]{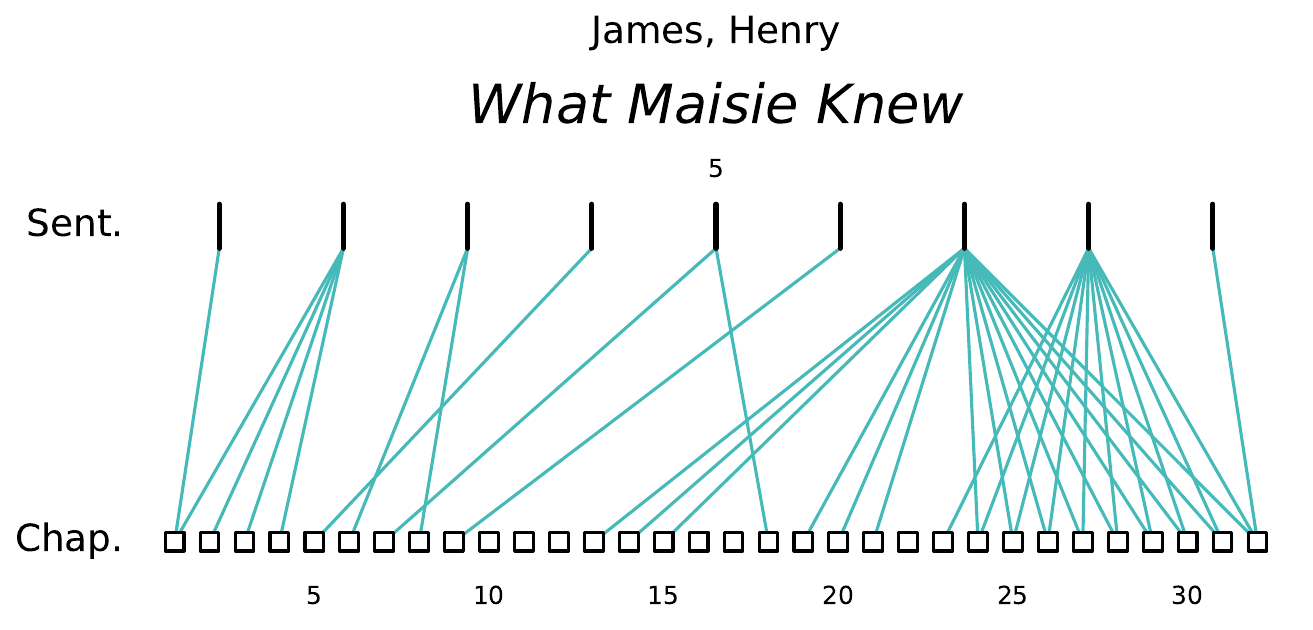}
    {\captionsetup{hypcap=false, labelformat=simple, labelsep=none, textformat=empty}\captionof{figure}{\label{fig:pg7118-bipartite}}}
  \end{minipage}\hfill
  \begin{minipage}[t]{0.48\linewidth}\centering
    \includegraphics[width=\linewidth]{images/pg4300_bipartite.pdf}
    {\captionsetup{hypcap=false, labelformat=simple, labelsep=none, textformat=empty}\captionof{figure}{\label{fig:pg4300-bipartite}}}
  \end{minipage}\bigskip

\noindent
  \begin{minipage}[t]{0.48\linewidth}\centering
    \includegraphics[width=\linewidth]{images/pg5200_bipartite.pdf}
    {\captionsetup{hypcap=false, labelformat=simple, labelsep=none, textformat=empty}\captionof{figure}{\label{fig:pg5200-bipartite}}}
  \end{minipage}\hfill
  \begin{minipage}[t]{0.48\linewidth}\centering
    \includegraphics[width=\linewidth]{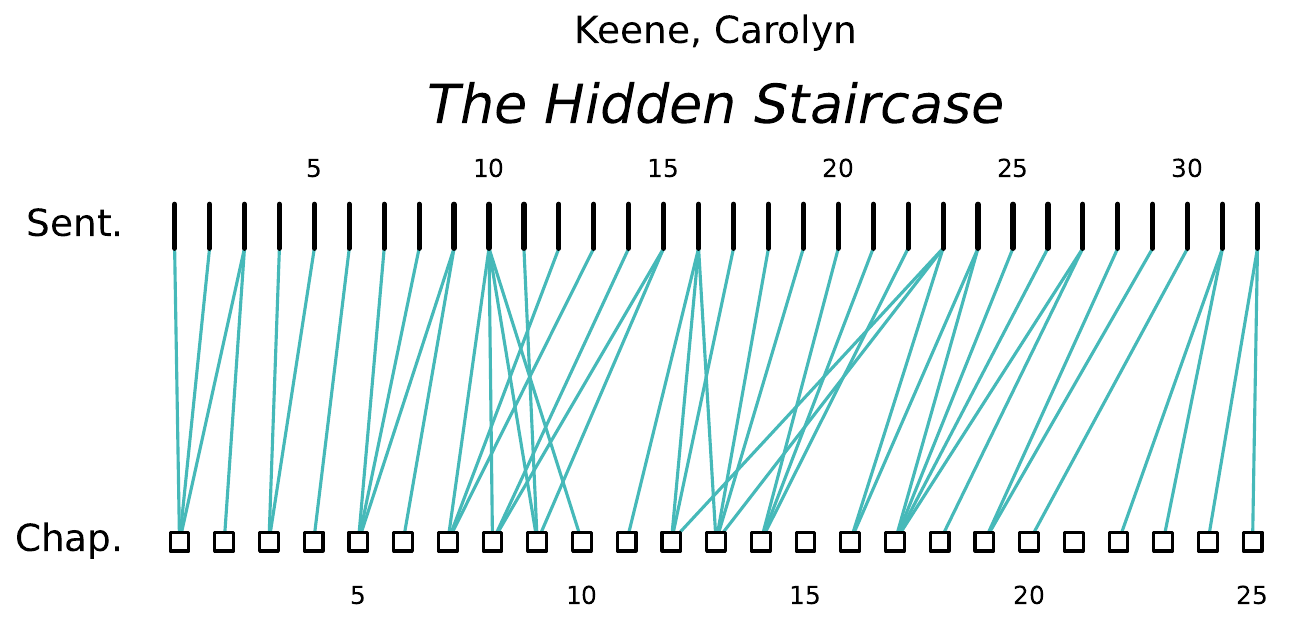}
    {\captionsetup{hypcap=false, labelformat=simple, labelsep=none, textformat=empty}\captionof{figure}{\label{fig:pg77602-bipartite}}}
  \end{minipage}\bigskip

\noindent
  \begin{minipage}[t]{0.48\linewidth}\centering
    \includegraphics[width=\linewidth]{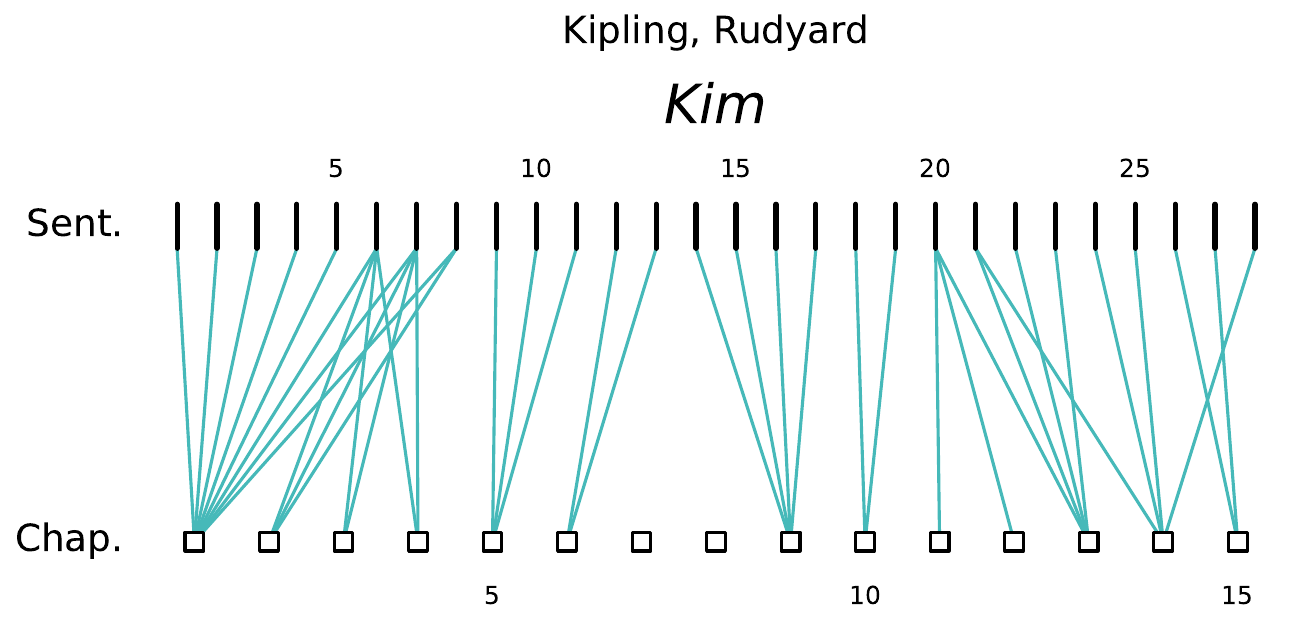}
    {\captionsetup{hypcap=false, labelformat=simple, labelsep=none, textformat=empty}\captionof{figure}{\label{fig:pg2226-bipartite}}}
  \end{minipage}\hfill
  \begin{minipage}[t]{0.48\linewidth}\centering
    \includegraphics[width=\linewidth]{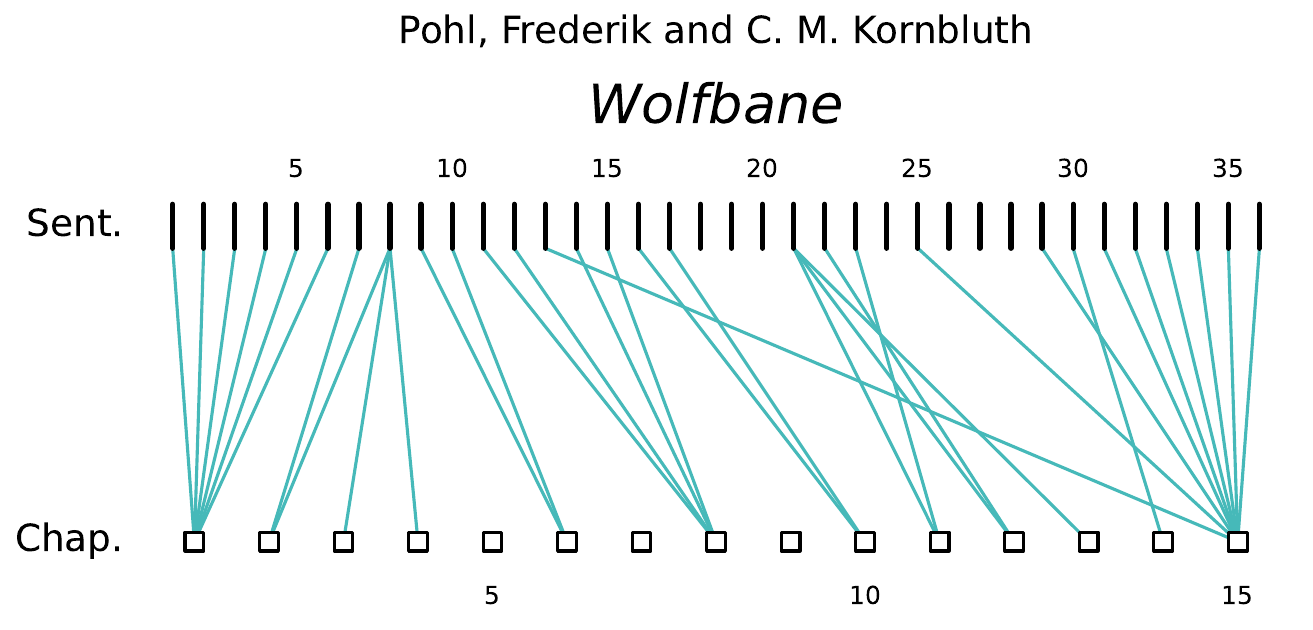}
    {\captionsetup{hypcap=false, labelformat=simple, labelsep=none, textformat=empty}\captionof{figure}{\label{fig:pg51845-bipartite}}}
  \end{minipage}\bigskip

\noindent
  \begin{minipage}[t]{0.48\linewidth}\centering
    \includegraphics[width=\linewidth]{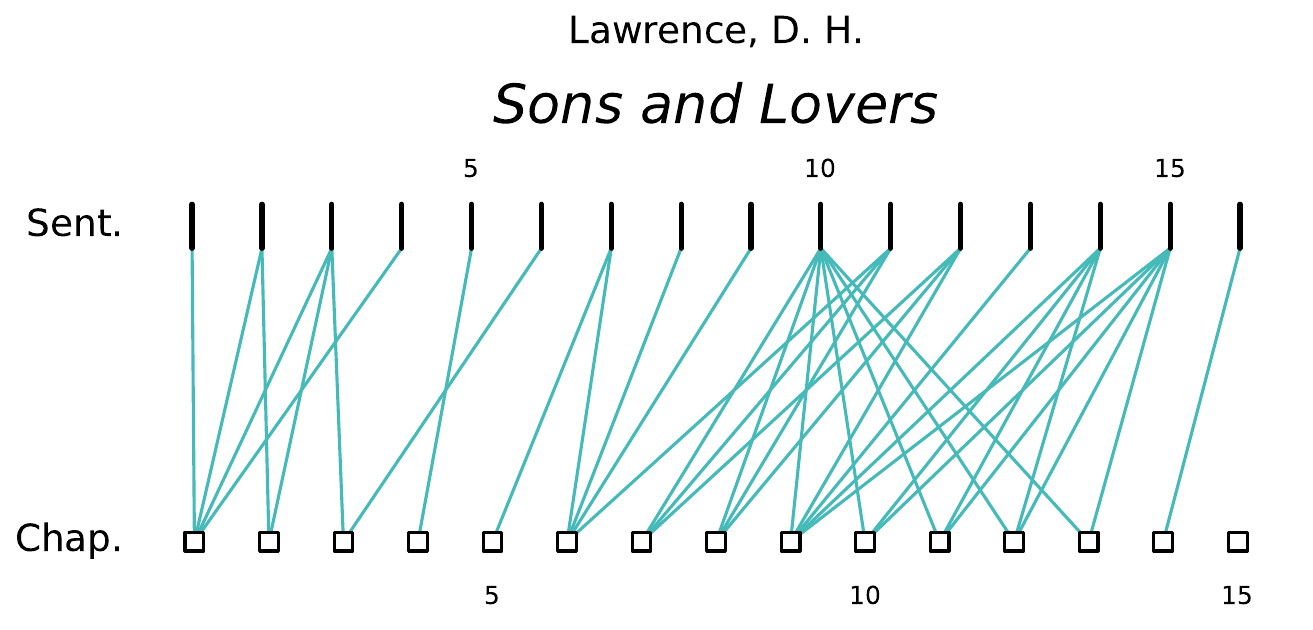}
    {\captionsetup{hypcap=false, labelformat=simple, labelsep=none, textformat=empty}\captionof{figure}{\label{fig:pg217-bipartite}}}
  \end{minipage}\hfill
  \begin{minipage}[t]{0.48\linewidth}\centering
    \includegraphics[width=\linewidth]{images/pg175_bipartite.pdf}
    {\captionsetup{hypcap=false, labelformat=simple, labelsep=none, textformat=empty}\captionof{figure}{\label{fig:pg175-bipartite}}}
  \end{minipage}\bigskip

\noindent
  \begin{minipage}[t]{0.48\linewidth}\centering
    \includegraphics[width=\linewidth]{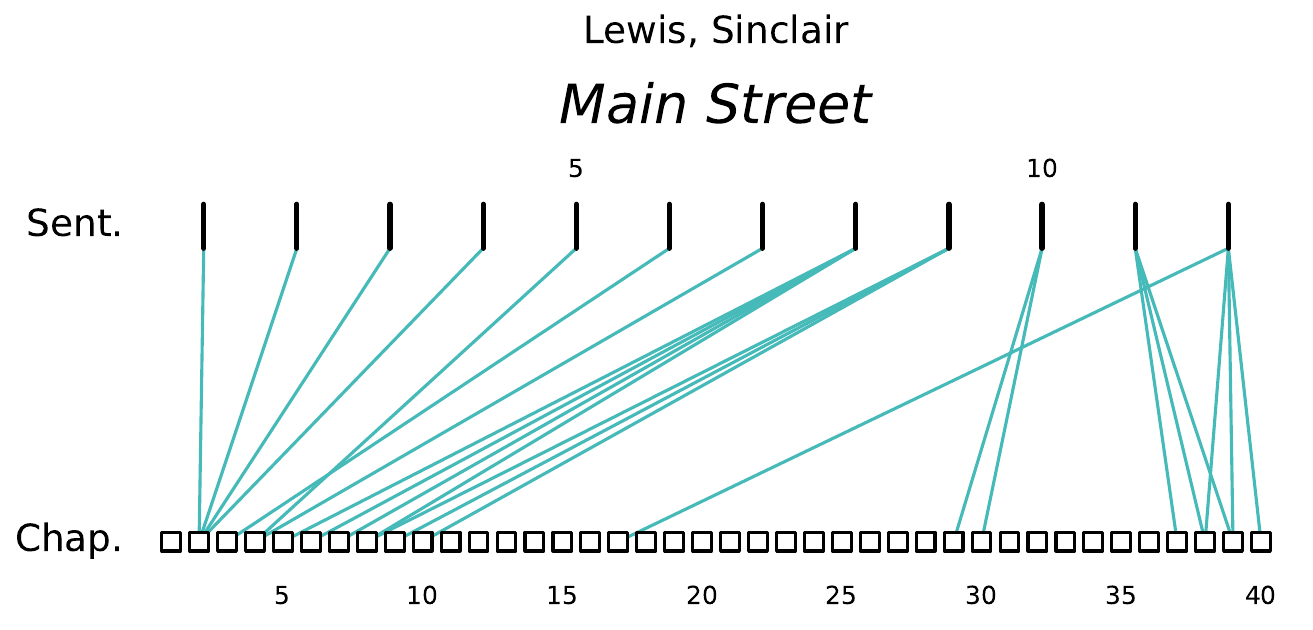}
    {\captionsetup{hypcap=false, labelformat=simple, labelsep=none, textformat=empty}\captionof{figure}{\label{fig:pg543-bipartite}}}
  \end{minipage}\hfill
  \begin{minipage}[t]{0.48\linewidth}\centering
    \includegraphics[width=\linewidth]{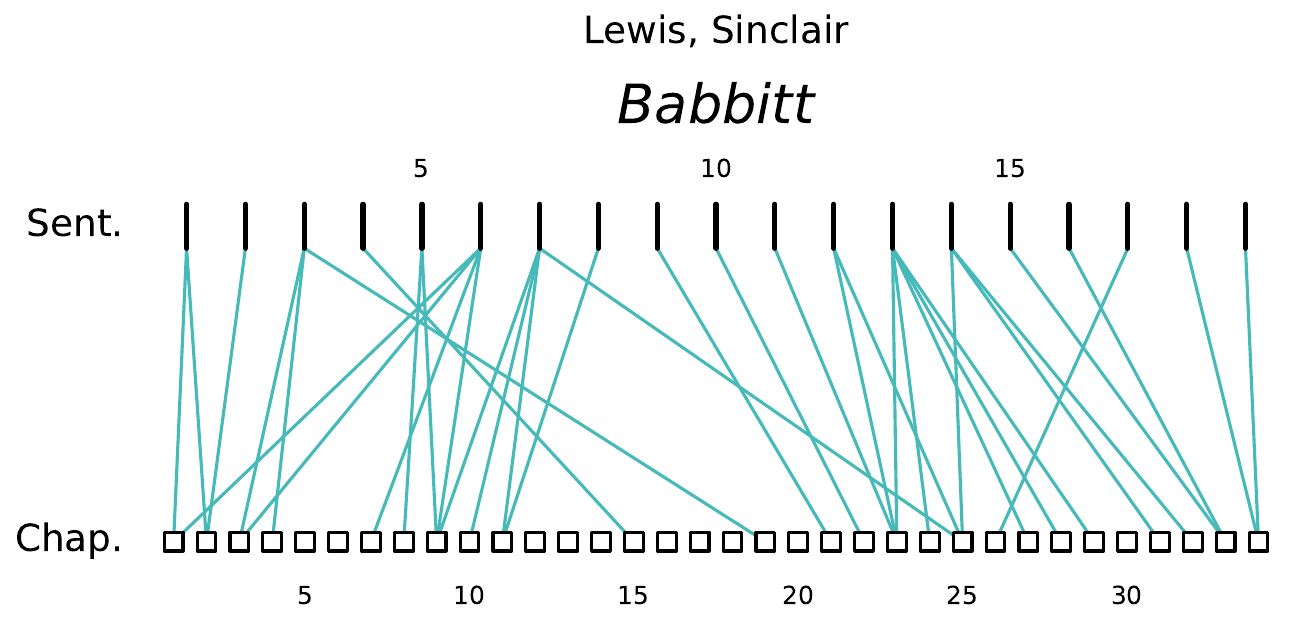}
    {\captionsetup{hypcap=false, labelformat=simple, labelsep=none, textformat=empty}\captionof{figure}{\label{fig:pg1156-bipartite}}}
  \end{minipage}\bigskip

\noindent
  \begin{minipage}[t]{0.48\linewidth}\centering
    \includegraphics[width=\linewidth]{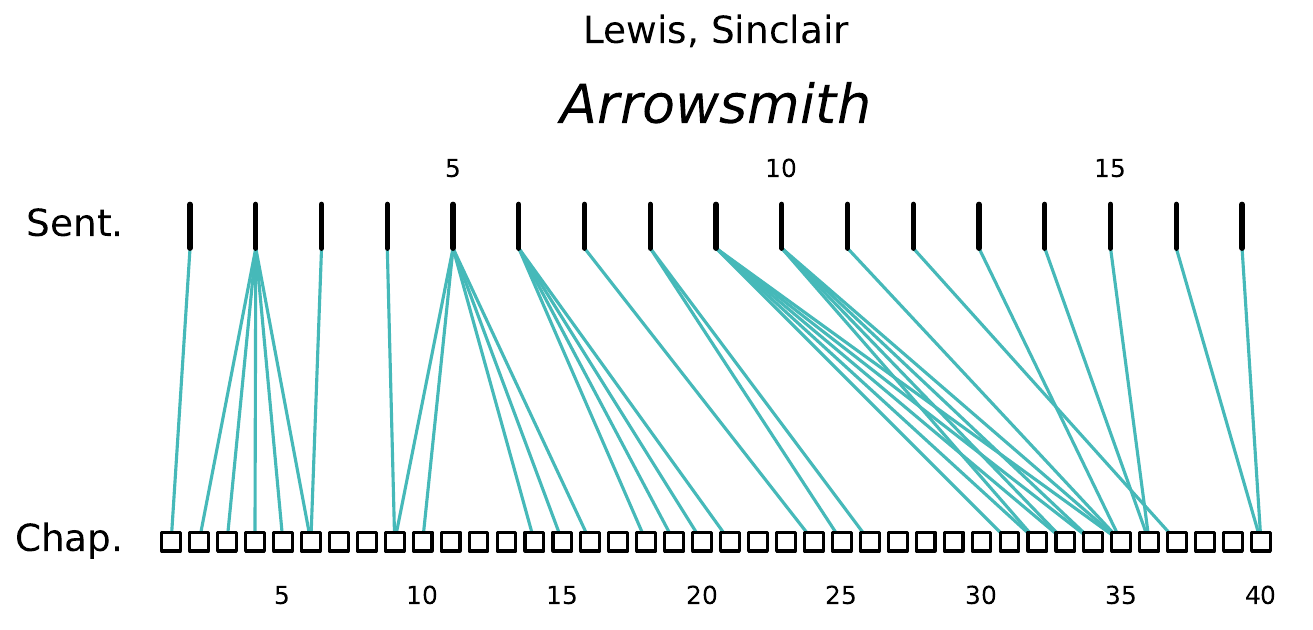}
    {\captionsetup{hypcap=false, labelformat=simple, labelsep=none, textformat=empty}\captionof{figure}{\label{fig:pg70875-bipartite}}}
  \end{minipage}\hfill
  \begin{minipage}[t]{0.48\linewidth}\centering
    \includegraphics[width=\linewidth]{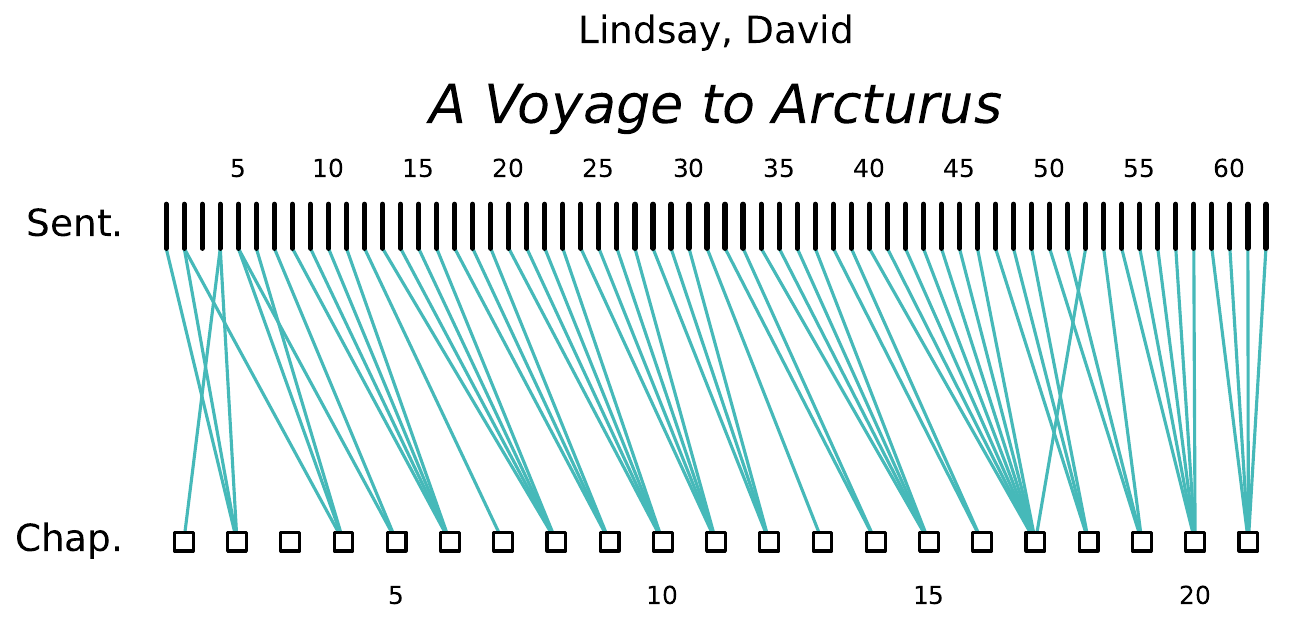}
    {\captionsetup{hypcap=false, labelformat=simple, labelsep=none, textformat=empty}\captionof{figure}{\label{fig:pg1329-bipartite}}}
  \end{minipage}\bigskip

\noindent
  \begin{minipage}[t]{0.48\linewidth}\centering
    \includegraphics[width=\linewidth]{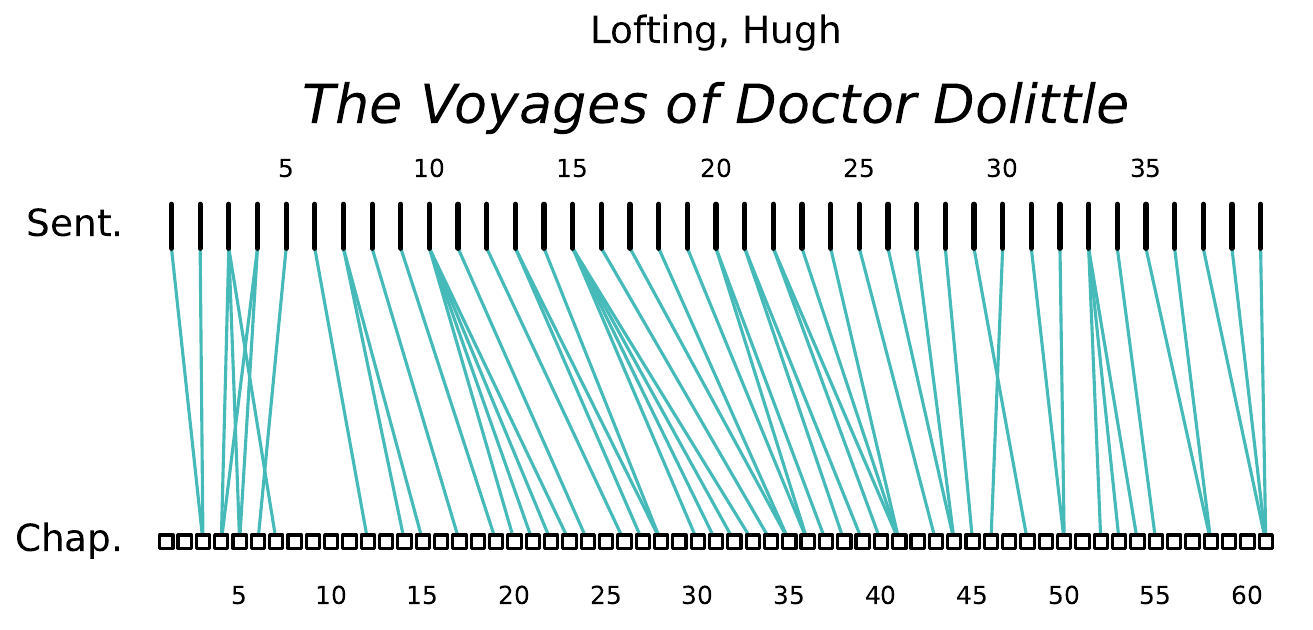}
    {\captionsetup{hypcap=false, labelformat=simple, labelsep=none, textformat=empty}\captionof{figure}{\label{fig:pg1154-bipartite}}}
  \end{minipage}\hfill
  \begin{minipage}[t]{0.48\linewidth}\centering
    \includegraphics[width=\linewidth]{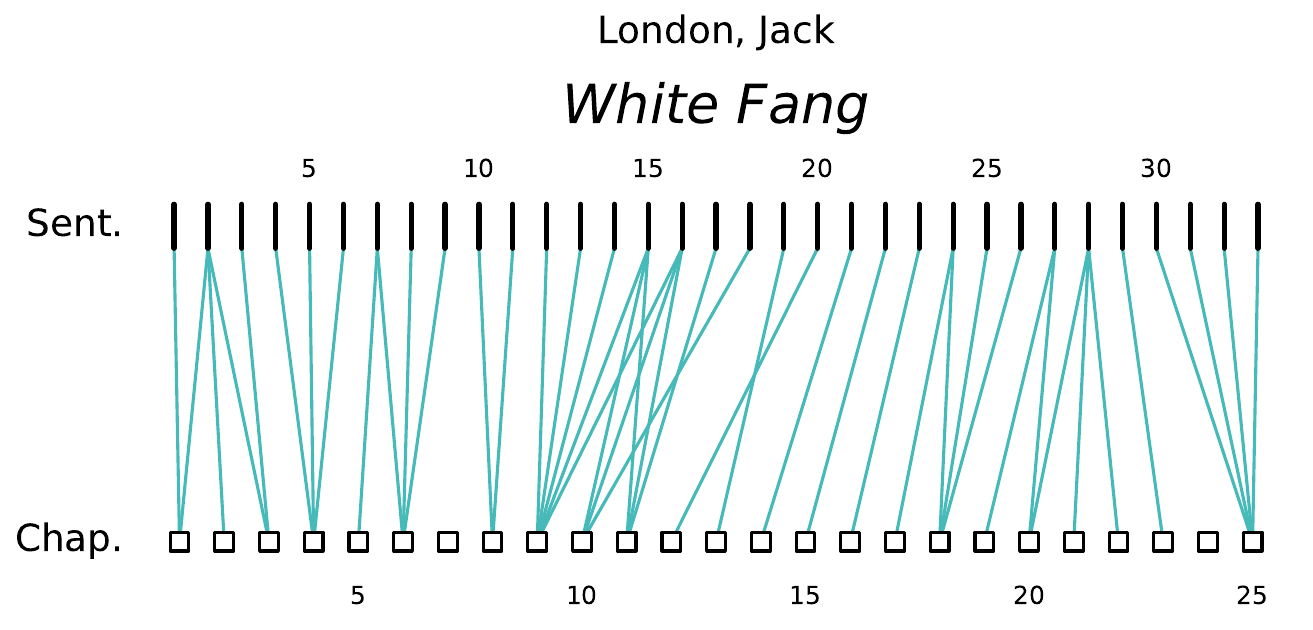}
    {\captionsetup{hypcap=false, labelformat=simple, labelsep=none, textformat=empty}\captionof{figure}{\label{fig:pg910-bipartite}}}
  \end{minipage}\bigskip

\noindent
  \begin{minipage}[t]{0.48\linewidth}\centering
    \includegraphics[width=\linewidth]{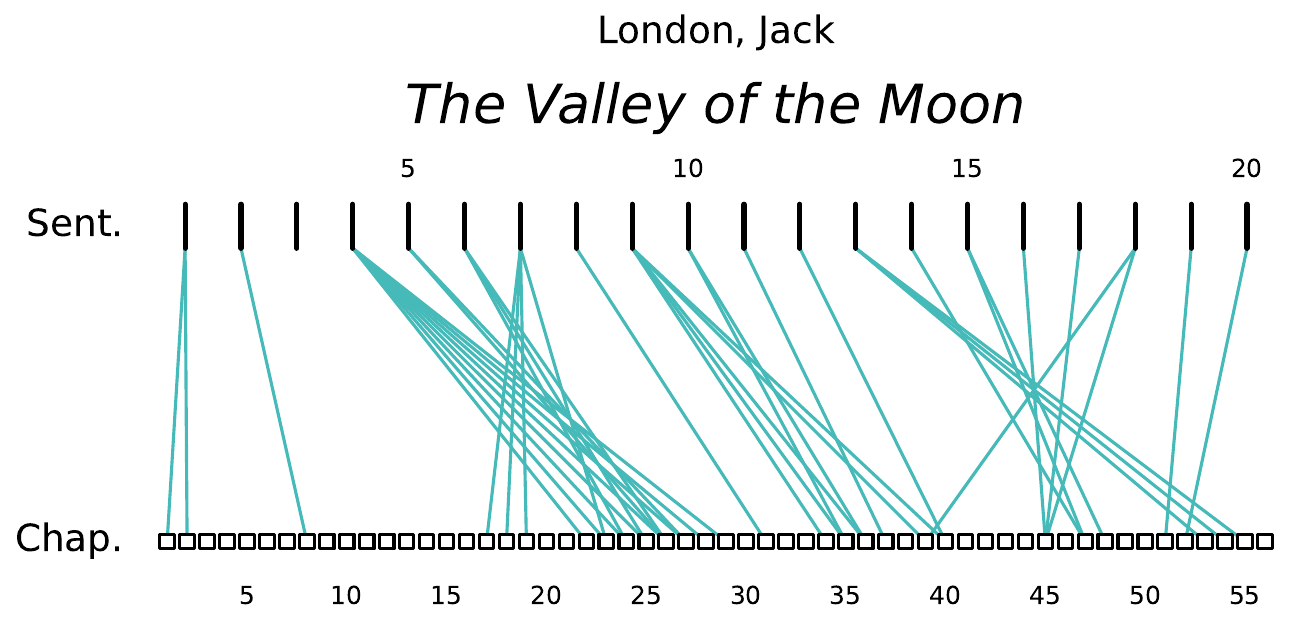}
    {\captionsetup{hypcap=false, labelformat=simple, labelsep=none, textformat=empty}\captionof{figure}{\label{fig:pg1449-bipartite}}}
  \end{minipage}\hfill
  \begin{minipage}[t]{0.48\linewidth}\centering
    \includegraphics[width=\linewidth]{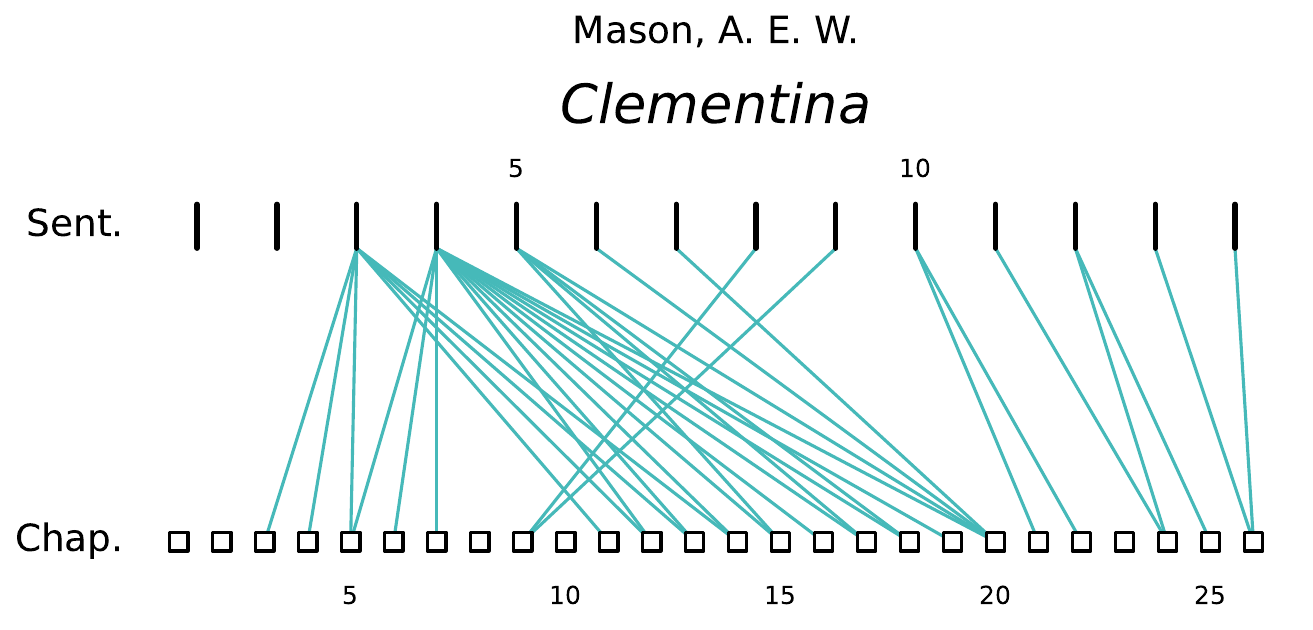}
    {\captionsetup{hypcap=false, labelformat=simple, labelsep=none, textformat=empty}\captionof{figure}{\label{fig:pg13567-bipartite}}}
  \end{minipage}\bigskip

\noindent
  \begin{minipage}[t]{0.48\linewidth}\centering
    \includegraphics[width=\linewidth]{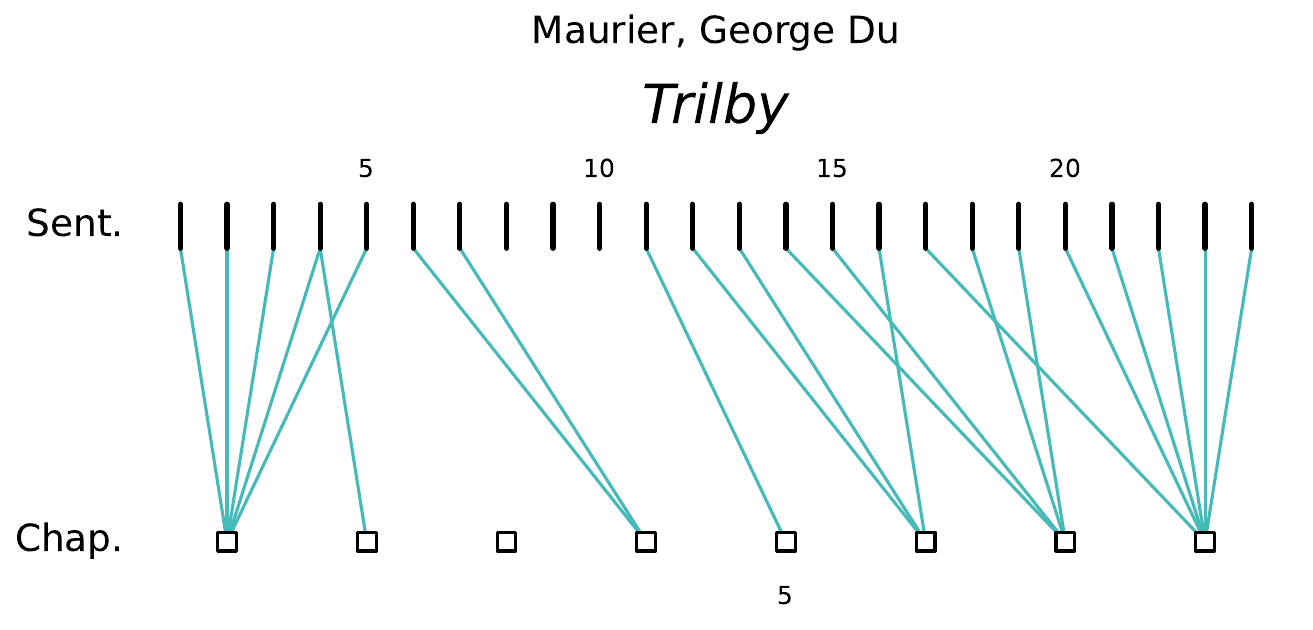}
    {\captionsetup{hypcap=false, labelformat=simple, labelsep=none, textformat=empty}\captionof{figure}{\label{fig:pg39858-bipartite}}}
  \end{minipage}\hfill
  \begin{minipage}[t]{0.48\linewidth}\centering
    \includegraphics[width=\linewidth]{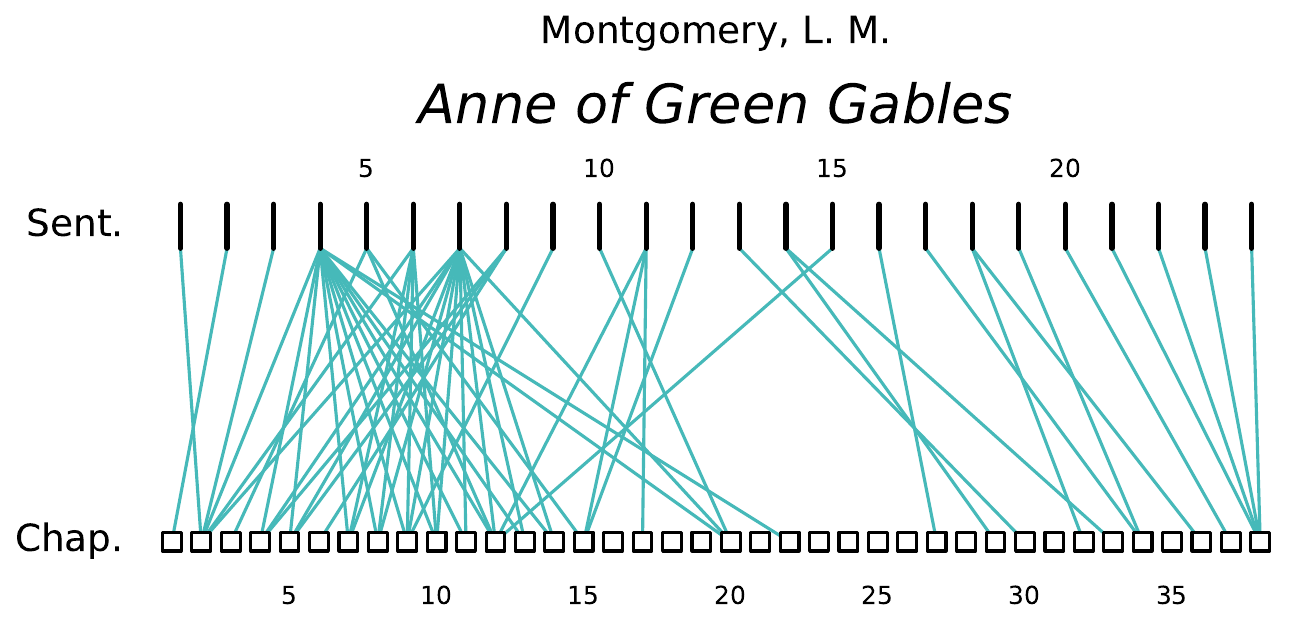}
    {\captionsetup{hypcap=false, labelformat=simple, labelsep=none, textformat=empty}\captionof{figure}{\label{fig:pg45-bipartite}}}
  \end{minipage}\bigskip

\noindent
  \begin{minipage}[t]{0.48\linewidth}\centering
    \includegraphics[width=\linewidth]{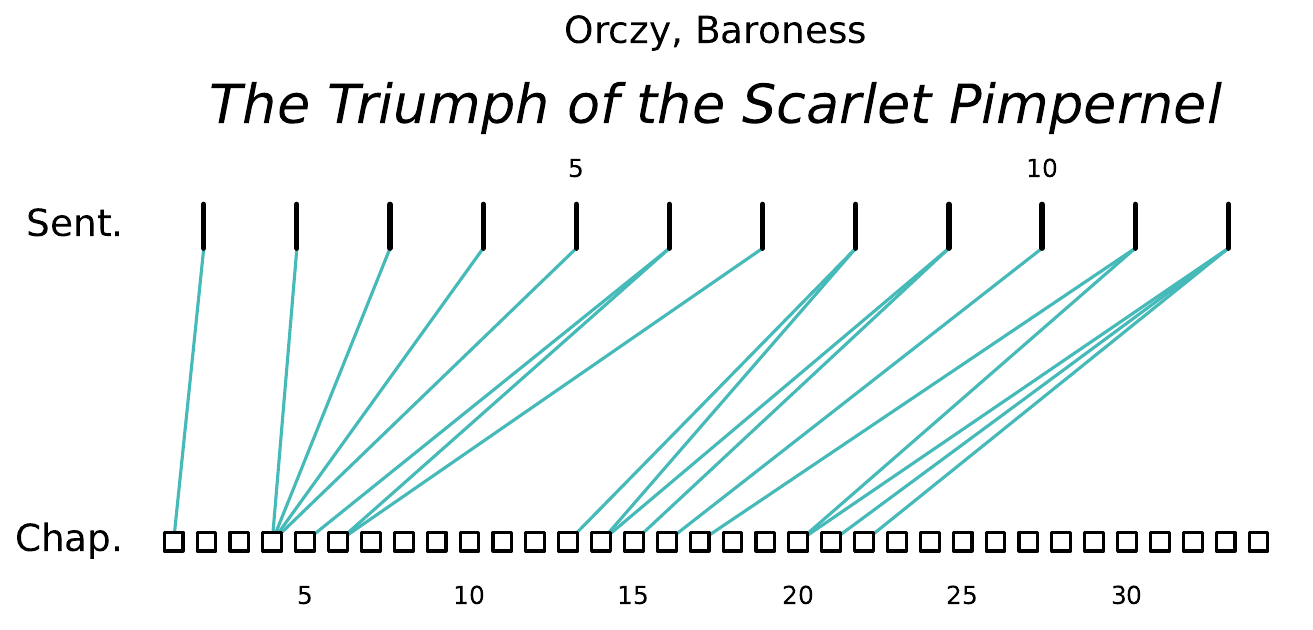}
    {\captionsetup{hypcap=false, labelformat=simple, labelsep=none, textformat=empty}\captionof{figure}{\label{fig:pg65695-bipartite}}}
  \end{minipage}\hfill
  \begin{minipage}[t]{0.48\linewidth}\centering
    \includegraphics[width=\linewidth]{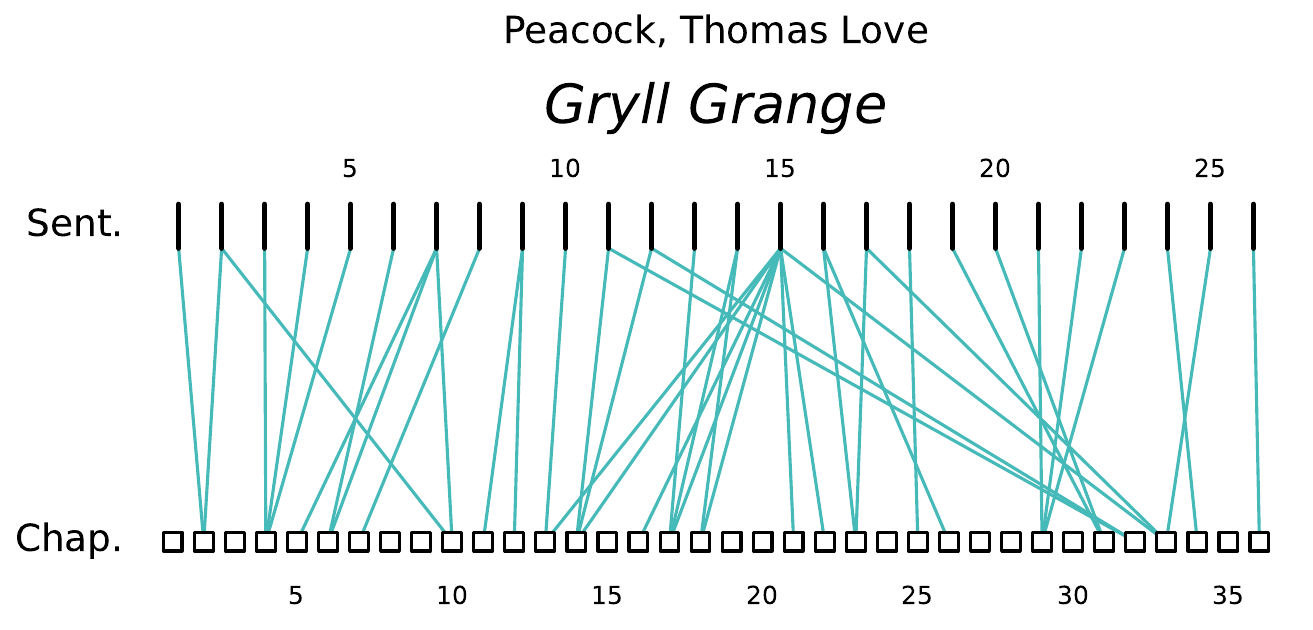}
    {\captionsetup{hypcap=false, labelformat=simple, labelsep=none, textformat=empty}\captionof{figure}{\label{fig:pg21514-bipartite}}}
  \end{minipage}\bigskip

\noindent
  \begin{minipage}[t]{0.48\linewidth}\centering
    \includegraphics[width=\linewidth]{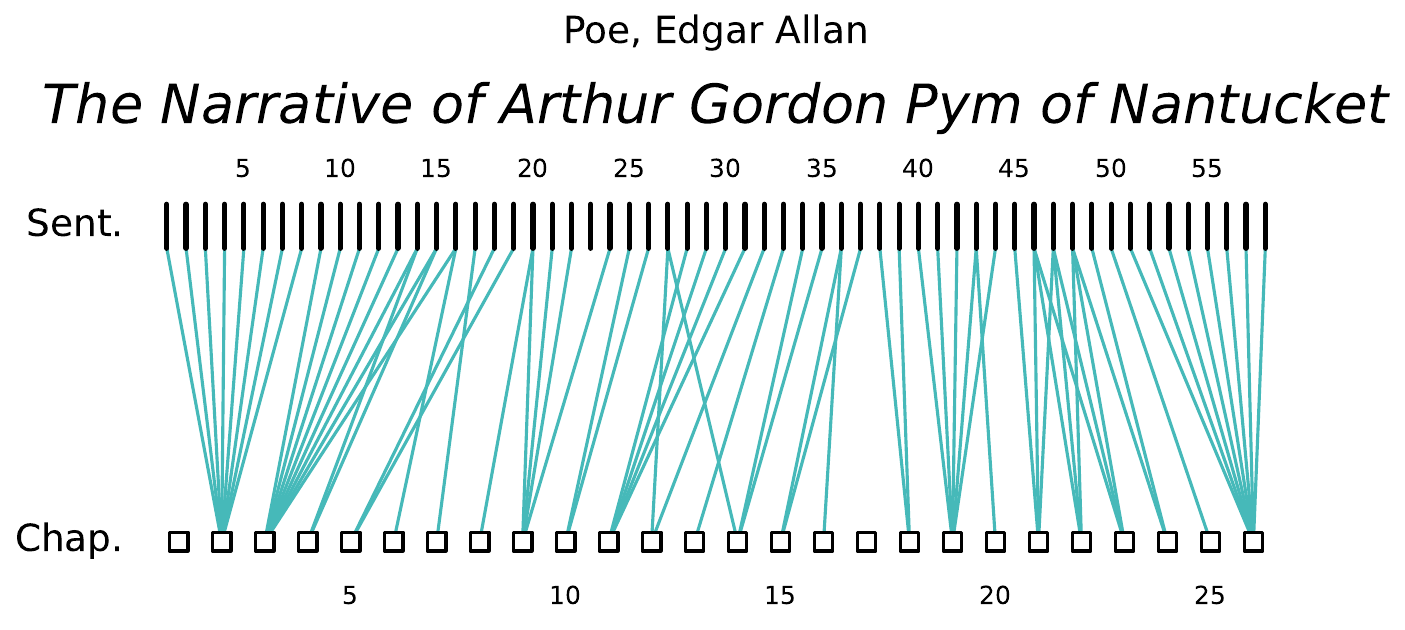}
    {\captionsetup{hypcap=false, labelformat=simple, labelsep=none, textformat=empty}\captionof{figure}{\label{fig:pg51060-bipartite}}}
  \end{minipage}\hfill
  \begin{minipage}[t]{0.48\linewidth}\centering
    \includegraphics[width=\linewidth]{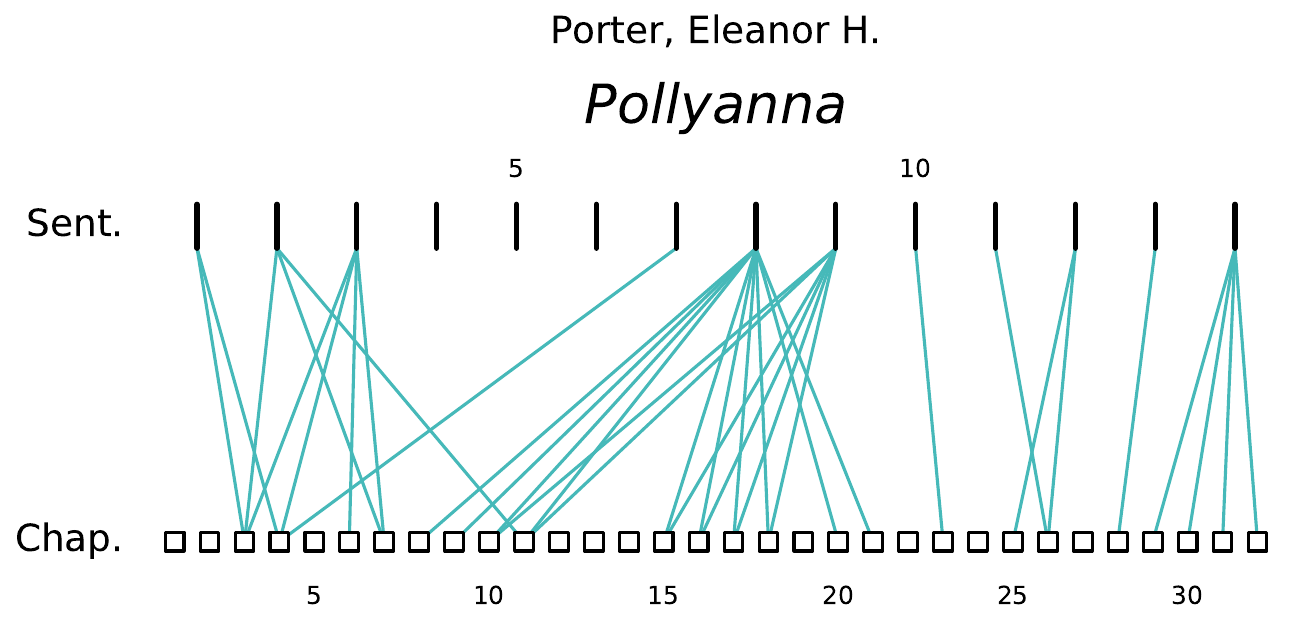}
    {\captionsetup{hypcap=false, labelformat=simple, labelsep=none, textformat=empty}\captionof{figure}{\label{fig:pg1450-bipartite}}}
  \end{minipage}\bigskip

\noindent
  \begin{minipage}[t]{0.48\linewidth}\centering
    \includegraphics[width=\linewidth]{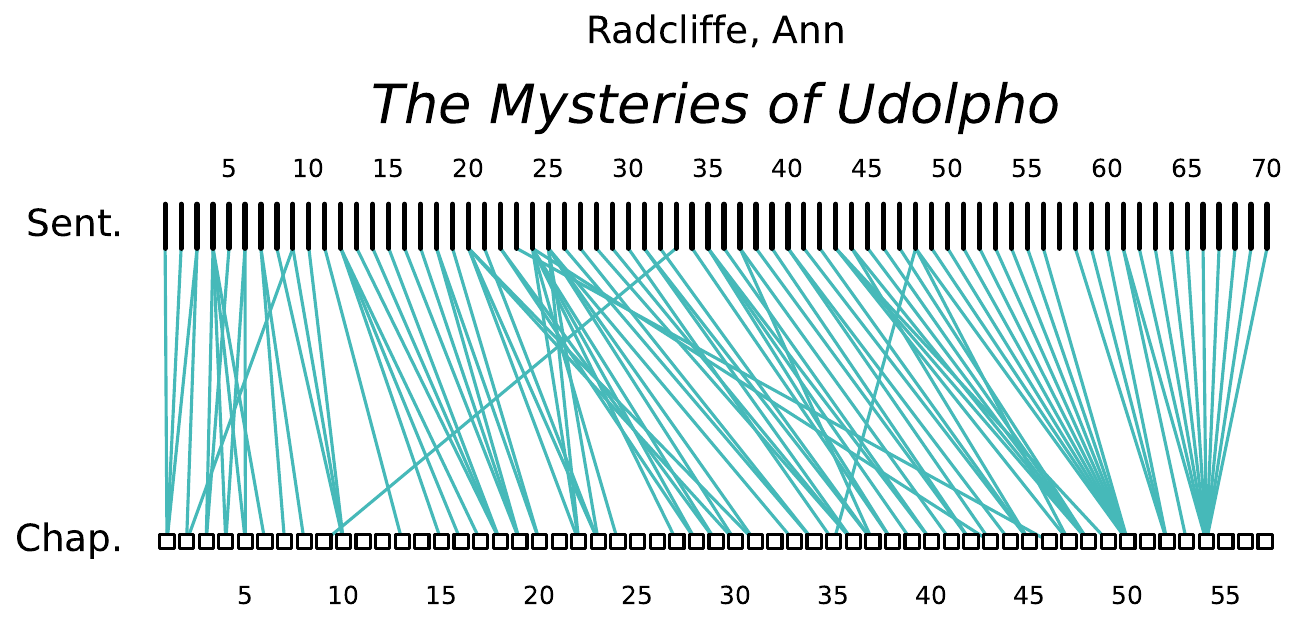}
    {\captionsetup{hypcap=false, labelformat=simple, labelsep=none, textformat=empty}\captionof{figure}{\label{fig:pg3268-bipartite}}}
  \end{minipage}\hfill
  \begin{minipage}[t]{0.48\linewidth}\centering
    \includegraphics[width=\linewidth]{images/pg1250_bipartite.pdf}
    {\captionsetup{hypcap=false, labelformat=simple, labelsep=none, textformat=empty}\captionof{figure}{\label{fig:pg1250-bipartite}}}
  \end{minipage}\bigskip

\noindent
  \begin{minipage}[t]{0.48\linewidth}\centering
    \includegraphics[width=\linewidth]{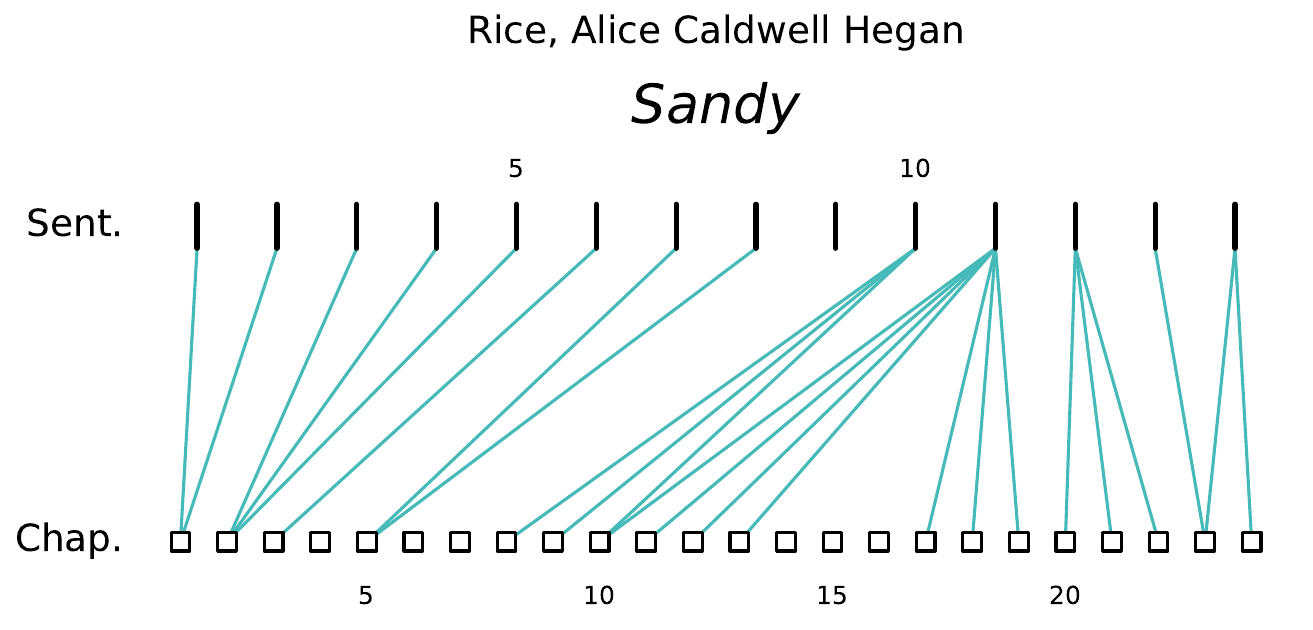}
    {\captionsetup{hypcap=false, labelformat=simple, labelsep=none, textformat=empty}\captionof{figure}{\label{fig:pg14079-bipartite}}}
  \end{minipage}\hfill
  \begin{minipage}[t]{0.48\linewidth}\centering
    \includegraphics[width=\linewidth]{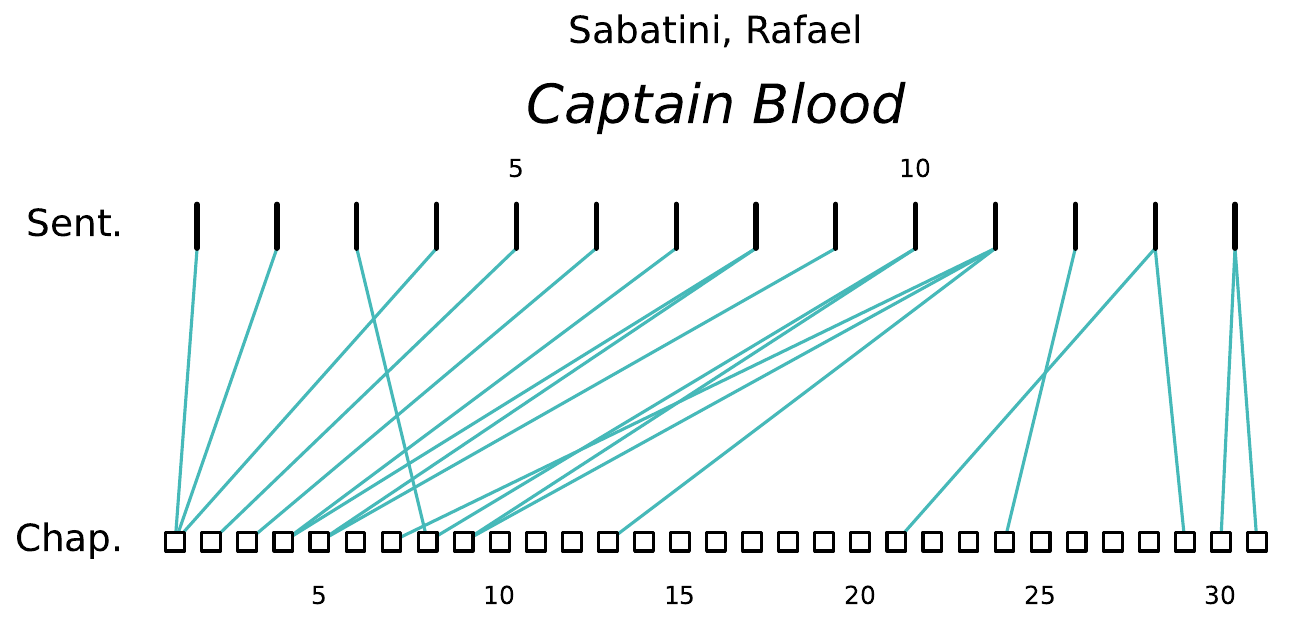}
    {\captionsetup{hypcap=false, labelformat=simple, labelsep=none, textformat=empty}\captionof{figure}{\label{fig:pg1965-bipartite}}}
  \end{minipage}\bigskip

\noindent
  \begin{minipage}[t]{0.48\linewidth}\centering
    \includegraphics[width=\linewidth]{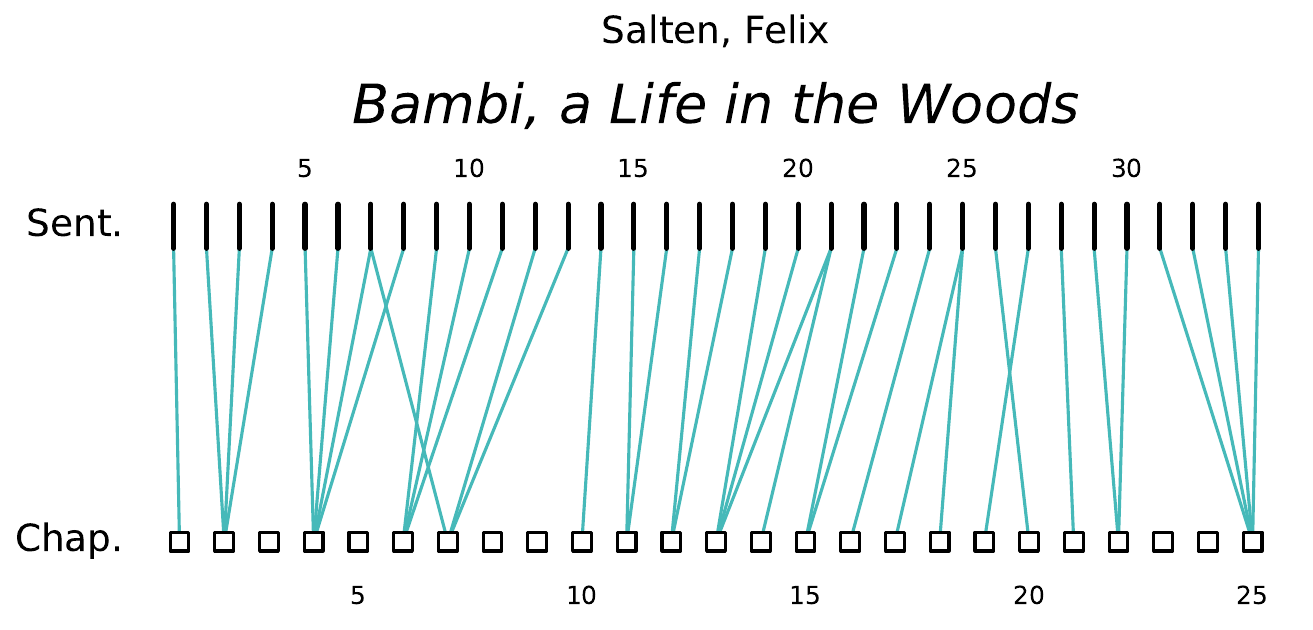}
    {\captionsetup{hypcap=false, labelformat=simple, labelsep=none, textformat=empty}\captionof{figure}{\label{fig:pg63849-bipartite}}}
  \end{minipage}\hfill
  \begin{minipage}[t]{0.48\linewidth}\centering
    \includegraphics[width=\linewidth]{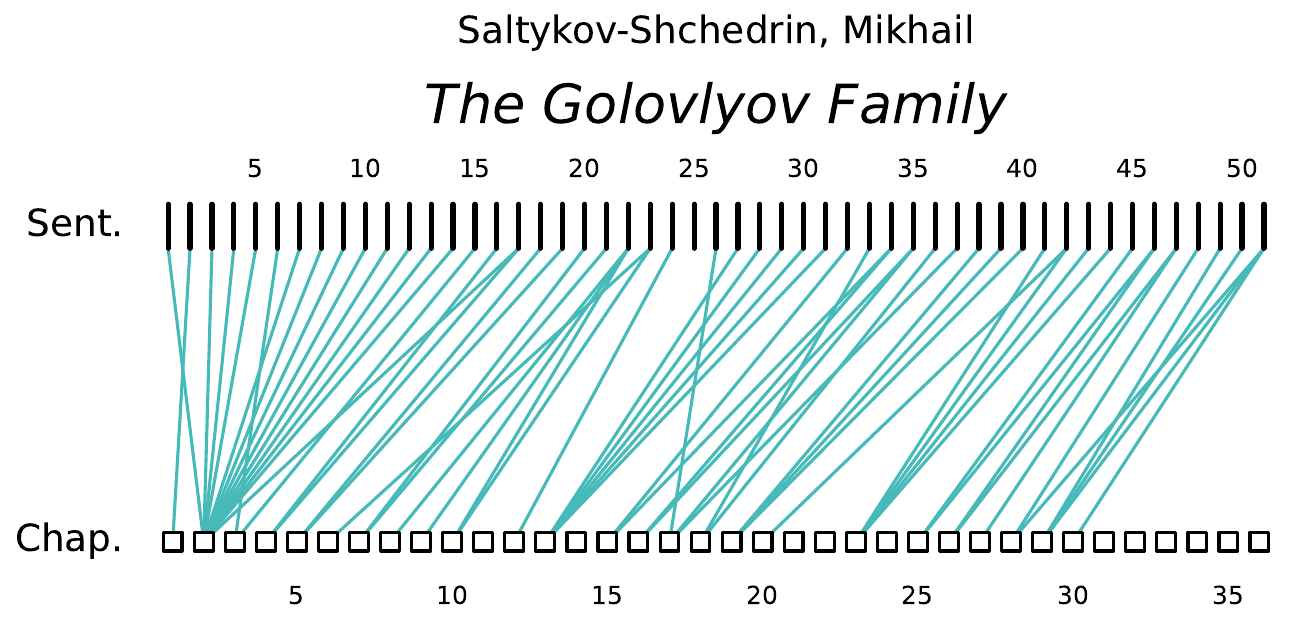}
    {\captionsetup{hypcap=false, labelformat=simple, labelsep=none, textformat=empty}\captionof{figure}{\label{fig:pg44237-bipartite}}}
  \end{minipage}\bigskip

\noindent
  \begin{minipage}[t]{0.48\linewidth}\centering
    \includegraphics[width=\linewidth]{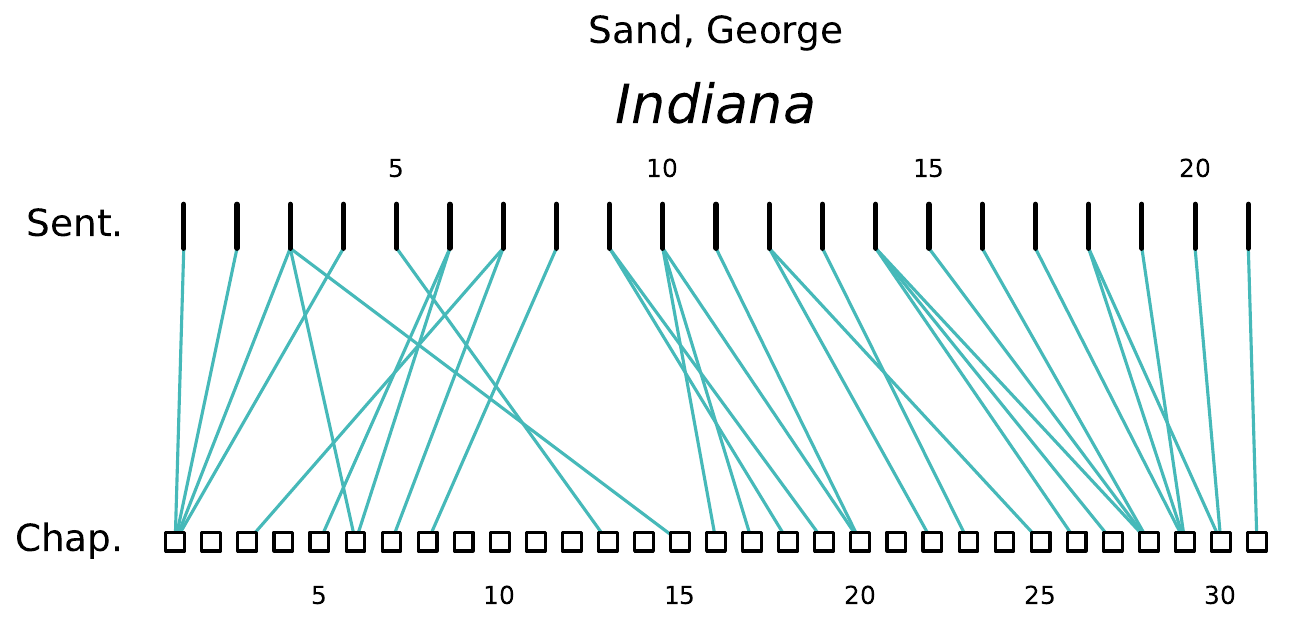}
    {\captionsetup{hypcap=false, labelformat=simple, labelsep=none, textformat=empty}\captionof{figure}{\label{fig:pg63445-bipartite}}}
  \end{minipage}\hfill
  \begin{minipage}[t]{0.48\linewidth}\centering
    \includegraphics[width=\linewidth]{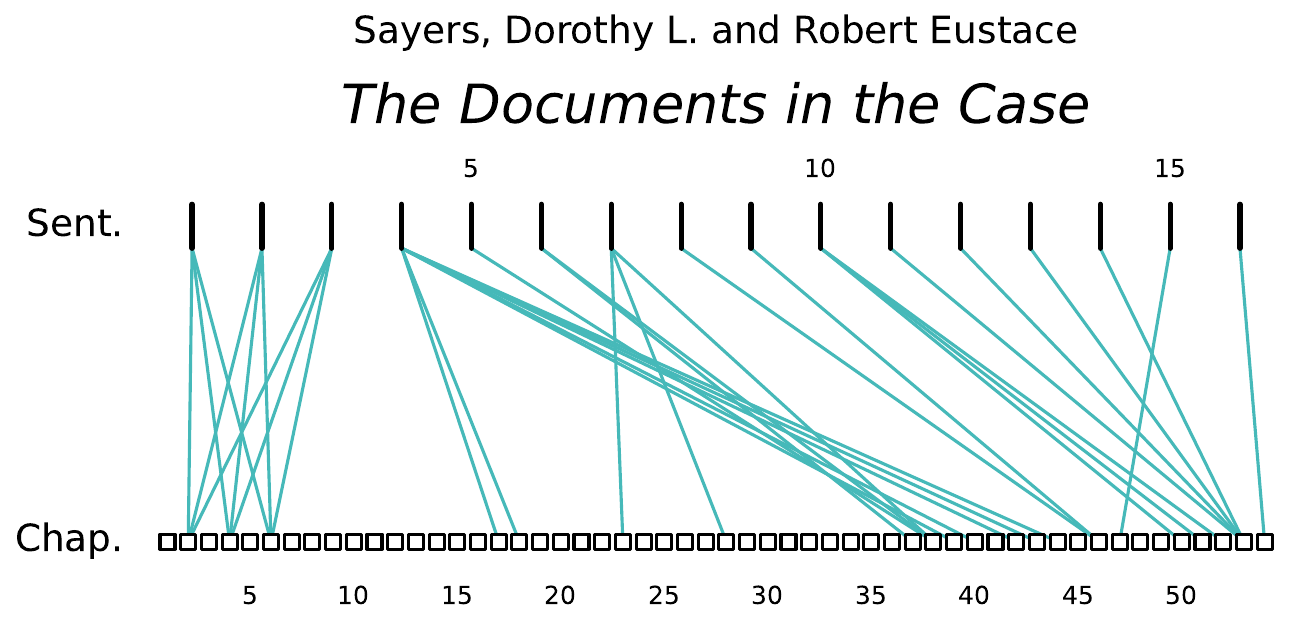}
    {\captionsetup{hypcap=false, labelformat=simple, labelsep=none, textformat=empty}\captionof{figure}{\label{fig:pg77601-bipartite}}}
  \end{minipage}\bigskip

\noindent
  \begin{minipage}[t]{0.48\linewidth}\centering
    \includegraphics[width=\linewidth]{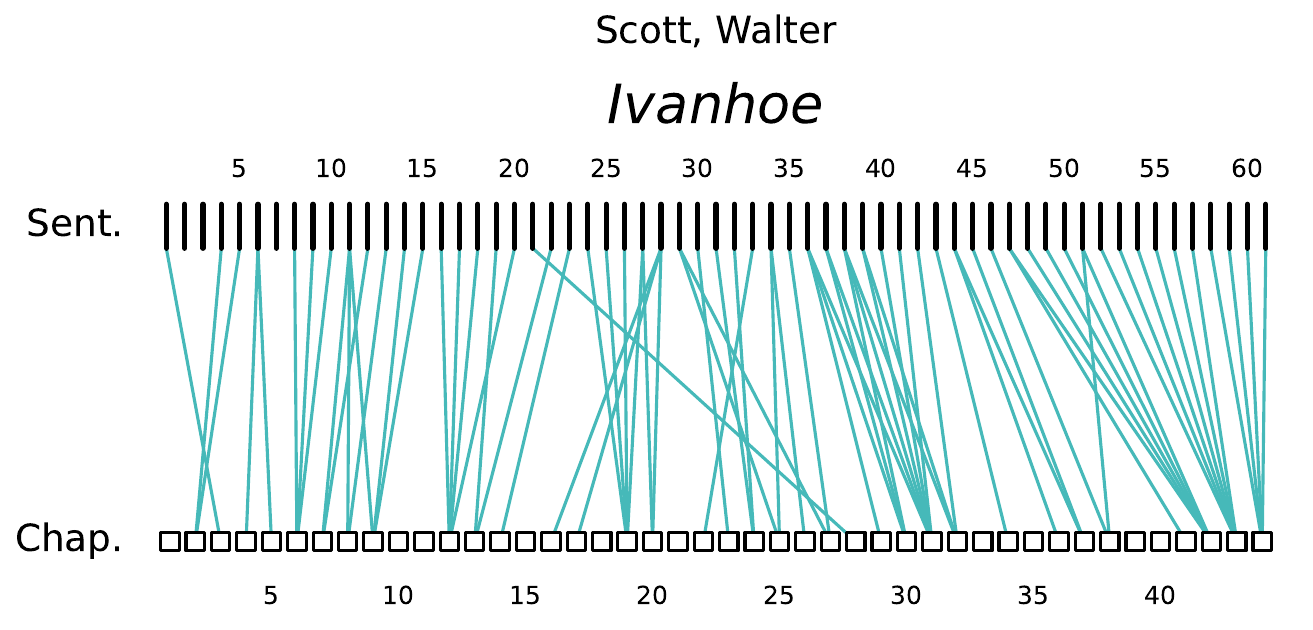}
    {\captionsetup{hypcap=false, labelformat=simple, labelsep=none, textformat=empty}\captionof{figure}{\label{fig:pg82-bipartite}}}
  \end{minipage}\hfill
  \begin{minipage}[t]{0.48\linewidth}\centering
    \includegraphics[width=\linewidth]{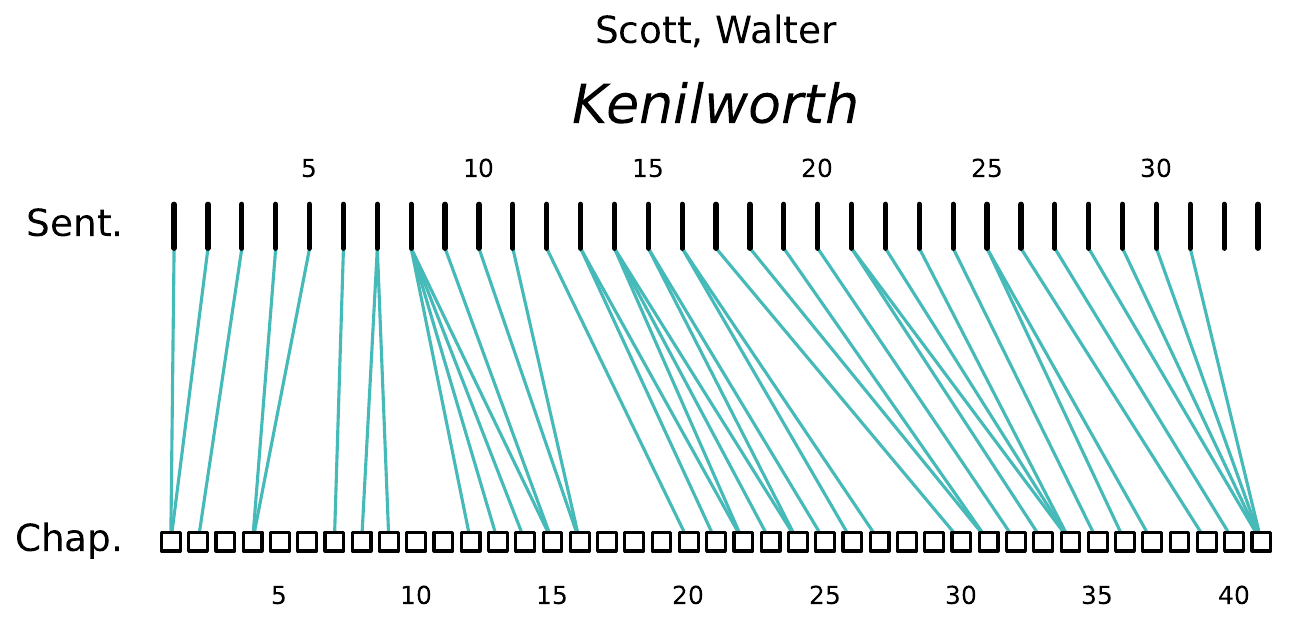}
    {\captionsetup{hypcap=false, labelformat=simple, labelsep=none, textformat=empty}\captionof{figure}{\label{fig:pg1606-bipartite}}}
  \end{minipage}\bigskip

\noindent
  \begin{minipage}[t]{0.48\linewidth}\centering
    \includegraphics[width=\linewidth]{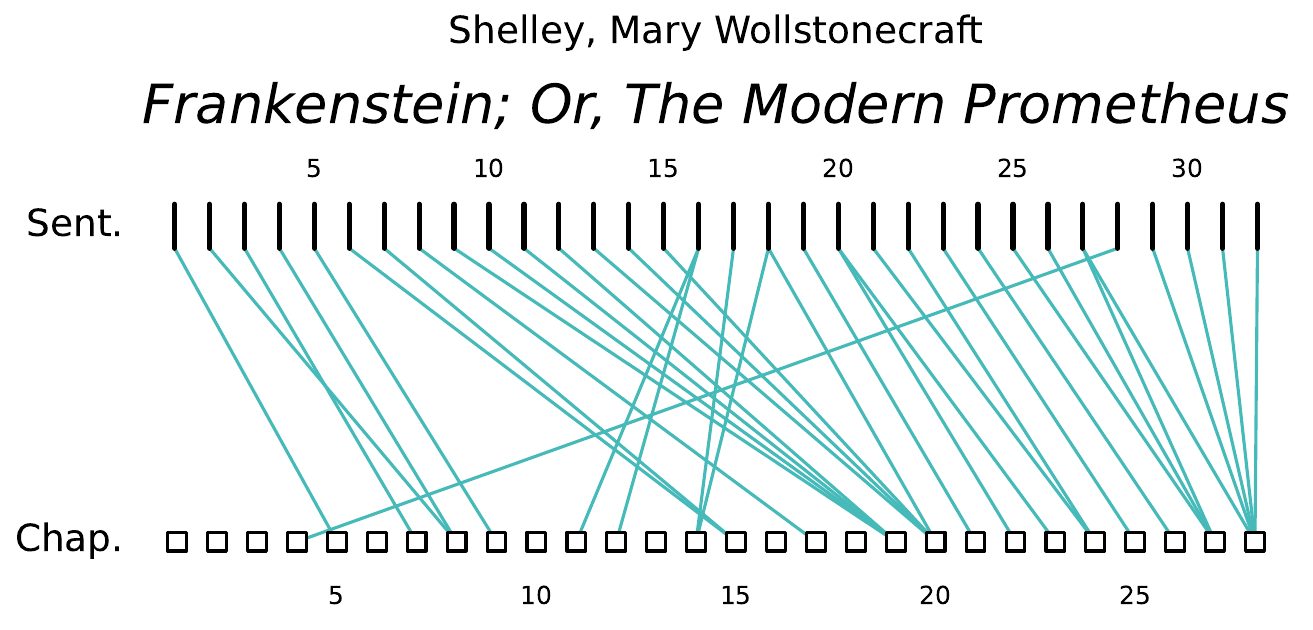}
    {\captionsetup{hypcap=false, labelformat=simple, labelsep=none, textformat=empty}\captionof{figure}{\label{fig:pg84-bipartite}}}
  \end{minipage}\hfill
  \begin{minipage}[t]{0.48\linewidth}\centering
    \includegraphics[width=\linewidth]{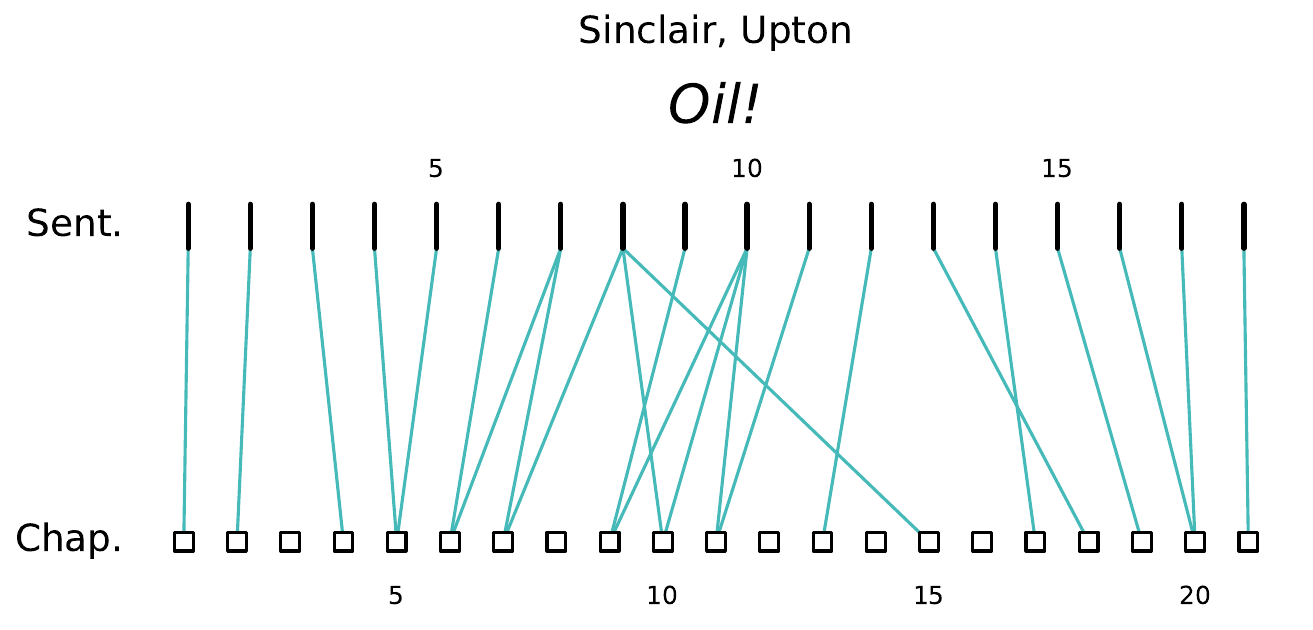}
    {\captionsetup{hypcap=false, labelformat=simple, labelsep=none, textformat=empty}\captionof{figure}{\label{fig:pg70379-bipartite}}}
  \end{minipage}\bigskip

\noindent
  \begin{minipage}[t]{0.48\linewidth}\centering
    \includegraphics[width=\linewidth]{images/pg2160_bipartite.pdf}
    {\captionsetup{hypcap=false, labelformat=simple, labelsep=none, textformat=empty}\captionof{figure}{\label{fig:pg2160-bipartite}}}
  \end{minipage}\hfill
  \begin{minipage}[t]{0.48\linewidth}\centering
    \includegraphics[width=\linewidth]{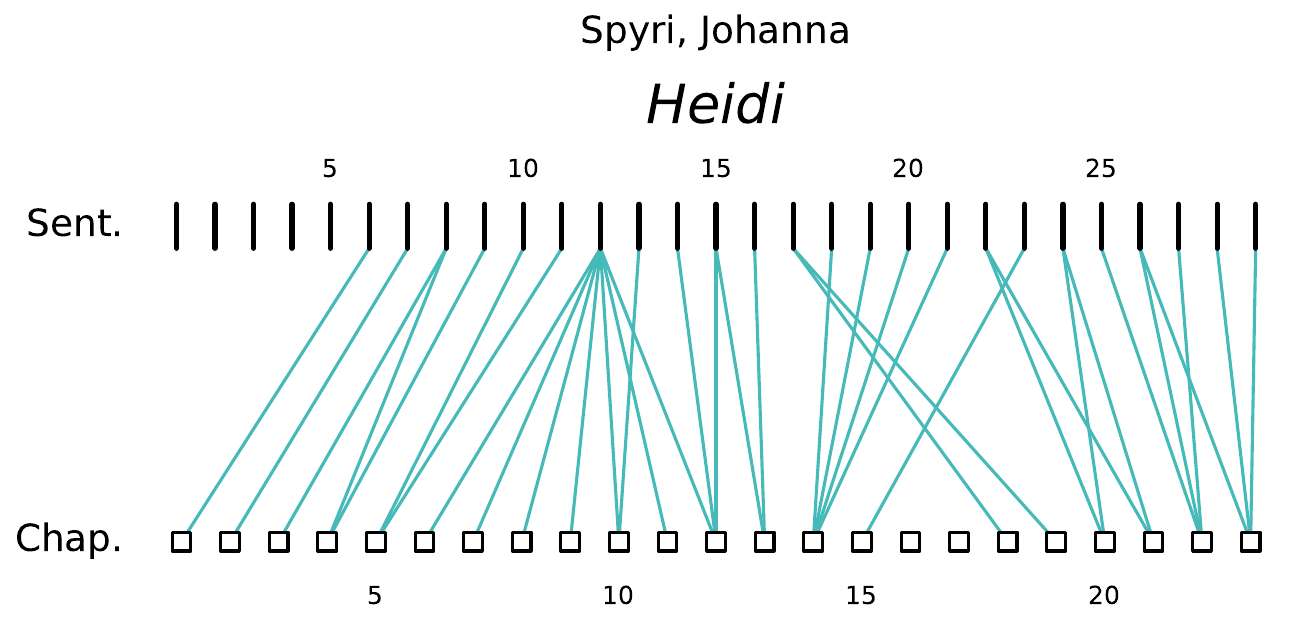}
    {\captionsetup{hypcap=false, labelformat=simple, labelsep=none, textformat=empty}\captionof{figure}{\label{fig:pg1448-bipartite}}}
  \end{minipage}\bigskip

\noindent
  \begin{minipage}[t]{0.48\linewidth}\centering
    \includegraphics[width=\linewidth]{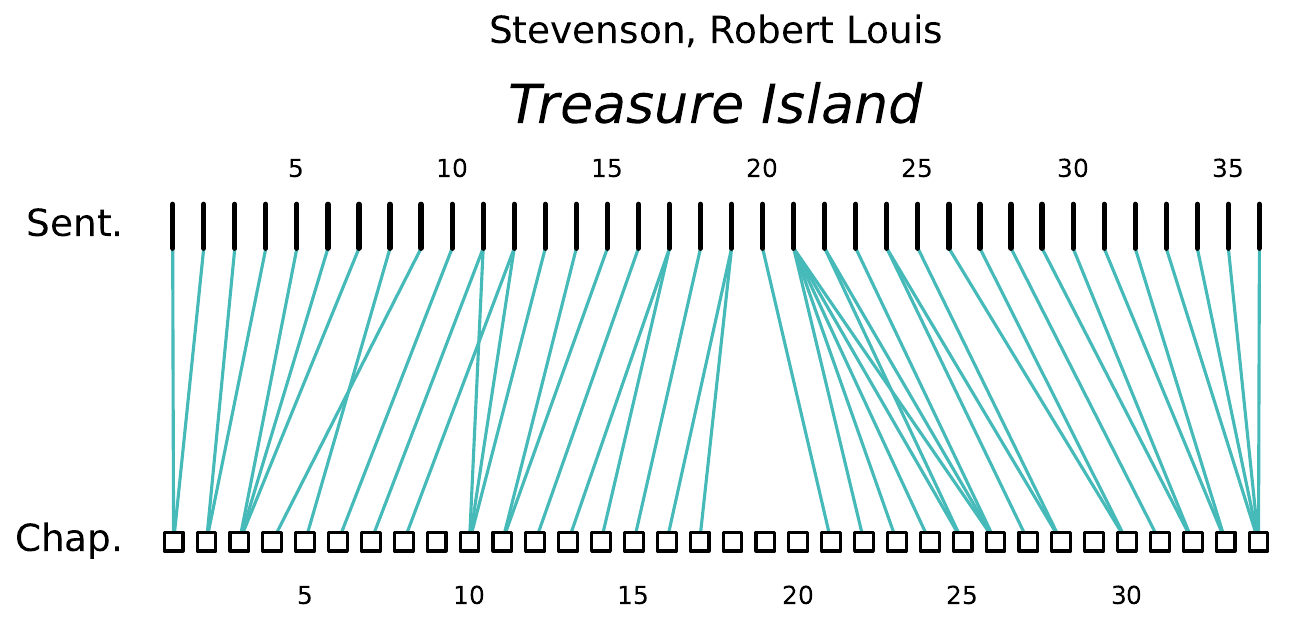}
    {\captionsetup{hypcap=false, labelformat=simple, labelsep=none, textformat=empty}\captionof{figure}{\label{fig:pg120-bipartite}}}
  \end{minipage}\hfill
  \begin{minipage}[t]{0.48\linewidth}\centering
    \includegraphics[width=\linewidth]{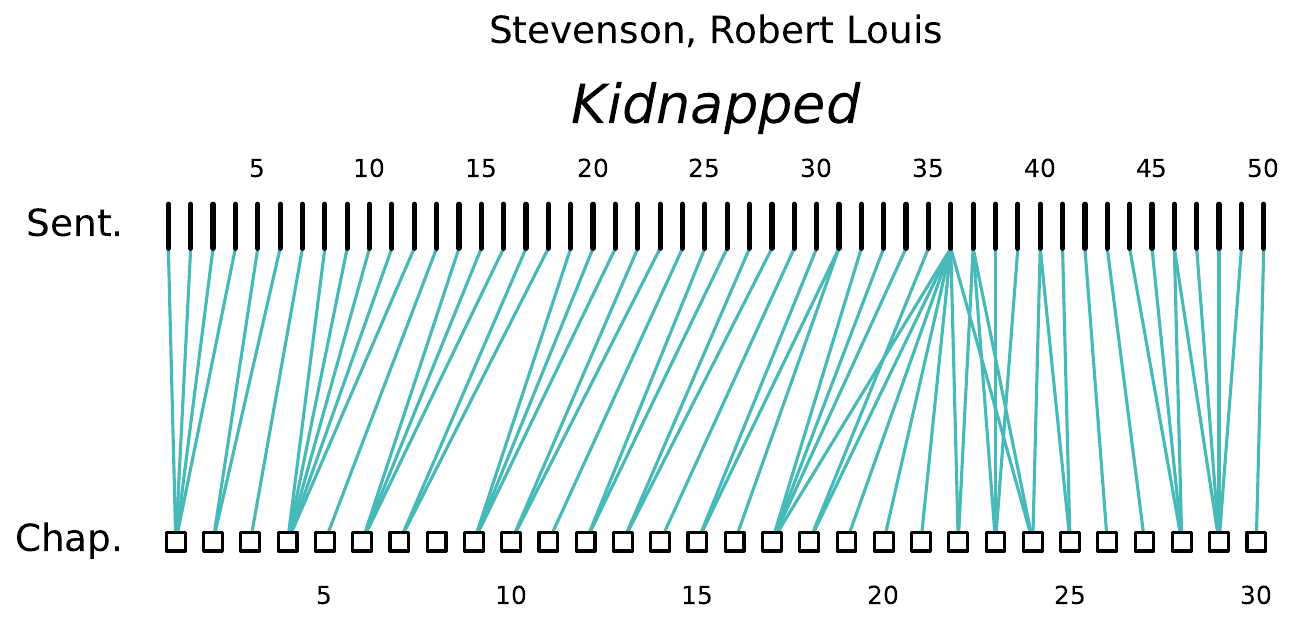}
    {\captionsetup{hypcap=false, labelformat=simple, labelsep=none, textformat=empty}\captionof{figure}{\label{fig:pg421-bipartite}}}
  \end{minipage}\bigskip

\noindent
  \begin{minipage}[t]{0.48\linewidth}\centering
    \includegraphics[width=\linewidth]{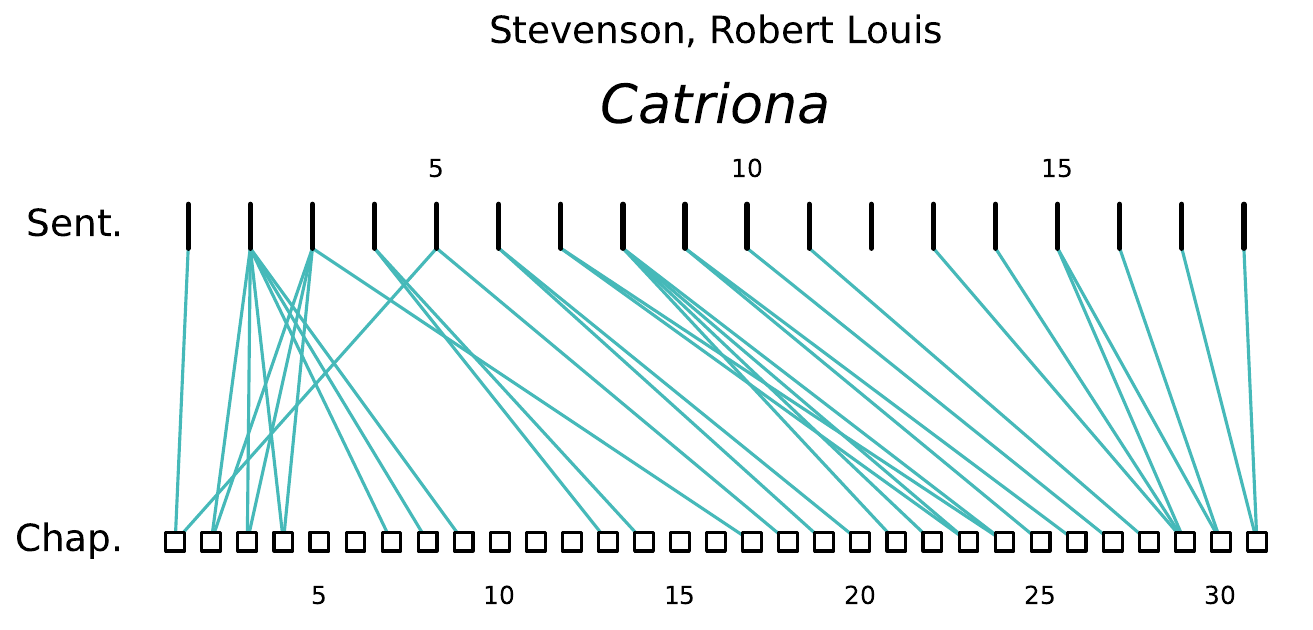}
    {\captionsetup{hypcap=false, labelformat=simple, labelsep=none, textformat=empty}\captionof{figure}{\label{fig:pg589-bipartite}}}
  \end{minipage}\hfill
  \begin{minipage}[t]{0.48\linewidth}\centering
    \includegraphics[width=\linewidth]{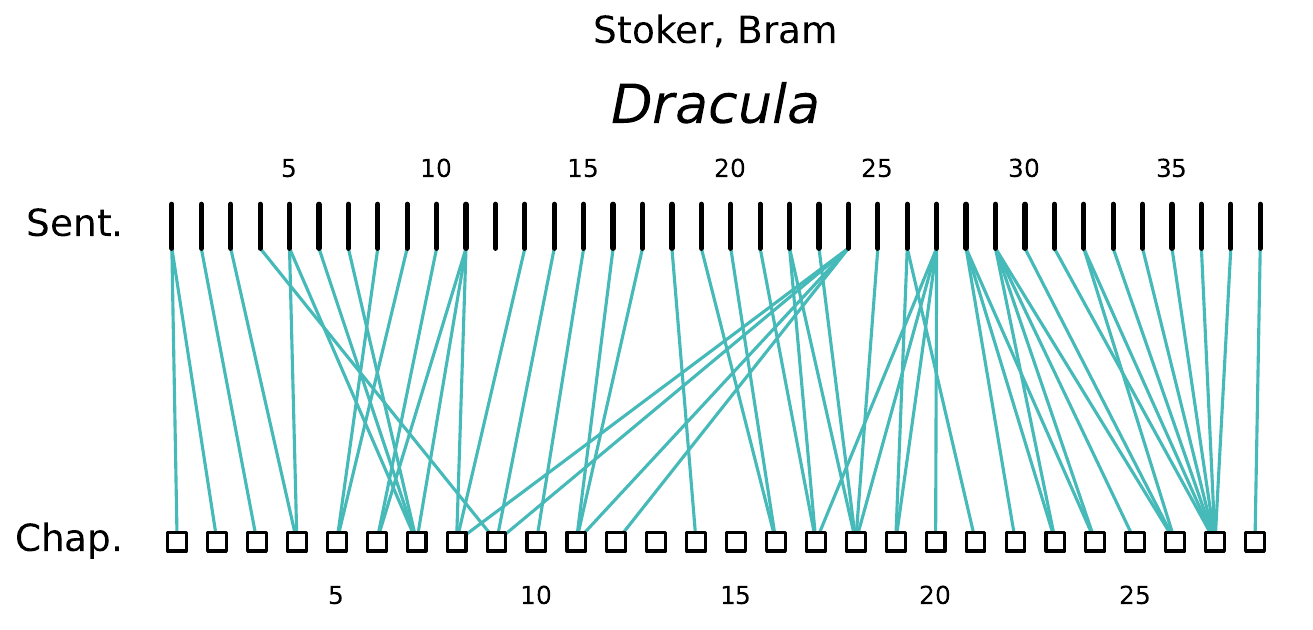}
    {\captionsetup{hypcap=false, labelformat=simple, labelsep=none, textformat=empty}\captionof{figure}{\label{fig:pg45839-bipartite}}}
  \end{minipage}\bigskip

\noindent
  \begin{minipage}[t]{0.48\linewidth}\centering
    \includegraphics[width=\linewidth]{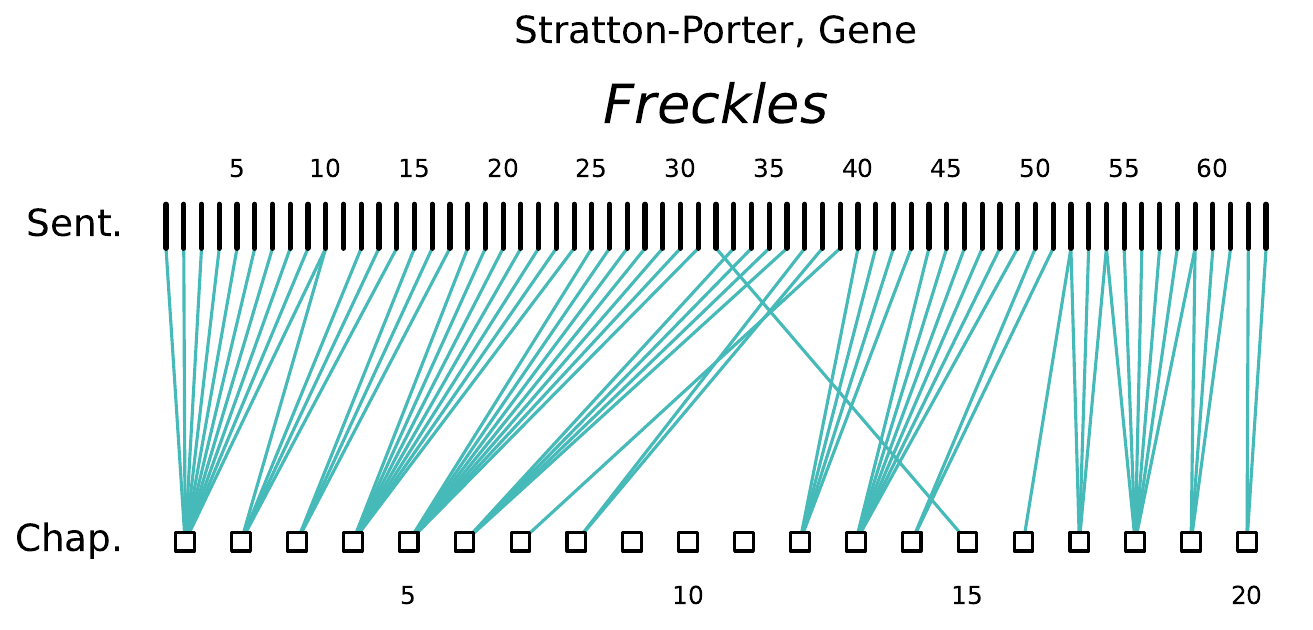}
    {\captionsetup{hypcap=false, labelformat=simple, labelsep=none, textformat=empty}\captionof{figure}{\label{fig:pg111-bipartite}}}
  \end{minipage}\hfill
  \begin{minipage}[t]{0.48\linewidth}\centering
    \includegraphics[width=\linewidth]{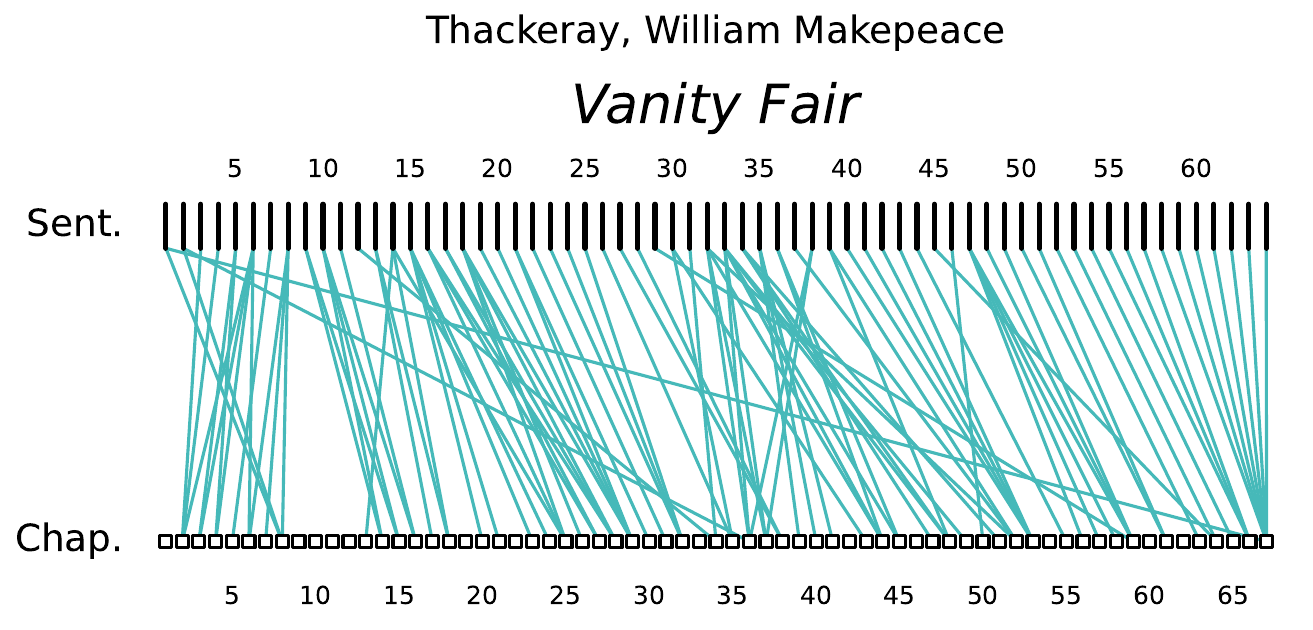}
    {\captionsetup{hypcap=false, labelformat=simple, labelsep=none, textformat=empty}\captionof{figure}{\label{fig:pg599-bipartite}}}
  \end{minipage}\bigskip

\noindent
  \begin{minipage}[t]{0.48\linewidth}\centering
    \includegraphics[width=\linewidth]{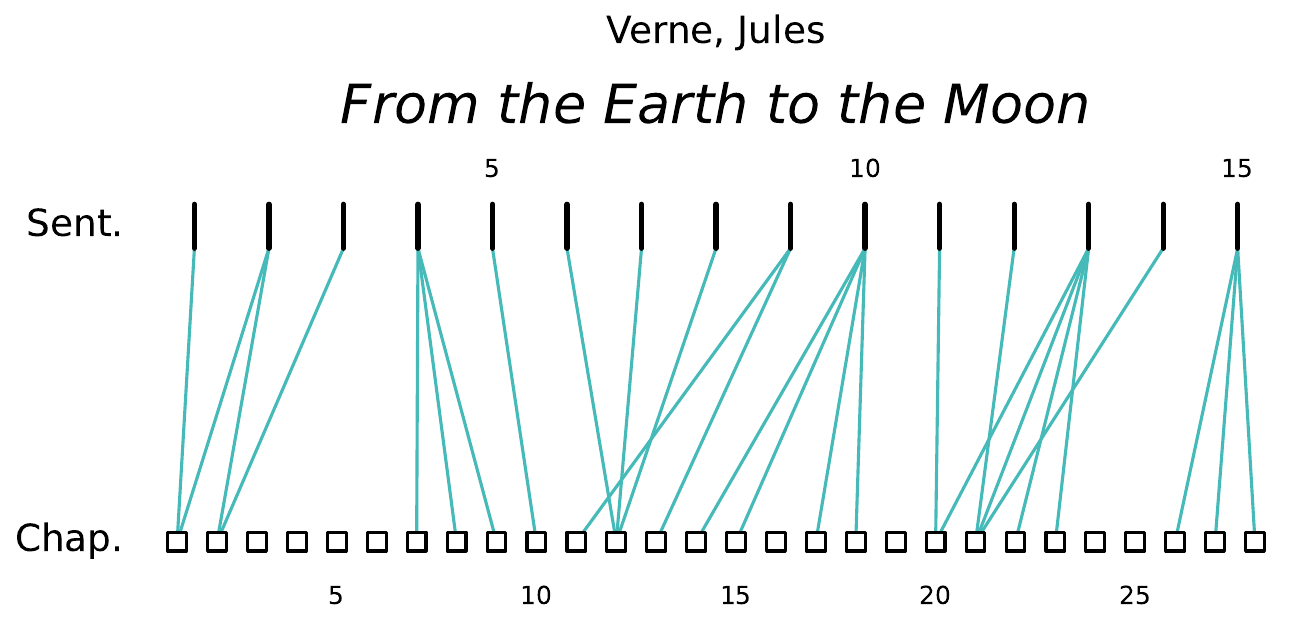}
    {\captionsetup{hypcap=false, labelformat=simple, labelsep=none, textformat=empty}\captionof{figure}{\label{fig:pg83-bipartite}}}
  \end{minipage}\hfill
  \begin{minipage}[t]{0.48\linewidth}\centering
    \includegraphics[width=\linewidth]{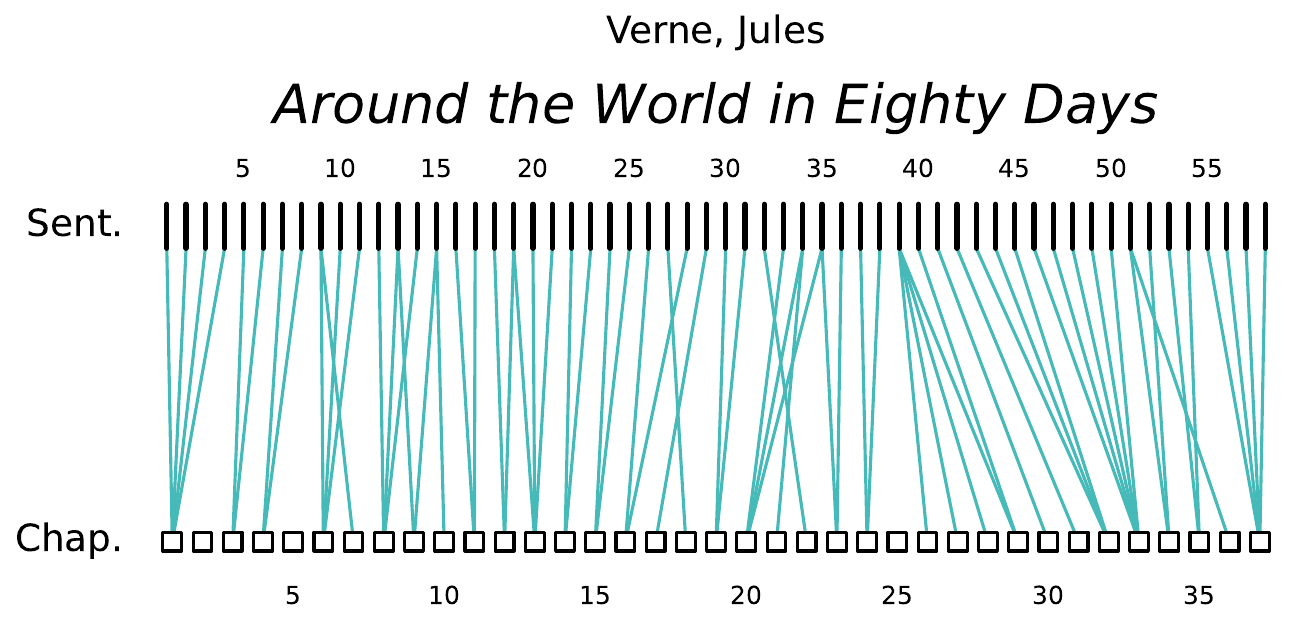}
    {\captionsetup{hypcap=false, labelformat=simple, labelsep=none, textformat=empty}\captionof{figure}{\label{fig:pg103-bipartite}}}
  \end{minipage}\bigskip

\noindent
  \begin{minipage}[t]{0.48\linewidth}\centering
    \includegraphics[width=\linewidth]{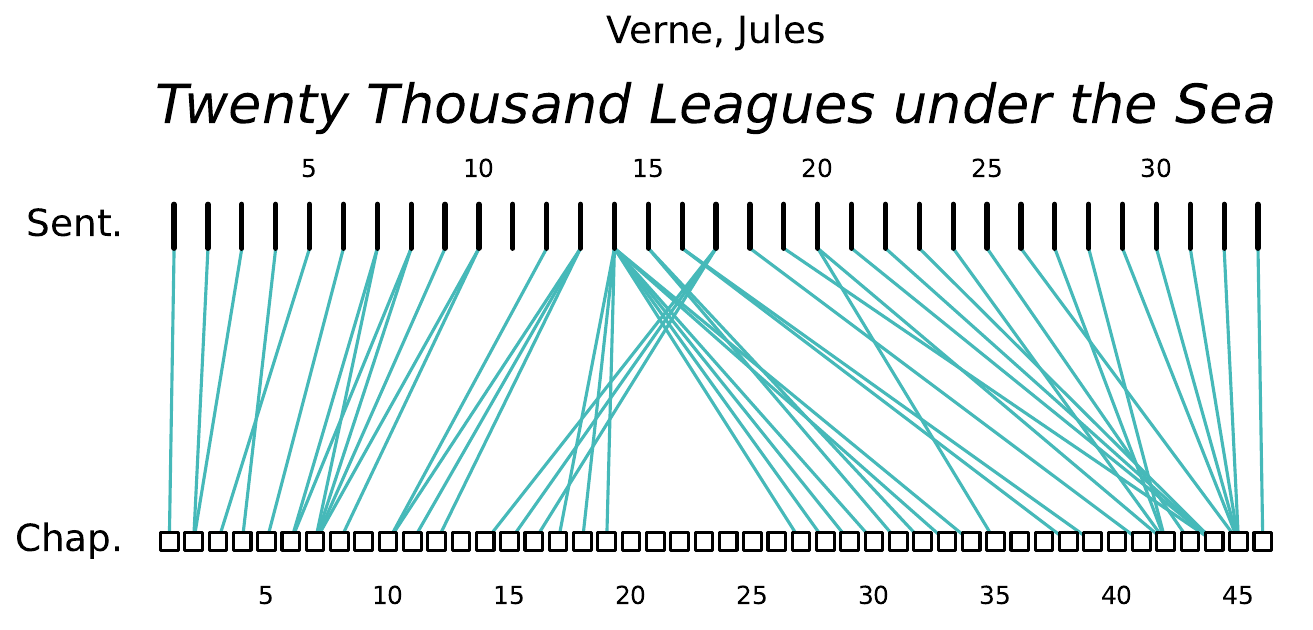}
    {\captionsetup{hypcap=false, labelformat=simple, labelsep=none, textformat=empty}\captionof{figure}{\label{fig:pg164-bipartite}}}
  \end{minipage}\hfill
  \begin{minipage}[t]{0.48\linewidth}\centering
    \includegraphics[width=\linewidth]{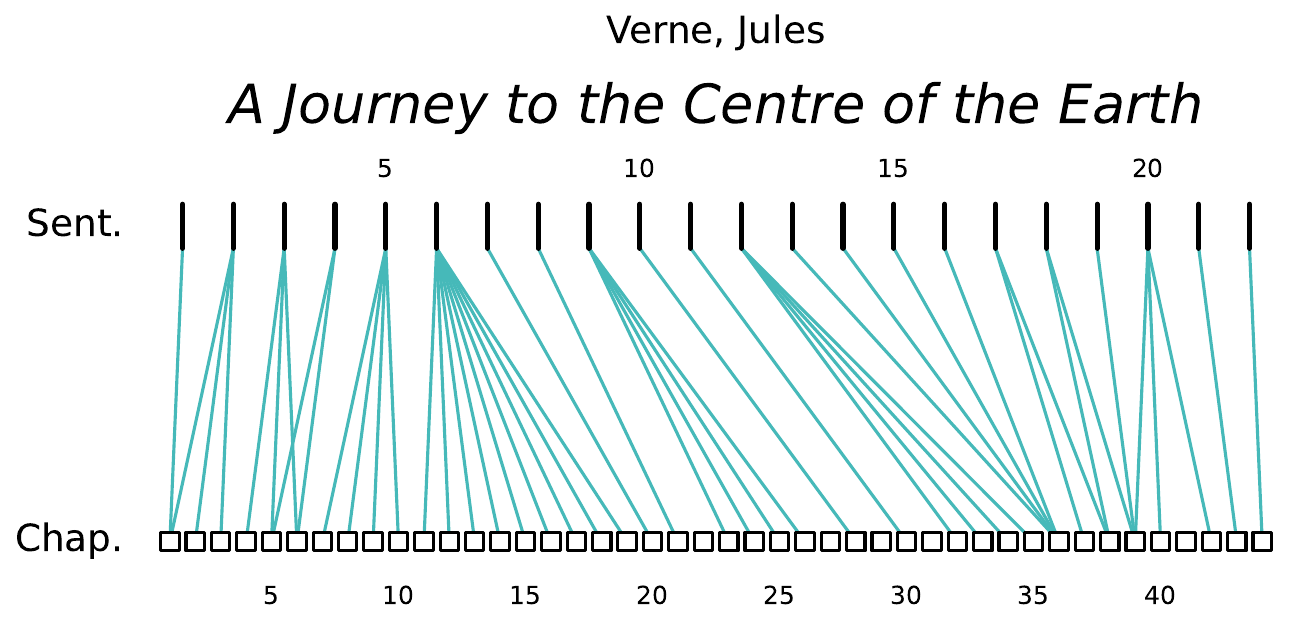}
    {\captionsetup{hypcap=false, labelformat=simple, labelsep=none, textformat=empty}\captionof{figure}{\label{fig:pg18857-bipartite}}}
  \end{minipage}\bigskip

\noindent
  \begin{minipage}[t]{0.48\linewidth}\centering
    \includegraphics[width=\linewidth]{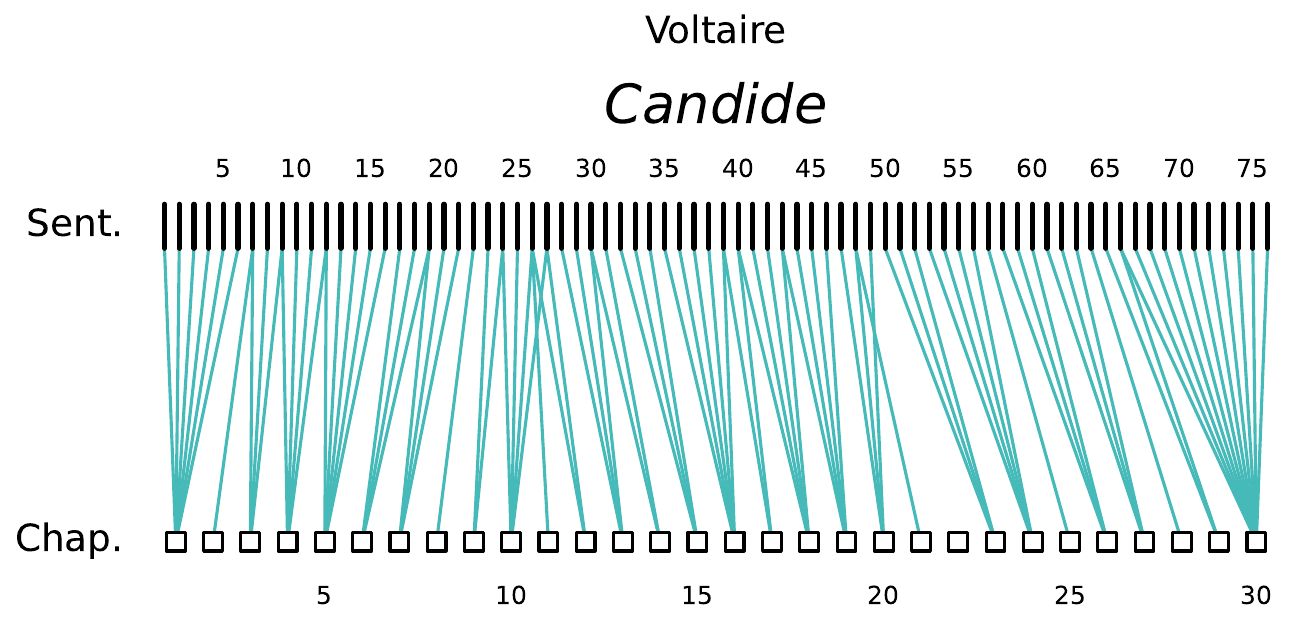}
    {\captionsetup{hypcap=false, labelformat=simple, labelsep=none, textformat=empty}\captionof{figure}{\label{fig:pg19942-bipartite}}}
  \end{minipage}\hfill
  \begin{minipage}[t]{0.48\linewidth}\centering
    \includegraphics[width=\linewidth]{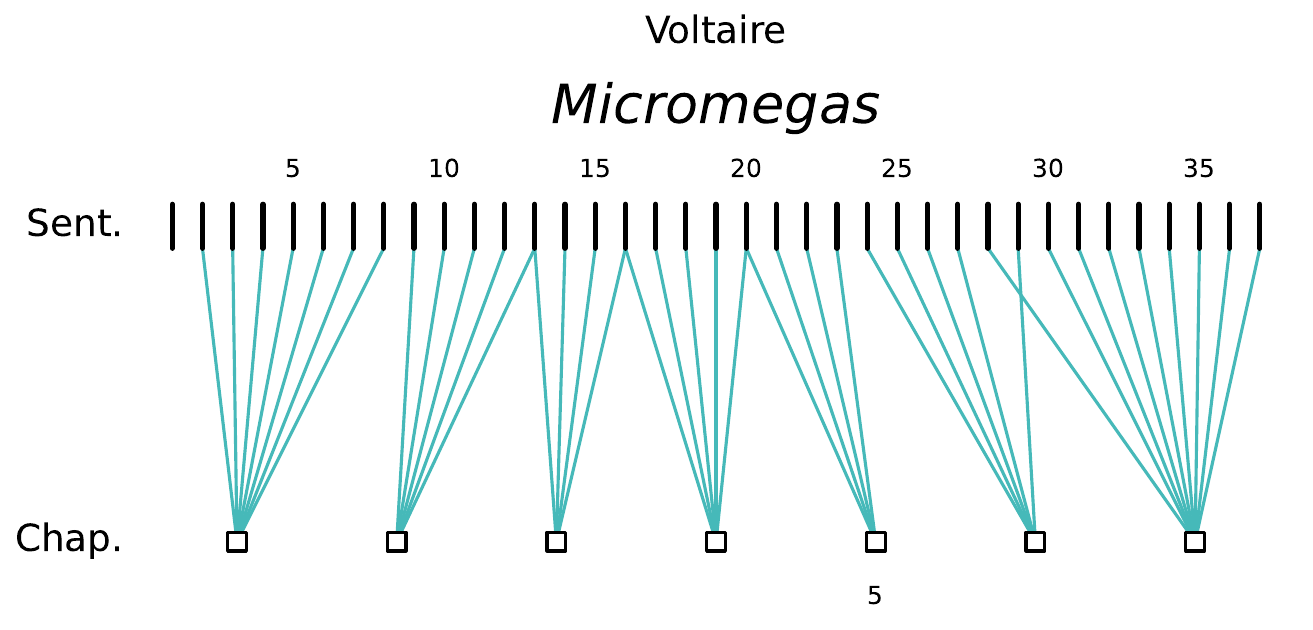}
    {\captionsetup{hypcap=false, labelformat=simple, labelsep=none, textformat=empty}\captionof{figure}{\label{fig:pg30123-bipartite}}}
  \end{minipage}\bigskip

\noindent
  \begin{minipage}[t]{0.48\linewidth}\centering
    \includegraphics[width=\linewidth]{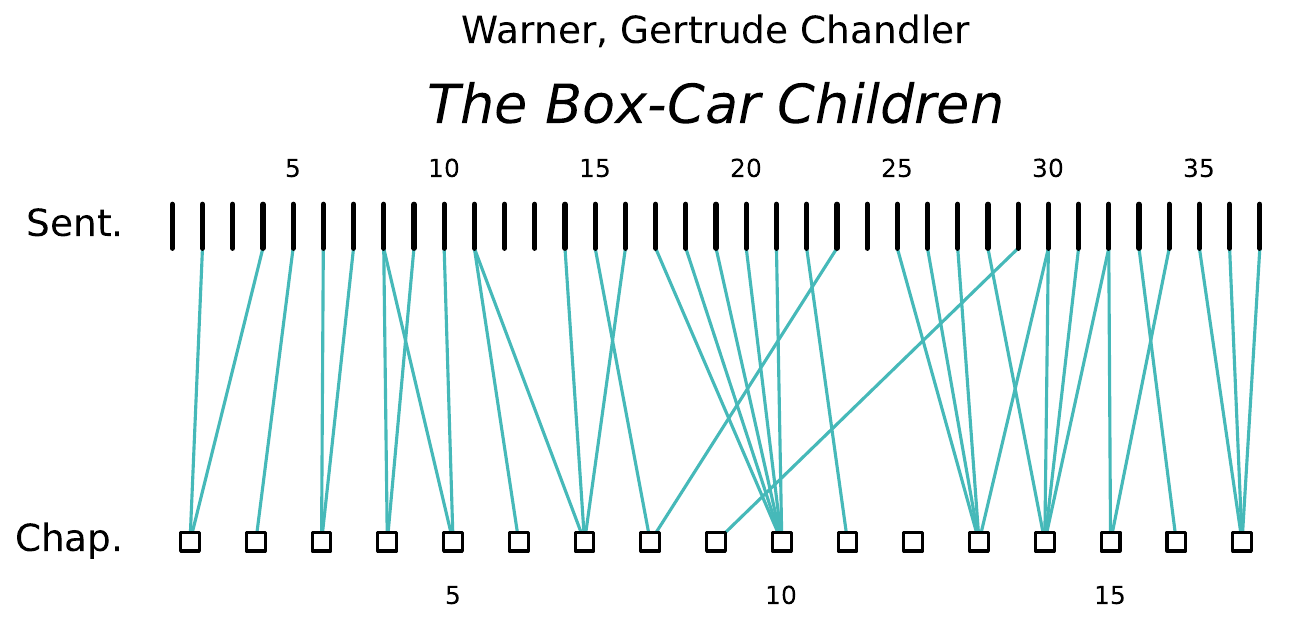}
    {\captionsetup{hypcap=false, labelformat=simple, labelsep=none, textformat=empty}\captionof{figure}{\label{fig:pg42796-bipartite}}}
  \end{minipage}\hfill
  \begin{minipage}[t]{0.48\linewidth}\centering
    \includegraphics[width=\linewidth]{images/pg77900_bipartite.pdf}
    {\captionsetup{hypcap=false, labelformat=simple, labelsep=none, textformat=empty}\captionof{figure}{\label{fig:pg77900-bipartite}}}
  \end{minipage}\bigskip

\noindent
  \begin{minipage}[t]{0.48\linewidth}\centering
    \includegraphics[width=\linewidth]{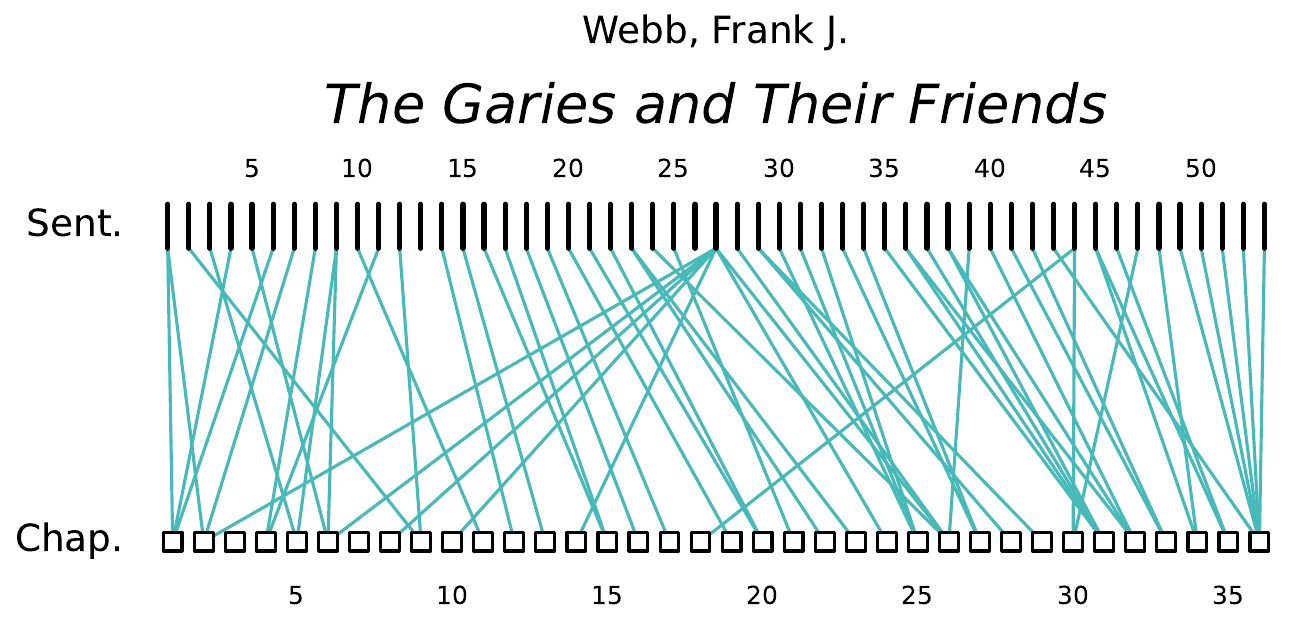}
    {\captionsetup{hypcap=false, labelformat=simple, labelsep=none, textformat=empty}\captionof{figure}{\label{fig:pg11214-bipartite}}}
  \end{minipage}\hfill
  \begin{minipage}[t]{0.48\linewidth}\centering
    \includegraphics[width=\linewidth]{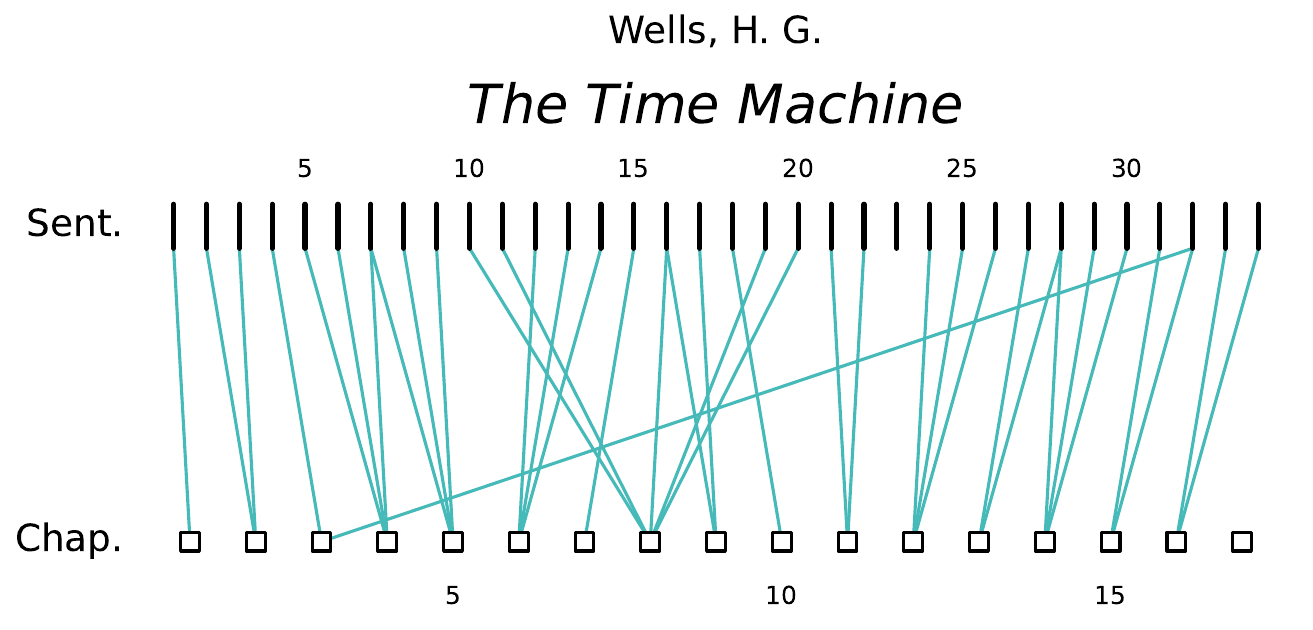}
    {\captionsetup{hypcap=false, labelformat=simple, labelsep=none, textformat=empty}\captionof{figure}{\label{fig:pg35-bipartite}}}
  \end{minipage}\bigskip

\noindent
  \begin{minipage}[t]{0.48\linewidth}\centering
    \includegraphics[width=\linewidth]{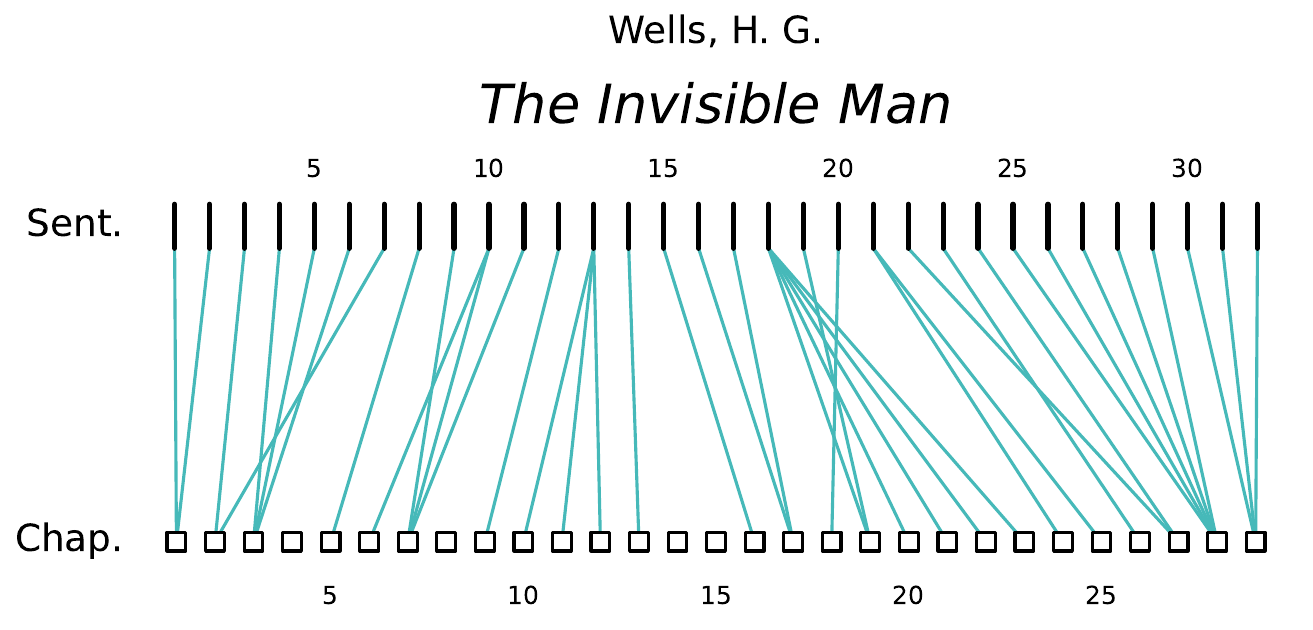}
    {\captionsetup{hypcap=false, labelformat=simple, labelsep=none, textformat=empty}\captionof{figure}{\label{fig:pg5230-bipartite}}}
  \end{minipage}\hfill
  \begin{minipage}[t]{0.48\linewidth}\centering
    \includegraphics[width=\linewidth]{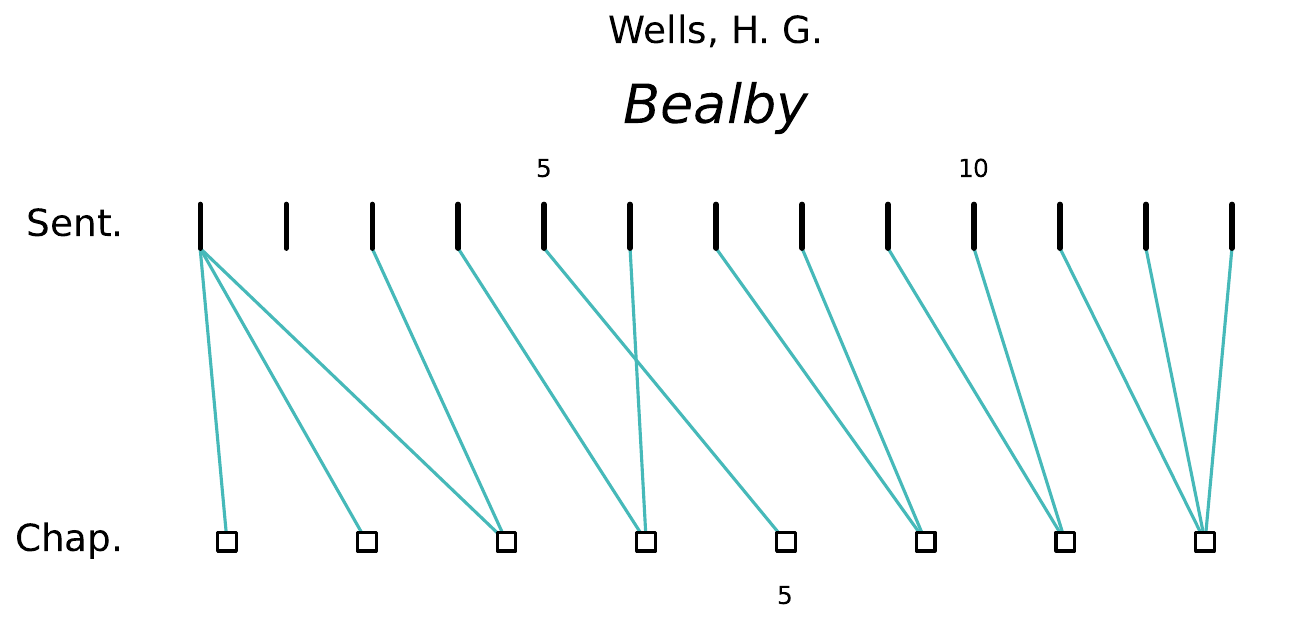}
    {\captionsetup{hypcap=false, labelformat=simple, labelsep=none, textformat=empty}\captionof{figure}{\label{fig:pg59769-bipartite}}}
  \end{minipage}\bigskip

\noindent
  \begin{minipage}[t]{0.48\linewidth}\centering
    \includegraphics[width=\linewidth]{images/pg284_bipartite.pdf}
    {\captionsetup{hypcap=false, labelformat=simple, labelsep=none, textformat=empty}\captionof{figure}{\label{fig:pg284-bipartite}}}
  \end{minipage}\hfill
  \begin{minipage}[t]{0.48\linewidth}\centering
    \includegraphics[width=\linewidth]{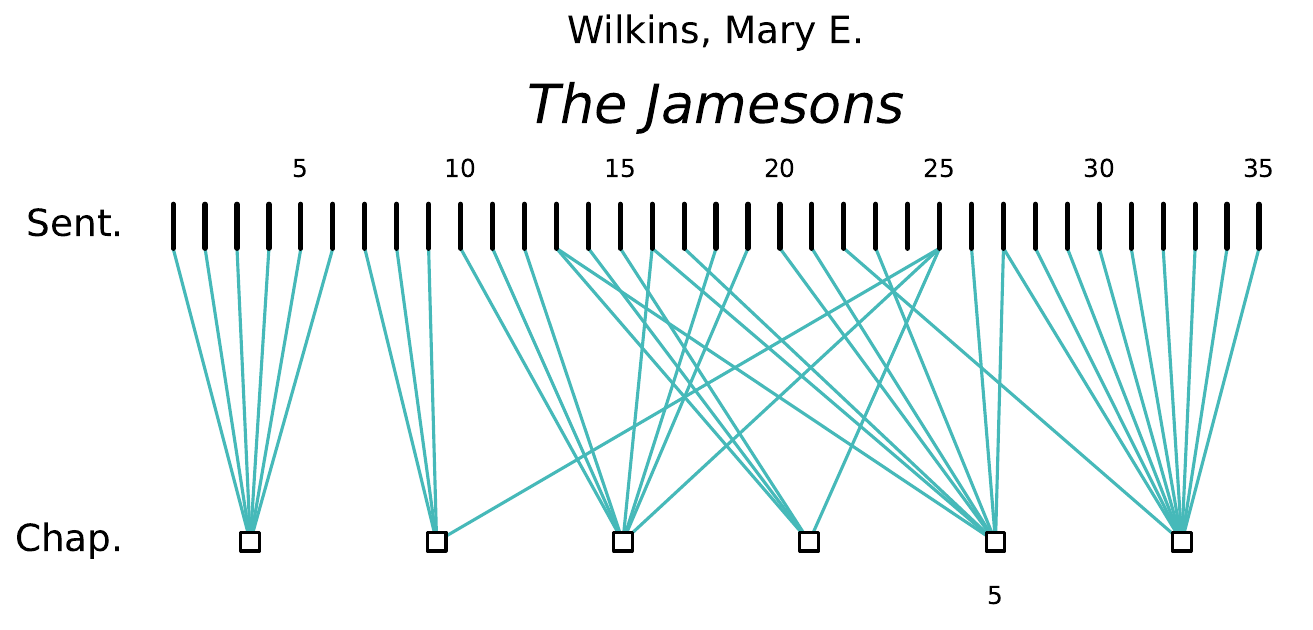}
    {\captionsetup{hypcap=false, labelformat=simple, labelsep=none, textformat=empty}\captionof{figure}{\label{fig:pg17792-bipartite}}}
  \end{minipage}\bigskip

\noindent
  \begin{minipage}[t]{0.48\linewidth}\centering
    \includegraphics[width=\linewidth]{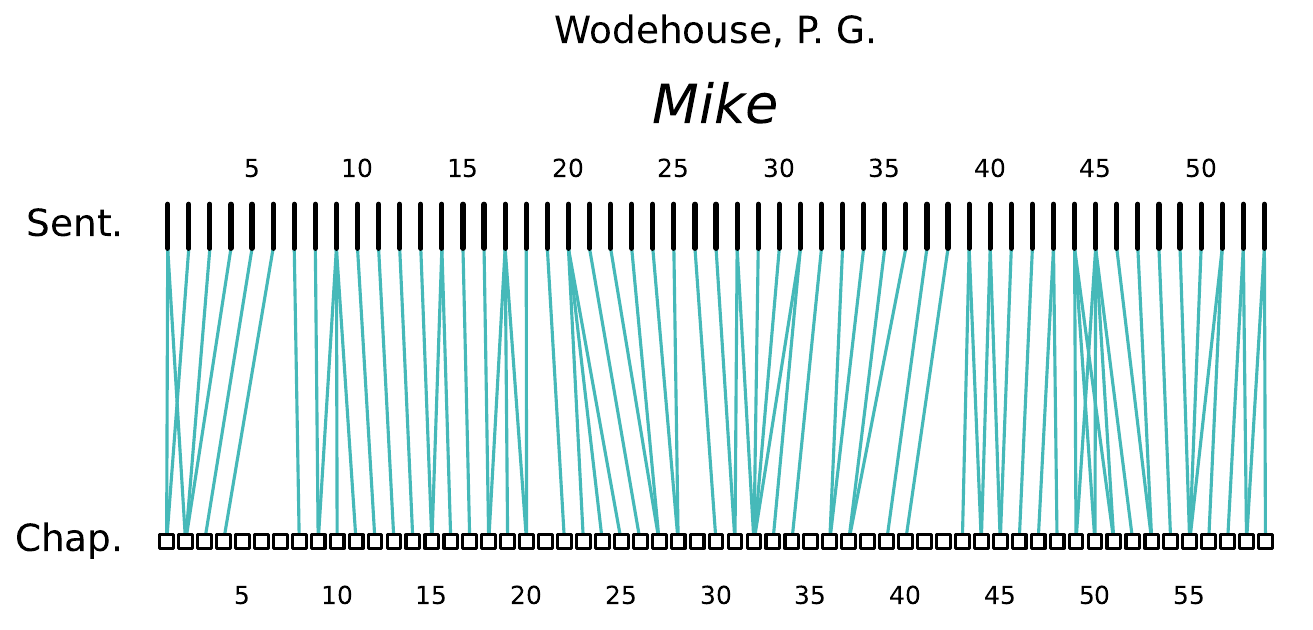}
    {\captionsetup{hypcap=false, labelformat=simple, labelsep=none, textformat=empty}\captionof{figure}{\label{fig:pg7423-bipartite}}}
  \end{minipage}\hfill
  \begin{minipage}[t]{0.48\linewidth}\centering
    \includegraphics[width=\linewidth]{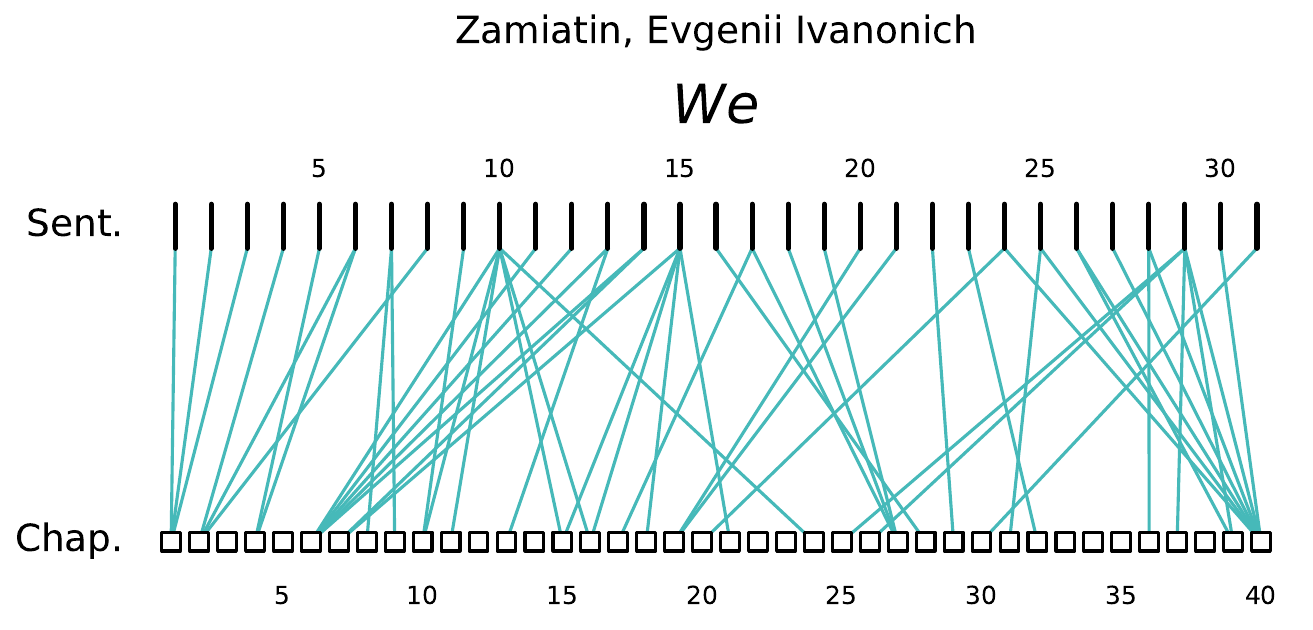}
    {\captionsetup{hypcap=false, labelformat=simple, labelsep=none, textformat=empty}\captionof{figure}{\label{fig:pg61963-bipartite}}}
  \end{minipage}\bigskip

\noindent
  \begin{minipage}[t]{0.48\linewidth}\centering
    \includegraphics[width=\linewidth]{images/pg8600_bipartite.pdf}
    {\captionsetup{hypcap=false, labelformat=simple, labelsep=none, textformat=empty}\captionof{figure}{\label{fig:pg8600-bipartite}}}
  \end{minipage}\hfill
  \begin{minipage}[t]{0.48\linewidth}\centering
    \includegraphics[width=\linewidth]{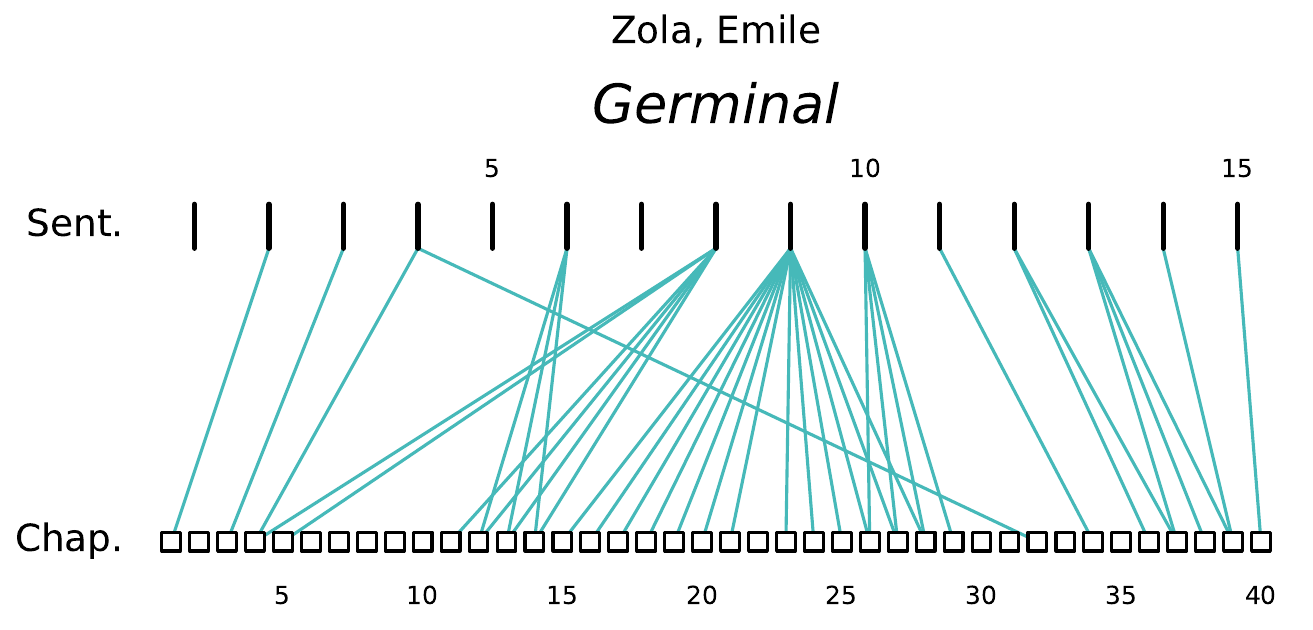}
    {\captionsetup{hypcap=false, labelformat=simple, labelsep=none, textformat=empty}\captionof{figure}{\label{fig:pg56528-bipartite}}}
  \end{minipage}\bigskip

\section{Summary Texts}
\label{sec:summaries}

This section includes the Wikipedia summaries referenced in the body of the paper in the order of appearance. Summary sentences that were not matched with any chapter of the text after alignment appear \textcolor{skipred}{in red}.

\subsection{\emph{Ivanhoe: A Romance} by Walter Scott}
\label{sec:82-summary}

\begin{Verbatim}[frame=single,commandchars=\\\{\},fontsize=\small]
1.  Protagonist Wilfred of Ivanhoe is disinherited by his father Cedric of
    Rotherwood for supporting the Norman King Richard and for falling in love
    with the Lady Rowena, a ward of Cedric and descendant of the Saxon Kings of
    England.
2.  \textcolor{skipred}{Cedric planned to have Rowena marry the powerful Lord Athelstane, a}
    \textcolor{skipred}{pretender to the Crown of England by his descent from the last Saxon King,}
    \textcolor{skipred}{Harold Godwinson.}
3.  \textcolor{skipred}{Ivanhoe accompanies King Richard on the Third Crusade, where he is said to}
    \textcolor{skipred}{have played a notable role in the Siege of Acre.}
4.  The book opens with a scene of Norman knights and prelates seeking the
    hospitality of Cedric.
5.  They are guided there by a pilgrim, known at that time as a palmer.
6.  That same night, Isaac of York, a Jewish moneylender, seeks refuge at
    Rotherwood on his way to the tournament at Ashby.
7.  \textcolor{skipred}{Following the night's meal, the palmer observes one of the Normans, the}
    \textcolor{skipred}{Templar Brian de Bois-Guilbert, issue orders to his Saracen soldiers to}
    \textcolor{skipred}{capture Isaac.}
8.  The palmer then assists in Isaac's escape from Rotherwood, with the
    additional aid of the swineherd Gurth.
9.  Isaac of York offers to repay his debt to the palmer with a suit of armour
    and a war horse to participate in the tournament at Ashby-de-la-Zouch
    Castle, on his inference that the palmer was secretly a knight.
10. The palmer is taken by surprise, but accepts the offer.
11. The tournament is presided over by Prince John.
12. Also in attendance are Cedric, Athelstane, Lady Rowena, Isaac of York, his
    daughter Rebecca, Robin of Locksley and his men, Prince John's advisor
    Waldemar Fitzurse, and numerous Norman knights.
13. On the first day of the tournament, in a bout of individual jousting, a
    mysterious knight, identifying himself only as "Desdichado" (described in
    the book as Spanish, taken by the Saxons to mean "Disinherited"), defeats
    Bois-Guilbert.
14. The masked knight declines to reveal himself despite Prince John's request,
    but is nevertheless declared the champion of the day and is permitted to
    choose the Queen of the Tournament.
15. He bestows this honour upon Lady Rowena.
16. On the second day, at a melee, Desdichado is the leader of one party,
    opposed by his former adversaries.
17. Desdichado's side is soon hard-pressed and he himself beset by multiple
    foes until rescued by a knight nicknamed Le Noir Faineant ('the Black
    Sluggard'), who thereafter departs in secret.
18. When forced to unmask himself to receive his coronet (the sign of
    championship), Desdichado is identified as Wilfred of Ivanhoe, returned
    from the Crusades.
19. This causes much consternation to Prince John and his court who now fear
    the imminent return of King Richard.
20. Ivanhoe is severely wounded in the competition yet his father does not move
    quickly to tend to him.
21. Instead, Rebecca, a skilled physician, tends to him while they are lodged
    near the tournament and then convinces her father to take Ivanhoe with them
    to their home in York when he is fit for that trip.
22. The conclusion of the tournament includes feats of archery by Locksley,
    such as splitting a willow reed with his arrow.
23. Prince John's dinner for the local Saxons ends in insults.
24. In the forests between Ashby and York, Isaac, Rebecca and the wounded
    Ivanhoe are abandoned by their guards, who fear bandits and take all of
    Isaac's horses.
25. Cedric, Athelstane and the Lady Rowena meet them and agree to travel
    together.
26. The party is captured by de Bracy and his companions and taken to
    Torquilstone, the castle of Front-de-BÏuf.
27. The swineherd Gurth and Wamba the jester manage to escape, and then
    encounter Locksley, who plans a rescue.
28. The Black Knight, having taken refuge for the night in the hut of local
    friar, the Holy Clerk of Copmanhurst, volunteers his assistance on learning
    about the captives from Robin of Locksley.
29. They then besiege the Castle of Torquilstone with Robin's own men,
    including the friar and assorted Saxon yeomen.
30. Inside Torquilstone, de Bracy expresses his love for the Lady Rowena but is
    refused.
31. Brian de Bois-Guilbert tries to rape Rebecca and is thwarted.
32. He then tries to seduce her and is rebuffed.
33. Front-de-BÏuf tries to wring a hefty ransom from Isaac of York, but Isaac
    refuses to pay unless his daughter is freed.
34. When the besiegers deliver a note to yield up the captives, their Norman
    captors demand a priest to administer the Final Sacrament to Cedric;
    whereupon Cedric's jester Wamba slips in disguised as a priest, and takes
    the place of Cedric, who escapes and brings important information to the
    besiegers on the strength of the garrison and its layout.
35. On his way out, Cedric meets the Saxon crone Ulrica, who vows revenge on
    Front-de-BÏuf and advises Cedric to tell the besiegers.
36. The besiegers storm the castle.
37. The castle is set aflame during the assault by Ulrica, the daughter of the
    original lord of the castle, Lord Torquilstone, as revenge for her father's
    death.
38. Front-de-BÏuf is killed in the fire while de Bracy surrenders to the Black
    Knight, who identifies himself as King Richard and releases de Bracy.
39. Bois-Guilbert escapes with Rebecca while Isaac is captured by the Clerk of
    Copmanhurst.
40. The Lady Rowena is saved by Cedric, while the still-wounded Ivanhoe is
    rescued from the burning castle by King Richard.
41. In the fighting, Athelstane is wounded and presumed dead while attempting
    to rescue Rebecca, whom he mistakes for Rowena.
42. Following the battle, Locksley plays host to King Richard.
43. Word is conveyed by de Bracy to Prince John of the King's return and the
    fall of Torquilstone.
44. In the meantime, Bois-Guilbert rushes with his captive to the nearest
    Templar Preceptory, where Lucas de Beaumanoir, the Grand Master of the
    Templars, takes umbrage at Bois-Guilbert's infatuation and subjects Rebecca
    to a trial for witchcraft.
45. At Bois-Guilbert's secret request, she claims the right to trial by combat;
    and Bois-Guilbert, who had hoped to fight as Rebecca's champion, is
    devastated when the Grand Master orders him to fight on behalf of the
    Templestowe.
46. Rebecca then writes to her father to procure a champion for her.
47. Cedric organizes Athelstane's funeral at Coningsburgh, in the midst of
    which the Black Knight arrives with Ivanhoe.
48. Cedric, who had not been present at Locksley's carousal, is ill-disposed
    towards the knight upon learning his true identity, but Richard calms
    Cedric and reconciles him with his son.
49. During this conversation, Athelstane emerges – not dead, but laid in his
    coffin alive by monks desirous of the funeral money.
50. Over Cedric's renewed protests, Athelstane pledges his homage to the Norman
    King Richard and urges Cedric to allow Rowena to marry Ivanhoe, to which
    Cedric finally agrees.
51. Soon after this reconciliation, Ivanhoe receives word from Isaac beseeching
    him to fight on Rebecca's behalf.
52. Ivanhoe, riding day and night, arrives in time for the trial by combat;
    however, both horse and man are exhausted, with little chance of victory.
53. Bois-Guilbert refuses to fight but Ivanhoe accuses him of breaking his word
    and the Templar reacts fiercely.
54. His face becomes flushed and he is ready for combat.
55. The two knights make one charge at each other with lances, Bois-Guilbert
    appearing to have the advantage.
56. Ivanhoe and his horse go down, but Bois-Guilbert also falls though barely
    touched.
57. Ivanhoe quickly gets up to finish the fight with his sword, but
    Bois-Guilbert does not rise and dies a victim of his own contending
    passions.
58. Ivanhoe and Rowena marry and live a long and happy life together.
59. Fearing further persecution, Rebecca and her father plan to quit England
    for Granada.
60. Before leaving, Rebecca comes to Rowena shortly after the wedding to bid
    her a solemn farewell.
61. Ivanhoe's military service ends with the death of King Richard five years
    later.
\end{Verbatim}

\subsection{\emph{Oil!} by Upton Sinclair}
\label{sec:70379-summary}

\begin{Verbatim}[frame=single,commandchars=\\\{\},fontsize=\small]
1.  James Arnold "Dad" Ross and his son, James Jr. ("Bunny") are introduced as
    they drive through southern California to meet with the Watkins family, who
    are leasing out some oil property they own.
2.  They find out that the family is deadlocked about how the properties run
    and proceeds should be divided.
3.  While Dad and Bunny go quail hunting on the Watkins' goat ranch, they find
    oil.
4.  At Bunny's urging, Dad tries to prevent the elder Watkins from beating his
    daughter Ruth, trying to convince them that he has received a "third
    revelation" which prohibits parents from beating their children.
5.  The plan backfires when Eli, Ruth's brother, interjects himself into the
    discussion and claims that he has received the revelation.
6.  As drilling begins at the Watkins ranch, Bunny begins to realize his
    father's business methods are not entirely ethical.
7.  After a worker is killed in an accident and an oil well is destroyed in a
    blowout, Dad's workforce goes on strike.
8.  Bunny is torn between loyalty to Dad and his friendship to Ruth and her
    rebellious brother Paul, who support the workers.
9.  Paul is drafted into World War I and, when the conflict is over, remains in
    Siberia to fight the rising Bolsheviks.
10. Back home, Bunny enrolls in college, and he becomes increasingly involved
    with socialism through a classmate, Rachel Menzies.
11. Paul returns home and tells of his travels, explaining he has become a
    communist.
12. Bunny accompanies Dad to the seaside mansion of his business associate
    Vernon Roscoe.
13. Dad and Roscoe flee the country to avoid being subpoenaed by Congress in
    the Teapot Dome scandal.
14. Before Dad goes away, Bunny proposes parting ways with his father and
    earning his own way in the world; Dad is confused and hurt, but not
    unsupportive.
15. Overseas, Dad meets and marries Mrs. Olivier, a widow and spiritualist, but
    soon passes away from pneumonia.
16. Bunny decides to dedicate his life and inheritance to social justice while
    Roscoe moves to get control of the bulk of Dad's estate.
17. Bunny and his sister Bertie are swindled out of most of their inheritance
    by Roscoe and Mrs. Olivier.
18. Bunny marries Rachel and they dedicate themselves to establishing a
    socialist institution of learning; Eli, by now a successful evangelist,
    falsely claims that Paul underwent a deathbed conversion to Christianity.
\end{Verbatim}

\subsection{\emph{Carmilla} by Joseph Sheridan Le Fanu}
\label{sec:10007-summary}

\begin{Verbatim}[frame=single,commandchars=\\\{\},fontsize=\small]
1.  The story is presented as part of the casebook of Dr. Hesselius.
2.  A woman named Laura narrates, beginning with her childhood in a "picturesque 
    and solitary" castle amid an extensive forest in Styria, where she lives 
    with her father, a wealthy English widower retired from service to the 
    Austrian Empire.
3.  When she was six, Laura had a vision of a beautiful visitor in her 
    bedchamber.
4.  She later claims to have been punctured in her breast, although no wound 
    was found.
5.  All the household assure Laura that it was just a dream, but they step up 
    security as well and there is no subsequent vision or visitation.
6.  Twelve years later, Laura's father tells her of a letter from his friend, 
    General Spielsdorf.
7.  The General was supposed to visit them with his niece, Bertha Rheinfeldt, 
    but she died under mysterious circumstances.
8.  The General promises to discuss the circumstances in detail when they meet 
    later.
9.  Laura, saddened by the loss of a potential friend, longs for a companion.
10. A carriage accident outside Laura's home unexpectedly brings Carmilla, a 
    girl of Laura's age, into the family's care.
11. Both girls instantly recognise each other from the "dream" they both had 
    when they were young.
12. Carmilla appears injured after her carriage accident, but her mysterious 
    mother informs Laura's father that her journey is urgent and cannot be 
    delayed.
13. She arranges to leave Carmilla with Laura and her father until she can 
    return in three months.
14. Before leaving, she notes that Carmilla will not disclose any information 
    whatsoever about her family, her past, or herself.
15. Carmilla and Laura grow to be close friends, but occasionally Carmilla's 
    mood abruptly changes.
16. She sometimes makes romantic advances towards Laura.
17. Carmilla refuses to tell anything about herself, despite questioning by 
    Laura.
18. Her secrecy is not the only mysterious thing about Carmilla; she never 
    joins the household in its prayers, she sleeps much of the day, and she 
    seems to sleepwalk outside at night.
19. Meanwhile, young women and girls in the nearby towns have begun dying from 
    an unknown malady.
20. When the funeral procession of one such victim passes by the two girls, 
    Laura joins in the funeral hymn.
21. Carmilla bursts out in rage and scolds Laura, complaining that the hymn 
    hurts her ears.
22. When a shipment of restored heirloom paintings arrives, Laura finds a 
    portrait of her ancestor, Countess Mircalla Karnstein, dated 1698.
23. The portrait resembles Carmilla exactly, down to the mole on her neck.
24. Carmilla suggests that she might be descended from the Karnsteins, though 
    the family died out centuries before.
25. During Carmilla's stay, Laura has nightmares of a large, cat-like beast 
    entering her room.
26. The beast springs onto the bed and Laura feels something like two needles, 
    an inch or two apart, darting deep into her breast.
27. The beast then takes the form of a female figure and disappears through the 
    door without opening it.
28. In another nightmare, Laura hears a voice say, "Your mother warns you to 
    beware of the assassin," and a light reveals Carmilla standing at the foot 
    of her bed, her nightdress drenched in blood.
29. Laura's health declines, and her father has a doctor examine her.
30. He finds a small, blue spot, an inch or two below her collar, where the 
    creature in her dream bit her, and speaks privately with her father, only 
    asking that Laura never be unattended.
31. Her father sets out with Laura in a carriage for the ruined village of 
    Karnstein, three miles distant.
32. They leave a message behind asking Carmilla and a governess to follow once 
    the perpetually late-sleeping Carmilla awakes.
33. En route to Karnstein, Laura and her father encounter Spielsdorf, who tells 
    them his story.
34. At a costume ball, Spielsdorf and Bertha had met a beautiful young woman 
    named Millarca and her enigmatic mother.
35. Bertha was immediately taken with Millarca.
36. The mother convinced Spielsdorf that she was an old friend of his and asked 
    that Millarca be allowed to stay with them for three weeks while she 
    attended to a secret matter of great importance.
37. Bertha fell mysteriously ill, suffering the same symptoms as Laura.
38. After consulting with a specially ordered priestly doctor, Spielsdorf 
    realised that Bertha was being visited by a vampire.
39. He hid with a sword and waited until a large, black creature crawled onto 
    Bertha's bed and spread itself onto her throat.
40. He leapt from his hiding place and attacked the creature, which had then 
    taken the form of Millarca.
41. She fled through the locked door, unharmed.
42. Bertha died before the morning dawned.
43. Upon arriving at Karnstein, Spielsdorf asks a woodman where he can find 
    the tomb of Mircalla Karnstein.
44. The woodman says the tomb was relocated long ago by a Moravian nobleman 
    who vanquished the vampires haunting the region.
45. While Spielsdorf and Laura are alone in the ruined chapel, Carmilla 
    appears.
46. Spielsdorf attacks her with an axe.
47. Carmilla disarms Spielsdorf and disappears.
48. Spielsdorf explains that Carmilla is also Millarca, both anagrams for the 
    original name of the vampire Mircalla, Countess Karnstein.
49. The party is joined by Baron Vordenburg, the descendant of the hero who rid 
    the area of vampires.
50. Vordenburg, an authority on vampires, has discovered that his ancestor was 
    romantically involved with Mircalla before she died.
51. Using his forefather's notes, he locates Mircalla's hidden tomb.
52. An imperial commission exhumes the body of Mircalla.
53. Immersed in blood, it seems to be breathing faintly, its heart beating, its 
    eyes open.
54. A stake is driven through its heart, and it gives a corresponding shriek; 
    then, the head is struck off.
55. The body and head are burned to ashes, which are thrown into a river.
56. Afterwards, Laura's father takes his daughter on a year-long tour through 
    Italy to regain her health and recover from the trauma, but she never fully 
    does.
\end{Verbatim}

\subsection{\emph{L'Assommoir} by \'{E}mile Zola}
\label{sec:8600-summary}

\begin{Verbatim}[frame=single,commandchars=\\\{\},fontsize=\small]
1.  L'Assommoir begins with Gervaise and her two young sons being abandoned by 
    Lantier, who takes off for parts unknown with another woman.
2.  Though at first she swears off men altogether, eventually she gives in to 
    the advances of Coupeau, a teetotal roofer, and they are married.
3.  The marriage sequence is one of the most famous set-pieces of Zola's work; 
    the account of the wedding party's impromptu and chaotic trip to the Louvre 
    is one of the novelist's most famous passages.
4.  Through a combination of happy circumstances, Gervaise is able to realise 
    her dream and raise enough money to open her own laundry.
5.  The couple's happiness appears to be complete with the birth of a daughter, 
    Anna, nicknamed Nana (the protagonist of Zola's later novel of the same 
    title).
6.  However, later in the story, we witness the downward trajectory of 
    Gervaise's life from this happy high point.
7.  Coupeau is injured in a fall from the roof of a new hospital he is working 
    on, and during his lengthy convalescence he takes first to idleness, then 
    to gluttony, and eventually to drink.
8.  In only a few years, Coupeau becomes a vindictive, wife-beating alcoholic, 
    with no intention of trying to find more work.
9.  Gervaise struggles to keep her home together, but her excessive pride leads 
    her to a number of embarrassing failures and before long everything is 
    going downhill.
10. Gervaise becomes infected by her husband's newfound laziness and, in an 
    effort to impress others, spends her money on lavish feasts and accumulates 
    uncontrolled debt.
11. The home is further disrupted by the return of Lantier, who is warmly 
    welcomed by Coupeau - by this point losing interest in both Gervaise and 
    life itself, and becoming seriously ill.
12. The ensuing chaos and financial strain is too much for Gervaise, who loses 
    her laundry-shop and is sucked into a spiral of debt and despair.
13. Eventually, she too finds solace in drink and, like Coupeau, slides into 
    heavy alcoholism.
14. All this prompts Nana - already suffering from the chaotic life at home and 
    getting into trouble on a daily basis - to run away from her parents' home 
    and become a casual prostitute.
15. Gervaise's story is told against a backdrop of a rich array of other well-
    drawn characters with their own vices and idiosyncrasies.
16. Notable amongst these being Goujet, a young blacksmith, who spends his life 
    in unconsummated love for the hapless laundress.
17. Eventually, sunk by debt, hunger and alcohol, Coupeau and Gervaise both 
    die.
18. The latter's corpse lies for two days in her unkempt hovel before it is 
    noticed by her disdaining neighbors.
\end{Verbatim}

\subsection{\emph{Flatland: A Romance of Many Dimensions} by Edwin Abbott Abbott}
\label{sec:45506-summary}

\begin{Verbatim}[frame=single,commandchars=\\\{\},fontsize=\small]
1.  The story describes a two-dimensional world inhabited by geometric figures 
    (flatlanders); women are line segments, while men are polygons with various 
    numbers of sides.
2.  The narrator is a square, a member of the caste of gentlemen and 
    professionals, who guides the readers through some of the implications of 
    life in two dimensions.
3.  \textcolor{skipred}{The first half of the story goes through the practicalities of existing in a two-dimensional universe, as well as a history leading up to the year 1999 on the eve of the 3rd Millennium.}
4.  On New Year's Eve, the Square dreams of a visit to a one-dimensional world, 
    "Lineland", inhabited by men, who are lines, while the women are "lustrous 
    points".
5.  These points and lines are unable to see the Square as anything other than 
    a set of points on a line.
6.  Thus, the Square attempts to convince the realm's monarch of a second 
    dimension but cannot do so.
7.  In the end, the monarch of Lineland tries to kill the Square rather than 
    tolerate him any further.
8.  Following this vision, the Square is visited by a sphere.
9.  Similar to the "points" in Lineland, he is unable to see the three-
    dimensional object as anything other than a circle (more precisely, a 
    disk).
10. The Sphere then levitates up and down through Flatland, allowing the Square 
    to see the circle expand and contract between a great circle and small 
    circles.
11. The Sphere then tries further to convince the Square of the third dimension 
    by dimensional analogies (a point becomes a line, a line becomes a square).
12. The Square is still unable to comprehend the third dimension, so the 
    Sphere resorts to deeds: he gives information about the "insides" of the 
    house, moves a tablet through the third dimension, and even goes inside the 
    Square for a moment.
13. Still unable to comprehend the third dimension, the Square is taken by the 
    Sphere to the third dimension, Spaceland.
14. This Sphere visits Flatland at the turn of each millennium to introduce a 
    new apostle to the idea of a third dimension in the hope of eventually 
    educating the population of Flatland.
15. From the safety of Spaceland, they can oversee the leaders of Flatland, 
    acknowledging the Sphere's existence and prescribing the silencing.
16. After this proclamation is made, many witnesses are massacred or imprisoned 
    (according to caste), including the Square's brother.
17. After the Square's mind is opened to new dimensions, he tries to convince 
    the Sphere of the theoretical possibility of the existence of a fourth 
    dimension and higher spatial dimensions.
18. The Sphere at first scoffs at the idea of higher dimensions, just as the 
    Square had done, showing that his comprehension is not as broad as he had 
    thought.
19. Still, the Sphere returns his student to Flatland in disgrace.
20. The Square then has a dream in which the Sphere revisits him, this time to 
    introduce him to a zero-dimensional space, Pointland, of whom the Point 
    (sole inhabitant, monarch, and universe in one) perceives any communication 
    as a thought originating in his own mind (cf.
21. Solipsism).
22. The Square recognises the ignorance of the monarchs of Pointland and 
    Lineland correspond with his own (and the Sphere's) previous ignorance of 
    the existence of higher dimensions than their own.
23. Once returned to Flatland, the Square cannot convince anyone of Spaceland's 
    existence, especially after official decrees are announced that anyone 
    preaching the existence of three dimensions will be imprisoned (or 
    executed, depending on caste).
24. For example, he tries to convince his relative of the third dimension but 
    cannot move a square "upward," as opposed to forward or sideways.
25. Eventually, the Square himself is imprisoned for just this reason, with 
    only occasional contact with his brother, who is imprisoned in the same 
    facility.
26. He cannot convince his brother, even after all they have both seen.
27. Seven years after being imprisoned, "A. Square" writes out the book 
    Flatland as a memoir, hoping to keep it as posterity for a future 
    generation that can see beyond their two-dimensional existence.
\end{Verbatim}

\subsection{\emph{Adam Bede} by George Eliot}
\label{sec:507-summary}

\begin{Verbatim}[frame=single,commandchars=\\\{\},fontsize=\small]
1.  Adam, a local carpenter much admired for his integrity and intelligence, is 
    in love with Hetty.
2.  She is attracted to Arthur, the local squire's charming grandson and heir, 
    and falls in love with him.
3.  When Adam interrupts a tryst between them, Adam and Arthur fight.
4.  Arthur agrees to give up Hetty and leaves Hayslope to return to his militia.
5.  After he leaves, Hetty Sorrel agrees to marry Adam but shortly before their 
    marriage, discovers that she is pregnant.
6.  In desperation, she leaves in search of Arthur but cannot find him.
7.  \textcolor{skipred}{Unwilling to return to the village on account of the shame and ostracism she would have to endure, she delivers her baby with the assistance of a friendly woman she encounters.}
8.  \textcolor{skipred}{She subsequently abandons the infant in a field but not being able to bear the child's cries, she tries to retrieve the infant.}
9.  \textcolor{skipred}{However, she is too late, the infant having already died of exposure.}
10. Hetty is caught and tried for child murder.
11. She is found guilty and sentenced to hang.
12. Dinah enters the prison and pledges to stay with Hetty until the end.
13. Her compassion brings about Hetty's contrite confession.
14. When Arthur Donnithorne, on leave from the militia for his grandfather's 
    funeral, hears of her impending execution, he races to the court and has 
    the sentence commuted to penal transportation.
15. Ultimately, Adam and Dinah, who gradually become aware of their mutual 
    love, marry and live peacefully with his family.
\end{Verbatim}

\subsection{\emph{Anthem} by Ayn Rand}
\label{sec:1250-summary}

\begin{Verbatim}[frame=single,commandchars=\\\{\},fontsize=\small]
1.  Equality 7-2521, a 21-year-old man writing by candlelight in a tunnel under 
    the earth, tells the story of his life up to that point.
2.  He exclusively uses plural pronouns ("we", "our", "they") to refer to 
    himself and others.
3.  He was raised like all children in his society, away from his parents in 
    collective homes: the Home of Infants from birth until five years old, then 
    the Home of Students from five to fifteen.
4.  He believes he has a "curse" that makes him learn quickly and ask many 
    questions.
5.  He excels at the Science of Things and dreams of becoming a Scholar, but 
    when the Council of Vocations assigns his Life Mandate at fifteen, he is 
    assigned to be a Street Sweeper.
6.  Equality 7-2521 accepts his street sweeping assignment as penance for his 
    Transgression of Preference in secretly desiring to be a Scholar.
7.  He works with the handicapped Union 5-3992 and International 4-8818, the 
    latter of whom is Equality 7-2521's only friend (which is another 
    Transgression of Preference, because all are supposedly equal in their 
    society).
8.  Despite International 4-8818's protests that any exploration unauthorized 
    by a Council is forbidden, Equality 7-2521 explores an underground tunnel 
    near the City Theatre tent, and finds metal tracks.
9.  Equality 7-2521 believes the tunnel is from the Unmentionable Times of the 
    distant past.
10. He begins sneaking away from his community at night to use the tunnel as a 
    laboratory for scientific experiments, using garbage he has taken from the 
    Home of the Scholars.
11. He is using stolen paper from the Home of the Clerks to write his journal 
    by candlelight, using candles stolen from the larder at the Home of the 
    Street Sweepers.
12. While cleaning a road at the edge of the city, Equality 7-2521 meets 
    Liberty 5-3000, a 17-year-old Peasant girl who works in the fields.
13. He commits another transgression by thinking constantly of her, instead of 
    waiting to be assigned a woman at the annual Time of Mating, in which men 
    aged twenty and over, and women of eighteen and over, are assigned to each 
    other solely for breeding.
14. She has dark eyes and golden hair, and he names her "The Golden One".
15. When he speaks to her, he discovers that she also thinks of him.
16. He reveals his secret name for her, and Liberty 5-3000 tells Equality 
    7-2521 she has named him "The Unconquered".
17. Continuing his scientific work, Equality 7-2521 rediscovers electricity.
18. In the ruins of the tunnel, he finds a glass box with wires that gives off 
    light when he passes electricity through it.
19. He decides to take his discovery to the World Council of Scholars; he 
    thinks such a great gift to mankind will outweigh his many transgressions 
    and lead to him being made a Scholar.
20. However, one night, his absence from the Home of the Street Sweepers is 
    noticed.
21. He is whipped and held in the Palace of Corrective Detention.
22. The night before the World Council of Scholars is set to meet, he easily 
    escapes; essentially walking out of the hall as there are no guards because 
    no one has ever attempted escape before.
23. The next day, he presents his work to the World Council of Scholars.
24. Horrified that he has done unauthorized research, they assail him as a 
    "wretch" and a "gutter cleaner" and say he must be punished.
25. They want to destroy his discovery so it will not disrupt the plans of the 
    World Council and the Department of Candles.
26. Equality 7-2521 seizes the box, cursing the council before fleeing into the 
    Uncharted Forest that lies outside the city.
27. In the forest, Equality 7-2521 sees himself as damned for having left his 
    fellow men, but he enjoys his freedom.
28. No one will pursue him into this forbidden place.
29. He only misses Liberty 5-3000.
30. On his second day of living in the forest, Liberty 5-3000 appears; she 
    followed him into the forest and vows to stay with him forever.
31. They live together in the forest and try to express their love for one 
    another, but they lack the words to speak of love as individuals.
32. They find a house from the Unmentionable Times in the mountains and decide 
    to live in it.
33. While reading books from the house's library, Equality 7-2521 discovers the 
    word "I" and tells Liberty 5-3000 about it.
34. Her first words to him are, ÒI love you.
35. Ó Having rediscovered individuality, they give themselves new names from 
    the books: Equality 7-2521 becomes "Prometheus" and Liberty 5-3000 becomes 
    "Gaea".
36. Months later, Gaea is pregnant with Prometheus's child.
37. Prometheus wonders how men in the past could have given up their 
    individuality; he plans a future in which they will regain it.
\end{Verbatim}

\subsection{\emph{The Phantom of the Opera} by Gaston Leroux}
\label{sec:175-summary}

\begin{Verbatim}[frame=single,commandchars=\\\{\},fontsize=\small]
1.  In the 1880s, in Paris, the Palais Garnier Opera House is believed to be 
    haunted by an entity known as the 'Phantom of the Opera', or simply the 
    'Opera Ghost', after stagehand Joseph Buquet is found hanged, the noose 
    around his neck missing.
2.  At a gala performance for the retirement of the opera house's managers, a 
    young, little-known Swedish soprano, Christine Daae, is called upon to sing 
    in place of the opera's leading soprano, Carlotta, who is ill.
3.  Christine's performance is a success.
4.  Among the audience is the Vicomte Raoul de Chagny, who recognizes her as 
    his childhood playmate and recalls his love for her.
5.  He attempts to visit her backstage, where he hears a man complimenting her 
    from inside her dressing room.
6.  He investigates the room once Christine leaves, only to find it empty.
7.  At Perros-Guirec, Christine meets with Raoul, who confronts her about the 
    voice he heard in her room.
8.  Christine says she has been tutored by the "Angel of Music", whom her 
    father used to tell her and Raoul about.
9.  When Raoul suggests that she might be the victim of a prank, she storms 
    off.
10. Christine visits her father's grave one night, where a mysterious figure 
    appears and plays the violin for her.
11. Raoul attempts to confront the figure but is struck and knocked out in the 
    process.
12. Back at the Palais Garnier, the new managers receive a letter from the 
    Phantom demanding that they allow Christine to perform the lead role of 
    Marguerite in Faust and that box five be left empty for his use, lest they 
    perform in a house with a curse on it.
13. The managers assume his demands are a prank and ignore them.
14. Soon after, Carlotta ends up croaking like a toad, and a chandelier drops 
    into the audience, killing a spectator.
15. The Phantom, having abducted Christine from her dressing room, reveals 
    himself as a deformed man called Erik.
16. Erik intends to hold her prisoner in his lair with him for a few days.
17. Still, she causes him to change his plans when she unmasks him and, to the 
    horror of both, beholds his skull-like face.
18. Fearing that she will leave him, he decides to hold her permanently.
19. However, when Christine requests her release after two weeks, he agrees on 
    the condition that she wear his ring and be faithful to him.
20. On the roof of the Opera House, Christine tells Raoul about her abduction 
    and makes Raoul promise to take her away to where Erik can never find her,
    even if she resists.
21. Raoul says he will act on his promise the next day.
22. Unbeknownst to Christine and Raoul, Erik is watching them and overheard 
    their whole conversation.
23. The following night, the enraged and jealous Erik abducts Christine during 
    a production of Faust and tries to force her to marry him.
24. Raoul is led by a mysterious Opera House regular, 'the Persian', into 
    Erik's secret lair in the bowels of the building.
25. Still, they end up trapped in a mirrored room by Erik, who threatens that 
    unless Christine agrees to marry him, he will kill them and everyone in the 
    Opera House by using explosives.
26. Under duress, Christine agrees to marry Erik.
27. Erik initially tries to drown Raoul and the Persian, using the water which 
    would have been used to douse the explosives.
28. Still, Christine begs, promising him she would not kill herself after 
    becoming his bride.
29. Erik releases Raoul and 'the Persian' from his torture chamber.
30. When Erik is alone with Christine, he lifts his mask to kiss her on her 
    forehead and is eventually given a kiss back.
31. Erik reveals he has never kissed anyone, including his own mother, who 
    would run away if he ever tried to kiss her.
32. Moved, he and Christine cry together.
33. She also holds his hand and says, "Poor, unhappy Erik", which reduces him 
    to "a dog ready to die for her".
34. He allows 'the Persian' and Raoul to escape, though not before making 
    Christine promise that she will visit him on his death day and return the 
    ring he gave her.
35. He also makes 'the Persian' promise that afterward, he will go to the 
    newspaper and report his death, as he will die soon "of love."
36. Later, Christine returns to Erik's lair, and per his request, returns the 
    ring and buries him 'somewhere he will never be found'.
37. Afterward, a local newspaper runs the note: "Erik is dead".
38. Christine and Raoul then elope together, never to return.
39. The epilogue reveals that Erik was born deformed and is the son of a 
    construction business owner.
40. He ran away from his native Normandy to work in fairs and caravans, 
    schooling himself in the circus arts across Europe and Asia, and eventually 
    building trick palaces in Persia and Turkey.
41. Returning to France, he started his own construction business.
42. After being subcontracted to work on the Palais Garnier's foundations, Erik 
    discreetly built his secret lair with hidden passages and other tricks that 
    allowed him to spy on the managers.
\end{Verbatim}

\subsection{\emph{O Pioneers!} by Willa Cather}
\label{sec:24-summary}

\begin{Verbatim}[frame=single,commandchars=\\\{\},fontsize=\small]
1.  On a windy January day in Hanover, Nebraska, Alexandra Bergson is with her 
    five-year-old brother Emil, whose little kitten has climbed a telegraph 
    pole and is afraid to come down.
2.  Alexandra asks her neighbor and friend Carl Linstrum to retrieve the 
    kitten.
3.  Later, Alexandra finds Emil in the general store with Marie Tovesky.
4.  They are playing with the kitten.
5.  Marie lives in Omaha and is visiting her uncle Joe Tovesky.
6.  Alexandra's father is dying, and it is his wish that she run the farm after 
    he is gone.
7.  Alexandra and her brothers Oscar and Lou later visit Ivar, known as Crazy 
    Ivar because of his unorthodox views.
8.  For instance, he sleeps in a hammock, believes in killing no living thing 
    and goes barefoot summer and winter.
9.  But he is known for healing sick animals.
10. Alexandra is concerned about their hogs as the hogs of many of their 
    neighbors are dying.
11. Crazy Ivar advises her to keep their hogs clean rather than letting them 
    live in filth and to give them fresh, clean water and good food.
12. This simply confirms Oscar's and Lou's opinion that Ivar deserves the name 
    Crazy Ivar.
13. Alexandra, however, starts making plans for where she will relocate the hogs.
14. After years of crop failure, many of the Bergson's neighbors are selling out, even if it means taking a loss.
15. Then they learn the Linstrums have also decided to leave.
16. Oscar and Lou want to leave too, but neither their mother nor Alexandra will.
17. After visiting villages downwards to see how they are getting on, Alexandra talks her brothers into mortgaging the farm to buy more land, in hopes of ending up as rich landowners.
18. Sixteen years later, the farms are now prosperous.
19. Alexandra and her brothers have divided up their inheritance, and Emil has just returned from college.
20. The Linstrum farm has failed, and Marie, now married to Frank Shabata, has bought it.
21. That same day, the Bergsons are surprised by a visit from Carl Linstrum, whom they have not seen for thirteen years.[2] [Note: Carl says it has been sixteen years, but this is a textual error.
22. John Bergson died sixteen years earlier, and Carl's family left during the drought that occurred three years later.[citation needed]] Having failed at a job in Chicago, he is on his way to Alaska, but decides to stay with Alexandra for a while.
23. Carl notices the growing flirtatious relationship between Emil and Marie.
24. Lou and Oscar suspect that Carl wants to marry Alexandra, and are resentful at the idea that Carl might end up owning Alexandra's share of the farm, which they still view as belonging to them and which would otherwise be inherited by their children.
25. This causes problems between Alexandra and her brothers, and they stop speaking to each other.
26. Carl, recognizing a problem, decides to leave for Alaska.
27. At the same time, Emil announces he is leaving to travel through Mexico.
28. \textcolor{skipred}{Alexandra is left alone.}
29. Winter has settled down over the Divide again; the season in which Nature recuperates, in which she sinks to sleep between the fruitfulness of autumn and the passion of spring.
30. The birds have gone.
31. The teeming life that goes on down in the long grass is exterminated.
32. The prairie dog keeps his hole.
33. The rabbits run shivering from one frozen garden patch to another and are hard put to it to find frost-bitten cabbage stalks.
34. At night the coyotes roam the wintry waste, howling for food.
35. The variegated fields are all one color now; the pastures, the stubble, the roads, and the sky are the same leaden gray.
36. The hedgerows and trees are scarcely perceptible against the bare earth, whose slaty hue they have taken on.
37. The ground is frozen so hard that it bruises the foot to walk on the roads or in the plowed fields.
38. It is like an iron country, and the spirit is oppressed by its rigor and melancholy.
39. One could easily believe that in that dead landscape the germs of life and fruitfulness were extinct forever.
40. Alexandra spends the winter alone, except for occasional visits from Marie, whom she visits with Mrs. Lee, Lou's mother-in-law.
41. She also has an increased number of mysterious dreams she has had since girlhood.
42. These dreams are about a strong, god-like male figure who carries her over the fields.
43. Emil returns from Mexico City.
44. His best friend, Amedee, is now married with a young son.
45. At a fair at the French church, Emil and Marie kiss for the first time.
46. They later confess their illicit love, and Emil determines to leave for law school in Michigan.
47. Before he leaves, Amedee dies from a ruptured appendix, and as a result both Emil and Marie realize what they value most.
48. Before leaving for Michigan, Emil stops by Marie's farm to say one last goodbye, and they fall into a passionate embrace beneath the white mulberry tree.
49. They stay there for several hours, until Marie's husband, Frank, finds them and shoots them in a drunken rage.
50. He flees to Omaha, where he later turns himself in for the crime.
51. Ivar discovers Emil's abandoned horse, leading him to search for the boy and discover the bodies.
52. After Emil's death Alexandra is distraught, in shock, and slightly dazed.
53. She goes off in a rainstorm.
54. Ivar goes looking for her and brings her back home, where she sleeps fitfully and dreams about death.
55. She then decides to visit Frank in Lincoln where he is incarcerated.
56. While in town she walks by Emil's university campus, comes upon a polite young man who reminds her of Emil, and feels better.
57. The next day she talks to Frank in prison.
58. He is bedraggled and can barely speak properly, and she promises to do what she can to see him released; she bears no ill will toward him.
59. She then receives a telegram from Carl, telling her that he is back.
60. They decide to marry, unconcerned with the approval of her brothers.
\end{Verbatim}

\subsection{\emph{The Triumph of the Scarlet Pimpernel} by Baroness Orczy}
\label{sec:65695-summary}

\begin{Verbatim}[frame=single,commandchars=\\\{\},fontsize=\small]
1.  The story starts in Paris in April 1794, year II of the French Revolution.
2.  Theresia Cabarrus is a beautiful but shallow Spaniard who is betrothed to Citizen Tallien the popular Representative in the Convention and one of Robespierre's inner circle.
3.  She is credited with exercising a mellowing influence over Tallien, whom she met in Bordeaux but although she is engaged to be married to him, what little love she has appears to be lavished on another.
4.  Bertrand Moncrif is a good-looking but impulsive young man who appears determined to martyr himself in opposition to the revolutionary government.
5.  To this end, he has gathered the siblings of his long-term sweetheart, Regine de Serval, into his plan to denounce Robespierre at one of the Fraternal suppers.
6.  Despite warnings from Regine he insists on carrying through his plans which inevitably go awry and the wrath of the mob is soon turned towards the small group.
7.  After a timely intervention on the part of the Scarlet Pimpernel, using the guise of the coal heaver Rateau (who also appears in several short stories in The League of the Scarlet Pimpernel - The Cabaret de la Liberte, Needs Must and A Battle of Wits), the de Servals are saved from a lynching while Moncrif lies unconscious and unseen under a table.
8.  In England, Moncrif and the de Servals are finally free to resume an almost normal life.
9.  Theresia arrives at Dover dressed in men's clothes and claiming she has been driven out of France by her association with Bertrand, in fear of her life.
10. An obviously staged row between the Spaniard and Chauvelin outside Sir Percy's cottage fails to persuade our hero that she is up to anything but mischief, but he seems to relish the prospect of such an intelligent and wily adversary and promises not to reveal her true identity to anyone for he "is a lover of sport."
11. With her plans to seduce Percy scuppered, Theresia turns her attention to Sir Percy's wife Marguerite and uses an all too willing Bertrand to set the trap.
12. Lady Blakeney is kidnapped yet again and taken to France and imprisoned as bait for Sir Percy.
\end{Verbatim}

\subsection{\emph{Beauvallet} by Georgette Heyer}
\label{sec:75547-summary}

\begin{Verbatim}[frame=single,commandchars=\\\{\},fontsize=\small]
1.  The year is 1586 and 35-year-old Sir Nicholas Beauvallet (great-great-great-grandson of Simon the Coldheart) - 'El Beauvallet' to the Spanish and called 'Mad Nick' by his men - is one of the most daring pirates of the Elizabethan era.
2.  With the blessing of the Queen, this friend and former associate of Sir Walter Raleigh sails the seas in his ship The Venture with the intention of plundering any Spanish ships that come his way.
3.  It is while engaged in one such enterprise that he takes the galleon on which the retired and ailing Governor Don Manuel de Rada y Sylva is returning home, accompanied by his daughter Dominica.
4.  Having plundered their own ship, Beauvallet promises to take them with him the rest of the way to Spain.
5.  Dona Dominica's haughty spirit appeals to Beauvallet and he vows to come back to claim her within the year.
6.  Upon landing in England, he rides on a visit to his elder brother Gerard who, lacking heirs himself, reminds Nicholas that it is his duty to continue the family line.
7.  Three months later, Sir Nicholas visits a relation in France and then rides south towards the Spanish border, accompanied by his servant Joshua.
8.  There he meets the young Chevalier Claude de Guise, on a mission to deliver a secret message to the Spanish King.
9.  On catching the Chevalier trying to steal his horse, Beauvallet kills him in a sword fight and assumes his identity.
10. Travelling as a Frenchman provides Beauvallet with a convenient disguise in a rigidly Catholic land where the English are abhorred as Protestant heretics.
11. Having arrived in Madrid, he learns that Dominica's father has died and that she is now under the protection of her uncle, Don Rodriguez.
12. But even as he makes contact with Dominica's new family, Beauvallet arouses the suspicion of the French ambassador, M. de Lauviniere, who makes enquiries about him.
13. Henceforward Sir Nicholas must engineer his escape with his chosen bride while avoiding the clutch of the Inquisition, or imprisonment as a spy, and the jealous intrigues of Dominica's other suitors.
\end{Verbatim}

\subsection{\emph{The Golovlyov Family} by Mikhail Saltykov-Shchedrin}
\label{sec:44237-summary}

\begin{Verbatim}[frame=single,commandchars=\\\{\},fontsize=\small]
1.  Arina Petrova, matriarch of the Golovlyov family, runs a large estate (4,000 serfs) in Russia.
2.  She learns that her first born son, Stepan/Styopka/The Dolt has squandered the land and house she gave to him.
3.  She was a practical and strict noblewoman, and she banished her drunken husband Vladmir Mihailitch to his room for several decades while she ran the estate.
4.  Arina sent Stepan to college, where he was the class clown.
5.  He worked in a series of government jobs, but lost them all due to laziness.
6.  He returns home after losing his estate.
7.  Arina's second child is Anna, who ran off and married a musician named Ulanov.
8.  Anna has twin girls Anninka and Lubinka.
9.  Ulanov soon abandons his family, and Anna dies of an illness 3 months later.
10. Arina hoped to be rid of her children by giving them estates.
11. She was very upset when Anna died ('throwing her two brats on to my shoulders') and when Stepan returned.
12. Her third son is Porphyry/Iudushka/Bloodsucker; he is an obsequious, scheming son.
13. Her fourth son is Pavel; he is normal and unremarkable in any way.
14. She keeps her family on a very tight financial leash, and they live at poverty level despite their wealth.
15. Stepan, having nowhere to go, sadly travels back home.
16. Arina declares that she hates him, and says "he has been nothing but a worry and a disgrace to me all his life.
17. Ó She wonders who she is saving her money for.
18. Stepan is let back into the estate, but becomes depressed and runs away one winter evening.
19. He is found alive but never speaks again; he dies shortly thereafter.
20. Serfdom is abolished by the Tsar.
21. Vladimir Mihailitch dies and Arina divides her estate between her 2 remaining children.
22. Pavel dies from alcoholism having refused to make a will.
23. This means everything goes to Porphyry.
24. Arina leaves the big house and moves in with her orphaned twin granddaughters to the Pogorelka estate.
25. \textcolor{skipred}{Porphyry gets married and has 2 sons.}
26. He becomes very religious, but it is only for show.
27. While Arina is very strict with the girls, her energy for managing an estate is waning.
28. The girls demand to leave, and she lets them.
29. Arina becomes depressed living in an empty house in rural Russia.
30. She begins visiting her son for good food and conversation.
31. Porphyry's wife dies, and he takes a lover, Yevpraxeya.
32. The twins write back that they have both become successful provincial actresses and can support themselves.
33. Porphyry's first son, Volodya, kills himself.
34. Porphyry's second son, Pyotr/Petenka, arrives unexpectedly to beg for money.
35. He was an infantry treasurer that has gambled the unit's money away.
36. Pyotr explains that Volodya killed himself because Porphyry refused to support him and his new wife (Volodya informed, but didn't ask permission for the marriage).
37. Porphyry refuses to pay the debt, sending Pyotr away to await trial.
38. Arina falls ill.
39. Porphyry calls for the twins as Arina dies.
40. Prophyry's son Pyotr is banished to Siberia and he dies on the way there.
41. Anninka arrives to settle some paperwork since Arina died.
42. The twins want to continue their acting careers, and have no interest in moving back.
43. Porphyry has become a compulsive talker.
44. Anninka's visit with Porphyry is awful and she leaves as soon as the papers are signed.
45. Yevpraxeya becomes pregnant with Porphyry's child.
46. He is afraid of a scandal, so he ignores the pregnancy and denies any involvement.
47. She gives birth to a boy; Porphyry feels guilty for having a child out of wedlock, so he sends the baby to the orphanage without Yevpraxeya's knowledge.
48. Yevpraxeya, deprived of her child, decides to ruin Porphyry's life.
49. She begins to complain incessantly and refuses to listen to Porphyry's constant babbling.
50. She takes several lovers.
51. Porphyry becomes a recluse and begins to lose his mind.
\end{Verbatim}

\subsection{\emph{Anne of Green Gables} by L.~M.~Montgomery}
\label{sec:45-summary}

\begin{Verbatim}[frame=single,commandchars=\\\{\},fontsize=\small]
1.  Anne Shirley, a young orphan from the fictional community of Bolingbroke, Nova Scotia (based upon the real community of New London, Prince Edward Island), is sent to live with Marilla and Matthew Cuthbert, unmarried siblings in their fifties and sixties, after a childhood spent in strangers' homes and orphanages.
2.  Marilla and Matthew had originally sought to adopt a boy from the orphanage to help Matthew run their farm at Green Gables, which is set in the fictional town of Avonlea (based on Cavendish, Prince Edward Island).
3.  Through a misunderstanding, the orphanage sends Anne instead.
4.  Anne is fanciful, imaginative, eager to please, and dramatic.
5.  She is also adamant her name should always be spelled with an "e" at the end.
6.  However, she is defensive about her appearance, despising her red hair, freckles, and pale, thin frame, but liking her nose.
7.  She is talkative, especially when it comes to describing her fantasies and dreams.
8.  At first, stern Marilla says Anne must return to the orphanage, but after much observation and consideration, along with kind, quiet Matthew's encouragement, Marilla decides to let her stay.
9.  Anne takes much joy in life and adapts quickly, thriving in the close-knit farming village.
10. Her imagination and talkativeness soon brighten up Green Gables.
11. The book recounts Anne's struggles and joys in settling into Green Gables (the first real home she's ever known): the country school where she quickly excels in her studies; her friendship with Diana Barry, the girl living next door (her best or "bosom friend" as Anne fondly calls her); her budding literary ambitions; and her rivalry with her classmate Gilbert Blythe, who teases her about her red hair.
12. For that, he earns her instant hatred, although he apologizes several times.
13. As time passes, however, Anne realizes she no longer hates Gilbert, but her pride and stubbornness keep her from speaking to him.
14. The book also follows Anne's adventures in Avonlea.
15. Episodes include playtime with her friends Diana, calm, placid Jane Andrews, and beautiful, boy-crazy Ruby Gillis.
16. She has run-ins with the unpleasant Pye sisters, Gertie and Josie, and frequent domestic "scrapes" such as dyeing her hair green while intending to dye it black, and accidentally getting Diana drunk by giving her what she thinks is raspberry cordial but which turns out to be currant wine.
17. At sixteen, Anne goes to Queen's Academy to earn a teaching license, along with Gilbert, Ruby, Josie, Jane, and several other students, excluding Diana, much to Anne's dismay.
18. She obtains her license in one year instead of the usual two and wins the Avery Scholarship awarded to the top student in English.
19. This scholarship would allow her to pursue a Bachelor of Arts (B.A.) degree at the fictional Redmond College (based on the real Dalhousie College) on the mainland in Nova Scotia.
20. Near the end of the book, however, tragedy strikes when Matthew dies of a heart attack after learning that all of his and Marilla's money has been lost in a bank failure.
21. Out of devotion to Marilla and Green Gables, Anne gives up the scholarship to stay at home and help Marilla, whose eyesight is failing.
22. She plans to teach at the Carmody school, the nearest school available, and return to Green Gables on weekends.
23. In an act of friendship, Gilbert Blythe gives up his teaching position at the Avonlea School in favor of Anne, to work at the White Sands School instead, knowing that Anne wants to stay close to Marilla after Matthew's death.
24. After this kind act, Anne and Gilbert's friendship is cemented, and Anne looks forward to what life will bring next.
\end{Verbatim}

\subsection{\emph{David Copperfield} by Charles Dickens}
\label{sec:766-summary}

\begin{Verbatim}[frame=single,commandchars=\\\{\},fontsize=\small]
1.  \textcolor{skipred}{The story follows the life of David Copperfield from childhood to maturity.}
2.  David was born in Blunderstone, Suffolk, England, six months after the death of his father.
3.  David spends his early years residing in a small house called the Rookery.
4.  His loving and childish mother, Clara Copperfield, and their kindly housekeeper, Clara Peggotty, bring him up here; they call him Davy.
5.  When he is seven years old, his mother surprises David by marrying Edward Murdstone while the boy is visiting Peggotty's family in Yarmouth.
6.  Her brother, a fisherman named Mr Peggotty, lives in a beached barge, with his adopted niece and nephew Emily and Ham, and an elderly widow, Mrs Gummidge. "
7.  Little Em'ly" is somewhat spoiled by her fond foster father, and David is in love with her.
8.  Here he is known as "Master Copperfield.
9.  " On his return, David discovers his mother has married and is immediately given good reason to dislike his stepfather, Murdstone, who believes exclusively in stern, harsh parenting, calling it "firmness".
10. David has similar feelings for Murdstone's sister Jane, who moves into the house soon afterwards.
11. Between them, they tyrannise David and his poor mother, making their lives miserable.
12. When David falls behind in his studies, Murdstone thrashes him, and David bites his hand; in consequence, he is sent away to Salem House, a boarding school, under a ruthless headmaster named Mr Creakle.
13. There he is befriended by two quite different older boys, James Steerforth and Tommy Traddles.
14. He develops an impassioned admiration for Steerforth, perceiving him as someone noble, who could do great things if he would, and one who pays attention to him.
15. David goes home for the holidays to learn that his mother has given birth to a baby boy.
16. Shortly after David returns to Salem House, his mother and her baby die, and David returns home immediately.
17. Peggotty marries the local carrier, Mr Barkis.
18. Murdstone sends David to work for a wine merchant in London - a business of which Murdstone is a joint owner.
19. After some months, David's friendly but spendthrift landlord, Wilkins Micawber, is arrested for debt and sent to the King's Bench Prison, and the rest of Mr. Micawber's family soon moves to the Prison too.
20. David visits the Micawbers regularly at the Prison, and boards nearby.
21. When Micawber's release is imminent, the Micawbers decide they will soon move to Plymouth.
22. David realises that will leave him alone in London, where no one cares about him.
23. He makes up his mind to run away to Dover to find his only known remaining relative, his eccentric and kind-hearted great-aunt Betsey Trotwood.
24. She had come to Blunderstone at his birth, only to depart in ire upon learning that he was not a girl.
25. However, she takes it upon herself to raise David, despite Murdstone's attempt to regain custody of him.
26. She encourages him to 'be as like his sister, 'Betsey Trotwood' as he can be - that is, to meet the expectations she had for the girl who was never born.
27. David's great-aunt renames him "Trotwood Copperfield" and addresses him as "Trot", one of several names others call David in the novel.
28. David's aunt sends him to a better school than the last he attended.
29. It is run by kind Dr. Strong, whose methods inculcate honour and self-reliance in his pupils.
30. During term, David lodges with the lawyer Mr Wickfield and his daughter Agnes, who becomes David's friend and confidante.
31. Wickfield's clerk, Uriah Heep, also lives at the house.
32. By devious means, Uriah Heep gradually gains a complete ascendancy over the aging and alcoholic Wickfield, to Agnes's great sorrow.
33. Heep, as he maliciously confides to David, aspires to marry Agnes.
34. Ultimately with the aid of Micawber, who has been employed by Heep as a secretary, his fraudulent behaviour is revealed. (
35. At the end of the book, David encounters him in prison, convicted of attempting to defraud the Bank of England.)
36. After completing school, David apprentices to be a proctor.
37. During this time, due to Heep's fraudulent activities, his aunt's fortune has diminished.
38. David toils to make a living.
39. He works mornings and evenings for his former teacher Dr Strong as a secretary, and also starts to learn shorthand, with the help of his old school-friend Traddles, upon completion reporting parliamentary debate for a newspaper.
40. With considerable moral support from Agnes and his own great diligence and hard work, David ultimately finds fame and fortune as an author, writing fiction.
41. David's romantic but self-serving school friend, Steerforth, also re-acquainted with David, goes on to seduce and dishonour Emily, offering to marry her off to his manservant Littimer before deserting her in Europe.
42. Her uncle Mr Peggotty manages to find her with the help of Martha, who had grown up in their part of England and then settled in London.
43. Ham, who had been engaged to marry Emily before the tragedy, dies in a fierce storm off the coast while rescuing shipwreck victims.
44. Steerforth was aboard the ship and also dies.
45. Mr Peggotty takes Emily to a new life in Australia, accompanied by Mrs Gummidge and the Micawbers, where all eventually find security and happiness.
46. Meanwhile, David has fallen in love with Dora Spenlow, and marries her.
47. Their marriage proves troublesome for David in the sense of everyday practical affairs, but he never stops loving her.
48. Dora dies early in their marriage after a miscarriage.
49. After Dora's death, Agnes encourages David to return to normal life and his profession of writing.
50. While living in Switzerland to dispel his grief over so many losses, David realises that he loves Agnes.
51. Upon returning to Britain, after a failed attempt to conceal his feelings, David finds that Agnes loves him too.
52. They quickly marry, and in this marriage he finds true happiness.
53. David and Agnes then have at least five children, including a daughter named after his great-aunt, Betsey Trotwood.
\end{Verbatim}

\subsection{\emph{Main Street} by Sinclair Lewis}
\label{sec:543-summary}

\begin{Verbatim}[frame=single,commandchars=\\\{\},fontsize=\small]
1.  Carol Milford, the daughter of a judge, grew up in Mankato, Minnesota, and became an orphan in her teenage years.
2.  In college, she reads a book on village improvement in a sociology class and begins to dream of redesigning villages and towns.
3.  After college, she attends a library school in Chicago and is exposed to many radical ideas and lifestyles.
4.  She becomes a librarian in Saint Paul, Minnesota, the state capital, but finds the work unrewarding.
5.  She marries Will Kennicott, a doctor from the small town of Gopher Prairie.
6.  When they marry, Will convinces her to live in his hometown.
7.  Carol is filled with disdain for the town's physical ugliness and smug conservatism and immediately formulates plans to remake Gopher Prairie.
8.  She speaks with its members about progressive changes, joins women's clubs, distributes literature, and holds a party to liven up Gopher Prairie's inhabitants.
9.  Despite her efforts, these ventures are ineffective and she is constantly derided by the leading cliques.
10. She finds some comfort and companionship with a variety of social outsiders in the town, but these companions all fail to live up to her expectations.
11. After a political meeting of the Nonpartisan League is broken up by local authorities, Carol leaves her husband and moves to Washington, D.C. to become a clerk in a wartime government agency but she eventually returns.
12. Nevertheless, she does not feel defeated.
\end{Verbatim}

\subsection{\emph{Captain Blood} by Rafael Sabatini}
\label{sec:1965-summary}

\begin{Verbatim}[frame=single,commandchars=\\\{\},fontsize=\small]
1.  The protagonist is the sharp-witted Dr. Peter Blood, a fictional Irish physician who had had a wide-ranging career as a soldier and sailor (including a commission as a captain under the Dutch admiral Michiel de Ruyter) before settling down to practice medicine in the town of Bridgwater, Somerset.
2.  The story is told from the perspective of an omniscient narrator, who enables the reader to see the thoughts and views of many different characters.
3.  The narrator-perhaps meant to be Sabatini himself-claims to have acquired the story from the ship's logs of Blood's longtime companion Jeremy Pitt.
4.  The book opens with Blood attending to his geraniums while the town prepares to fight for James Scott, 1st Duke of Monmouth.
5.  He wants no part in the Monmouth Rebellion, but while attending to some of the rebels wounded at the Battle of Sedgemoor, Peter is arrested.
6.  During the Bloody Assizes, he is convicted by the infamous Judge Jeffreys of treason on the grounds that "if any person be in actual rebellion against the King, and another person-who really and actually was not in rebellion-does knowingly receive, harbour, comfort, or succour him, such a person is as much a traitor as he who indeed bore arms."
7.  The sentence for treason is death by hanging, but King James II, for purely financial reasons, has the sentence for Blood and other convicted rebels commuted to transportation to penal servitude in the Caribbean.
8.  Upon arrival on the island of Barbados, Blood is bought by Colonel William Bishop, initially for forced work in the Colonel's prison farms but later hired out by Bishop when Blood's skills as a physician prove superior to those of the local doctors.
9.  During his period of servitude, Blood wins the pity and sympathy of Arabella, Colonel Bishop's niece.
10. When a Spanish force attacks and raids Bridgetown, Blood escapes with a number of other convicts (including former shipmaster Jeremy Pitt, the one-eyed giant Edward Wolverstone, former gentleman Nathaniel Hagthorpe and two Royal Navy veterans, former petty officer Nicholas Dyke and former master gunner Ned Ogle).
11. The escapees capture the Spaniards' ship and sail away to become some of the most successful pirates in the Caribbean, hated and feared by the Spanish but always sparing English ships.
12. Colonel Bishop, humiliated by Blood's superior abilities and daring escape, devotes himself to capturing and executing Blood.
13. After the Glorious Revolution, Blood is pardoned.
14. As a reward for saving the colony of Jamaica from a French assault, he is appointed its governor in place of Colonel Bishop, who had abandoned his post to hunt for Blood, and the novel ends with the implication that Blood will not only marry Arabella but will also generously forgive Bishop.
\end{Verbatim}

\subsection{\emph{The Wind in the Willows} by Kenneth Grahame}
\label{sec:289-summary}

\begin{Verbatim}[frame=single,commandchars=\\\{\},fontsize=\small]
1.  The novel's characters are anthropomorphized animals.
2.  With the arrival of spring and fine weather outside, the good-natured Mole loses patience with spring cleaning, exclaiming, "Hang spring cleaning!"
3.  He leaves his underground home and comes up at the bank of the river, which he has never seen before.
4.  Here he meets Water Rat, a water vole, who takes Mole for a ride in his row boat.
5.  They get along well and spend many more days boating. "
6.  Ratty" teaches Mole the ways of the river, and the two friends live together in Ratty's riverside home.
7.  One summer day, Rat and Mole disembark near the grand Toad Hall and pay a visit to Toad.
8.  Toad is wealthy, jovial, friendly and kindhearted, but sometimes arrogant and rash; he regularly becomes obsessed with fads, only to abandon them abruptly.
9.  His current craze is his horse-drawn caravan.
10. When a passing automobile scares his horse and causes the caravan to overturn into a ditch, Toad's craze for caravan travel is immediately replaced by an obsession with motorcars.
11. Mole goes to the Wild Wood on a snowy winter's day, hoping to meet the elusive but wise and virtuous Badger.
12. Mole becomes lost in the woods, succumbs to fright and hides among the sheltering roots of a tree.
13. Rat finds him as snow begins to fall in earnest.
14. As they attempt to find their way home, Mole barks his shin on the bootscraper on Badger's doorstep.
15. Badger welcomes Rat and Mole into his large, cosy underground home, providing them with hot food, dry clothes and reassuring conversation.
16. Badger learns from his visitors that Toad has crashed seven cars, has been in hospital three times, and has been issued numerous fines resulting in great expense.
17. With the arrival of spring, the three of them place Toad under house arrest with themselves as guards.
18. However, Toad pretends to be sick, tricking Ratty into leaving for a doctor, and escapes.
19. Badger and Mole continue residing at Toad Hall in the hope that Toad may return.
20. Toad orders lunch at The Red Lion Inn and then sees a motorcar pull into the courtyard.
21. He takes the car, drives it recklessly, and is caught by the police.
22. He is sent to prison for 20 years, most of which is for "gross impertinence" to his captors.
23. In prison, Toad gains the sympathy of the gaoler's daughter.
24. With her help, he escapes, disguised as a washerwoman.
25. After a long series of misadventures he returns to the hole of the Water Rat.
26. Rat hauls Toad inside and informs him that Toad Hall has been taken over by weasels, stoats and ferrets from the Wild Wood, who have driven out Mole and Badger.
27. Armed to the teeth, Badger, Rat, Mole and Toad enter through the tunnel and pounce upon the unsuspecting Wild-Wooders, who are holding a celebratory party.
28. Having driven away the intruders, Toad holds a banquet to mark his return, during which he behaves both quietly and humbly.
29. He makes up for his earlier excesses by seeking out and compensating those he has wronged, and the four friends live happily ever after.
30. In addition to the main narrative, the book contains several independent short stories featuring Rat and Mole, such as an encounter with the wild god Pan while they are searching for Otter's son Portly, and Ratty's meeting with a Sea Rat.
31. \textcolor{skipred}{These appear for the most part between the chapters chronicling Toad's adventures, and they are often omitted from adaptations of the story as well as from some abridged versions.}
\end{Verbatim}

\subsection{\emph{Nostromo: A Tale of the Seaboard} by Joseph Conrad}
\label{sec:2021-summary}

\begin{Verbatim}[frame=single,commandchars=\\\{\},fontsize=\small]
1.  Charles Gould is a native Costaguanero of English descent who owns an important silver-mining concession near the key port of Sulaco.
2.  He is tired of the political instability in Costaguana and its concomitant corruption, and uses his wealth to support Ribiera's government, which he believes will finally bring stability to the country after years of misrule and tyranny by self-serving dictators.
3.  \textcolor{skipred}{Instead, Gould's refurbished silver mine and the wealth it has generated inspires a new round of revolutions and self-proclaimed warlords, plunging Costaguana into chaos.}
4.  Among others, the forces of the revolutionary General Montero invade Sulaco after securing the inland capital.
5.  Gould, adamant that his silver mine should not become spoil for his enemies, orders Nostromo, the trusted "Capataz de Cargadores" (Head Longshoreman) of Sulaco, to take the mine's most recent load of silver offshore, and arranges for the mine complex to be destroyed by dynamite if the coup leaders try to take it.
6.  Nostromo is an Italian expatriate who has risen to his position through his bravery and daring exploits. ("
7.  \textcolor{skipred}{Nostromo" is Italian for "shipmate" or "boatswain", but the name could also be considered a corruption of the Italian phrase "nostro uomo" or "nostr'uomo", meaning "our man").}
8.  \textcolor{skipred}{Nostromo's real name is Giovanni Battista Fidanza ÑFidanza meaning "trust" in archaic Italian.}
9.  Nostromo is a commanding figure in Sulaco, respected by the wealthy Europeans and seemingly limitless in his abilities to command power among the local population.
10. \textcolor{skipred}{He is, however, never admitted to become a part of upper-class society, but is instead viewed by the rich as their useful tool.}
11. He is believed by Charles Gould and his own employers to be incorruptible, and it is for this reason that Nostromo is entrusted with removing the silver from Sulaco to keep it from the revolutionaries.
12. Accompanied by the young journalist Martin Decoud, Nostromo sets off to smuggle the silver out of Sulaco.
13. However, the lighter on which the silver is being transported is struck at night in the waters off Sulaco by a transport carrying the invading revolutionary forces under the command of Colonel Sotillo.
14. Nostromo and Decoud manage to save the silver by putting the lighter ashore on Great Isabel.
15. Decoud and the silver are deposited on the deserted island of Great Isabel in the expansive bay off Sulaco, while Nostromo scuttles the lighter and manages to swim back to shore undetected.
16. Back in Sulaco, Nostromo's power and fame continues to grow as he daringly rides over the mountains to summon the army which ultimately saves Sulaco's powerful leaders from the revolutionaries and ushers in the independent state of Sulaco.
17. In the meantime, left alone on the deserted island, Decoud eventually loses his mind.
18. He takes the small lifeboat out to sea and there shoots himself, after first weighing his body down with some of the silver ingots so that he would sink into the sea.
19. His exploits during the revolution do not bring Nostromo the fame he had hoped for, and he feels slighted and used.
20. Feeling that he has risked his life for nothing, he is consumed by resentment, which leads to his corruption and ultimate destruction, for he has kept secret the true fate of the silver after all others believed it lost at sea.
21. He finds himself becoming a slave of the silver and its secret, even as he slowly recovers it ingot by ingot during nighttime trips to Great Isabel.
22. The fate of Decoud is a mystery to Nostromo, which combined with the fact of the missing silver ingots only adds to his paranoia.
23. Eventually a lighthouse is constructed on Great Isabel, threatening Nostromo's ability to recover the treasure in secret.
24. The ever resourceful Nostromo manages to have a close acquaintance, the widower Giorgio Viola, named as its keeper.
25. Nostromo is in love with Giorgio's younger daughter, but ultimately becomes engaged to his elder daughter Linda.
26. One night while attempting to recover more of the silver, Nostromo is shot and killed, mistaken for a trespasser by old Giorgio.
\end{Verbatim}

\subsection{\emph{This Side of Paradise} by F.~Scott Fitzgerald}
\label{sec:805-summary}

\begin{Verbatim}[frame=single,commandchars=\\\{\},fontsize=\small]
1.  Amory Blaine, a young Midwesterner, believes that he has a great destiny, but the precise nature of this destiny eludes him.
2.  He attends a preparatory school where he becomes a football quarterback.
3.  He grows estranged from his eccentric mother Beatrice Blaine and becomes the protege of Monsignor Thayer Darcy, a Catholic priest.
4.  During his sophomore year at Princeton, he returns to Minneapolis over Christmas break and falls in love with Isabelle Borge, a wealthy debutante whom he first met as a boy.
5.  Amory and Isabelle embark upon a romance.
6.  While at Princeton, Amory deluges Isabelle with letters, but she becomes disenchanted with him due to his criticism, and they break up on Long Island.
7.  Following their separation, Amory accompanies a Princeton classmate to an apartment occupied by two New York showgirls of easy virtue.
8.  He considers staying the night with the showgirls, but his conscience and an apparition compel him to leave.
9.  After four years at Princeton, he enlists in the United States Army amid World War I. He ships overseas to serve in the muddy trenches of the Western front.
10. While overseas, his mother Beatrice dies, and most of his family's wealth disappears due to a series of failed investments.
11. \textcolor{skipred}{After the armistice with Imperial Germany in November 1918, Amory settles in New York City amid the flowering of the Jazz Age.}
12. Rebounding from Isabelle, he becomes infatuated with Rosalind Connage, a cruel and narcissistic flapper.
13. Desperate for a job, Amory obtains employment with an advertising agency but detests the work.
14. His relationship with Rosalind deteriorates as she prefers a rival suitor, Dawson Ryder, a man of wealth and status.
15. Rejected by Rosalind due to his lack of financial prospects, Amory quits his advertising job and goes on a drinking binge until the start of prohibition in the United States.
16. When Amory travels to visit an uncle in Maryland, he meets Eleanor Savage, a beautiful and reckless atheist.
17. Eleanor chafes under the religious conformity and gender limitations imposed on her by contemporary society in Wilsonian America.
18. Amory and Eleanor spend a lazy summer conversing about love.
19. On their final night together before Amory returns to New York City, Eleanor attempts suicide by riding her horse over a cliff in order to prove her disbelief in any deity.
20. At the last moment, she leaps to safety as her horse plummets over the precipice, and Amory realizes that he does not love her.
21. Returning to New York City, Amory learns of Rosalind's engagement to his wealthy rival Dawson Ryder, and he declares that Rosalind is now dead to him.
22. The death of his beloved mentor, Monsignor Darcy, further dispirits him.
23. Homeless, Amory wanders from New York City to his alma mater Princeton.
24. Accepting a car ride from a wealthy upper-class man driven by his working-class chauffeur in a Locomobile, Amory speaks out in favor of socialism in the United States-although he admits he is still formulating his thoughts as he is talking.
25. While riding in the Locomobile, Amory continues his argument about their time's societal ills and articulates his disillusionment with the current historical era.
26. He announces his hope to stand alongside those in the younger generation and to bring forth a new age in America.
27. Both the upper-class and working-class men in the car denounce his views, but when Amory discovers that the upper-class man is the father of a Princeton classmate who died in World War I, they reconcile.
28. Amory parts ways with his travel companions, and the upper-class man tells him: "Good luck to you and bad luck to your theories."
29. Approaching Princeton, Amory recognizes his selfishness as well as his overindulgence in drink and beauty.
30. He realizes that he must transcend these flaws to become a better man.
31. Wandering through a graveyard at twilight, he reflects upon his inevitable mortality and finds solace in the fact that future generations may one day ponder his life.
32. He thinks of the next generation-inheriting disillusionment and a loss of faith, yet still chasing love and success.
33. After midnight, he stands alone gazing at Princeton's gothic towers and feels a newfound freedom.
34. He stretches out his arms and proclaims, "I know myself . . .
35. but that is all."
\end{Verbatim}

\subsection{\emph{Heart of Darkness} by Joseph Conrad}
\label{sec:219-summary}

\begin{Verbatim}[frame=single,commandchars=\\\{\},fontsize=\small]
1.  The novella opens on "the sea-reach of the Thames" where Charles Marlow tells his friends that "when the Romans first came here, nineteen hundred years ago" they would have sensed "the savagery, the utter savagery" surrounding them.
2.  Marlow then relates how he became captain of a river steamboat for an ivory trading company.
3.  He tells of his fascination as a child for "the blank spaces" on maps, particularly in Africa.
4.  The image of a river on the map particularly drew his attention.
5.  In a flashback, Marlow makes his way to Africa by taking passage on a steamer.
6.  He travels 30 miles (50 km) up the river to where his company's station is.
7.  Work on a railway is taking place.
8.  Marlow explores a narrow ravine, and is horrified to find himself in a place full of critically ill Africans who worked on the railroad and are now dying.
9.  Marlow must wait for ten days in the company's devastated Outer Station.
10. Marlow meets the company's chief accountant, who tells him of a Mr. Kurtz, who is in charge of a very important trading post, and is described as a respected first-class agent.
11. The accountant predicts that Kurtz will go far.
12. Marlow departs with 60 men to travel to the Central Station, where the steamboat that he will command is based.
13. At the station, he learns that his steamboat has been wrecked in an accident.
14. The general manager informs Marlow that he could not wait for Marlow to arrive, and tells him of a rumour that Kurtz is ill.
15. Marlow fishes his boat out of the river and spends months repairing it.
16. Delayed by the lack of tools and replacement parts, Marlow is frustrated by the time it takes to perform the repairs.
17. He learns that Kurtz is resented, not admired, by the manager.
18. Once underway, the journey to Kurtz's station takes two months.
19. The journey pauses for the night about eight miles (13 km) below the Inner Station.
20. In the morning the boat is enveloped by a thick fog.
21. The steamboat is later attacked by a barrage of arrows, and the helmsman is killed.
22. Marlow sounds the steam whistle repeatedly, frightening the attackers away.
23. After landing at Kurtz's station, a man boards the steamboat: a Russian wanderer who strayed into Kurtz's camp.
24. Marlow learns that the natives worship Kurtz and that he has been very ill.
25. The Russian tells of how Kurtz opened his mind and how he admires Kurtz for his power and his willingness to use it.
26. Marlow suspects that Kurtz has gone mad.
27. Marlow observes the station and sees a row of posts topped with the severed heads of natives.
28. Around the corner of the house, Kurtz appears with supporters who carry him as a ghost-like figure on a stretcher.
29. The area fills with natives ready for battle, but Kurtz shouts something and they retreat.
30. His entourage carries Kurtz to the steamer and lays him in a cabin.
31. The manager tells Marlow that Kurtz has harmed the company's business in the region because his methods are "unsound".
32. The Russian reveals that Kurtz believes the company wants to kill him, and Marlow confirms that hangings were discussed.
33. After midnight, Kurtz returns to shore.
34. Marlow finds Kurtz crawling back to the station house.
35. Marlow threatens to harm Kurtz if he raises an alarm, but Kurtz only laments that he did not accomplish more.
36. The next day they prepare to journey back down the river.
37. Kurtz's health worsens during the trip.
38. The steamboat breaks down, and while stopped for repairs, Kurtz gives Marlow a packet of papers, including his commissioned report and a photograph, telling him to keep them from the manager.
39. When Marlow next speaks with him, Kurtz is near death; Marlow hears him weakly whisper, "The horror!
40. The horror!"
41. A short while later, the manager's boy announces to the crew that Kurtz has died (the famous line "Mistah Kurtz-he dead" would become the epigraph of T. S. Eliot's poem "The Hollow Men").
42. The next day Marlow pays little attention to Kurtz's pilgrims as they bury "something" in a muddy hole.
43. Returning to Europe, Marlow is embittered and contemptuous of the "civilised" world.
44. Several callers come to retrieve the papers Kurtz entrusted to him, but Marlow withholds them or offers papers he knows they have no interest in.
45. He gives Kurtz's report to a journalist, for publication if he sees fit.
46. Marlow is left with some personal letters and a photograph of Kurtz's fiancee.
47. When Marlow visits her, she is deep in mourning although it has been more than a year since Kurtz's death.
48. She presses Marlow for information, asking him to repeat Kurtz's final words.
49. Marlow tells her that Kurtz's final word was her name.
\end{Verbatim}

\subsection{\emph{Fanshawe} by Nathaniel Hawthorne}
\label{sec:7085-summary}

\begin{Verbatim}[frame=single,commandchars=\\\{\},fontsize=\small]
1.  Dr. Melmoth, the president of fictional Harley College, takes into his care Ellen Langton, the daughter of his friend, Mr. Langton, who is at sea.
2.  Ellen is a young, beautiful girl and attracts the attentions of the college boys, especially Edward Walcott, a strapping though immature student, and Fanshawe, a reclusive, meek intellectual.
3.  While out walking, the three young people meet a nameless character called 'the angler', a name he gets for appearing an expert fisherman.
4.  The angler asks for a word with Ellen, tells her something in secret, and apparently flusters her.
5.  Walcott and Fanshawe become suspicious of his intentions.
6.  We learn that the angler is an old friend of the reformed Inn owner, Hugh Crombie.
7.  The two had been at sea together, where Mr. Langton had been the angler's mentor and caretaker.
8.  Langton and the angler had a falling out, however, and, thinking that Langton has been killed at sea, the angler undertakes to marry Ellen in order to inherit her father's considerable wealth.
9.  \textcolor{skipred}{Thus in his secret meeting with Ellen, the angler instructs her to sneak out of Melmoth's home and follow him, telling her he has information about her father's whereabouts.}
10. His real aim, though, is to kidnap her, to tell her of her father's death, and to manipulate her into marrying him.
11. When the various men (Melmoth, Edward, Fanshawe) learn that she is not in her chamber, they go searching for her.
12. The search reveals the nature of each: Melmoth, an aged scholar unused to physical labor, enlists the help of Walcott, who is the most skilled rider and the most likely to be able to contend with the angler in a fight.
13. Fanshawe, who lags behind the search because of his weak constitution and his slow horse, is given information by an old woman in a cabin (where another old woman, Widow Butler, who turns out to be the angler's mother, has just died) that allows him to reach the angler and Ellen first.
14. The angler has taken Ellen to a craggy cliff and cave, where he intends to hold her captive.
15. Ellen has finally realized the angler's intentions.
16. When Fanshawe arrives, he stands above them, looking over the edge of the cliff.
17. The angler begins to climb up the cliff to fight Fanshawe but grabs a twig too weak to support him and tumbles to his death.
18. Fanshawe awakens Ellen from a faint, and they travel back to town together.
19. Fanshawe loves Ellen but knows that he will die young because of his shut-in lifestyle.
20. When Langton offers Ellen's hand in marriage to Fanshawe in exchange for rescuing her, he refuses, sacrificing his happiness so as not to subject her to a life of widowhood.
21. He also knows that Ellen has affections for Walcott.
22. Fanshawe dies at 20.
23. Ellen and Walcott marry four years later.
24. The narrator states that Walcott grows out of his childish ways (drunkenness, impulsiveness, the suggestion of teenage affairs) and becomes content with Ellen.
25. They are, according to the narrator, happy, but the book ends on an ambivalent note, stating that the couple did not produce children.
\end{Verbatim}

\subsection{\emph{The Sound and the Fury} by William Faulkner}
\label{sec:75170-summary}

\begin{Verbatim}[frame=single,commandchars=\\\{\},fontsize=\small]
1.  \textcolor{skipred}{The Compsons, a once-prosperous white family, live in Jefferson, Mississippi.}
2.  \textcolor{skipred}{They have three sons: Quentin, Benjamin ("Benjy"), and Jason, along with a daughter named Candace ("Caddy").}
3.  \textcolor{skipred}{The family also employs a black servant, Dilsey Gibson, who lives on their property with her family.}
4.  Early in their childhoods, the Compson children's grandmother ("Damuddy") dies.
5.  Caddy climbs a tree to look inside her grandmother's room and observe her funeral, while her brothers look up at her mud-soaked underwear.
6.  Later, when Benjy is five years old, his parents realize he's severely mentally disabled and incapable of speech; his mother, who originally named him after her brother Maury, renames him Benjamin.
7.  \textcolor{skipred}{In 1909, Caddy has sex with a boy named Dalton Ames.}
8.  \textcolor{skipred}{Quentin, distraught, attempts to defend her honor.}
9.  \textcolor{skipred}{He proposes a joint suicide with Caddy, tries to fight Ames, and tells his father that he, not Ames, had sex with his sister.}
10. \textcolor{skipred}{The Compsons send Quentin to Harvard University, paying for his tuition by selling Benjy's share of the family's land.}
11. \textcolor{skipred}{His parents then take Caddy, now pregnant, to French Lick, Indiana, where she meets a banker named Herbert Head.}
12. \textcolor{skipred}{Caddy and Head marry in 1910, and Head promises Jason Compson a job at his bank.}
13. \textcolor{skipred}{However, Caddy's father reveals that she was pregnant before meeting Head, leading Head to divorce her.}
14. On June 2, 1910, Quentin plans to kill himself.
15. After buying flat irons, he meets a young Italian girl who speaks no English at a bakery and tries to return her home, only for her brother to interpret him as a kidnapper.
16. He reports Quentin to the authorities, who make him pay a fine.
17. Later, Quentin attacks his boorish classmate Gerald Bland, who reminds him of Dalton Ames, but Bland knocks him unconscious.
18. Quentin returns to his dorm to clean himself up, then uses the irons to drown himself in the Charles River.
19. \textcolor{skipred}{Caddy names her daughter Quentin ("Miss Quentin") after her late brother.}
20. \textcolor{skipred}{Over the following years, Mr. Compson dies from alcoholism, leaving Jason and his mother to run the house.}
21. \textcolor{skipred}{Roskus, Dilsey's husband, also dies.}
22. Benjy's former plot of land, sold to pay for Quentin's tuition, becomes a golf course.
23. \textcolor{skipred}{Benjy is castrated after chasing a schoolgirl he believes to be Caddy.}
24. \textcolor{skipred}{Caddy leaves Miss Quentin with Jason and her mother, but they refuse to let Caddy live with them.}
25. Instead, Caddy sends them money every month, which Jason keeps for himself.
26. Most of the novel takes place over Easter weekend in 1928.
27. On Good Friday, Jason torments Miss Quentin, whom he blames for costing him the job at Head's bank, and unsuccessfully uses Caddy's money to short-sell cotton futures.
28. On Holy Saturday, Benjy celebrates his 33rd birthday.
29. Dilsey's grandson Luster looks after him while searching for a quarter so he can attend a traveling carnival; he ultimately receives one from Miss Quentin.
30. That night, Miss Quentin steals Jason's money.
31. She then escapes the Compson house by climbing down the same tree Caddy climbed as a child.
32. On Easter, Dilsey takes her family and Benjy to a black church, where they listen to a sermon from Reverend Shegog.
33. Meanwhile, Jason attempts to chase after Miss Quentin, who has left town with a man from the carnival.
34. After catching up with the carnival in the nearby town of Mottson, Jason provokes a fight with one of the workers and injures himself in the confrontation.
35. Having failed to find Miss Quentin, he hires a black man to drive him home.
36. Back in Jefferson, Luster takes Benjy to the town cemetery, but travels the wrong way around a Confederate monument in the town square.
37. Jason returns to find Benjy crying over this disruption to his routine; he intervenes and takes Benjy the right way.
\end{Verbatim}

\subsection{\emph{Pollyanna} by Eleanor H.~Porter}
\label{sec:1450-summary}

\begin{Verbatim}[frame=single,commandchars=\\\{\},fontsize=\small]
1.  The title character is Pollyanna Whittier, an eleven-year-old orphan who goes to live in the fictional town of Beldingsville, Vermont, with her wealthy but stern and cold spinster Aunt Polly Harrington, who does not want to take in Pollyanna but feels it is her duty to her late sister Jennie.
2.  Pollyanna's philosophy of life centers on what she calls "The Glad Game", an optimistic and positive attitude she learned from her father.
3.  The game consists of finding something to be glad about in every situation, no matter how bleak it may be.
4.  \textcolor{skipred}{It originated in an incident one Christmas when Pollyanna, who was hoping for a doll in the missionary barrel, found only a pair of crutches inside.}
5.  \textcolor{skipred}{Making the game up on the spot, Pollyanna's father taught her to look at the good side of thingsÑin this case, to be glad about the crutches because she did not need to use them.}
6.  \textcolor{skipred}{With this philosophy, and her own sunny personality and sincere, sympathetic soul, Pollyanna brings so much gladness to her aunt's dispirited New England town that she transforms it into a pleasant place to live.}
7.  The Glad Game shields her from her aunt's stern attitude: when Aunt Polly puts her in a stuffy attic room without carpets or pictures, she exults at the beautiful view from the high window; when she tries to "punish" her niece for being late to dinner by sentencing her to a meal of bread and milk in the kitchen with the servant Nancy, Pollyanna thanks her rapturously because she likes bread and milk, and she likes Nancy.
8.  Soon Pollyanna teaches some of Beldingsville's most troubled inhabitants to "play the game" as well, from Mrs. Snow, a querulous invalid, to Mr. Pendleton, a miserly bachelor who lives all alone in a cluttered mansion.
9.  Aunt Polly, too-finding herself helpless before Pollyanna's buoyant refusal to be downcast-gradually begins to thaw, although she resists the Glad Game longer than anyone else.
10. Eventually, however, even Pollyanna's robust optimism is put to the test when she is struck by a car and loses the use of her legs.
11. At first, she does not realize the seriousness of her injury, but her spirits plummet when she learns she will probably never walk again.
12. After that, she lies in bed, unable to find anything to be glad about.
13. Then the townspeople begin calling at Aunt Polly's house, eager to let Pollyanna know how much her encouragement has improved their lives; and Pollyanna decides she can still be glad that she at least had her legs.
14. The novel ends with Aunt Polly marrying her former lover Dr. Chilton and Pollyanna being sent to a specialist in spinal injuries, where she learns to walk again and is able to appreciate the use of her legs far more as a result of being temporarily disabled and unable to walk well.
\end{Verbatim}

\subsection{\emph{The Seven Dials Mystery} by Agatha Christie}
\label{sec:75288-summary}

\begin{Verbatim}[frame=single,commandchars=\\\{\},fontsize=\small]
1.  Sir Oswald and Lady Coote host a party at the stately home Chimneys, which they have rented for the season.
2.  The guest list includes Gerry Wade, Jimmy Thesiger, Ronny Devereux, Bill Eversleigh and Rupert "Pongo" Bateman.
3.  Since Wade has a bad habit of oversleeping, the others play a joke on him by placing eight alarm clocks in his room and timing them to go off at intervals.
4.  The next morning, a footman finds Wade dead in his bed, with chloral on his nightstand.
5.  Thesiger notices that one of the eight alarm clocks is missing.
6.  \textcolor{skipred}{It is later found in a hedge.}
7.  Lord Caterham and his daughter Lady Eileen 'Bundle' Brent move back into Chimneys.
8.  When Bundle drives to London to see Eversleigh, Ronny Devereux jumps out in front of her car.
9.  Before he dies, Devereux mutters "Seven Dials..." and "Tell...Jimmy Thesiger."
10. Bundle gets his body to a doctor, who tells her that her car did not hit Devereux; he was shot.
11. \textcolor{skipred}{Seven Dials turns out to be a seedy nightclub and gambling den in the Seven Dials area of London.}
12. Bundle recognises the doorman as Alfred, a footman from Chimneys.
13. Alfred tells her that he left Chimneys for far higher wages offered by Mosgorovsky, owner of the club.
14. Alfred takes Bundle into a secret room, where she hides in a cupboard and witnesses a meeting of six people wearing hoods with clock faces.
15. They talk of the always-missing "Number Seven", and about an upcoming party at Wyvern Abbey, where a scientist called Eberhard will offer a secret formula for sale to the British Air Minister.
16. At the party, the formula is stolen, but retrieved by Wade's stepsister; and Jimmy Thesiger is shot in his right arm.
17. Thesiger tells how he fought a man who climbed down the ivy.
18. The next morning, Battle finds a charred left-handed glove with teeth marks in the fireplace.
19. He theorises that the thief threw the gun onto the lawn from the terrace and then climbed back into the house via the ivy.
20. Bundle's father reports that Bauer, the footman who replaced Alfred, is missing.
21. Thesiger rings up Bundle and Gerry Wade's stepsister Loraine and tells them to meet him and Eversleigh at the Seven Dials club.
22. Bundle shows Thesiger the room where the Seven Dials meet.
23. Loraine finds Eversleigh unconscious in the car and they take him into the club.
24. Thesiger says that he will go for a doctor.
25. Someone knocks Bundle unconscious and she comes round in Eversleigh's arms.
26. Mr Mosgorovsky takes them into the meeting of the Seven Dials, where Number Seven is revealed as Superintendent Battle.
27. He reveals that they are a group of people doing secret service work for the government.
28. Battle tells Bundle that the association has succeeded with their main target, an international criminal whose stock in trade is the theft of secret formulae: Jimmy Thesiger was arrested that afternoon with his accomplice, Loraine Wade.
29. \textcolor{skipred}{Battle explains that Thesiger killed Wade and Devereux when they got onto his track.}
30. \textcolor{skipred}{Devereux took the eighth clock from Wade's room to see if anyone reacted to there being "seven dials".}
31. \textcolor{skipred}{At Wyvern Abbey, Thesiger stole the formula, passed it to Loraine, then shot himself in his right arm and disposed of his left-hand glove using his teeth.}
32. \textcolor{skipred}{Eversleigh feigned unconsciousness in the car outside the Seven Dials club.}
33. \textcolor{skipred}{Thesiger never went for a doctor, but hid in the club, and knocked Bundle unconscious.}
34. Bundle takes Wade's place in the Seven Dials and marries Bill Eversleigh.
\end{Verbatim}

\subsection{\emph{Wolfbane} by Frederik Pohl and C.~M.~Kornbluth}
\label{sec:51845-summary}

\begin{Verbatim}[frame=single,commandchars=\\\{\},fontsize=\small]
1.  The story opens in the town of Wheeling, West Virginia.
2.  We first see Roget Germyn, a banker and model Citizen.
3.  An Eye forms over him while he is meditating, but he is interrupted and the Eye vanishes.
4.  He then returns to a more Earthly concern - whether or not the 'sun' will be regenerated, ending the current period of hunger and cold.
5.  Attention then shifts to Glenn Tropile, who lives among the Citizens, but regards himself as one of the feared Wolves.
6.  He has even managed to slowly coach his wife, Gala, toward a more Wolfish outlook.
7.  Despite this rebellious attitude, he maintains a genuine interest in meditation.
8.  Glenn Tropile is exposed as a Wolf while stealing bread.
9.  He escapes execution and is collected by a community of Wolves living in the town of Princeton.
10. They find he doesn't entirely fit in there either, but hope he may get collected by an 'Eye', giving them a chance to measure this process in detail.
11. This eventually happens and they find, as expected, that his disappearance was facilitated by the Pyramid on Everest.
12. Glenn Tropile has been sent to the Pyramid's planet.
13. We then learn his fate.
14. To the Pyramids, the human race is nothing more than a useful source of 'Components' for a complex world-machine devoted mostly to feeding these artificial and semi-organic beings.
15. Tropile is suspended in a fluid-filled tank and 'wired in' to the vast computer system.
16. Later, he is linked to seven other humans as a 'Snowflake' - eight minds joined together to facilitate more complex tasks than a single human Component could manage.
17. In this condition, Tropile wakes, manages to retain his sanity, and wakes the other humans.
18. \textcolor{skipred}{They eventually merge with one another to form a sophisticated collective mind.}
19. \textcolor{skipred}{The freed Snowflake then spies on the Pyramids, finding that they have been traveling for some two million years, and have collected many species as Components, but seem locked in meaningless rituals surrounding an alien creature at the world's North Pole.}
20. \textcolor{skipred}{They later realize that this is the last survivor of the race that created the Pyramids.}
21. In the meantime, they have modified the collection process of human Components so that it selects persons known to at least one of the eight people composing the Snowflake (which has become almost as ruthless and inhuman as the alien Pyramids).
22. These humans are intended as 'mice,' disrupters of the planet, and later as an army with which to fight the Pyramids.
23. Roget Germyn is one of them, as is Tropile's wife.
24. \textcolor{skipred}{Facing a philosophical dead-end, the 'Snowflake' decides to separate its component minds to study the problem.}
25. Restored to individual identity, Glenn Tropile becomes horrified at what he has done.
26. \textcolor{skipred}{However, the majority of the others want to carry on.}
27. \textcolor{skipred}{While they are arguing, one of their number is taken over by the mind of the alien at the North Pole, who warns them that the Pyramids have noticed them and plan to wipe out half the planet to get rid of them.}
28. \textcolor{skipred}{They initially re-merge their personalities as the Snowflake and expedite their plans.}
29. Tropile decides he must physically disconnect himself from the Snowflake and leave to lead the humans.
30. They manage to defeat the Pyramids, but not before the remainder of the Snowflake is also destroyed.
31. The humans free many of the other Components and ship them back to Earth.
32. Tropile now finds he is a hero of sorts, but does not fit the role (though he never truly fit any role in which he was placed - Sheep, Wolf, or Component).
33. He also sees that there is a need for someone to wire themselves back into the alien planet's surviving systems, to re-kindle the 'sun' every five years and perhaps return the Earth to its original orbit.
34. He doesn't want to do it alone, but most of the people he knows are either unwilling or unsuitable.
35. On the last page, though, his wife agrees to join him.
36. He expects that there will later be others, that "[t]he ring of fire [will] grow."
\end{Verbatim}

\subsection{\emph{Germinal} by \'{E}mile Zola}
\label{sec:56528-summary}

\begin{Verbatim}[frame=single,commandchars=\\\{\},fontsize=\small]
1.  \textcolor{skipred}{The novel's central character is Etienne Lantier, previously seen in L'Assommoir (1877), and originally to have been the central character in Zola's "murder on the trains" thriller La Bote humaine (1890) before the overwhelmingly positive reaction to Germinal persuaded him otherwise.}
2.  The young migrant worker arrives at the forbidding coal mining town of Montsou in the bleak area of the far north of France to earn a living as a miner.
3.  Sacked from his previous job on the railways for assaulting a superior, Etienne befriends the veteran miner Maheu, who finds him somewhere to stay and gets him a job pushing the carts down the pit.
4.  Etienne is portrayed as a hard-working idealist but also a naive youth; Zola's genetic theories come into play as Etienne is presumed to have inherited his Macquart ancestors' traits of hotheaded impulsiveness and an addictive personality capable of exploding into rage under the influence of drink or strong passions.
5.  \textcolor{skipred}{Zola keeps his theorizing in the background and Etienne's motivations are much more natural as a result.}
6.  He embraces socialist principles, reading large amounts of working class movement literature and fraternizing with Souvarine, a Russian anarchist and political emigre who has also come to Montsou to seek a living in the pits, and Rasseneur, a pub owner.
7.  \textcolor{skipred}{Etienne's simplistic understanding of socialist politics and their rousing effect on him are very reminiscent of the rebel Silvere in the first novel in the cycle, La Fortune des Rougon (1871).}
8.  While this is going on, Etienne also falls for Maheu's daughter Catherine, also employed pushing carts in the mines, and he is drawn into the relationship between her and her brutish lover Chaval, a prototype for the character of Buteau in Zola's later novel La Terre (1887).
9.  The complex tangle of the miners' lives is played out against a backdrop of severe poverty and oppression, as their working and living conditions continue to worsen throughout the novel; eventually, pushed to breaking point, the miners decide to strike and Etienne, now a respected member of the community and recognized as a political idealist, becomes the leader of the movement.
10. While the anarchist Souvarine preaches violent action, the miners and their families hold back, their poverty becoming ever more disastrous, until they are sparked into a ferocious riot, the violence of which is described in explicit terms by Zola, as well as providing some of the novelist's best and most evocative crowd scenes.
11. The rioters are eventually confronted by police and the army that repress the revolt in a violent and unforgettable episode.
12. Disillusioned, the miners go back to work, blaming Etienne for the failure of the strike; then, Souvarine sabotages the entrance shaft of one of the Montsou pits, trapping Etienne, Catherine and Chaval at the bottom.
13. The ensuing drama and the long wait for rescue are among some of Zola's best scenes, and the novel draws to a dramatic close.
14. After Chaval is killed by Etienne, Catherine and Etienne are finally able to be lovers before Catherine dies in his arms.
15. Etienne is eventually rescued and fired but he goes on to live in Paris with Pluchart, an organizer for the International.
\end{Verbatim}

\subsection{\emph{Ulysses} by James Joyce}
\label{sec:4300-summary}

\begin{Verbatim}[frame=single,commandchars=\\\{\},fontsize=\small]
1.  At 8 a.m., Malachi "Buck" Mulligan, a boisterous medical student, calls an aspiring writer, Stephen Dedalus, up to the roof of the Sandycove Martello tower, where they live.
2.  There is tension between Stephen and Mulligan: Stephen overheard Mulligan make a cruel remark about Stephen's recently deceased mother, and Mulligan has invited an English student, Haines, whom Stephen dislikes, to stay with them.
3.  The three men eat breakfast and walk to the shore, where Mulligan demands from Stephen the key to the tower and a loan.
4.  The three make plans to meet at a pub, The Ship, at 12:30pm.
5.  Departing, Stephen decides that he will not return to the tower that night, as Mulligan, the "usurper", has taken it over.
6.  Stephen is teaching a history class on the victories of Pyrrhus of Epirus.
7.  After class, one student, Cyril Sargent, stays behind so that Stephen can show him how to do a set of algebraic exercises.
8.  Stephen looks at Sargent's ugly face and tries to imagine Sargent's mother's love for him.
9.  He then visits unionist school headmaster Garrett Deasy, from whom he collects his pay.
10. Deasy asks Stephen to take his long-winded letter about foot-and-mouth disease to a newspaper office for printing.
11. The two discuss Irish history and Deasy lectures on what he believes is the role of Jews in the economy.
12. As Stephen leaves, Deasy jokes that Ireland has "never persecuted the Jews" because the country "never let them in".
13. This episode is the source of some of the novel's best-known lines, such as Dedalus's claim that "history is a nightmare from which I am trying to awake" and that God is "a shout in the street".
14. Stephen walks along Sandymount Strand for some time, mulling various philosophical concepts, his family, his life as a student in Paris, and his mother's death.
15. As he reminisces he lies down among some rocks, watches a couple whose dog urinates behind a rock, scribbles some ideas for poetry and picks his nose.
16. This chapter is characterised by a stream of consciousness narrative style that changes focus wildly.
17. Stephen's education is reflected in the many obscure references and foreign phrases employed in this episode, which have earned it a reputation for being one of the book's most difficult chapters.
18. \textcolor{skipred}{The narrative shifts abruptly.}
19. The time is again 8 a.m., but the action has moved across the city and to the second protagonist of the book, Leopold Bloom, a part-Jewish advertising canvasser.
20. The episode opens with the line "Mr. Leopold Bloom ate with relish the inner organs of beasts and fowls."
21. After starting to prepare breakfast, Bloom decides to walk to a butcher to buy a mutton kidney.
22. Returning home, he prepares breakfast and brings it with the mail to his wife Molly as she lounges in bed.
23. One of the letters is from her concert manager Blazes Boylan, with whom she is having an affair.
24. Bloom reads a letter from their daughter Milly Bloom, who tells him about her progress in the photography business in Mullingar.
25. The episode closes with Bloom reading a magazine story titled "Matcham's Masterstroke", by Mr. Philip Beaufoy, while defecating in the outhouse.
26. While making his way to Westland Row post office Bloom is tormented by the knowledge that Molly will welcome Boylan into her bed later that day.
27. At the post office he surreptitiously collects a love letter from one 'Martha Clifford' addressed to his pseudonym, 'Henry Flower'.
28. He meets an acquaintance, and while they chat, Bloom attempts to ogle a woman wearing stockings, but is prevented by a passing tram.
29. Next, he reads the letter from Martha Clifford and tears up the envelope in an alley.
30. He wanders into a Catholic church during a service and muses on theology.
31. The priest has the letters I.N.R.I. or I.H.S. on his back; Molly had told Bloom that they meant I have sinned or I have suffered, and Iron nails ran in.
32. He buys a bar of lemon soap from a chemist.
33. He then meets another acquaintance, Bantam Lyons, who mistakenly takes him to be offering a racing tip for the horse Throwaway.
34. Finally, Bloom heads towards the baths.
35. The episode begins with Bloom entering a funeral carriage with three others, including Stephen's father.
36. They drive to Paddy Dignam's funeral, making small talk on the way.
37. The carriage passes both Stephen and Blazes Boylan.
38. There is discussion of various forms of death and burial.
39. Bloom is preoccupied by thoughts of his dead infant son, Rudy, and the suicide of his own father.
40. They enter the chapel for the service and subsequently leave with the coffin cart.
41. Bloom sees a mysterious man wearing a Mackintosh raincoat during the burial.
42. Bloom continues to reflect upon death, but at the end of the episode rejects morbid thoughts to embrace "warm fullblooded life".
43. At the office of the Freeman's Journal, Bloom attempts to place an ad.
44. Although initially encouraged by the editor, he is unsuccessful.
45. Stephen arrives bringing Deasy's letter about foot-and-mouth disease, but Stephen and Bloom do not meet.
46. Stephen leads the editor and others to a pub, relating an anecdote on the way about "two Dublin vestals".
47. The episode is broken into short segments by newspaper-style headlines, and is characterised by an abundance of rhetorical figures and devices.
48. Bloom's thoughts are peppered with references to food as lunchtime approaches.
49. He meets an old flame, hears news of Mina Purefoy's labour, and helps a blind boy cross the street.
50. He enters the restaurant of the Burton Hotel, where he is revolted by the sight of men eating like animals.
51. He goes instead to Davy Byrne's pub, where he consumes a gorgonzola cheese sandwich and a glass of burgundy, and muses upon the early days of his relationship with Molly and how the marriage has declined: "Me.
52. And me now."
53. Bloom's thoughts touch on what goddesses and gods eat and drink.
54. He ponders whether the statues of Greek goddesses in the National Museum have anuses as do mortals.
55. On leaving the pub Bloom heads toward the museum, but spots Boylan across the street and, panicking, rushes into the gallery across the street from the museum.
56. At the National Library, Stephen explains to some scholars his biographical theory of the works of Shakespeare, especially Hamlet, which he argues are based largely on the posited adultery of Shakespeare's wife.
57. Buck Mulligan arrives and interrupts to read out the telegram that Stephen had sent him indicating that he would not make their planned rendezvous at The Ship.
58. Bloom enters the National Library to look up an old copy of the ad he has been trying to place.
59. He passes in between Stephen and Mulligan as they exit the library at the end of the episode.
60. In this episode, nineteen short vignettes depict the movements of various characters, major and minor, through the streets of Dublin.
61. The episode begins by following Father Conmee, a Jesuit priest, on his trip north, and ends with an account of the cavalcade of the Lord Lieutenant of Ireland, William Ward, Earl of Dudley, through the streets, which is encountered by several characters from the novel.
62. In this episode, dominated by motifs of music, Bloom has dinner with Stephen's uncle at the Ormond hotel, while Molly's lover, Blazes Boylan, proceeds to his rendezvous with her.
63. While dining, Bloom listens to the singing of Stephen's father and others, watches the seductive barmaids, and composes a reply to Martha Clifford's letter.
64. This episode is narrated by an unnamed denizen of Dublin who works as a debt collector.
65. The narrator goes to Barney Kiernan's pub where he meets a character referred to only as "The Citizen".
66. This character is believed to be a satirisation of Michael Cusack, a founder member of the Gaelic Athletic Association.
67. When Leopold Bloom enters the pub, he is berated by the Citizen, who is a fierce Fenian and anti-Semite.
68. The episode ends with Bloom reminding the Citizen that his Saviour was a Jew.
69. As Bloom leaves the pub, the Citizen throws a biscuit tin at Bloom's head, but misses.
70. The episode is marked by extended tangents made in voices other than that of the unnamed narrator; these include streams of legal jargon, a report of a boxing match, Biblical passages, and elements of Irish mythology.
71. All the action of the episode takes place on the rocks of Sandymount Strand, the shoreline that Stephen visited in Episode 3.
72. A young woman, Gerty MacDowell, is seated on the rocks with her two friends, Cissy Caffrey and Edy Boardman.
73. The girls are taking care of three children, a baby, and four-year-old twins named Tommy and Jacky.
74. Gerty contemplates love, marriage and femininity as night falls.
75. The reader is gradually made aware that Bloom is watching her from a distance.
76. Gerty teases the onlooker by exposing her legs and underwear, and Bloom, in turn, masturbates.
77. Bloom's masturbatory climax is echoed by the fireworks at the nearby bazaar.
78. As Gerty leaves, Bloom realises that she has a lame leg, and believes this is the reason she has been "left on the shelf".
79. After several mental digressions he decides to visit Mina Purefoy at the maternity hospital.
80. It is uncertain how much of the episode is Gerty's thoughts, and how much is Bloom's sexual fantasy.
81. Some believe that the episode is divided into two halves: the first half the highly romanticized viewpoint of Gerty, and the other half that of the older and more realistic Bloom.
82. Joyce himself said, however, that "nothing happened between [Gerty and Bloom].
83. It all took place in Bloom's imagination".
84. Nausicaa attracted immense notoriety while the book was being published in serial form.
85. It has also attracted great attention from scholars of disability in literature.
86. The style of the first half of the episode borrows from (and parodies) romance magazines and novelettes.
87. Bloom's contemplation of Gerty parodies Dedalus's vision of the wading girl at the seashore in A Portrait of the Artist as a Young Man.
88. Bloom visits the maternity hospital where Mina Purefoy is giving birth, and finally meets Stephen, who has been drinking with his medical student friends and is awaiting the promised arrival of Buck Mulligan.
89. As the only father in the group of men, Bloom is concerned about Mina Purefoy in her labour.
90. He starts thinking about his wife and the births of his two children.
91. He also thinks about the loss of his only 'heir', Rudy.
92. The young men become boisterous, and start discussing such topics as fertility, contraception and abortion.
93. There is also a suggestion that Milly, Bloom's daughter, is in a relationship with one of the young men, Bannon.
94. They continue on to a pub to continue drinking, following the successful birth of a son to Mina Purefoy.
95. This chapter is remarkable for Joyce's wordplay, which, among other things, recapitulates the entire history of the English language.
96. After a short incantation, the episode starts with latinate prose, Anglo-Saxon alliteration, and moves on through parodies of, among others, Malory, the King James Bible, Bunyan, Pepys, Defoe, Sterne, Walpole, Gibbon, Dickens, and Carlyle, before concluding in a Joycean version of contemporary slang.
97. The development of the English language in the episode is believed to be aligned with the nine-month gestation period of the foetus in the womb.
98. Episode 15 is written as a play script, complete with stage directions.
99. The plot is frequently interrupted by "hallucinations" experienced by Stephen and Bloom-fantastic manifestations of the fears and passions of the two characters.
100. Stephen and his friend Lynch walk into Nighttown, Dublin's red-light district.
101. Bloom pursues them and eventually finds them at Bella Cohen's brothel where, in the company of her workers including Zoe Higgins, Florry Talbot and Kitty Ricketts, he has a series of hallucinations regarding his sexual fetishes, fantasies and transgressions.
102. In one of these hallucinations, Bloom is put in the dock to answer charges by a variety of sadistic, accusing women including Mrs Yelverton Barry, Mrs Bellingham and the Hon Mrs Mervyn Talboys.
103. In another of Bloom's hallucinations, he is crowned king of his own city, which is called Bloomusalem-Bloom imagines himself being loved and admired by Bloomusalem's citizens, but then imagines himself being accused of various charges.
104. As a result, he is burnt at the stake and several citizens pay their respects to him as he dies.
105. Then the hallucination ends, Bloom finds himself next to Zoe, and the two talk.
106. After they talk, Bloom continues to encounter other miscellaneous hallucinations, including one in which he converses with his grandfather Lipoti Virag, who lectures him about sex, among other things.
107. At the end of the hallucination, Bloom is speaking with some prostitutes when he hears a sound coming from downstairs.
108. He hears heels clacking on the staircase, and he observes what appears to be a male form passing down the staircase.
109. He speaks with Zoe and Kitty for a moment, and then sees Bella Cohen come into the brothel.
110. He observes her appearance and talks with her for a little while.
111. But this conversation subsequently begins another hallucination, in which Bloom imagines Bella to be a man named Mr. Bello and Bloom imagines himself to be a woman.
112. In this fantasy, Bloom imagines himself (or "herself", in the hallucination) being dominated by Bello, who both sexually and verbally humiliates Bloom.
113. Bloom also interacts with other imaginary characters in this scene before the hallucination ends.
114. After the hallucination ends, Bloom sees Stephen overpay at the brothel, and decides to hold onto the rest of Stephen's money for safekeeping.
115. Stephen hallucinates that his mother's rotting cadaver has risen up from the floor to confront him.
116. He cries Non serviam!,
117. uses his walking stick to smash a chandelier, and flees the room.
118. Bloom quickly pays Bella for the damage, then runs after Stephen.
119. He finds Stephen engaged in an argument with an English soldier, Private Carr, who, after hearing Stephen utter a perceived insult to King Edward VII, punches him.
120. The police arrive and the crowd disperses.
121. As Bloom tends to Stephen, he has a hallucination of his deceased son, Rudy, as an 11-year-old.
122. Bloom takes Stephen to a cabman's shelter near Butt Bridge to restore him to his senses.
123. There, they encounter a drunken sailor, D. B. Murphy (W. B. Murphy in the 1922 text).
124. The episode is dominated by the motif of confusion and mistaken identity, with Bloom, Stephen and Murphy's identities being repeatedly called into question.
125. The narrative's rambling and laboured style in this episode reflects the protagonists' nervous exhaustion and confusion.
126. Bloom returns home with Stephen, makes him a cup of cocoa, discusses cultural and linguistic differences between them, considers the possibility of publishing Stephen's parable stories, and offers him a place to stay for the night.
127. Stephen refuses Bloom's offer and is ambiguous in response to Bloom's proposal of future meetings.
128. The two men urinate in the backyard, Stephen departs and wanders off into the night, and Bloom goes to bed, where Molly is sleeping.
129. She awakens and questions him about his day.
130. The episode is written in the form of a rigidly organised and "mathematical" catechism of 309 questions and answers, and was reportedly Joyce's favourite episode in the novel.
131. The deep descriptions range from questions of astronomy to the trajectory of urination and include a list of 25 men that purports to be the "preceding series" of Molly's suitors and Bloom's reflections on them.
132. While describing events apparently chosen randomly in ostensibly precise mathematical or scientific terms, the episode is rife with errors made by the undefined narrator, many or most of which are intentional by Joyce.
133. The final episode consists of Molly Bloom's thoughts as she lies in bed next to her husband.
134. The episode uses a stream-of-consciousness technique in eight paragraphs and lacks punctuation.
135. Molly thinks about Boylan and Bloom, her past admirers, including Lieutenant Stanley G. Gardner, the events of the day, her childhood in Gibraltar, and her curtailed singing career.
136. She also hints at a lesbian relationship in her youth, with a childhood friend, Hester Stanhope.
137. These thoughts are occasionally interrupted by distractions, such as a train whistle or the need to urinate.
138. Molly is surprised by the early arrival of her menstrual period, which she ascribes to her vigorous sex with Boylan.
139. The episode concludes with Molly's remembrance of Bloom's marriage proposal, and of her acceptance: "he asked me would I yes to say yes my mountain flower and first I put my arms around him yes and drew him down to me so he could feel my breasts all perfume yes and his heart was going like mad and yes I said yes I will Yes."
\end{Verbatim}

\subsection{\emph{Micromegas} by Voltaire}
\label{sec:30123-summary}

\begin{Verbatim}[frame=single,commandchars=\\\{\},fontsize=\small]
1.  \textcolor{skipred}{The story is organized into seven brief chapters.}
2.  The first describes Micromegas, whose name literally means "small-large", an inhabitant of a planet orbiting the star Sirius.
3.  Micromegas stands 120,000 royal feet (38.9 km) tall and his circumference at the waist is 50,000 royal feet (16.24 km).
4.  The Sirian's home world is calculated to be 21.6 million times greater in circumference than Earth using mathematical ratios in a passage intended to relativize Man's home on a cosmic scale.
5.  When he is almost 450 years old, approaching the end of what the inhabitants of the planet orbiting Sirius consider his childhood and having already solved over fifty of Euclid's problems (eighteen more than Blaise Pascal) before the age of 250 years while studying at his planet's Jesuit college, Micromegas writes a scientific book examining the insects on his planet, which at 100 royal feet (32.5 m) are too small to be detected by ordinary Sirian microscopes.
6.  This book is considered heresy by his country's mufti, and after a 200-year trial, he is banished from the court for a term of 800 years.
7.  Micromegas takes this as an opportunity to travel between the various planets in a quest to develop his heart and his mind.
8.  Micromegas proceeds to begin his journey, traveling by taking advantage of gravity and "the forces of repulsion and attraction" (a reference endorsing the work of Sir Isaac Newton), and after extensive celestial travels he arrives on Saturn, where he befriends the native population and developed an intimate friendship with the secretary of the Academy of Saturn, a man less than a twentieth of his size (a "dwarf" standing only 6,000 royal feet or 1.95 km tall) and described as being clever but lacking the capacity for true genius.
9.  In the second chapter, they discuss the differences between their planets.
10. The Saturnian has 72 senses while the Sirian has almost 1,000.
11. The Saturnian lives for 15,000 Saturnian years while the Sirian lives 700 times longer; Micromegas reports that he has visited worlds where people live much longer than this, but who still consider their lifespans too short.
12. All of this further relativizes the size of the Earth in relation to the extraterrestrials, but Micromegas also engages the Saturnian philosophically and found him disappointing.
13. At the end of their conversation, they decide to take a philosophical journey together, and, in a comedic passage that begins chapter three, the Saturnian's mistress arrives with the intent of preventing her lover's departure.
14. The Secretary woos her and she leaves to console herself with a local dandy.
15. The two aliens set off from Saturn in pursuit of knowledge, visiting Saturn's ring, its moons, Jupiter's moons, Jupiter itself (for one Earth-year), and Mars, which they find so small that they fear that they cannot even lay down.
16. Eventually, they arrive on Earth on July 5, 1737 at the end of the third chapter and pause only to eat some mountains for lunch at the start of chapter four before circumnavigating the globe in 36 hours with the Saturnian only getting his lower legs wet in the deepest ocean and the Sirian barely wetting his ankles.
17. The Saturnian decides that the planet must be devoid of life, since he had as of yet seen none but Micromegas chastises him, resisting the temptation to make hasty conclusions and using his reason to direct his search.
18. The Sirian fashions a magnifying glass from a diamond in his necklace measuring 160 royal feet in diameter and spots a tiny speck in the Baltic sea which he discovers is a whale.
19. The Saturnian proceeds to ask many questions, including how such a tiny "atom" could move, if it was sentient, and many others which embarrassed the Sirian.
20. As they examine it, Micromegas finds a boatful of philosophers on their return from the Arctic Circle and carefully picks their ship up.
21. In chapter five, the space travelers examine the boat and notice the men aboard only upon their driving a pole into his finger.
22. It is here that Voltaire breaks with the narrative to briefly relativize Man's diminutive size using the ratio of a man's height to the size of the Earth and uses the moment to perform the same calculus on the scale of human conflict.
23. Using their magnifying-glass, the travelers become able to see the humans.
24. In chapter six, the Secretary hastily concludes that the tiny beings are too small to be of any intelligence or spirit, and Micromegas reasons with him to convince his companion that what he sees is the humans speaking with each other.
25. Still, they cannot yet hear them and the travelers devise a hearing tube made with the clippings of Micromegas's fingernails in order to hear the tiny voices.
26. After listening for a while, they come to discern the words spoken and to understand French.
27. In order to establish communication while fearing that their full voices might deafen the humans, they devise a method in which they carry their suppressed voices through toothpicks to the men on the Sirian's finger.
28. They begin a conversation, wherein they are shocked to discover the breadth of the human intellect but also are exposed to human vanity and philosophy, which the travelers come to mock.
29. The travelers first are amazed at the humans' ability to measure their visitors, establishing an equality of the mind at all scales, and informs the travelers that such creatures as bees exist and that animals exist that are equally as small to bees as men are to the Micromegas.
30. The seventh and final chapter sees the humans testing the philosophies of Aristotle, Descartes, Malebranche, Leibniz and Locke against the travelers' wisdom.
31. Beginning the deeper conversation, one of the human philosophers explains to the extraterrestrial visitors that Mankind had not found lasting happiness and that, to the contrary, hundreds of thousands of men will go to war against each other for, in the novella's relativization, insignificant quarrels.
32. At this, the Saturnian is impassioned with anger, entertaining the thought of stamping out the armies with three steps.
33. The conversation shifts upon the travelers' learning the occupation of their interlocutors towards the scientific prowess of Man, which ends when philosophical questions are asked.
34. Each philosopher espouses the teachings that he follows, and Micromegas finds fault in each theory save for that of the disciple of Locke, who exhibits philosophical modesty.
35. When the travelers hear the theory of Aquinas from his Summa Theologica that the universe was made uniquely for mankind, they fall into an enormous fit of laughter which causes the ship and its philosophers to fall in the Sirian's pocket.
36. Micromegas then is angry with the arrogance of Mankind and, taking pity on the humans, the Sirian decides to write them a book that will explain everything to them philosophically.
37. When the volume is presented to the French Academy of Sciences, the Academy's secretary opens the book only to find blank pages.
\end{Verbatim}

\section{Per-Book Metric Values}
\begin{center}
\newcolumntype{R}{>{\RaggedRight\arraybackslash}X}
\renewcommand{\arraystretch}{1.3}
\rowcolors{2}{gray!15}{white}
\footnotesize
\begin{xltabular}{\linewidth}{Rrrrrrrr}
    \toprule
    \textbf{Title} & \textbf{$l$} & \textbf{$p_c$} & \textbf{$p_s$} & \textbf{$\mu_m$} & \textbf{$\mu_{ODD}$} & \textbf{$G_c$} & \textbf{$G_s$} \\
    \midrule
    \endfirsthead        % header appears on the FIRST page only
    \endhead             % (empty -> no repeated header on continuation pages)
    A Christmas Carol & 1.00 & 1.00 & 1.00 & 0.52 & 0.07 & 0.07 & 0.03 \\
    A Farewell to Arms & 0.91 & 0.95 & 0.99 & 0.56 & 0.10 & 0.32 & 0.18 \\
    A High Wind in Jamaica & 0.88 & 0.90 & 0.92 & 0.62 & 0.12 & 0.30 & 0.10 \\
    A Journey to the Centre of the Earth & 1.00 & 0.89 & 1.00 & 0.50 & 0.09 & 0.17 & 0.38 \\
    A Little Princess & 0.89 & 1.00 & 1.00 & 0.48 & 0.08 & 0.29 & 0.21 \\
    A Room with a View & 0.98 & 0.90 & 0.98 & 0.44 & 0.07 & 0.36 & 0.07 \\
    A Study in Scarlet & 1.00 & 0.93 & 0.98 & 0.56 & 0.06 & 0.36 & 0.02 \\
    A Tale of Two Cities & 0.98 & 0.62 & 0.99 & 0.61 & 0.10 & 0.35 & 0.09 \\
    A Voyage to Arcturus & 0.98 & 0.95 & 0.98 & 0.58 & 0.09 & 0.25 & 0.06 \\
    Adam Bede & 1.00 & 0.41 & 0.80 & 0.55 & 0.16 & 0.04 & 0.49 \\
    Alice's Adventures in Wonderland & 0.98 & 1.00 & 1.00 & 0.48 & 0.08 & 0.25 & 0.10 \\
    Anne of Green Gables & 0.58 & 0.71 & 1.00 & 0.38 & 0.13 & 0.31 & 0.49 \\
    Anthem & 0.95 & 0.92 & 1.00 & 0.38 & 0.13 & 0.44 & 0.05 \\
    Armadale & 0.80 & 0.70 & 0.90 & 0.41 & 0.17 & 0.26 & 0.52 \\
    Around the World in Eighty Days & 0.99 & 0.92 & 1.00 & 0.51 & 0.04 & 0.26 & 0.13 \\
    Arrowsmith & 0.96 & 0.65 & 1.00 & 0.55 & 0.15 & 0.21 & 0.37 \\
    Babbitt & 0.82 & 0.71 & 1.00 & 0.51 & 0.11 & 0.23 & 0.33 \\
    Bambi, a Life in the Woods & 0.97 & 0.76 & 1.00 & 0.47 & 0.05 & 0.29 & 0.07 \\
    Bealby: A Holiday & 0.96 & 1.00 & 0.92 & 0.59 & 0.13 & 0.20 & 0.20 \\
    Beauvallet & 0.98 & 0.73 & 1.00 & 0.35 & 0.32 & 0.26 & 0.46 \\
    Candide & 1.00 & 0.97 & 1.00 & 0.51 & 0.05 & 0.31 & 0.13 \\
    Captain Blood & 0.85 & 0.45 & 1.00 & 0.33 & 0.27 & 0.21 & 0.21 \\
    Carmilla & 0.90 & 0.94 & 1.00 & 0.53 & 0.08 & 0.32 & 0.06 \\
    Catriona & 0.88 & 0.77 & 0.94 & 0.55 & 0.21 & 0.19 & 0.36 \\
    Clementina & 0.77 & 0.81 & 0.86 & 0.57 & 0.24 & 0.21 & 0.57 \\
    David Copperfield & 0.84 & 0.61 & 0.98 & 0.38 & 0.15 & 0.28 & 0.20 \\
    Deathworld & 0.96 & 0.86 & 0.86 & 0.58 & 0.12 & 0.40 & 0.27 \\
    Demos & 0.80 & 0.73 & 1.00 & 0.46 & 0.12 & 0.34 & 0.27 \\
    Dracula & 0.88 & 0.93 & 0.97 & 0.54 & 0.08 & 0.32 & 0.28 \\
    Emma & 0.88 & 0.67 & 1.00 & 0.55 & 0.10 & 0.25 & 0.18 \\
    Equality & 0.72 & 0.71 & 1.00 & 0.49 & 0.16 & 0.24 & 0.49 \\
    Evelina & 0.94 & 0.62 & 0.95 & 0.50 & 0.09 & 0.25 & 0.38 \\
    Fanshawe & 0.98 & 0.90 & 0.96 & 0.62 & 0.14 & 0.32 & 0.11 \\
    Flatland: A Romance of Many Dimensions & 0.99 & 0.59 & 1.08 & 0.71 & 0.19 & 0.24 & 0.13 \\
    Frankenstein; Or, The Modern Prometheus & 0.74 & 0.71 & 1.00 & 0.66 & 0.21 & 0.32 & 0.10 \\
    Freckles & 0.98 & 0.85 & 0.98 & 0.41 & 0.12 & 0.35 & 0.07 \\
    From the Earth to the Moon & 0.96 & 0.71 & 1.00 & 0.50 & 0.07 & 0.21 & 0.31 \\
    Germinal & 0.87 & 0.72 & 0.80 & 0.53 & 0.13 & 0.16 & 0.58 \\
    Green Mansions: A Romance of the Tropical Forest & 0.97 & 0.96 & 1.00 & 0.59 & 0.10 & 0.34 & 0.15 \\
    Greenmantle & 1.00 & 1.00 & 1.00 & 0.51 & 0.04 & 0.28 & 0.15 \\
    Gryll Grange & 0.80 & 0.67 & 1.00 & 0.50 & 0.09 & 0.24 & 0.30 \\
    Heart of Darkness & 1.00 & 1.00 & 1.00 & 0.62 & 0.22 & 0.31 & 0.00 \\
    Heidi & 0.94 & 0.91 & 0.83 & 0.55 & 0.09 & 0.26 & 0.40 \\
    Herland & 0.84 & 1.00 & 1.00 & 0.45 & 0.10 & 0.32 & 0.17 \\
    Howards End & 0.98 & 0.73 & 1.00 & 0.50 & 0.07 & 0.23 & 0.24 \\
    Huntingtower & 0.93 & 0.88 & 0.97 & 0.58 & 0.09 & 0.26 & 0.16 \\
    Indiana & 0.89 & 0.71 & 1.00 & 0.55 & 0.13 & 0.25 & 0.24 \\
    Ivanhoe: A Romance & 0.95 & 0.77 & 0.95 & 0.56 & 0.06 & 0.33 & 0.22 \\
    Jane Eyre & 0.98 & 0.89 & 0.96 & 0.49 & 0.06 & 0.30 & 0.14 \\
    Jude the Obscure & 0.92 & 0.77 & 1.00 & 0.55 & 0.12 & 0.23 & 0.28 \\
    Kazan & 0.99 & 0.70 & 1.00 & 0.48 & 0.06 & 0.33 & 0.09 \\
    Kenilworth & 1.00 & 0.73 & 0.94 & 0.56 & 0.12 & 0.22 & 0.26 \\
    Kidnapped & 0.99 & 0.97 & 1.00 & 0.48 & 0.07 & 0.29 & 0.17 \\
    Kim & 0.91 & 0.87 & 1.00 & 0.44 & 0.08 & 0.30 & 0.23 \\
    L'Assommoir & 0.46 & 0.85 & 1.00 & 0.47 & 0.20 & 0.31 & 0.31 \\
    Lady Audley's Secret & 0.92 & 0.73 & 0.97 & 0.45 & 0.06 & 0.26 & 0.17 \\
    Little Dorrit & 0.84 & 0.63 & 0.99 & 0.55 & 0.15 & 0.40 & 0.22 \\
    Little Women & 0.87 & 0.68 & 0.97 & 0.54 & 0.07 & 0.29 & 0.14 \\
    Madame Bovary & 0.97 & 0.94 & 1.00 & 0.54 & 0.06 & 0.23 & 0.32 \\
    Maggie: A Girl of the Streets & 0.98 & 0.89 & 1.00 & 0.44 & 0.06 & 0.20 & 0.20 \\
    Main Street & 0.94 & 0.40 & 1.00 & 0.40 & 0.27 & 0.23 & 0.32 \\
    Mansfield Park & 0.95 & 0.88 & 1.00 & 0.45 & 0.06 & 0.22 & 0.31 \\
    Marianela & 0.91 & 0.64 & 0.95 & 0.57 & 0.11 & 0.24 & 0.23 \\
    Mary Barton & 0.95 & 0.61 & 1.00 & 0.53 & 0.06 & 0.23 & 0.23 \\
    Metamorphosis & 1.00 & 1.00 & 1.00 & 0.50 & 0.14 & 0.04 & -0.00 \\
    Micromegas & 0.98 & 1.00 & 0.97 & 0.53 & 0.10 & 0.15 & 0.10 \\
    Middlemarch & 0.75 & 0.71 & 1.00 & 0.49 & 0.13 & 0.19 & 0.38 \\
    Mike & 1.00 & 0.85 & 1.00 & 0.53 & 0.02 & 0.21 & 0.22 \\
    Moods & 0.80 & 0.86 & 0.95 & 0.55 & 0.11 & 0.22 & 0.29 \\
    Moonfleet & 1.00 & 0.89 & 1.00 & 0.51 & 0.13 & 0.39 & 0.11 \\
    My Ántonia & 0.81 & 0.73 & 1.00 & 0.42 & 0.12 & 0.31 & 0.22 \\
    Ninety-Three & 0.91 & 0.70 & 0.96 & 0.51 & 0.08 & 0.26 & 0.37 \\
    Northanger Abbey & 0.98 & 0.74 & 1.00 & 0.61 & 0.13 & 0.30 & 0.15 \\
    Nostromo: A Tale of the Seaboard & 0.72 & 0.45 & 0.85 & 0.67 & 0.15 & 0.35 & 0.34 \\
    O Pioneers! & 0.92 & 0.83 & 0.98 & 0.48 & 0.08 & 0.43 & 0.11 \\
    Oblomov & 0.75 & 0.61 & 0.83 & 0.55 & 0.12 & 0.18 & 0.25 \\
    Oil! & 0.92 & 0.76 & 1.00 & 0.48 & 0.07 & 0.17 & 0.19 \\
    Oliver Twist & 0.96 & 0.66 & 1.00 & 0.54 & 0.12 & 0.23 & 0.23 \\
    Persuasion & 0.92 & 1.00 & 1.00 & 0.49 & 0.07 & 0.40 & 0.17 \\
    Pollyanna & 0.83 & 0.69 & 0.79 & 0.46 & 0.11 & 0.23 & 0.54 \\
    Pride and Prejudice & 0.98 & 0.61 & 1.00 & 0.52 & 0.06 & 0.29 & 0.13 \\
    Ramona & 0.84 & 0.85 & 1.00 & 0.47 & 0.13 & 0.36 & 0.25 \\
    Rinkitink in Oz & 0.97 & 0.96 & 1.00 & 0.50 & 0.05 & 0.28 & 0.15 \\
    Romola & 0.82 & 0.70 & 0.98 & 0.53 & 0.07 & 0.26 & 0.28 \\
    Rookwood & 0.71 & 0.57 & 1.00 & 0.46 & 0.17 & 0.20 & 0.33 \\
    Sandy & 1.00 & 0.75 & 0.93 & 0.47 & 0.17 & 0.19 & 0.41 \\
    She & 0.98 & 0.79 & 1.00 & 0.44 & 0.09 & 0.33 & 0.19 \\
    Shirley & 0.91 & 0.78 & 0.97 & 0.53 & 0.06 & 0.39 & 0.22 \\
    Siddhartha & 0.98 & 1.00 & 1.00 & 0.60 & 0.10 & 0.32 & 0.06 \\
    Sister Carrie: A Novel & 0.95 & 0.77 & 0.98 & 0.54 & 0.03 & 0.25 & 0.22 \\
    Sons and Lovers & 0.74 & 0.93 & 1.00 & 0.46 & 0.15 & 0.25 & 0.37 \\
    The Black Moth: A Romance of the XVIIIth Century & 0.62 & 0.68 & 0.93 & 0.42 & 0.14 & 0.27 & 0.27 \\
    The Box-Car Children & 0.92 & 0.94 & 0.86 & 0.54 & 0.05 & 0.28 & 0.22 \\
    The Cave Girl & 0.98 & 1.00 & 0.97 & 0.51 & 0.03 & 0.24 & 0.09 \\
    The Deerslayer & 1.00 & 0.31 & 1.00 & 0.57 & 0.13 & 0.17 & 0.17 \\
    The Documents in the Case & 0.78 & 0.39 & 1.00 & 0.58 & 0.23 & 0.27 & 0.35 \\
    The Expedition of Humphry Clinker & 0.10 & 0.79 & 1.00 & 0.49 & 0.35 & 0.41 & 0.51 \\
    The Further Adventures of Robinson Crusoe & 0.97 & 0.88 & 0.97 & 0.53 & 0.07 & 0.31 & 0.11 \\
    The Garies and Their Friends & 0.80 & 0.94 & 0.96 & 0.57 & 0.13 & 0.30 & 0.22 \\
    The Golovlyov Family & 0.95 & 0.69 & 0.98 & 0.35 & 0.19 & 0.36 & 0.17 \\
    The Good Soldier & 0.68 & 0.84 & 0.97 & 0.61 & 0.20 & 0.36 & 0.27 \\
    The Hidden Staircase & 0.94 & 0.92 & 1.00 & 0.44 & 0.08 & 0.28 & 0.23 \\
    The Hound of the Baskervilles & 0.97 & 0.93 & 1.00 & 0.52 & 0.06 & 0.26 & 0.08 \\
    The House of Mirth & 0.91 & 0.83 & 0.82 & 0.51 & 0.08 & 0.41 & 0.27 \\
    The House of the Seven Gables & 0.95 & 0.86 & 1.00 & 0.53 & 0.13 & 0.35 & 0.16 \\
    The Invisible Man & 0.96 & 0.86 & 1.00 & 0.54 & 0.10 & 0.28 & 0.18 \\
    The Jamesons & 0.82 & 1.00 & 0.97 & 0.57 & 0.15 & 0.18 & 0.14 \\
    The Last of the Mohicans & 0.96 & 0.85 & 1.00 & 0.59 & 0.09 & 0.29 & 0.13 \\
    The Maltese Falcon & 0.95 & 0.75 & 0.95 & 0.58 & 0.11 & 0.28 & 0.11 \\
    The Marrow of Tradition & 0.51 & 0.68 & 0.94 & 0.54 & 0.22 & 0.17 & 0.29 \\
    The Mill on the Floss & 0.82 & 0.71 & 0.94 & 0.49 & 0.17 & 0.26 & 0.50 \\
    The Murder of Roger Ackroyd & 0.89 & 0.56 & 1.00 & 0.42 & 0.15 & 0.36 & 0.10 \\
    The Mysteries of Udolpho & 0.92 & 0.81 & 0.99 & 0.56 & 0.13 & 0.33 & 0.28 \\
    The Narrative of Arthur Gordon Pym of Nantucket & 0.97 & 0.92 & 0.98 & 0.49 & 0.06 & 0.38 & 0.17 \\
    The Phantom of the Opera & 0.92 & 0.71 & 1.00 & 0.60 & 0.13 & 0.42 & 0.22 \\
    The Red Badge of Courage: An Episode of the American Civil War & 0.85 & 0.67 & 1.00 & 0.60 & 0.06 & 0.30 & 0.28 \\
    The Return of the Native & 0.89 & 0.73 & 0.97 & 0.48 & 0.09 & 0.36 & 0.20 \\
    The Rise of Silas Lapham & 0.71 & 0.85 & 0.92 & 0.47 & 0.15 & 0.29 & 0.38 \\
    The Scarlet Letter & 0.93 & 0.92 & 0.95 & 0.46 & 0.15 & 0.29 & 0.27 \\
    The Secret Agent: A Simple Tale & 0.92 & 1.00 & 0.98 & 0.46 & 0.07 & 0.34 & 0.07 \\
    The Seven Dials Mystery & 0.87 & 0.59 & 0.79 & 0.52 & 0.12 & 0.26 & 0.40 \\
    The Sound and the Fury & 0.52 & 1.00 & 0.59 & 0.53 & 0.23 & 0.17 & 0.47 \\
    The Three Musketeers & 0.97 & 0.87 & 1.00 & 0.47 & 0.08 & 0.30 & 0.26 \\
    The Time Machine & 0.83 & 0.94 & 0.97 & 0.48 & 0.07 & 0.22 & 0.12 \\
    The Triumph of the Scarlet Pimpernel & 1.00 & 0.35 & 1.00 & 0.32 & 0.28 & 0.25 & 0.21 \\
    The Turn of the Screw & 0.98 & 0.67 & 1.00 & 0.47 & 0.14 & 0.34 & 0.19 \\
    The Valley of Fear & 0.88 & 0.93 & 1.00 & 0.49 & 0.11 & 0.35 & 0.24 \\
    The Valley of the Moon & 0.70 & 0.52 & 0.95 & 0.57 & 0.17 & 0.21 & 0.40 \\
    The Vicar of Wakefield & 0.77 & 0.66 & 1.00 & 0.47 & 0.17 & 0.27 & 0.29 \\
    The Voyages of Doctor Dolittle & 0.99 & 0.69 & 1.00 & 0.53 & 0.07 & 0.18 & 0.23 \\
    The Way of All Flesh & 0.92 & 0.63 & 1.00 & 0.59 & 0.18 & 0.17 & 0.45 \\
    The Wind in the Willows & 0.83 & 0.92 & 0.97 & 0.39 & 0.16 & 0.31 & 0.09 \\
    The Wonderful Wizard of Oz & 0.99 & 0.88 & 1.00 & 0.50 & 0.06 & 0.37 & 0.13 \\
    This Side of Paradise & 0.99 & 0.90 & 0.97 & 0.67 & 0.17 & 0.36 & 0.06 \\
    Treasure Island & 0.99 & 0.85 & 1.00 & 0.52 & 0.09 & 0.25 & 0.19 \\
    Trilby & 0.89 & 0.88 & 0.88 & 0.60 & 0.13 & 0.32 & 0.16 \\
    Twenty Thousand Leagues under the Sea & 0.87 & 0.74 & 0.97 & 0.54 & 0.16 & 0.27 & 0.32 \\
    Ulysses & 0.97 & 1.00 & 0.99 & 0.56 & 0.06 & 0.30 & 0.01 \\
    Vanity Fair & 0.85 & 0.81 & 1.00 & 0.55 & 0.12 & 0.31 & 0.30 \\
    Victory: An Island Tale & 0.73 & 0.51 & 0.96 & 0.45 & 0.20 & 0.26 & 0.31 \\
    Vile Bodies & 0.95 & 1.00 & 0.96 & 0.52 & 0.05 & 0.18 & 0.14 \\
    Villette & 0.92 & 0.83 & 0.98 & 0.46 & 0.11 & 0.33 & 0.22 \\
    Virginia & 0.91 & 0.76 & 0.96 & 0.58 & 0.12 & 0.22 & 0.22 \\
    We & 0.75 & 0.75 & 1.00 & 0.48 & 0.13 & 0.32 & 0.31 \\
    What Maisie Knew & 0.83 & 0.81 & 1.00 & 0.60 & 0.13 & 0.19 & 0.54 \\
    Where Angels Fear to Tread & 0.95 & 1.00 & 1.00 & 0.49 & 0.07 & 0.21 & 0.13 \\
    White Fang & 0.98 & 0.92 & 1.00 & 0.46 & 0.05 & 0.30 & 0.21 \\
    Wolfbane & 0.91 & 0.80 & 0.81 & 0.55 & 0.14 & 0.40 & 0.29 \\
    Work: A Story of Experience & 1.00 & 0.90 & 1.00 & 0.49 & 0.04 & 0.29 & 0.04 \\
    Wuthering Heights & 0.97 & 0.88 & 0.98 & 0.45 & 0.12 & 0.31 & 0.22 \\
    \bottomrule
\end{xltabular}
\end{center}

% \end{wordcountsections}  
\end{document}